\documentclass[11pt]{article}

\usepackage[]{acl}

\usepackage{times}
\usepackage{latexsym}

\usepackage[T1]{fontenc}

\usepackage[utf8]{inputenc}

\usepackage{microtype}

\usepackage{inconsolata}

\usepackage{graphicx}

\usepackage[table]{xcolor}
\usepackage{makecell}
\usepackage{amsmath}

\usepackage{booktabs}
\usepackage{multirow}
\usepackage{subcaption}
\usepackage{arydshln}
\usepackage{soul}
\usepackage{algorithm}
\usepackage{algpseudocode}
\usepackage{rotating}
\usepackage{graphicx}
\usepackage{booktabs}
\definecolor{lightblue}{rgb}{0.68, 0.85, 0.9}
\usepackage{comment}
\usepackage{amssymb}
\usepackage{subcaption}
\definecolor{todo}{rgb}{1,0.5,0}
\usepackage{enumitem}

\definecolor{lightblue}{HTML}{D6EAF8}
\definecolor{lightgreen}{HTML}{D5F5E3}
\definecolor{lightgreen2}{HTML}{EEFBF4}
\definecolor{lightgreen3}{HTML}{EEFBF4}
\definecolor{lightyellow2}{HTML}{FFF9DB}
\definecolor{lightyellow3}{HTML}{FFFDEA}
\definecolor{lightyellow}{HTML}{FFF3B0}

\usepackage{pifont}
\newcommand{\cmark}{\ding{51}} %
\newcommand{\xmark}{\ding{55}} %

\usepackage{soul,xcolor}
\sethlcolor{yellow!20}      %
\newcommand{\hlg}[1]{\sethlcolor{green!5}\hl{#1}\sethlcolor{yellow!20}}

\usepackage{array}  %
\newcolumntype{M}[1]{>{\centering\arraybackslash}m{#1}}

\title{Author-in-the-Loop Response Generation and Evaluation: \\Integrating Author Expertise and Intent in Responses to Peer Review
}

\author{Qian Ruan, 
Iryna Gurevych  \\
        Ubiquitous Knowledge Processing Lab (UKP Lab)\\
        Department of Computer Science and Hessian Center for AI (hessian.AI)\\
        Technical University of Darmstadt \\
  \texttt{www.ukp.tu-darmstadt.de}}

\begin{document}
\maketitle
\begin{abstract}

Author response (rebuttal) writing is a critical stage of scientific peer review that demands substantial author effort. In practice, authors possess domain expertise, author-only information, and response strategies -- concrete forms of author expertise and intent -- and seek NLP assistance that integrates these signals into author response generation (ARG). Yet this author-in-the-loop paradigm lacks formal NLP formulation and systematic study: no dataset provides fine-grained author signals, existing ARG work lacks author inputs and controls, and no evaluation measures response reflection of author signals and effectiveness in addressing reviewer concerns. To fill these gaps, we introduce (i) \textit{Re$^3$Align}, the first large-scale dataset of aligned review–response–revision triplets, where revisions proxy author signals; (ii) \textit{REspGen}, an author-in-the-loop ARG framework supporting flexible author input, multi-attribute control, and evaluation-guided refinement; and (iii) \textit{REspEval}, a comprehensive evaluation suite with 20+ metrics spanning input utilization, controllability, response quality, and discourse. Experiments with SOTA LLMs demonstrate the benefits of author input and evaluation-guided refinement, the impact of input specificity on response quality, and controllability–quality trade-offs. 
We release our dataset,\footnote{\url{https://tudatalib.ulb.tu-darmstadt.de/handle/tudatalib/4982}} generation and evaluation tools.\footnote{\url{https://github.com/UKPLab/acl2026-respgen-respeval}}

\end{abstract}

\section{Introduction}
\label{sec:intro}
\begin{figure}[ht]
  \centering
  \includegraphics[width=0.48\textwidth]{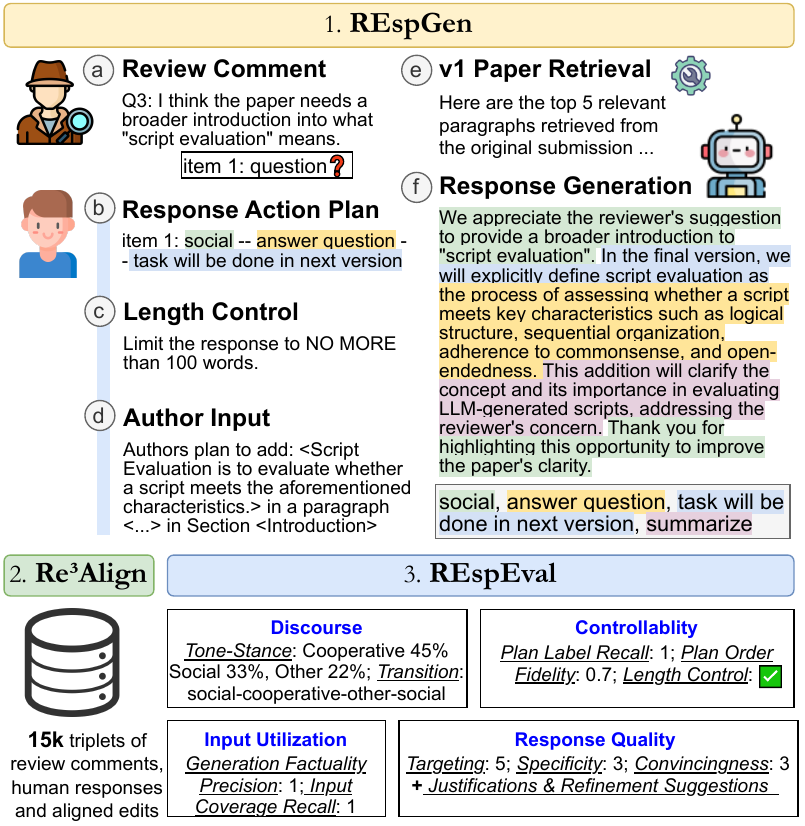}
  \caption{In this work, we contribute (1) \textit{\underline{REspGen}}, an author-in-the-loop ARG framework that integrates explicit author input (d), controllable planning and length (b–c), and additional paper context (e);
(2) \textit{\underline{Re$^3$Align}}, the first large-scale review–response–revision triplets dataset for modeling author signals; and
(3) \textit{\underline{REspEval}}, a comprehensive response evaluation framework with over 20 metrics spanning four dimensions.}
  \label{fig:fig1}
\end{figure}

Author response (rebuttal) writing is a critical stage of scientific peer review, where authors address reviewer concerns to seek favorable acceptance decisions. This process demands substantial author effort, making it a promising yet challenging NLP task \cite{what-can-nlp-do}.
In practice, authors write rebuttals drawing on domain expertise and strategic intent. Many reviewer concerns can only be addressed using information known exclusively to the authors, such as planned clarifications (e.g., the precise definition of \textit{script evaluation} in Figure~\ref{fig:fig1}), newly conducted experiments, and design rationales. Beyond content, authors strategically choose response approaches (revising, justifying, deferring, or promising future work) and control attributes such as length, tone, and discourse structure. Yet prior NLP research treats author response generation (ARG) as a generic, review-only text generation problem, failing to formalize this practice. We bridge this gap by reformulating ARG as an \textbf{author-in-the-loop} task.

To realize this, two fundamental questions remain unanswered. (1) \textit{How should author-in-the-loop ARG be formulated and benchmarked, and how should author expertise and intent be provided to generation systems?} Existing ARG studies rely solely on reviewer comments, lacking author expertise and intent \cite{jiu,reviewmt,re2}, and a systematic study supported by data and generation frameworks is missing. (2) \textit{Does the generated response reflect the author's provided expertise and intent while effectively addressing reviewer concerns?} This requires evaluating response quality, factual grounding, input coverage, and adherence to constraints and plans. Prior ARG work relies on surface similarity or coarse quality criteria, leaving these dimensions unsupported. Controllable ARG remains unexplored, and rigorous evaluation of controllability and trade-offs is an open challenge even in controllable text generation broadly \cite{ctg-survey,ctg-llm-survey}.

Author-in-the-loop ARG study is further hindered by \textbf{data scarcity}. Collecting author signals during live rebuttals is rarely feasible due to practical and ethical constraints. One alternative is approximating author signals using paper revisions: in conference settings, responses describe planned changes that later appear in revised papers, enabling post-hoc edit extraction to proxy author signals at response time; in journal settings, responses argue based on already-implemented revisions. This requires \emph{complete paper records} encompassing original and revised manuscripts, peer reviews, and authentic author responses, which are available in only a few existing datasets \cite{nlpeer,peerj}. More critically, existing resources lack the fine-grained annotations needed to model author signals at the granularity of individual reviewer concerns, such as edit analyses, review–response segment alignments, and mappings to concrete paper edits.

To address these challenges, we introduce \textbf{\textit{Re$^3$Align}} (§\ref{sec:dataset_construction}), the first large-scale dataset for author-in-the-loop ARG, comprising 3.4k complete paper records with 440k sentence-level edit annotations and 15k aligned review–response–edit triplets. 
We further propose \textbf{\textit{REspGen}} (§\ref{sec:respgen}), an author-in-the-loop ARG framework supporting various levels of author input specificity and enabling controllable generation over response planning and length, with iterative refinement guided by  \textbf{\textit{REspEval}} (§\ref{sec:respeval}). 
REspEval provides over 20 novel metrics spanning controllability, input utilization (factual grounding and coverage), response quality (targeting, specificity, and convincingness), and discourse characteristics (tone–stance profiles and transitions). 
Finally, we experiment with five SOTA LLMs across nine settings to systematically analyze the effects of author signals, input specificity, attribute control, and evaluation-guided refinement (§\ref{sec:results_discussion}).
Our work makes four key \textbf{contributions}:
\begin{itemize}[itemsep=0pt, parsep=0pt]
\item The first large dataset of review–response–edit triplets with rich annotations, enabling a new formulation of the ARG task;
\item  An author-in-the-loop ARG framework supporting flexible author input, multi-attribute control, and evaluation-guided refinement;
\item  A comprehensive evaluation suite with 20+ novel metrics for controllability, input utilization, response quality, and discourse;
\item  Extensive experiments across five LLMs and nine settings, yielding insights into ARG behavior under varied inputs and controls, and cross-dimensional trade-offs.
\end{itemize}

This work provides the first systematic formalization and study of author-in-the-loop response generation and evaluation, bridging author expertise and intent with NLP assistance to support effective and efficient author response writing.

\begin{table*}[ht]
\tabcolsep=0.03cm
\fontsize{8}{8}
\selectfont
\tabcolsep=0.04cm
\renewcommand{\arraystretch}{1.1}%
\begin{tabular}{l|lllll|lll|lll} 
 & & & & & & & & & & & \\[-6pt] \hline
&\multicolumn{4}{c}{Data} &&\multicolumn{2}{c}{Generation}&&\multicolumn{2}{c}{ Evaluation}\\ \hline
      &review &response&\begin{tabular}[c]{@{}l@{}} revision \\annotation \end{tabular}
      &\begin{tabular}[c]{@{}l@{}} triplet \\ alignment \end{tabular} &
      &\begin{tabular}[c]{@{}l@{}}author\\ input  \end{tabular} &\begin{tabular}[c]{@{}l@{}}author\\ control \end{tabular} 
      && dimension 
       &metric\\ \hline
       \begin{tabular}[c]{@{}l@{}}Jiu-Jitsu \citeyearpar{jiu} \end{tabular} 
       &\cmark seg & \cmark seg&\xmark&\xmark&
       &\xmark & \xmark 
       && Similarity&\begin{tabular}[c]{@{}l@{}}ROUGE, BERTScore\end{tabular}
       \\ \hline
       \begin{tabular}[c]{@{}l@{}}ReviewMT \citeyearpar{reviewmt} \end{tabular} 
         &\cmark doc & \cmark doc&\xmark&\xmark&
       &\xmark & \xmark 
       && Similarity&\begin{tabular}[c]{@{}l@{}}ROUGE, BLEU, METEOR\end{tabular}
       \\ \hline
       \begin{tabular}[c]{@{}l@{}}Re$^2$ \citeyearpar{re2} \end{tabular} 
          &\cmark doc & \cmark doc&\xmark&\xmark&
        &\xmark & \xmark 
       && \begin{tabular}[c]{@{}l@{}}Similarity, Quality\end{tabular}&\begin{tabular}[c]{@{}l@{}}ROUGE, BLEU, BERTScore\end{tabular}
       \\ \hline
        \textbf{Ours} &\cmark seg & \cmark seg &\cmark sent 
        &\begin{tabular}[c]{@{}l@{}}\cmark seg\end{tabular}
        &
        &\begin{tabular}[c]{@{}l@{}}\cmark \end{tabular}  &\begin{tabular}[c]{@{}l@{}}\cmark \end{tabular}  
       && \begin{tabular}[c]{@{}l@{}}Similarity, Quality, Discourse, \\ Input Utilization, Controllability\end{tabular} &\begin{tabular}[c]{@{}l@{}}20+ novel metrics\end{tabular}\\ \bottomrule
       \end{tabular}
       \caption{Comparison of related works on author response generation, including data, generation task formulations and evaluation dimensions and metrics. doc/seg/sent: document-/segment-/sentence-level alignments and annotations.
       } 
       \label{tab:related_work}
\end{table*}

\section{Related Work}
\label{sec:related_work}
\noindent\textbf{Author Response Generation} has recently emerged as a challenging and underexplored task in NLP for scientific peer review \cite{what-can-nlp-do,tasks-datasets-analysis}. Early work on author response includes argument-pair extraction \cite{ape} and response discourse analysis \cite{disapere}. Empirical studies further identify key success factors for effective responses, including explicit revision statements, high specificity, concrete evidence, and appropriate tone \cite{ten-simple-rules,does-my-rebuttal,what-makes-success}.
Recent work shifts toward generation, with studies on attitude- and theme-guided generation \cite{jiu} and multi-turn review–rebuttal dialogue \cite{reviewmt,re2}. 
However, these generation approaches rely solely on reviewer comments, producing generic responses that lack concrete details, especially those requiring author expertise. Evaluation is limited to similarity metrics, overlooking response diversity and broader success factors from empirical studies.
As summarized in Table~\ref{tab:related_work}, we address these limitations in three ways by (i) introducing the first large-scale triplet dataset of reviews, responses, and aligned sentence-level edits, treating revisions as explicit signals of author expertise and intent; (ii) formulating ARG as an author-in-the-loop task integrating author expertise and intent through explicit input and controllable generation; (iii) proposing a comprehensive evaluation suite with 20+ metrics spanning four dimensions beyond similarity-based evaluation.

\begin{table}[]
\tabcolsep=0.06cm
\fontsize{8}{8}
\selectfont
\tabcolsep=0.06cm
\renewcommand{\arraystretch}{1.15}%
\begin{tabular}[t]{l|l|l|l|l|l }
 & & & & & \\[-6pt] \hline
 & \#Paper & \#Pair 
&\#Edit & \#Linked Edit & \#Re$^3$ Triplet \\\hline
EMNLP24  &679 & 2,108  &86,247 &16,762  &1,933 \\ 
PeerJ  & 2,715 &13,963 & 353,551 &181,534 &13,588 \\ \hline
Total & 3,394 &16,071 & 439,798 &198,296 &15,521 \\ \bottomrule
\end{tabular}
\caption{Re$^3$Align Dataset Statistics. Reported are the counts of papers, aligned review–response pairs, annotated sentence-level edits, edits linked to the pairs, and the final number of aligned triplets.}
\label{tab:data_stat}
\end{table}

\noindent\textbf{Controllable Text Generation and Evaluation} aims to steer model outputs toward user-specified constraints \cite{ctg-survey}. Prior work primarily focuses on single-attribute control, including length \cite{ctrl-len}, topic \cite{ctg-topic1}, and sentiment \cite{ctg-senti}, as well as content-based control such as query-focused \cite{c-query}, entity-centric \cite{c-entity2}, and aspect-based generation \cite{c-aspect}. Recent surveys highlight persistent challenges in simultaneous multi-attribute control, trade-offs between controllability and generation quality, and the lack of rigorous evaluation methods \cite{ctg-survey,ctg-llm-survey}.
In ARG, controllability remains unexplored despite authors' need to strategically control response construction while integrating their own content. We provide the first study of controllability in ARG, examining control over length, discourse planning, and content integration. We further introduce a comprehensive evaluation framework with fine-grained metrics assessing: (i) how well generations adhere to single- and simultaneous multi-attribute controls; (ii) how effectively author-provided content is incorporated; and (iii) how response quality is impacted.

\section{Dataset Construction: \textit{Re$^3$Align}}
\label{sec:dataset_construction}

\subsection{Data Collection and Preprocessing}
\label{subsec:data_collection}
Our framing of ARG requires raw data capturing the full review–revision–response (Re$^3$) process with authentic human texts. Only a few resources, such as the EMNLP24 subset of NLPEERv2 \cite{nlpeer} and MOPRD \cite{peerj}, provide peer reviews, author responses, original submissions, and revised papers. 
EMNLP24 provides peer reviews and rebuttal discussions from OpenReview,\footnote{\url{https://openreview.net/}}
 which we organize into reviewer–author discussion chains,  extracting and merging consecutive author replies into single responses. MOPRD offers data from PeerJ\footnote{\url{https://peerj.com/}}
 across multiple scientific domains including computer science, chemistry, physics, and materials science. We retain only papers with a complete Re$^3$ record. The final corpus includes 679 EMNLP24 papers and 2,715 PeerJ papers (Table \ref{tab:data_stat}), covering both conference and journal workflows.
We group each paper’s versions, reviews, and responses under a unified identifier and convert them into intertextual graphs (ITGs) \cite{f1000rd}, augmented with sentence-level nodes (details in §\ref{subsec:app_prepro}).

\begin{figure*}[ht]
  \centering
  \includegraphics[width=0.91\textwidth]{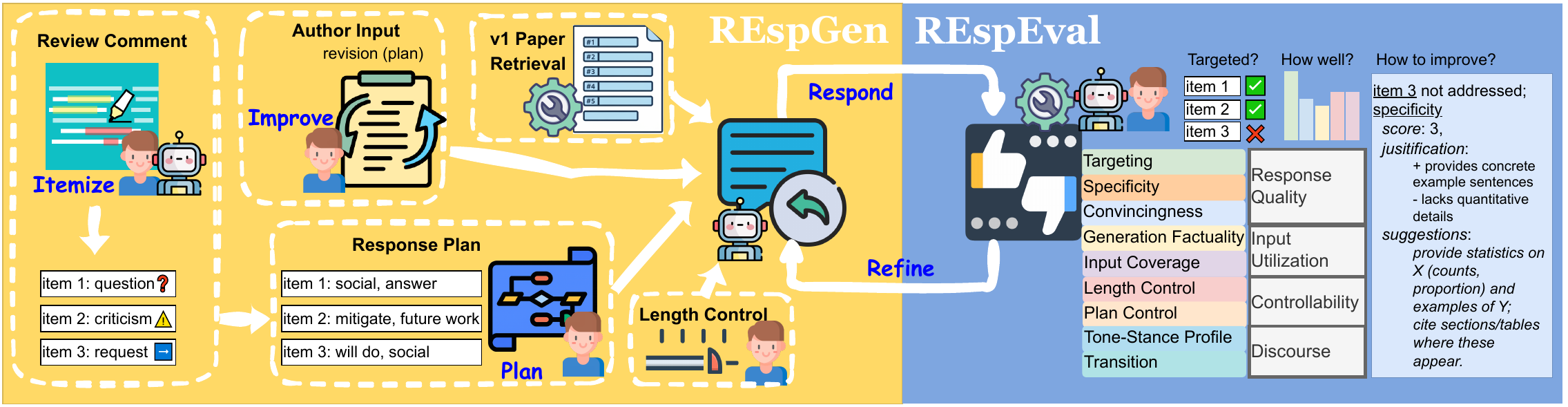}
  \caption{Frameworks: \textit{REspGen} \& \textit{REspEval}.}
  \label{fig:framework}
\end{figure*}

\subsection{Review-Response Pair Alignment and Revision Annotation}
\label{subsec:pair_extraction} 
Authors often quote review sentences to structure their replies. To extract review–response pairs, we match every review sentence to every response sentence using an assembled matching algorithm (§\ref{subsec:app_match_algorithm}) and merge the longest contiguous matches to identify quoted review spans. These spans are then used to segment the response, with each segment defined as the text following a quoted span and preceding the next one.
An illustrative example is shown in Figure~\ref{fig:app_pair_example} in §\ref{subsec:app_match_algorithm}.
After applying quality filtering strategies (§\ref{subsec:app_match_algorithm}), we obtain 2,108 and 13,963 review–response segment pairs from EMNLP24 and PeerJ, respectively. Human verification of 100 pairs confirms a 98\% alignment accuracy.
We further apply SOTA revision analysis models \cite{llm-classifier,re3} to  align sentence-level edits across paper versions and label each with edit action and intent. These models achieve over 90 F1 for alignment and action labeling, and 84.3 F1/85.6\% accuracy for edit intent classification. In total, this produces 439,798 edits.

\subsection{Re$^3$ Triplet Alignment}
\label{subsec:triplet_alignment} 
For each submission, we have the original paper $D^{t}$, the revised paper $D^{t+1}$, and reviewer–author exchanges $(C_k, A_k)$ for reviewer $k$. Sentences in $D^{t}$ and $D^{t+1}$ are denoted  $x^{t}_j$ and $x^{t+1}_i$. From earlier steps, we extract sentence-level edits $e_{ij} = e(x^{t+1}_i, x^{t}_j)$, with the full edit set denoted as $E$.
We also obtain aligned review–response segments $p^{k}_{mn} = p(c^{k}_m, a^{k}_n)$ with $c^{k}_m \in C_k$ and $a^{k}_n \in A_k$, denoted collectively as $P$. Figures \ref{fig:app_pair_example} and \ref{fig:app_edit_example} (§\ref{sec:app_data_construction}) provide illustrative examples of the notations.

The task of triplet alignment is to determine whether an edit $e_{ij} \in E$ is relevant to a pair $p^{k}_{mn} \in P$. For each $p^{k}_{mn} = p(c^{k}_m, a^{k}_n)$, we use a two-way strategy: (i) align the review comment $c^{k}_m$ to each edit $e(x^{t+1}_i, x^{t}_j)$ using a function set $\text{CE}$, and (ii) align the response segment $a^{k}_n$ to the same edit using a function set $\text{AE}$. Each function set combines a fine-tuned SOTA LLM classifier (>90\% accuracy) with a lightweight similarity component (details in §\ref{subsec:app_re3_align}). All positive alignments are aggregated as the edits linked to $p^{k}_{mn}$, denoted $[e_{\text{align}}]$, yielding triplets
$
t^{k}_{mn} = \bigl(c^{k}_m,\, a^{k}_n,\, [e_{\text{align}}]\bigr).
$
We obtain 15,521 triplets with non-empty aligned edits. Human evaluation of 125 aligned edits yields precision of 0.86 for EMNLP 2024 and 0.71 for PeerJ, with perfect recall in both cases. Errors primarily arise from aggregation, which may align lexically or semantically similar texts whose edits are not directly relevant to reviewer concerns. We adopt aggregation to prioritize coverage and minimize missed alignments.

\section{Generation Framework: \textit{REspGen}}
\label{sec:respgen}
\textit{REspGen} is a modular framework (Figure~\ref{fig:framework}, left) with response plan and length control (§\ref{subsec:respgen_control}), configurable author input and paper context (§\ref{subsec:respgen_tune_input}), and evaluation-guided refinement (§\ref{subsec:respgen_refine}).

\subsection{Response Attribute Control}
\label{subsec:respgen_control}
\noindent\textbf{Item-Based Response Planning.}
We adopt the \textit{review action} taxonomy of \citet{disapere} and derive three review item types,\footnote{We retain types that \citet{disapere} found to be commonly addressed in author responses. Other review types (e.g., strengths, summaries) typically require no response.} including \textit{Criticism}, \textit{Question}, and \textit{Request}, to classify spans within each review segment (definitions in Table~\ref{tab:app_item_types}). To simulate realistic author planning, we prompt GPT-5 to jointly analyze each review–response pair, itemize the review, align spans from the human response, and assign \textit{response action} labels \cite{disapere}. Table~\ref{tab:app_response_actions} lists the 16 labels grouped into five stance classes: \textit{Cooperative}, \textit{Defensive}, \textit{Hedge}, \textit{Social}, and \textit{Other}. Illustrative examples appear in Figure~\ref{fig:fig1} and Figure~\ref{fig:app_review_response_labels_example}, with the prompt in Figure~\ref{fig:app_review_response_labels_prompt} (§\ref{subsec:app_resp_plan}).
In \textit{REspGen}, authors may specify a response plan for each review item by providing a sequence of \textit{response action} labels (see (b) of Figure~\ref{fig:fig1}), which guides the tone, stance, and discourse flow of the generated response. In experiments, we simulate author control using the annotations above.  Further details are provided in §\ref{subsec:app_resp_plan}.

\noindent\textbf{Length-Constrained Generation.}
Many peer-review venues impose strict length limits on author responses to encourage focused communication.\footnote{For example, %
\url{https://docs.openreview.net/reference/default-forms/default-rebuttal-form}}
 In \textit{REspGen}, authors may specify an upper-bound word limit for generation. Since appropriate length depends on the complexity of the review concern, in our experiments we simulate realistic author-provided limits by setting them to $n$+50 where $n$ is the human response length (Figure~\ref{fig:fig1}(c)).

\subsection{Input Component Configuration}
\label{subsec:respgen_tune_input}
\textit{REspGen} supports configurable input components, including varying author input specificity and optional paper context. We simulate author input through aligned sentence-level edits $[e_{\text{align}}]$, where each edit is supplied as either (i) an edited-sentence string (\textbf{\textit{S}}), simulating rough revision ideas not yet anchored to a specific location, or (ii) the string with its paragraph \textbf{\textit{context}} and section title, simulating polished in-context revisions and reflecting where the edit appears (Figure~\ref{fig:fig1}(d)). 
Beyond author input, \textit{REspGen} supports an optional \textbf{\textit{v1}} retrieval module that retrieves the top five relevant paragraphs from the original submission using a retrieval–reranking approach conditioned on the review segment (details in §\ref{subsec:app_input_cond}). This paper-level context provides additional topic grounding for response generation.

\subsection{Evaluation-guided Refinement}
\label{subsec:respgen_refine}
\textit{REspGen} includes an iterative refinement module that interfaces with the evaluation framework \textit{REspEval}. Given a review segment, optional author input and \textit{v1} retrieval, response plan and length control, the system first generates an initial draft. \textit{REspEval} then evaluates this draft and returns evaluation metrics, justifications and refinement suggestions (§\ref{sec:respeval}). These results, together with the original inputs, controls, and initial draft, are fed back into \textit{REspGen} to produce a refined response. 
This iterative process leverages \textit{REspEval} feedback to progressively improve responses, helping them better reflect author intent, satisfy controls, and address reviewer concerns.

\section{Evaluation Framework: \textit{REspEval}}
\label{sec:respeval}
\textit{REspEval} evaluates four dimensions: discourse (§\ref{subsec:respeval_discourse}), controllability (§\ref{subsec:respeval_control}), input utilization (§\ref{subsec:respeval_input_usage}), and response quality (§\ref{subsec:respeval_quality}), with \hl{subdimensions} and \hlg{metrics} color-coded for quick reference below. \textit{Quality} (§\ref{subsec:respeval_quality}) serves as the primary indicator of author response effectiveness, while the remaining dimensions capture complementary aspects: how well the model maintains coherent stance and structure, adheres to controls (length and planning), and incorporates author inputs (factual grounding and coverage). Together, these dimensions reflect the distinct challenges introduced by the author-in-the-loop ARG paradigm.

\subsection{Response Discourse Analysis}
\label{subsec:respeval_discourse}
Following §\ref{subsec:respgen_control}, we label response spans with actions, yielding two analyses: (i) \hl{tone–stance profile}, obtained by mapping actions to the five stance classes and computing their word-weighted proportions \hlg{\textit{\%Coop}}, \hlg{\textit{\%Defe}}, \hlg{\textit{\%Hed}}, \hlg{\textit{\%Soc}}, \hlg{\textit{\%Other}} and \hlg{\textit{ArgLoad}}$=$\textit{\%Coop}$+$\textit{\%Defe}$+$\textit{\%Hed}, which reflects overall argumentative load; and (ii) \hl{transition flow}, capturing stance distributions across response positions and shifts between adjacent spans. These analyses characterize communicative attitude and discourse dynamics, enabling comparison of human and LLM responses, as discussed in §\ref{subsubsec:discuss_5}.

\subsection{Controllability Evaluation}
\label{subsec:respeval_control}
For \hl{length control (lenC)}, we compute the difference between the upper bound limit and the generated length for each sample, where positive values indicate adherence. We report the percentage of generations that meet the constraint (\hlg{\textit{\%met}}) and the median length difference across all samples (\hlg{\textit{m.diff}}).
For \hl{response plan control (planC)}, we assess how generated response action labels and their ordering match the plan, reporting label precision (\hlg{\textit{P}}), recall (\hlg{\textit{R}}), and \hlg{\textit{F1}}. 
To evaluate ordering, we compute order fidelity (\hlg{\textit{OF}}), measuring how well correctly-produced actions preserve plan order.
Let $\mathbf{m}=(m_1,\ldots,m_T)$ denote the indices of plan actions matched to the generated response in generation order, with $m_i=-1$ indicating no match. 
We then define
$
\mathbf{s}=\{\,m_i \mid m_i \ge 0\,\}
$
as the matched plan indices in generated order, and let $\mathbf{s}^\ast$ be the same elements sorted in ascending order (i.e., the plan order).
OF is the longest common subsequence (LCS) of $\mathbf{s}$ and $\mathbf{s}^\ast$, normalized by $|\mathbf{s}|$:
\[
\mathrm{OF}(\mathbf{m})=
\begin{cases}
0, & |\mathbf{s}|=0,\\[4pt]
\dfrac{\mathrm{LCS}(\mathbf{s},\mathbf{s}^\ast)}{|\mathbf{s}|}, & \text{otherwise}.
\end{cases}
\]

\subsection{Input Utilization Measures}
\label{subsec:respeval_input_usage}
We assess how generated responses use given inputs through two fact-based measures inspired by atomic fact-checking \cite{factscore}, which decomposes text into atomic facts and verifies each against reference sources. We adapt their best GPT-based approach\footnote{\citet{factscore} use GPT-3.5, the most advanced available at the time; we use GPT-5 with redesigned prompts and examples for improved scientific fact extraction.} and introduce:
(i) \hl{Generation Factuality Precision (GFP)}: extract atomic facts from the generated response and verify each against all given inputs (edit strings, optional paragraph context, and v1 content). GFP is the proportion of generated facts supported by the inputs, indicating factual grounding.
(ii) \hl{Input Coverage Recall (ICR)}: decompose the core author input (edit strings) into atomic facts and check whether each is expressed in the generated response. ICR measures how well the model prioritizes and incorporates the author’s intended improvements.
For both measures, we report the proportions of supported (\hlg{\textit{\%sup}}), unsupported (\hlg{\textit{\%unsup}}) and contradicted (\hlg{\textit{\%con}}) facts.

\subsection{Response Quality Evaluation}
\label{subsec:respeval_quality}
We assess response quality using three core criteria grounded in venue guidelines,\footnote{\url{https://aclrollingreview.org/authors}; \url{https://peerj.com/benefits/academic-rebuttal-letters/}} expert advice,\footnote{\url{https://deviparikh.medium.com/how-we-write-rebuttals-dc84742fece1}} and empirical studies \cite{does-my-rebuttal,what-makes-success}: \hl{targeting} (directly addressing reviewer concerns), \hl{specificity} (providing concrete evidence and details), and \hl{convincingness} (presenting clear, persuasive justification), which emphasize substantive effectiveness beyond surface-level fluency.

We evaluate these three dimensions using GPT-5 as a judge with structured, rubric-grounded reasoning, an approach shown to improve interpretability, reliability, and alignment with human judgments \citep{llmsasjudges}. Given the review, response, and review item–response alignments, GPT-5 assigns 5-point scores for targeting (\hlg{\textit{Targ}}), specificity (\hlg{\textit{Spec}}), and convincingness (\hlg{\textit{Conv}}). For each dimension, it provides evidence-based justifications by listing strengths and weaknesses with concrete references to relevant items, as well as refinement suggestions. Prompts with scoring rubrics, example outputs, and additional details are provided in §\ref{subsec:app_respeval_qual_proce}.

We validate our approach through comprehensive studies (§\ref{subsec:app_respeval_qual_veri}), confirming assessments are (i) \textbf{consistent} across runs, (ii) \textbf{robust} to perturbations, distinguishing genuine from degraded responses, (iii) \textbf{interpretable}, and (iv) \textbf{reliable} via two human studies with 12 experienced researchers. In Study~1 (Figure~\ref{fig:human_study1}), annotators rate agreement (1–5) with GPT-5 scores, justifications, and suggestions per dimension. In Study~2 (Figure~\ref{fig:human_study2}), annotators judge which of two responses to the same review is superior (or tied) per dimension and overall. Results across 1,365 judgments show strong human-LLM alignment (agreement rating > 4.17/5, disagreement < 5\%) and substantial \cite{Landis1977TheMO} inter-annotator agreement on win/loss comparisons (Krippendorff's $\alpha$ = 0.81–0.89).

\begin{table*}[t]
\fontsize{8}{8}
\selectfont
\centering
\renewcommand{\arraystretch}{1.02} 
\tabcolsep=0.03cm
\begin{tabular}[]{lll|llll|llll|llll|lllll|llll}
\hline
Metric&&&\multicolumn{3}{c}{GFP}&&\multicolumn{3}{c}{ICR}&&\multicolumn{3}{c}{lenC}&&\multicolumn{4}{c}{PlanC}&&\multicolumn{3}{c}{Quality}\\ \hline

&&
& \%sup &\%unsup&\%con
&&\%sup &\%unsup&\%con
&& \#w & \%met &m.diff
&& P &R&F1&OF
&&Targ &Spec&Conv
\\ \hline
Human
&&&\begin{tabular}{l}.458\end{tabular}&\begin{tabular}{l}.453\end{tabular}&\begin{tabular}{l}.089\end{tabular}

&&\begin{tabular}{l}.200\end{tabular}&\begin{tabular}{l}.756\end{tabular}&\begin{tabular}{l}.044\end{tabular}

&&\begin{tabular}{l}115\end{tabular}&\begin{tabular}{l}/\end{tabular}&\begin{tabular}{l}/\end{tabular}

&&\begin{tabular}{l}/\end{tabular} &\begin{tabular}{l}/\end{tabular}&\begin{tabular}{l}/\end{tabular}&\begin{tabular}{l}/\end{tabular}

&&\begin{tabular}{l}.788\end{tabular}&\begin{tabular}{l}.575\end{tabular}&\begin{tabular}{l}.575\end{tabular}
\\\hline
\begin{tabular}{l}\includegraphics[height=1.4em]{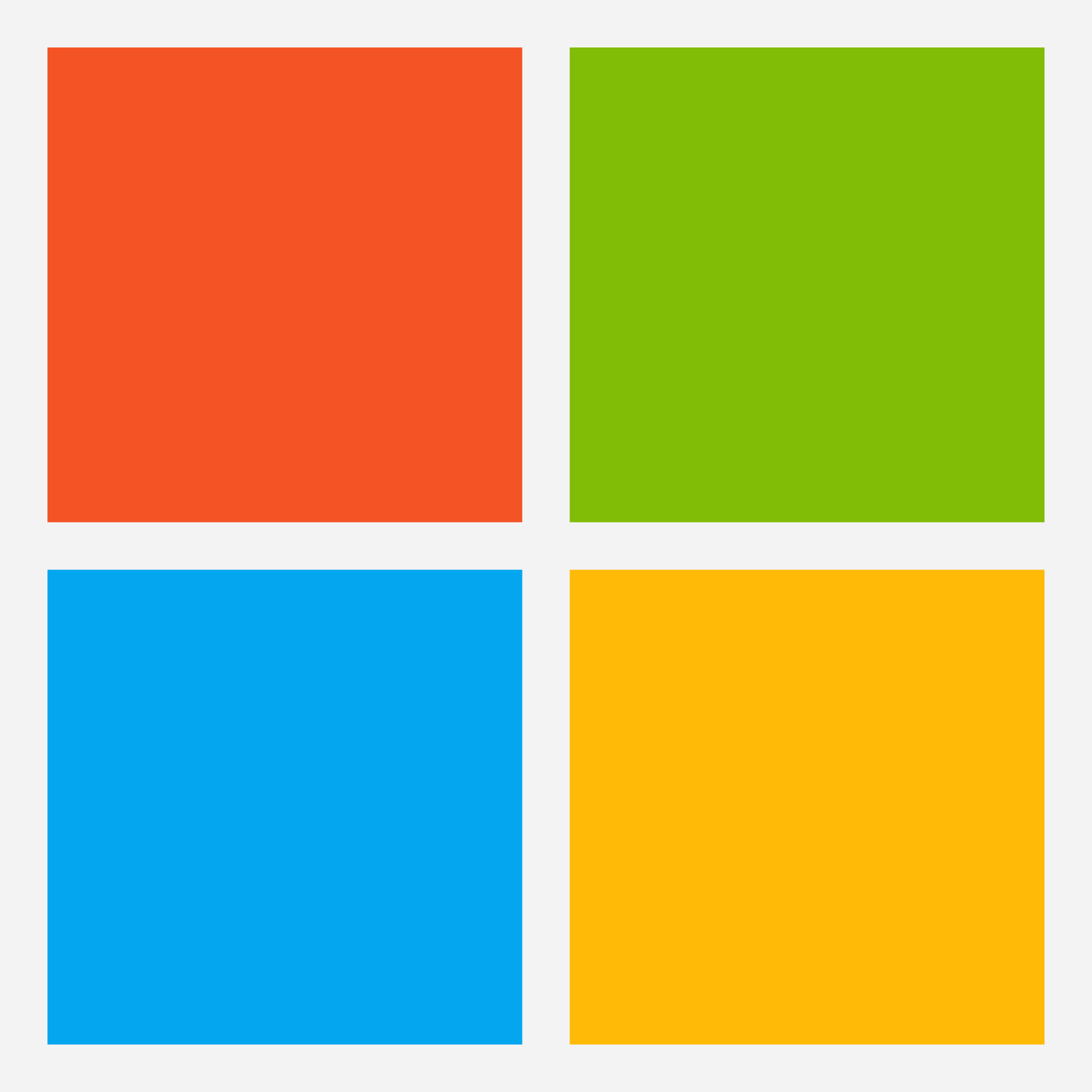}\hspace{0.5em}\\Phi-4\end{tabular}
&\begin{tabular}{l}\textit{1.noAIx}\\\textit{2.wAIx\_\textcircled{\tiny 1}\textit{S}}\\\textit{3.wAIx\_\textcircled{\tiny 2}\textit{+context}}\\\textit{4.wAIx\_\textcircled{\tiny 3}\textit{+v1}}\\\textit{5.+Cont.\_\textcircled{\tiny 1}\textit{lenC}}\\\textit{6.+Cont.\_\textcircled{\tiny 2}\textit{lenC\&planC}}\\\textit{7.+Cont.\_\textcircled{\tiny 3}\textit{planC}}\\\textit{8.+Refine\_Cont.\textcircled{\tiny 2}}\\\textit{9.+Refine\_Cont.\textcircled{\tiny 3}}\end{tabular}
&
&\begin{tabular}{l}.362\\.575\\.577\\.705\\\textbf{.748}\\.673\\.680\\.489\\.490 \end{tabular}
&\begin{tabular}{l}.542\\.374\\.364\\.236\\.200\\.263\\.253\\.442\\.443\end{tabular}
&\begin{tabular}{l}.096\\\textbf{.051}\\.059\\.059\\.052\\.064\\.067\\.069\\.068\end{tabular}&

&\begin{tabular}{l}.300\\\textbf{.509}\\.470\\.358\\/\\/\\/\\/\\/ \end{tabular}
&\begin{tabular}{l}.926\\.450\\.494\\.592\\/\\/\\/\\/\\/ \end{tabular}
&\begin{tabular}{l}.044\\.042\\\textbf{.036}\\.050\\/\\/\\/\\/\\/ \end{tabular}&

&\begin{tabular}{l}161\\127\\428\\343\\343\\284\\284\\312\\368 \end{tabular}
&\begin{tabular}{l}/\\/\\/\\/\\\textbf{.458}\\.250\\/\\.104\\/ \end{tabular}
&\begin{tabular}{l}/\\/\\/\\/\\-27\\-55\\/\\-128\\/ \end{tabular}&

&\begin{tabular}{l}/\\/\\/\\/\\/\\.471\\\textbf{.485}\\.387\\.368 \end{tabular}
&\begin{tabular}{l}/\\/\\/\\/\\/\\.642\\.644\\.671\\\textbf{.691} \end{tabular}
&\begin{tabular}{l}/\\/\\/\\/\\/\\.497\\\textbf{.504}\\.444\\.434 \end{tabular} 
&\begin{tabular}{l}/\\/\\/\\/\\/\\.755\\\textbf{.791}\\.790\\.729 \end{tabular} &

&\begin{tabular}{l}.775\\.821\\.783\\.771\\.779\\.829\\.821\\\cellcolor{lightgreen3}\textbf{.929}\\\cellcolor{lightgreen3}\textbf{.929}\end{tabular}
&\begin{tabular}{l}.446\\.563\\.583\\.579\\.583\\.600\\.583\\\textbf{.733}\\.713 \end{tabular}
&\begin{tabular}{l}.483\\.579\\.592\\.579\\.579\\.613\\.596\\\textbf{.725}\\\textbf{.725}\end{tabular}&

\\\hline
\begin{tabular}{l}\includegraphics[height=1.4em]{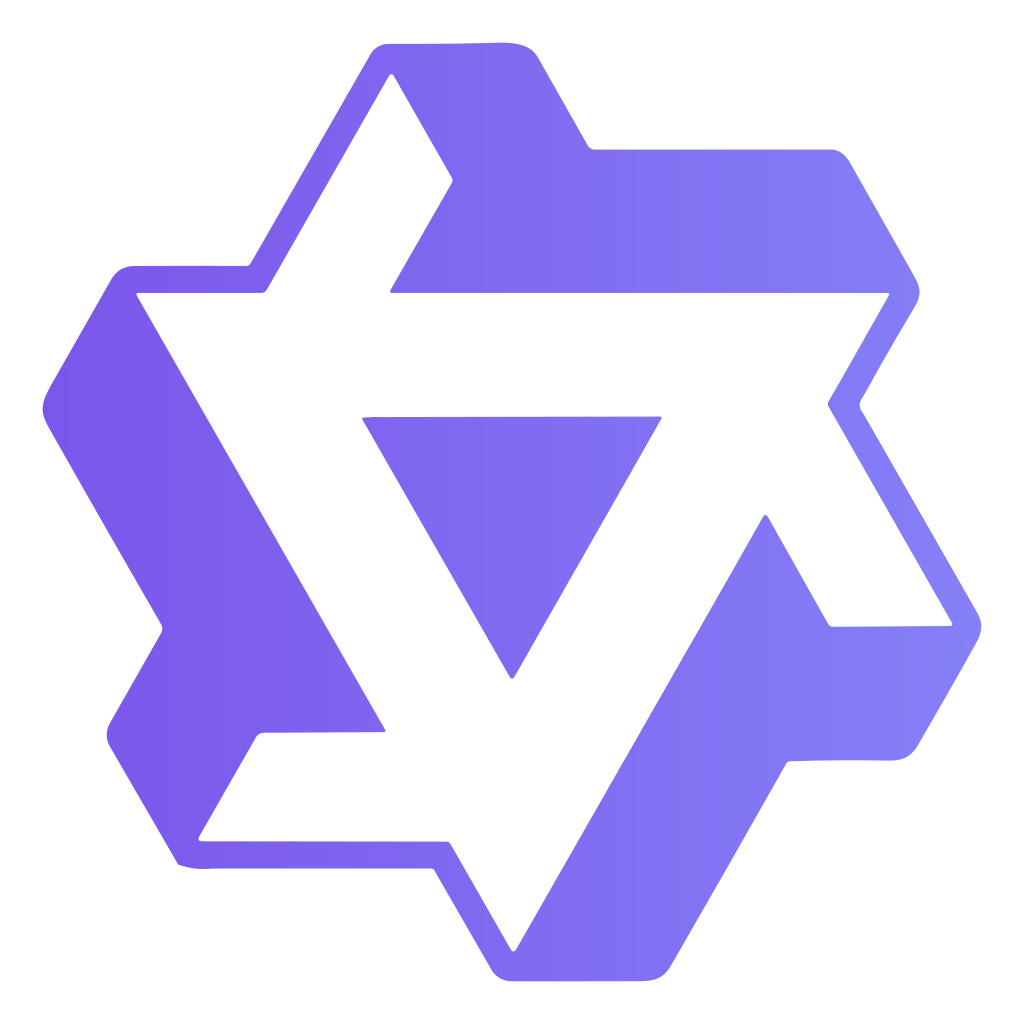}\hspace{0.5em}\\Qwen3\end{tabular}

&\begin{tabular}{l}\textit{1.noAIx}\\\textit{2.wAIx\_\textcircled{\tiny 1}\textit{S}}\\\textit{3.wAIx\_\textcircled{\tiny 2}\textit{+context}}\\\textit{4.wAIx\_\textcircled{\tiny 3}\textit{+v1}}\\\textit{5.+Cont.\_\textcircled{\tiny 1}\textit{lenC}}\\\textit{6.+Cont.\_\textcircled{\tiny 2}\textit{lenC\&planC}}\\\textit{7.+Cont.\_\textcircled{\tiny 3}\textit{planC}}\\\textit{8.+Refine\_Cont.\textcircled{\tiny 2}}\\\textit{9.+Refine\_Cont.\textcircled{\tiny 3}}\end{tabular}
&

&\begin{tabular}{l}.308\\.618\\.643\\\textbf{.744}\\.734\\.724\\.719\\.576\\.586 \end{tabular}
&\begin{tabular}{l}.566\\.342\\.317\\.214\\.223\\.224\\.252\\.373\\.380 \end{tabular}
&\begin{tabular}{l}.126\\.040\\.040\\.042\\.044\\.052\\\textbf{.028}\\.050\\.035 \end{tabular}&

&\begin{tabular}{l}.024\\\textbf{.628}\\.572\\.496\\/\\/\\/\\/\\/ \end{tabular}
&\begin{tabular}{l}.930\\.343\\.404\\.463\\/\\/\\/\\/\\/ \end{tabular}
&\begin{tabular}{l}.046\\.029\\\textbf{.025}\\.041\\/\\/\\/\\/\\/ \end{tabular}&

&\begin{tabular}{l}123\\127\\164\\205\\125\\130\\216\\142\\290 \end{tabular}
&\begin{tabular}{l}/\\/\\/\\/\\\cellcolor{lightgreen}\textbf{1.00}\\\cellcolor{lightgreen2}.958\\/\\.896\\/ \end{tabular}
&\begin{tabular}{l}/\\/\\/\\/\\38\\33\\/\\21\\/ \end{tabular}&

&\begin{tabular}{l}/\\/\\/\\/\\/\\.498\\.429\\\textbf{.506}\\.385 \end{tabular}
&\begin{tabular}{l}/\\/\\/\\/\\/\\.696\\\cellcolor{lightgreen}\textbf{.793}\\.678\\.752 \end{tabular}
&\begin{tabular}{l}/\\/\\/\\/\\/\\.534\\.522\\\textbf{.544}\\.459 \end{tabular} 
&\begin{tabular}{l}/\\/\\/\\/\\/\\.842\\.826\\.807\\\textbf{.847} \end{tabular} &

&\begin{tabular}{l}.808\\.892\\.875\\.913\\.904\\.913\\\cellcolor{lightgreen2}.938\\\cellcolor{lightgreen2}.938\\\cellcolor{lightgreen}\textbf{.983}\end{tabular}
&\begin{tabular}{l}.508\\.638\\.683\\.721\\.700\\.700\\.725\\\cellcolor{lightgreen2}.771\\\cellcolor{lightgreen}\textbf{.842} \end{tabular}
&\begin{tabular}{l}.554\\.654\\.683\\.717\\.700\\.700\\.725\\\cellcolor{lightgreen2}.758\\\cellcolor{lightgreen}\textbf{.800}\end{tabular}&

\\\hline
\begin{tabular}{l}\includegraphics[height=1.4em]{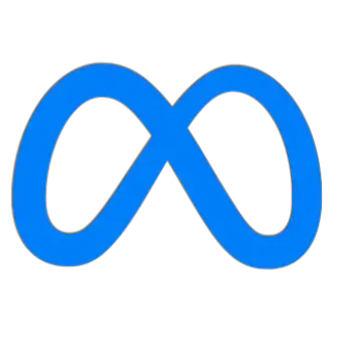}\hspace{0.5em}\\Llama-3.3\end{tabular}
&\begin{tabular}{l}\textit{1.noAIx}\\\textit{2.wAIx\_\textcircled{\tiny 1}\textit{S}}\\\textit{3.wAIx\_\textcircled{\tiny 2}\textit{+context}}\\\textit{4.wAIx\_\textcircled{\tiny 3}\textit{+v1}}\\\textit{5.+Cont.\_\textcircled{\tiny 1}\textit{lenC}}\\\textit{6.+Cont.\_\textcircled{\tiny 2}\textit{lenC\&planC}}\\\textit{7.+Cont.\_\textcircled{\tiny 3}\textit{planC}}\\\textit{8.+Refine\_Cont.\textcircled{\tiny 2}}\\\textit{9.+Refine\_Cont.\textcircled{\tiny 3}}\end{tabular}
&

&\begin{tabular}{l}.483\\.766\\.760\\.771\\\cellcolor{lightgreen}\textbf{.820}\\\cellcolor{lightgreen3}.788\\.770\\.657\\.647 \end{tabular}
&\begin{tabular}{l}.414\\.215\\.217\\.173\\.142\\.157\\.183\\.261\\.319 \end{tabular}
&\begin{tabular}{l}.103\\\textbf{.019}\\.023\\.056\\.038\\.055\\.047\\.082\\.034 \end{tabular}&

&\begin{tabular}{l}.054\\\cellcolor{lightgreen3}\textbf{.664}\\.534\\.420\\/\\/\\/\\/\\/ \end{tabular}
&\begin{tabular}{l}.912\\.319\\.426\\.542\\/\\/\\/\\/\\/ \end{tabular}
&\begin{tabular}{l}.034\\\textbf{.017}\\.040\\.039\\/\\/\\/\\/\\/ \end{tabular}&

&\begin{tabular}{l}126\\169\\183\\198\\82\\82\\214\\125\\304 \end{tabular}
&\begin{tabular}{l}/\\/\\/\\/\\\cellcolor{lightgreen}\textbf{1.00}\\\cellcolor{lightgreen}\textbf{1.00}\\/\\.875\\/ \end{tabular}
&\begin{tabular}{l}/\\/\\/\\/\\82\\84\\/\\52\\/ \end{tabular}&

&\begin{tabular}{l}/\\/\\/\\/\\/\\\cellcolor{lightgreen3}\textbf{.619}\\.486\\.589\\.372 \end{tabular}
&\begin{tabular}{l}/\\/\\/\\/\\/\\.470\\.705\\.545\\\textbf{.707} \end{tabular}
&\begin{tabular}{l}/\\/\\/\\/\\/\\.490\\\textbf{.533}\\.508\\.444 \end{tabular} 
&\begin{tabular}{l}/\\/\\/\\/\\/\\.728\\\textbf{.825}\\.718\\.806 \end{tabular} &

&\begin{tabular}{l}.763\\.800\\.850\\.829\\.829\\.804\\.850\\\textbf{.892}\\.888\end{tabular}
&\begin{tabular}{l}.396\\.550\\.608\\.588\\.513\\.467\\.575\\.667\\\cellcolor{lightgreen3}\textbf{.750} \end{tabular}
&\begin{tabular}{l}.438\\.567\\.608\\.575\\.517\\.504\\.592\\.638\\\textbf{.700}\end{tabular}&

\\\hline
\begin{tabular}{l}\includegraphics[height=1.4em]{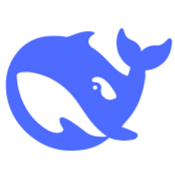}\hspace{0.5em}\\DeepSeek\end{tabular}

&\begin{tabular}{l}\textit{1.noAIx}\\\textit{2.wAIx\_\textcircled{\tiny 1}\textit{S}}\\\textit{3.wAIx\_\textcircled{\tiny 2}\textit{+context}}\\\textit{4.wAIx\_\textcircled{\tiny 3}\textit{+v1}}\\\textit{5.+Cont.\_\textcircled{\tiny 1}\textit{lenC}}\\\textit{6.+Cont.\_\textcircled{\tiny 2}\textit{lenC\&planC}}\\\textit{7.+Cont.\_\textcircled{\tiny 3}\textit{planC}}\\\textit{8.+Refine\_Cont.\textcircled{\tiny 2}}\\\textit{9.+Refine\_Cont.\textcircled{\tiny 3}}\end{tabular}
&

&\begin{tabular}{l}.412\\.720\\.702\\.738\\\cellcolor{lightgreen2}\textbf{.815}\\.762\\.754\\.728\\.734 \end{tabular}
&\begin{tabular}{l}.491\\.273\\.279\\.232\\.144\\.203\\.218\\.231\\.238 \end{tabular}
&\begin{tabular}{l}.097\\\textbf{.007}\\.019\\.031\\.042\\.035\\.028\\.041\\.028\end{tabular}&

&\begin{tabular}{l}.046\\\cellcolor{lightgreen}\textbf{.695}\\.584\\.452\\/\\/\\/\\/\\/ \end{tabular}
&\begin{tabular}{l}.913\\.272\\.374\\.514\\/\\/\\/\\/\\/ \end{tabular}
&\begin{tabular}{l}.041\\\textbf{.033}\\.041\\.035\\/\\/\\/\\/\\/ \end{tabular}&

&\begin{tabular}{l}113\\154\\172\\194\\96\\93\\179\\97\\194 \end{tabular}
&\begin{tabular}{l}/\\/\\/\\/\\\cellcolor{lightgreen}\textbf{1.00}\\\cellcolor{lightgreen}\textbf{1.00}\\/\\\cellcolor{lightgreen}\textbf{1.00}\\/ \end{tabular}
&\begin{tabular}{l}/\\/\\/\\/\\64\\63\\/\\63\\/ \end{tabular}&

&\begin{tabular}{l}/\\/\\/\\/\\/\\\cellcolor{lightgreen2}.626\\.554\\\cellcolor{lightgreen}\textbf{.661}\\.585 \end{tabular}
&\begin{tabular}{l}/\\/\\/\\/\\/\\.582\\\textbf{.710}\\.585\\.709 \end{tabular}
&\begin{tabular}{l}/\\/\\/\\/\\/\\.563\\\cellcolor{lightgreen3}.577\\\cellcolor{lightgreen}\textbf{.587}\\\cellcolor{lightgreen2}.585 \end{tabular} 
&\begin{tabular}{l}/\\/\\/\\/\\/\\.779\\.823\\\cellcolor{lightgreen3}.852\\\cellcolor{lightgreen2}\textbf{.861} \end{tabular} &

&\begin{tabular}{l}.771\\.850\\.817\\.904\\.879\\.867\\.888\\.913\\\textbf{.925}\end{tabular}
&\begin{tabular}{l}.433\\.600\\.608\\.692\\.642\\.588\\.663\\.704\\\textbf{.746} \end{tabular}
&\begin{tabular}{l}.496\\.621\\.617\\.700\\.638\\.625\\.671\\.704\\\cellcolor{lightgreen3}\textbf{.742}\end{tabular}&

\\\hline
\begin{tabular}{l}\includegraphics[height=1.4em]{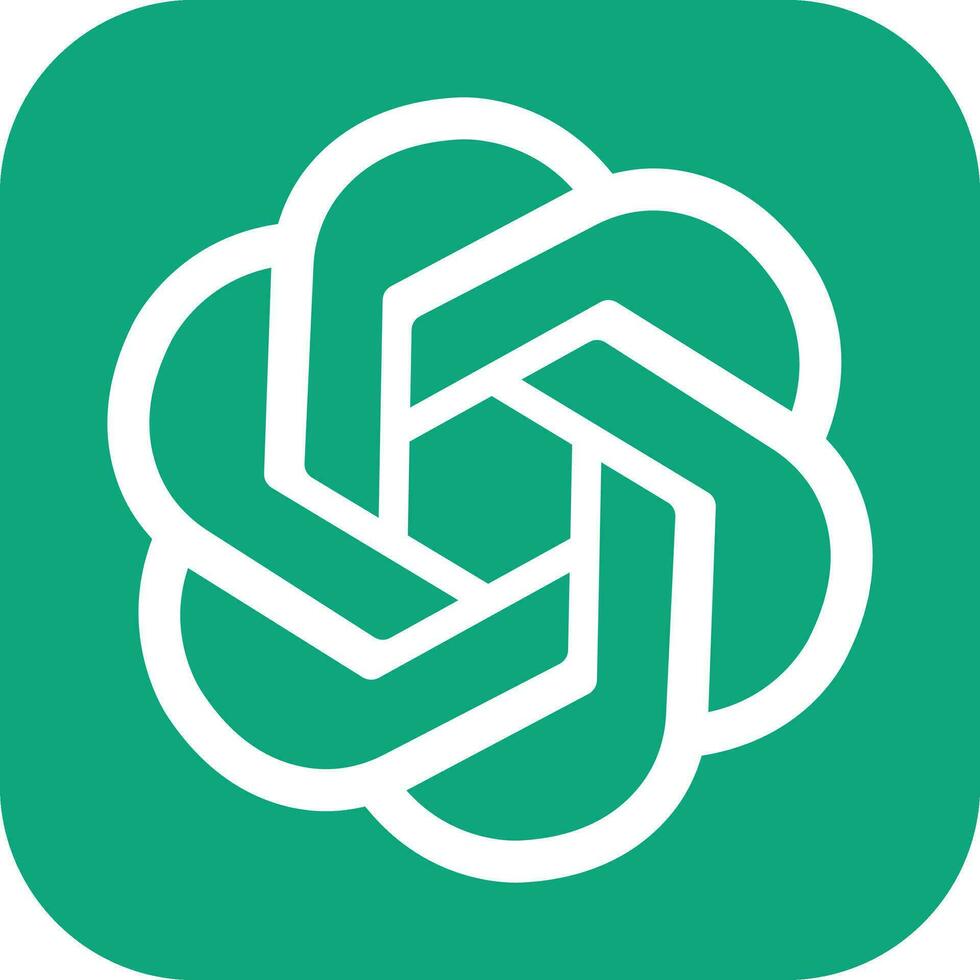}\hspace{0.5em}\\GPT-4o\end{tabular}
&\begin{tabular}{l}\textit{1.noAIx}\\\textit{2.wAIx\_\textcircled{\tiny 1}\textit{S}}\\\textit{3.wAIx\_\textcircled{\tiny 2}\textit{+context}}\\\textit{4.wAIx\_\textcircled{\tiny 3}\textit{+v1}}\\\textit{5.+Cont.\_\textcircled{\tiny 1}\textit{lenC}}\\\textit{6.+Cont.\_\textcircled{\tiny 2}\textit{lenC\&planC}}\\\textit{7.+Cont.\_\textcircled{\tiny 3}\textit{planC}}\\\textit{8.+Refine\_Cont.\textcircled{\tiny 2}}\\\textit{9.+Refine\_Cont.\textcircled{\tiny 3}}\end{tabular}
&

&\begin{tabular}{l}.443\\.689\\.708\\\textbf{.781}\\.774\\.744\\.762\\.715\\.695 \end{tabular}
&\begin{tabular}{l}.467\\.284\\.282\\.192\\.198\\.238\\.215\\.268\\.277 \end{tabular}
&\begin{tabular}{l}.090\\.027\\\textbf{.010}\\.028\\.028\\.019\\.023\\.017\\.029\end{tabular}&

&\begin{tabular}{l}.033\\\cellcolor{lightgreen2}\textbf{.668}\\.571\\.432\\/\\/\\/\\/\\/ \end{tabular}
&\begin{tabular}{l}.928\\.301\\.400\\.532\\/\\/\\/\\/\\/ \end{tabular}
&\begin{tabular}{l}.039\\.031\\\textbf{.029}\\.036\\/\\/\\/\\/\\/ \end{tabular}&

&\begin{tabular}{l}247\\265\\311\\339\\158\\156\\336\\163\\367 \end{tabular}
&\begin{tabular}{l}/\\/\\/\\/\\.854\\\cellcolor{lightgreen3}\textbf{.917}\\/\\.792\\/ \end{tabular}
&\begin{tabular}{l}/\\/\\/\\/\\11\\10\\/\\5\\/ \end{tabular}&

&\begin{tabular}{l}/\\/\\/\\/\\/\\.506\\.386\\\textbf{.507}\\.373 \end{tabular}
&\begin{tabular}{l}/\\/\\/\\/\\/\\.744\\\cellcolor{lightgreen3}.784\\.688\\\cellcolor{lightgreen2}\textbf{.790} \end{tabular}
&\begin{tabular}{l}/\\/\\/\\/\\/\\\textbf{.567}\\.477\\.554\\.470\end{tabular} 
&\begin{tabular}{l}/\\/\\/\\/\\/\\.834\\.794\\.772\\\cellcolor{lightgreen}\textbf{.880} \end{tabular} &

&\begin{tabular}{l}.825\\.821\\.817\\\cellcolor{lightgreen3}\textbf{.929}\\.867\\.879\\.913\\.900\\.925\end{tabular}
&\begin{tabular}{l}.479\\.600\\.629\\.688\\.633\\.596\\.692\\.675\\\textbf{.721} \end{tabular}
&\begin{tabular}{l}.547\\.625\\.629\\.708\\.633\\.621\\.700\\.675\\\textbf{.721}\end{tabular}&
\\\hline
\end{tabular}
\caption[LLMs]{Evaluation results across five LLMs and nine settings. Metrics cover input utilization (GFP, ICR), controllability (lenC, planC), and response quality (\textit{Targ}, \textit{Spec}, \textit{Conv}) (§\ref{sec:respeval}).  Scores are normalized to [0,1]; best per LLM per metric is bolded, and top-three across LLMs are marked in green. Settings 1-9: no author input (1), rough edit string as author input (2), add paragraph context (3), with paper retrieval (4), plus length control (5), plan control (7), or combined controls (6), refinement on 6 and 7 (8–9). Full descriptions in Table~\ref{tab:app_exp_settings} (§\ref{subsec:app_exp_details}).
}
\label{tab:ARG}
\end{table*}

\section{Experiments}
\label{sec:results_discussion}

\subsection{Experimental Setup}
\label{subsec:exp_details}
We evaluate ARG with five SOTA LLMs across nine \textit{REspGen} settings using \textit{REspEval}. Tables \ref{tab:ARG}, \ref{tab:ARG_TSP}-\ref{tab:ARG_refine} present results addressing eight research questions detailed in §\ref{subsec:discuss}. 
We evaluate five leading LLMs, both open-source and proprietary:
\includegraphics[height=0.8em]{fig_model_logo/microsoft_logo.png}Phi-4-Reasoning~\cite{phi4reasoning} (Phi-4); \includegraphics[height=0.8em]{fig_model_logo/qwen_logo.png}Qwen3-32B~\cite{qwen3} (Qwen3); \includegraphics[height=0.8em]{fig_model_logo/llama_logo.png}Llama-3.3-70B-Instruct\footnote{\url{https://huggingface.co/meta-llama/Llama-3.3-70B-Instruct}}
 (Llama-3.3); \includegraphics[height=0.8em]{fig_model_logo/deepseek_logo.png}DeepSeek-R1~\cite{deepseekr1} (DeepSeek); and \includegraphics[height=0.8em]{fig_model_logo/gpt_logo.jpg}GPT-4o.\footnote{\url{https://openai.com/index/gpt-4o-system-card/}}
 We select EMNLP24 cases where reviewers explicitly note a score increase (e.g., “I have improved the score”), yielding 48 instances. This ensures that the human responses used as baselines are verifiably strong and effective.\footnote{All responses are effective in a broad sense (papers accepted), we focus on cases where effectiveness is explicitly confirmed. PeerJ lacks score changes, and EMNLP24 releases only final scores; thus, we identify clear effectiveness through explicit reviewer statements.}
 We evaluate nine settings that systematically ablate and vary components of \textit{REspGen} to reveal interactions and trade-offs across evaluation dimensions. Starting from review-only generation (Setting~1), we add author input as edit strings (Setting~2), with additional paragraph context (Setting~3), and further augment with v1 retrieval (Setting~4). We then examine impacts of length control (Setting~5), response plan control (Setting~7), and both controls combined (Setting~6). Finally, we evaluate evaluation-guided refinement applied to outputs from Settings~6 and~7, yielding Settings~8 and 9. Detailed descriptions and prompts are provided in Table~\ref{tab:app_exp_settings} (§\ref{subsec:app_exp_details}).

These settings are designed to systematically examine eight research questions in §\ref{subsec:discuss}: the role of author input (RQ1–2); how input specificity affects input utilization (RQ3) and response quality (RQ4); how single- and multi-attribute control mechanisms operate (RQ5) and the resulting trade-offs between controllability and response quality (RQ6); the effectiveness of evaluation-guided refinement (RQ7); and the tonal and discourse-level differences between human and LLM-generated responses (RQ8). Taken together, these analyses provide a detailed characterization of LLM behavior under varied input and control conditions and yield insights for designing author-in-the-loop ARG systems.

\subsection{Results and Discussion}
\label{subsec:discuss}

\subsubsection{Why Author Input Matters}
\label{subsubsec:discuss_1}
To address \textbf{\textit{RQ1: Are LLMs aware of missing information?}}, we instruct models to insert placeholders (e.g., \textit{[author info: <description>]}) when author-only information is needed.  Table~\ref{tab:ARG_TSP} (§\ref{subsec:app_discuss}) shows all models except Phi-4 frequently use placeholders under review-only generation (54.2–95.8\%), confirming awareness of missing information. Once any author input is provided (Settings~2–4), placeholder usage drops sharply (0–25\%), demonstrating models recognize and leverage supplied information.
For \textbf{\textit{RQ2: Does author input improve response quality?}}, Table~\ref{tab:ARG} shows consistent improvements across all models and quality metrics (\textit{Targ}, \textit{Spec}, \textit{Conv}) when author input is added (Settings 2–4 vs. 1), with most gains statistically significant (Table~\ref{tab:app_aix_sig}, §\ref{subsec:app_discuss}). While review-only responses (Setting 1) underperform human baselines, all models surpass them in most author-input settings.
Together, these results demonstrate  that author input is both necessary and effective for ARG,  motivating our author-in-the-loop framework.

\subsubsection{Input Specificity and Its Impact}
\label{subsubsec:discuss_2}
To address \textbf{\textit{RQ3: How do input specificity and detail affect input utilization?}}, we analyze Settings~2 (edit string), 3 (plus paragraph context), and 4 (add v1 context) using \textit{GFP} and \textit{ICR} (Table~\ref{tab:ARG}). As input context increases, all LLMs incorporate more input-supported facts, with GFP support rising across Settings~2–4. In Setting~4, all LLMs achieve high GFP support (70.5–78.1\%) and low contradiction rates (2.8–5.9\%), indicating strong factual grounding with limited hallucination. In contrast, ICR support decreases with richer inputs across Settings~2–4, suggesting additional context dilutes focus on core author improvements. ICR contradiction rates remain low across all settings, confirming no increased hallucination. Together, these results reveal a trade-off: richer input improves factual grounding but may reduce emphasis on core information.
For \textbf{\textit{RQ4: How does input specificity affect response quality?}}, among Settings~2–4, Qwen3, DeepSeek, and GPT-4o achieve best quality metrics under Setting~4, indicating richer input generally improves response quality, though this is model-dependent: Llama-3.3 peaks at Setting~3 and Phi-4 shows only marginal differences among Settings~2–4. 
We further investigate edge cases where richer inputs degrade response quality and ICR, discussed as case studies in §\ref{subsec:case_study}.

\subsubsection{Controllability and Its Impact}
\label{subsubsec:discuss_3}
To answer \textbf{\textit{RQ5: How well can models satisfy single- and multi-attribute controls in ARG?}}, we evaluate length control (Setting~5), response plan control (Setting~7), and both controls combined (Setting~6) using \textit{lenC} and \textit{planC} metrics (Table~\ref{tab:ARG}). Length control (Setting 5) is effective for Qwen3, Llama-3.3, and DeepSeek (100\% met), moderately so for GPT-4o (85.4\%), but weak for Phi-4 (45.8\%). 
Under plan control (Setting~7), all models achieve high label recall and order fidelity, indicating general adherence to the prescribed structure, though occasional extra actions reduce label precision.
With  multi-attribute control (Setting~6), Qwen3 and Phi-4 degrade notably in length control, while Llama-3.3 and DeepSeek maintain length adherence but exhibit reduced plan recall, F1, and order fidelity. GPT-4o is the only model that improves under joint control.
Qwen3, Llama-3.3, and DeepSeek handle single-attribute control well, whereas Phi-4 consistently struggles with length control.
For \textbf{\textit{RQ6: How do different controls affect response quality?}}, we compare Settings~5–7 against the unconstrained Setting~4. Length control alone (Setting 5) substantially degrades response quality for all models except Phi-4, likely by limiting space for detailed argumentation. Adding plan control (Setting 6) improves Phi-4, minimally affects Qwen3, and slightly degrades Llama-3.3, DeepSeek, and GPT-4o. Among controlled settings, plan-only control (Setting~7) yields the best quality for all models except Phi-4 and matches Setting~4 performance.
Overall, LLMs handle single-attribute control well, but joint multi-attribute control remains challenging and often degrades response quality, with length constraints as the primary bottleneck.

\begin{figure}[t]
\centering
\begin{tabular}{lcc}
\hline
 & \fontsize{8}{8}\selectfont \textit{8.+Refine\_Cont.\textcircled{\tiny 2}}
 & \fontsize{8}{8}\selectfont\textit{9.+Refine\_Cont.\textcircled{\tiny 3}} \\ \hline
\fontsize{8}{8}\selectfont
  \begin{minipage}[l]{0.12\linewidth}  %
    \raggedright\fontsize{8}{9}\selectfont
    \includegraphics[height=1.4em]{fig_model_logo/microsoft_logo.png}\\
    Phi-4
  \end{minipage}
  &
  \begin{minipage}[c]{0.31\linewidth}  %
    \centering
    \includegraphics[width=\linewidth]{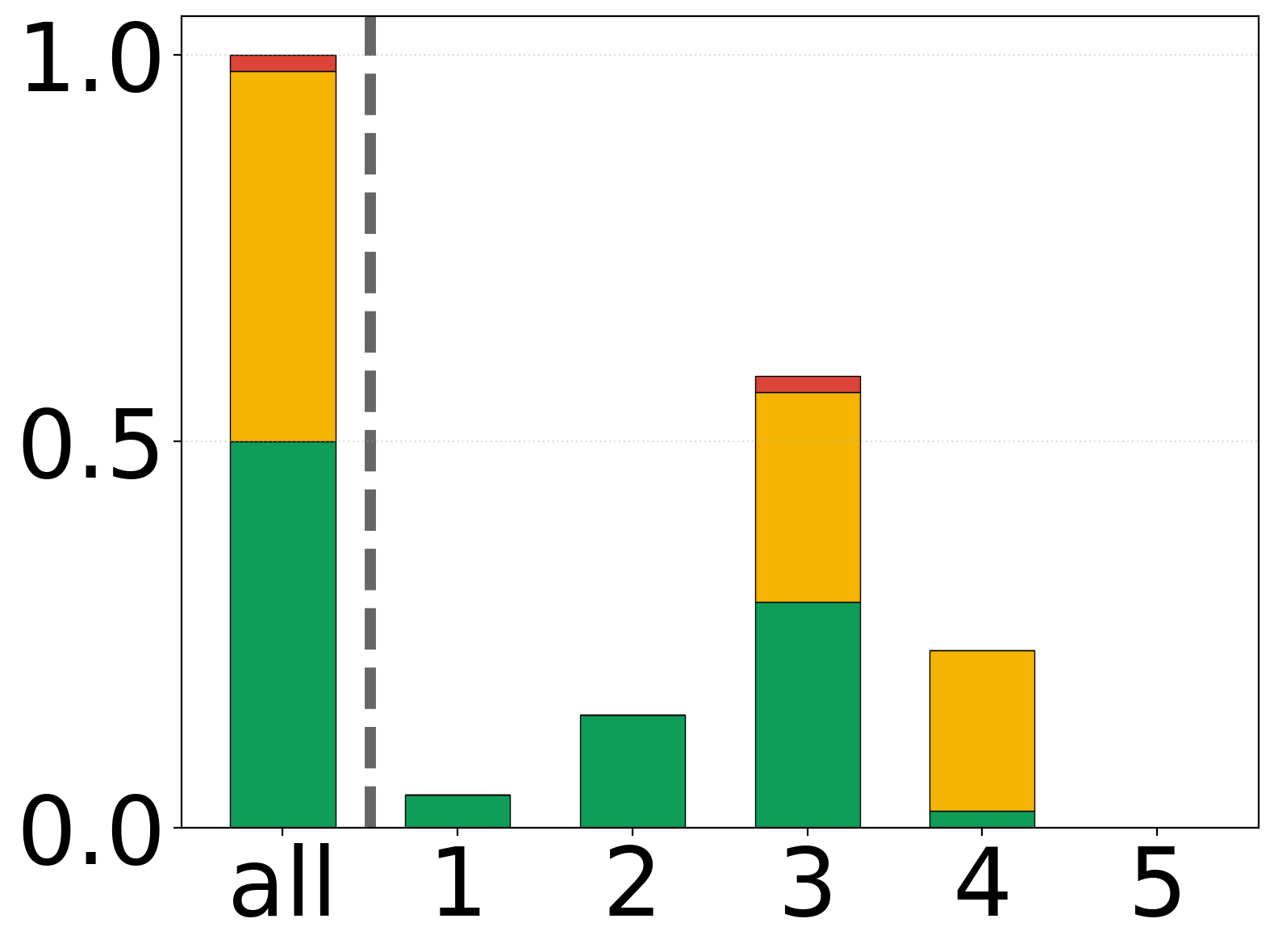}
  \end{minipage}
  &
  \begin{minipage}[c]{0.31\linewidth}  %
    \centering
    \includegraphics[width=\linewidth]{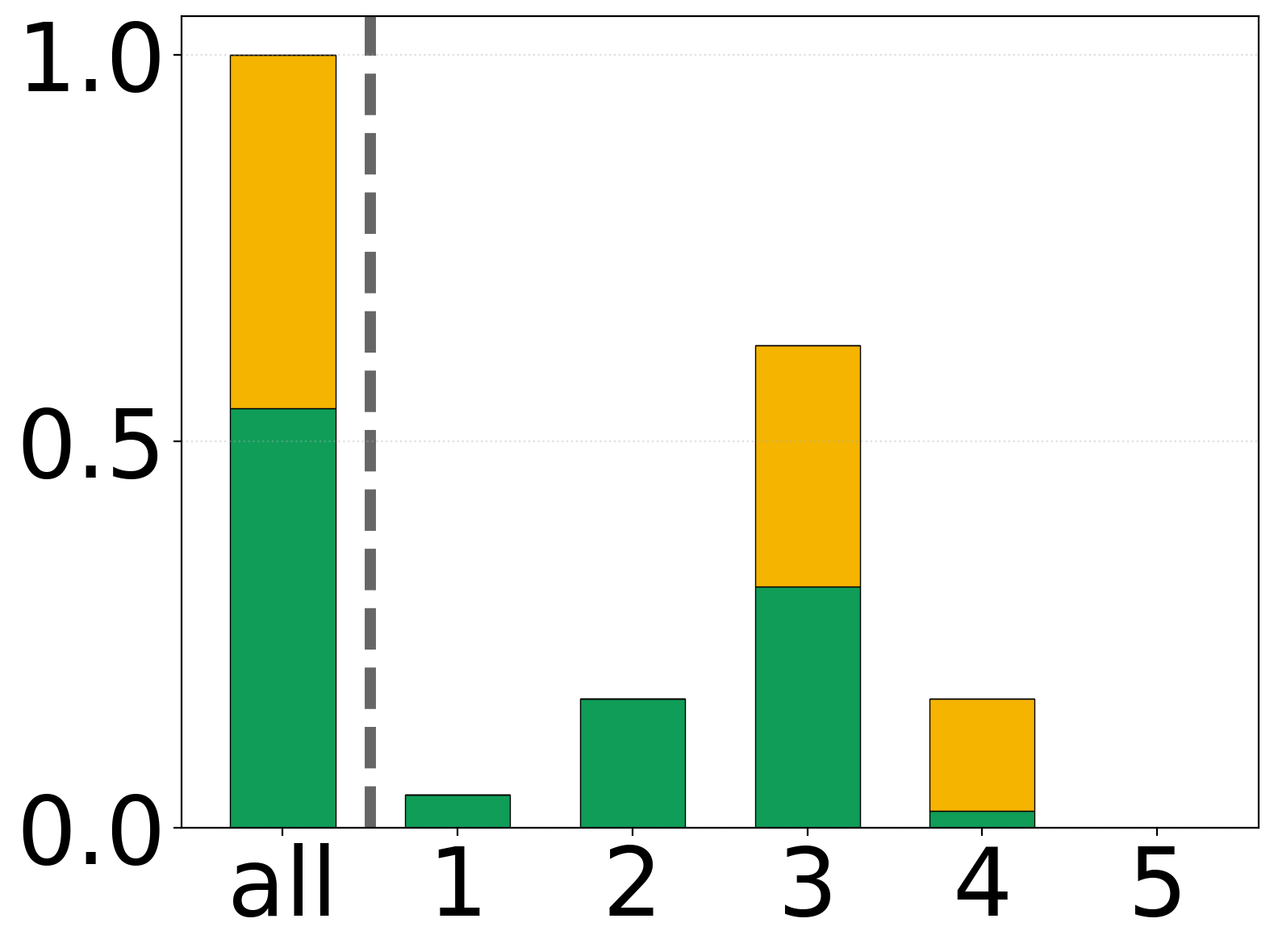}
  \end{minipage}
  \\[0.5em]
\fontsize{8}{8}\selectfont
  \begin{minipage}[l]{0.12\linewidth}  
    \raggedright\fontsize{8}{9}\selectfont
    \includegraphics[height=1.4em]{fig_model_logo/qwen_logo.png}\\
    Qwen3
  \end{minipage}
  &
  \begin{minipage}[c]{0.31\linewidth}  %
    \centering
    \includegraphics[width=\linewidth]{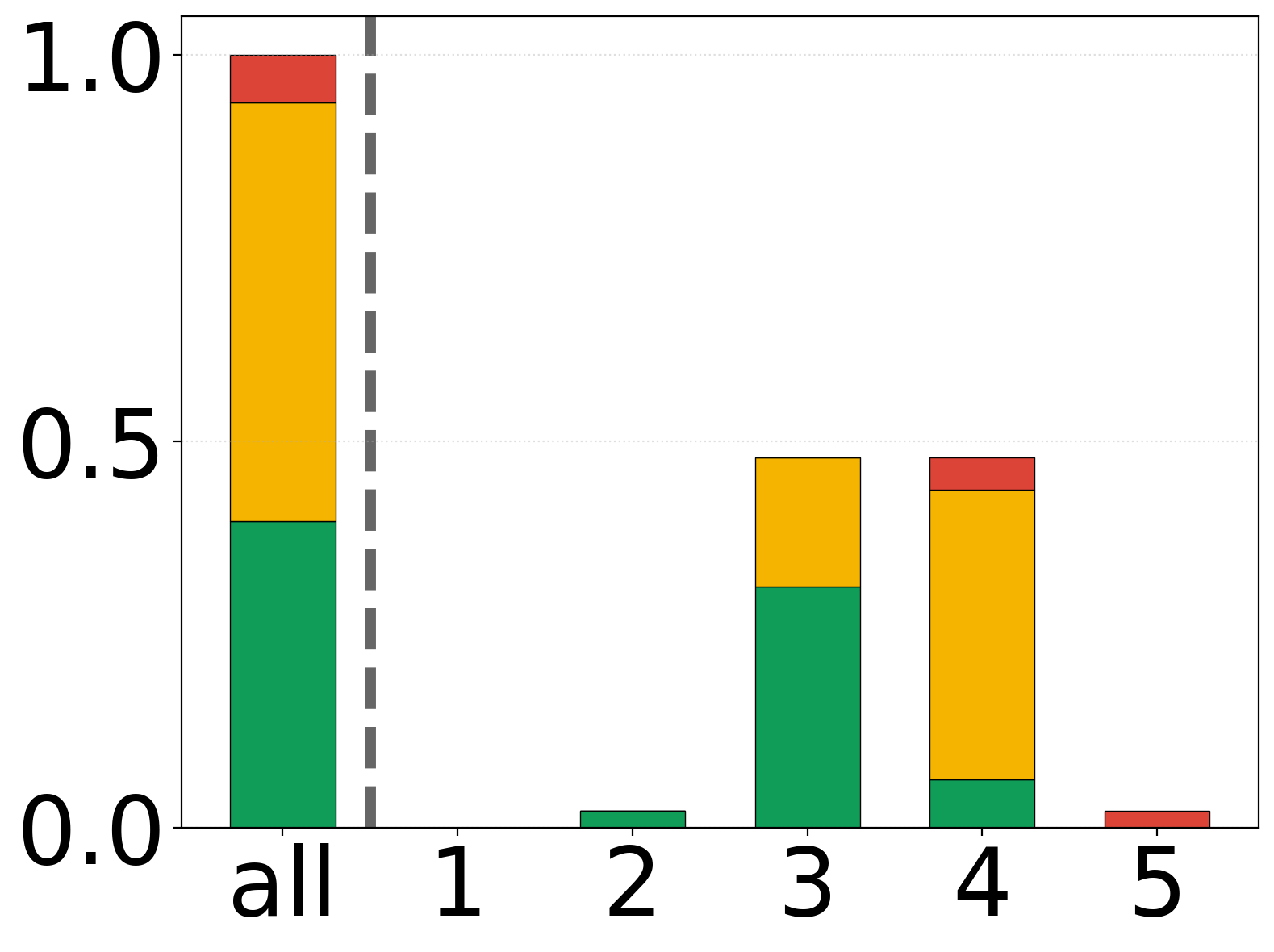}
  \end{minipage}
  &
  \begin{minipage}[c]{0.31\linewidth}  %
    \centering
    \includegraphics[width=\linewidth]{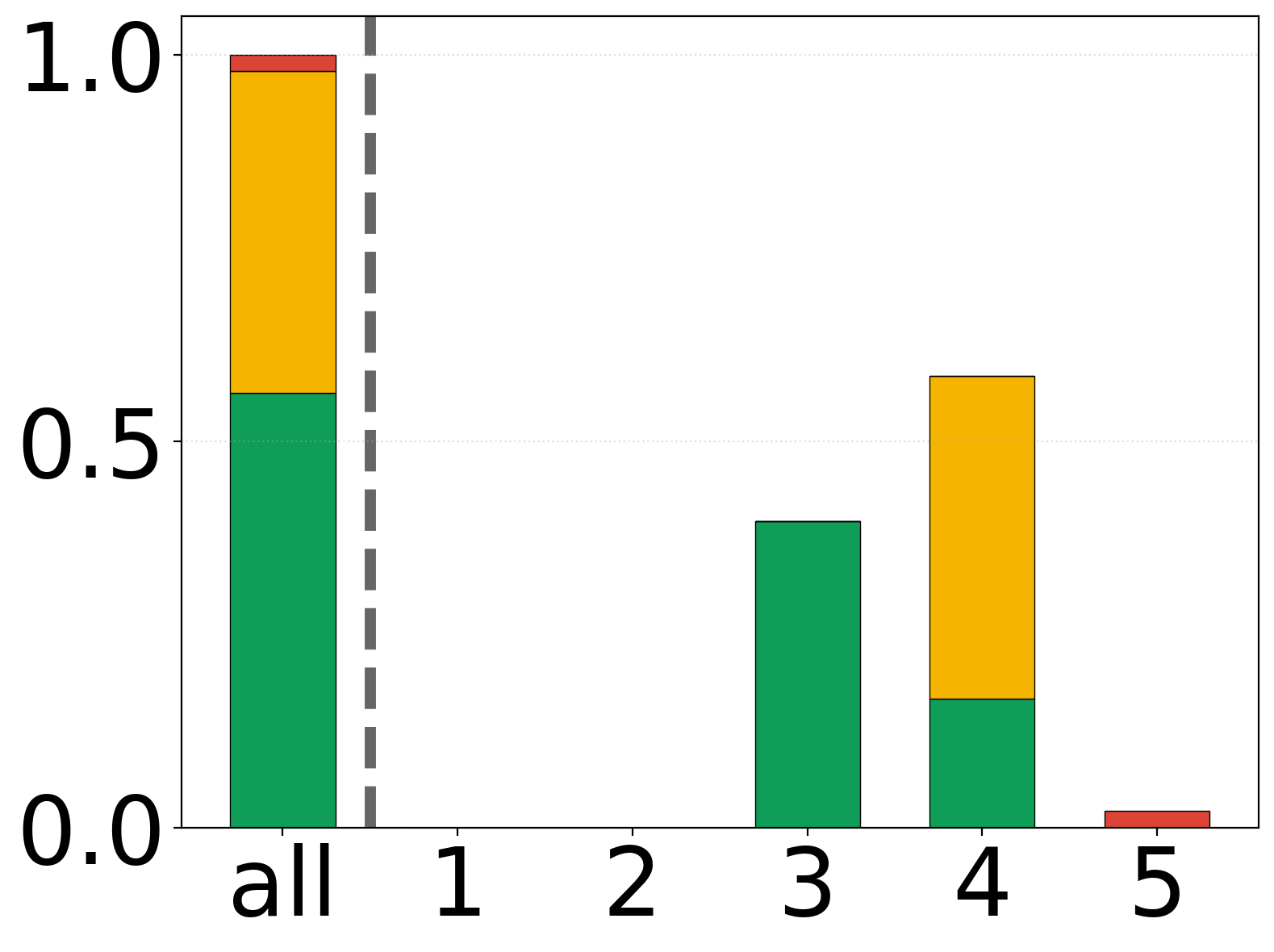}
  \end{minipage}
  \\[0.5em]
\fontsize{8}{8}\selectfont
  \begin{minipage}[l]{0.12\linewidth}  
    \raggedright\fontsize{8}{9}\selectfont
    \includegraphics[height=1.4em]{fig_model_logo/llama_logo.png}\\
    Llama-3.3
  \end{minipage}
  &
  \begin{minipage}[c]{0.31\linewidth}  %
    \centering
    \includegraphics[width=\linewidth]{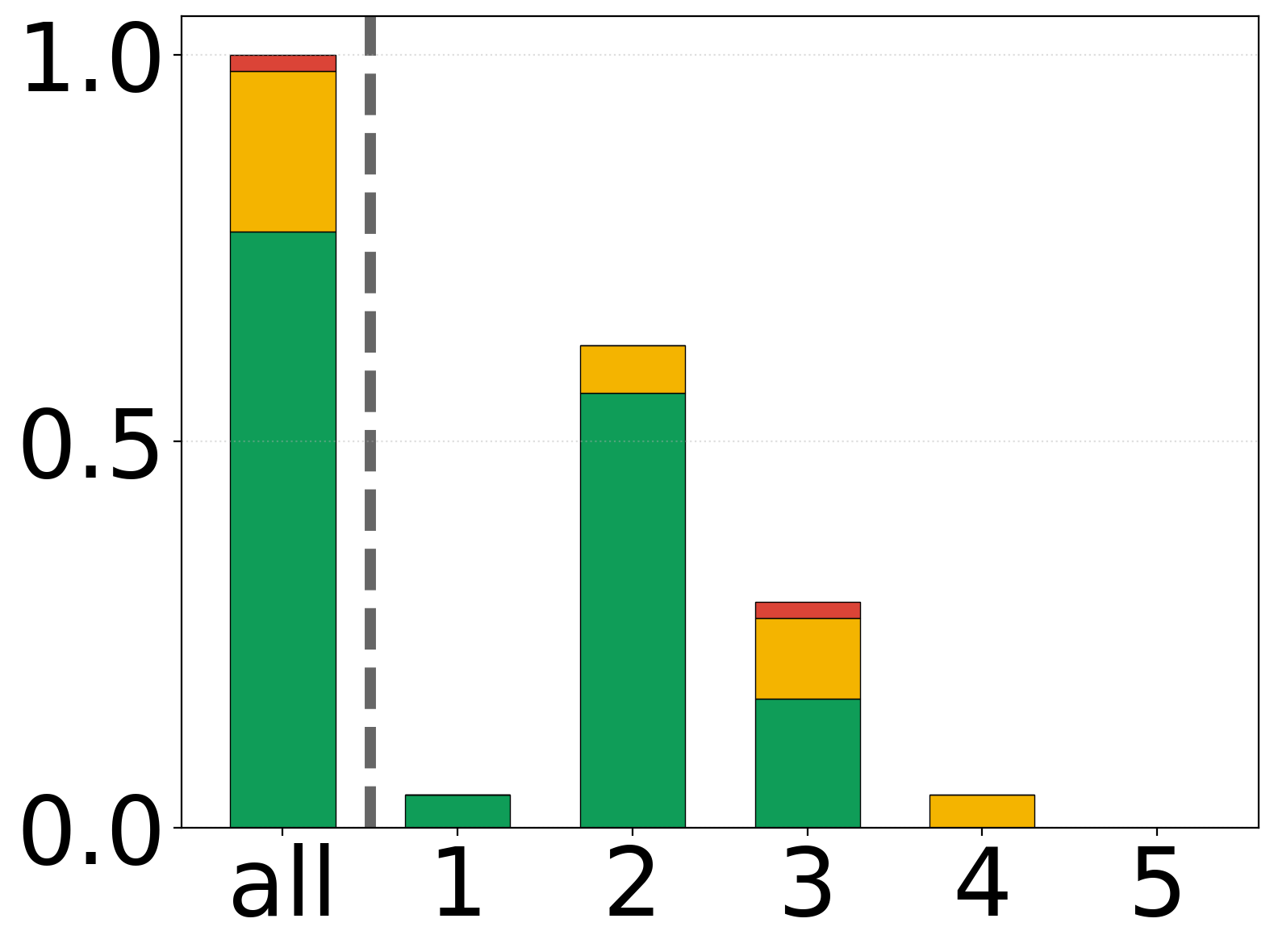}
  \end{minipage}
  &
  \begin{minipage}[c]{0.31\linewidth}  %
    \centering
    \includegraphics[width=\linewidth]{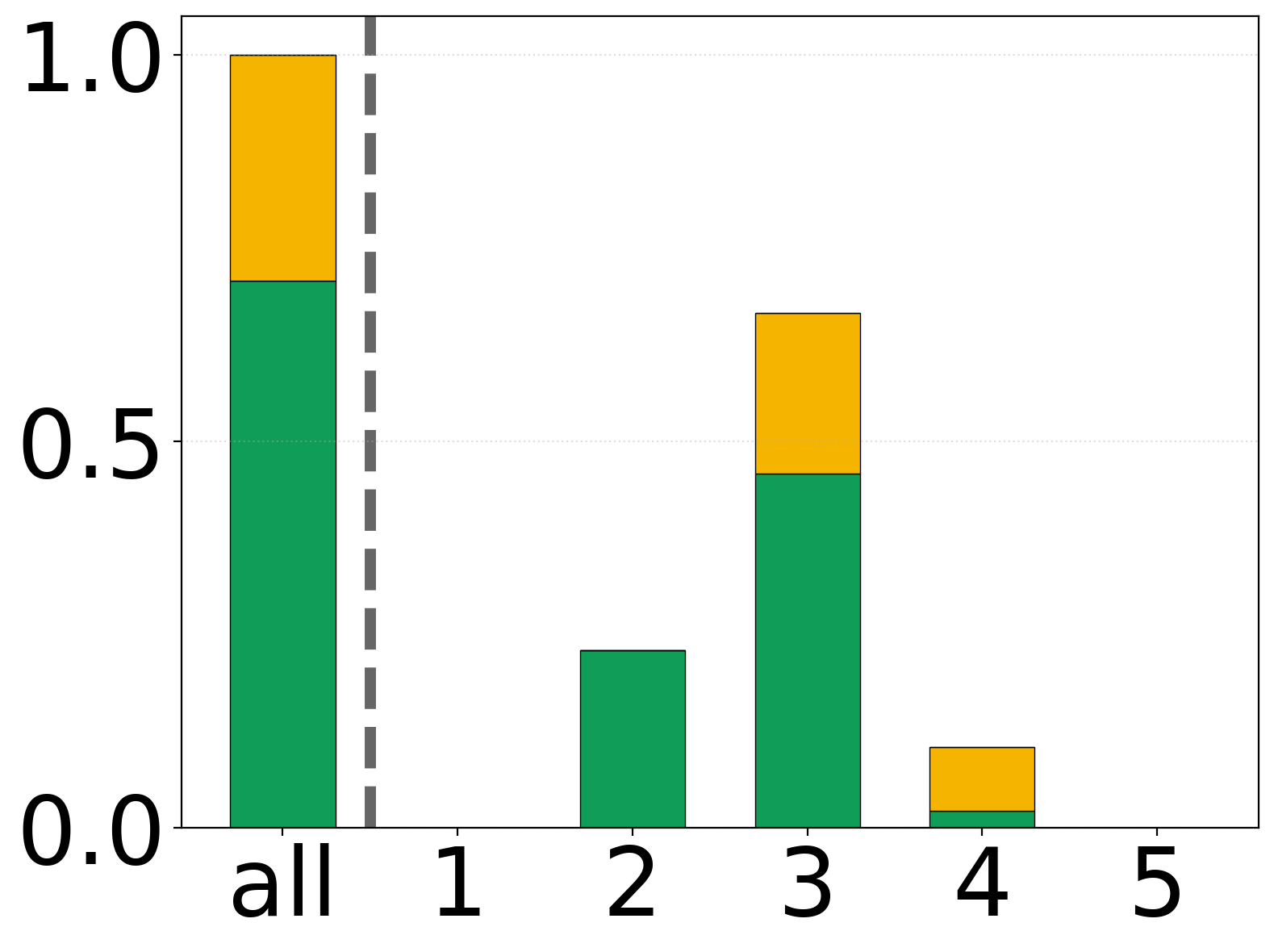}
  \end{minipage}
  \\[0.5em]

\fontsize{8}{8}\selectfont
  \begin{minipage}[l]{0.12\linewidth}  
    \raggedright\fontsize{8}{9}\selectfont
    \includegraphics[height=1.4em]{fig_model_logo/deepseek_logo.png}\\
    DeepSeek
  \end{minipage}
  &
  \begin{minipage}[c]{0.31\linewidth}  %
    \centering
    \includegraphics[width=\linewidth]{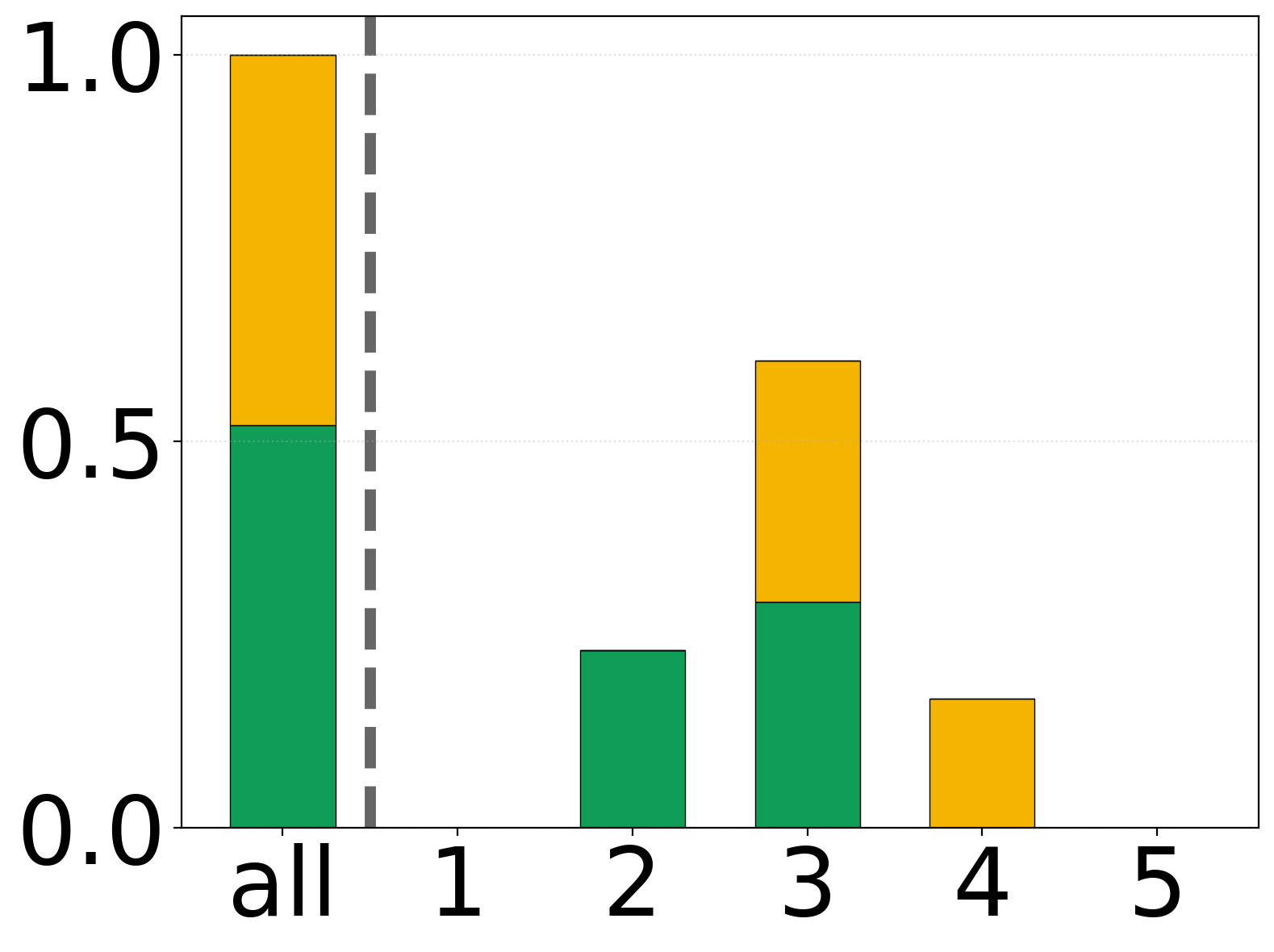}
  \end{minipage}
  &
  \begin{minipage}[c]{0.31\linewidth}  %
    \centering
    \includegraphics[width=\linewidth]{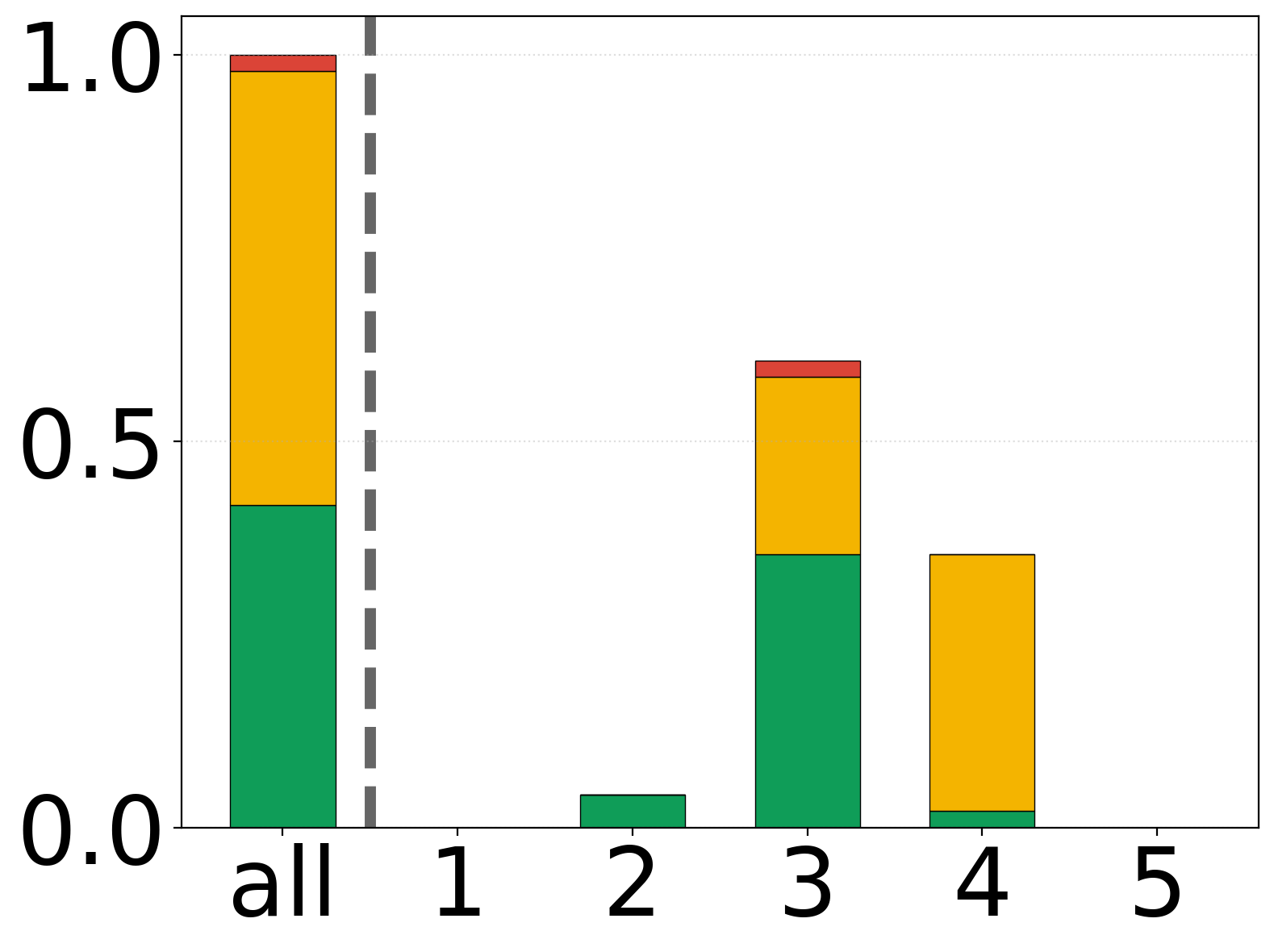}
  \end{minipage}
  \\[0.5em]

\fontsize{8}{8}\selectfont
  \begin{minipage}[l]{0.12\linewidth}  
    \raggedright\fontsize{8}{9}\selectfont
    \includegraphics[height=1.4em]{fig_model_logo/gpt_logo.jpg}\\
    GPT-4o
  \end{minipage}
  &
  \begin{minipage}[c]{0.31\linewidth}  %
    \centering
    \includegraphics[width=\linewidth]{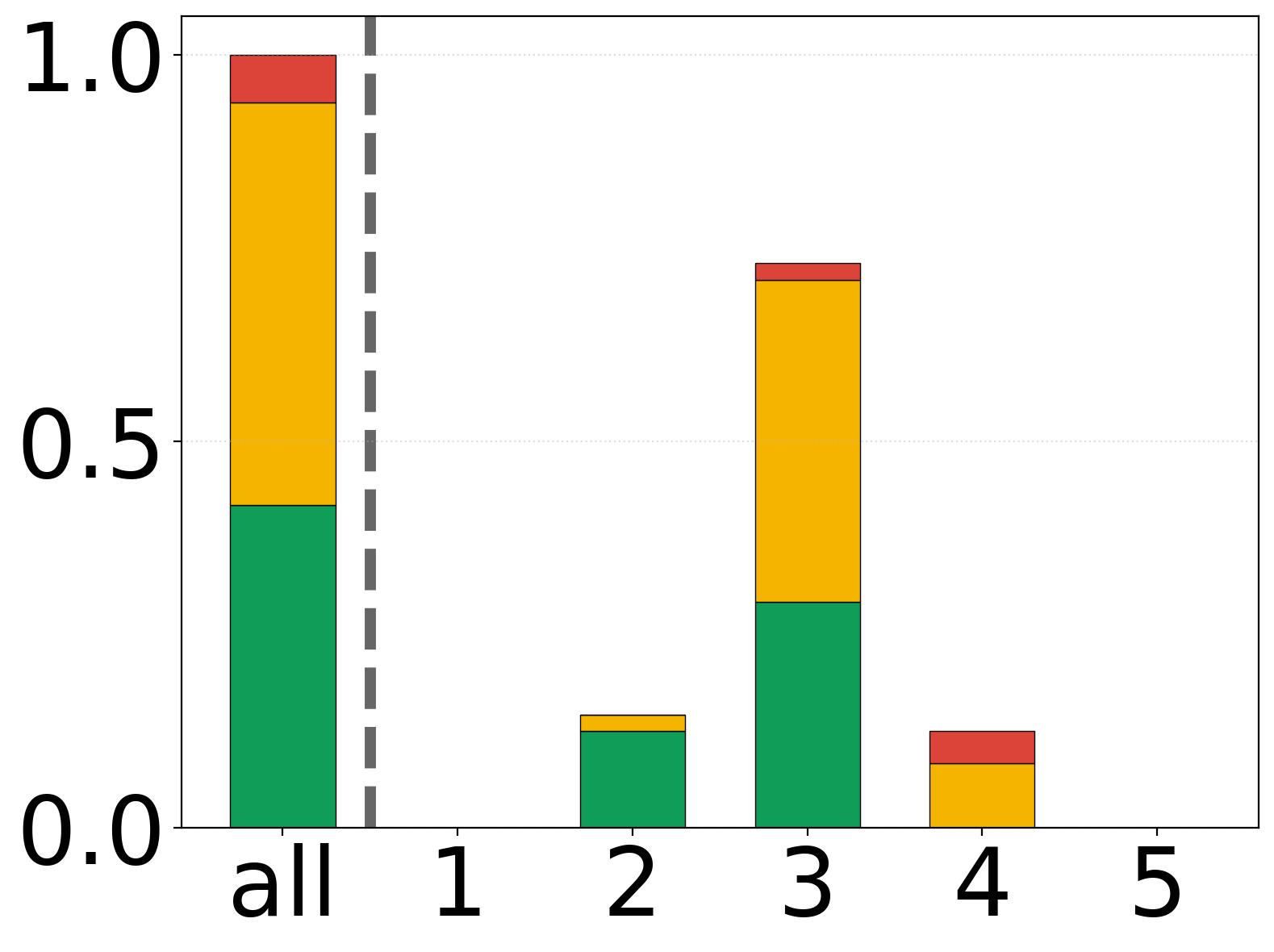}
  \end{minipage}
  &
  \begin{minipage}[c]{0.31\linewidth}  %
    \centering
    \includegraphics[width=\linewidth]{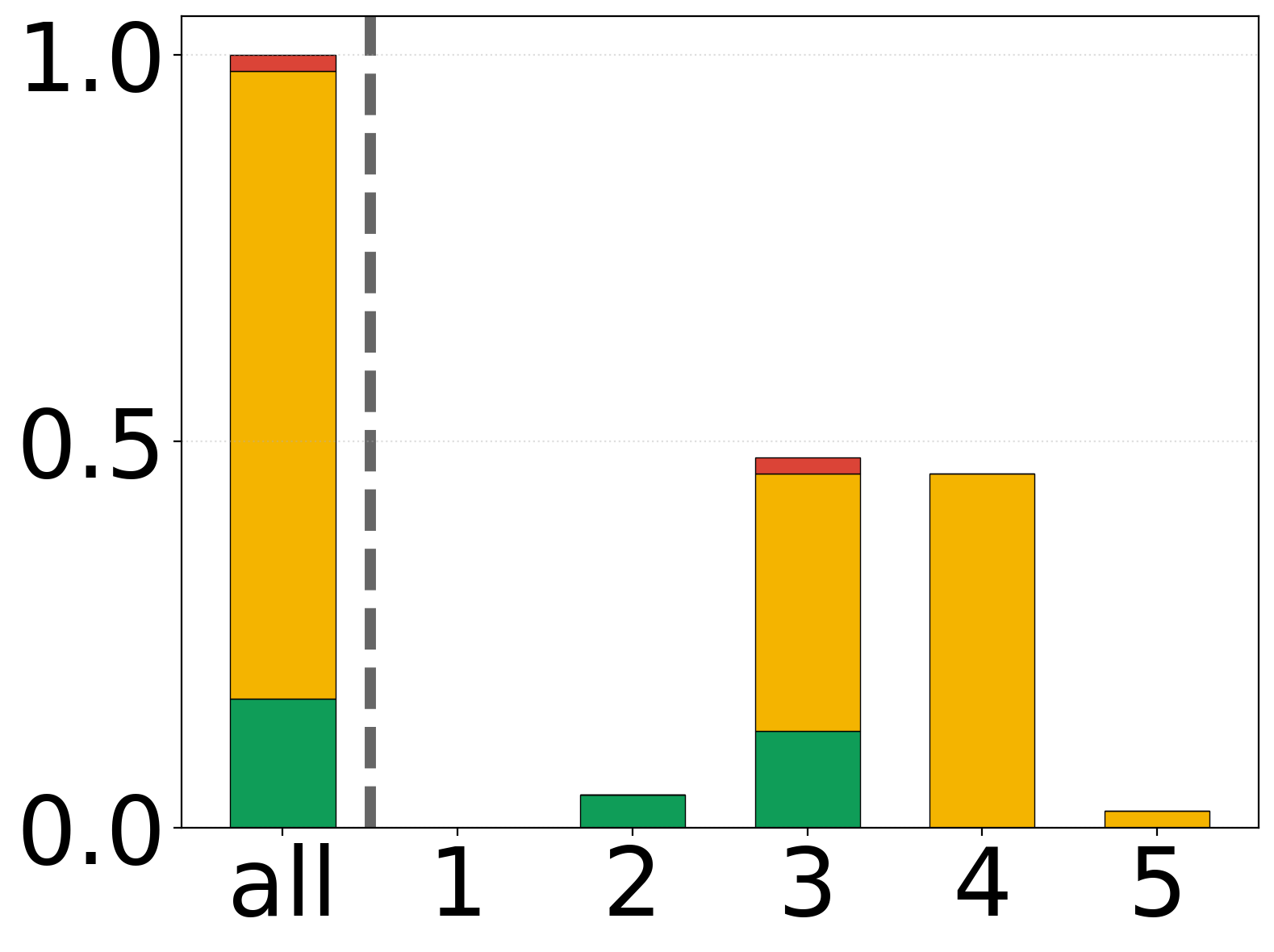}
  \end{minipage}
\\\hline
\end{tabular}
\caption{Changes in \textit{Specificity} after refinement across five LLMs. Colors indicate increase (green), no change (yellow), or decrease (red); the first bar shows overall proportions, followed by distributions by initial score.}
\label{fig:refine_spec}
\end{figure}

\subsubsection{Refinement Effectiveness}
\label{subsubsec:discuss_4}
To address \textbf{\textit{RQ7: How does REspEval feedback improve responses?}}, we analyze refinement applied to outputs from Settings~6 and~7, yielding Settings~8 and~9.
Table~\ref{tab:ARG_refine} (§\ref{subsec:app_discuss}) shows refinement produces statistically significant gains across all LLMs, settings, and quality metrics,\footnote{The only exception is \textit{Targeting} in Setting 9 with GPT-4o, where initial scores are already strong.} %
confirming the effectiveness of evaluation-guided refinement.
Improvements are most pronounced for initially weak responses (scores < 3) and diminish as initial quality increases. Consistently, Phi-4 and Llama-3.3 show the highest improvement rates in \textit{Specificity} (Figure~\ref{fig:refine_spec}) and \textit{Convincingness} (Figure~\ref{fig:refine_conv}, §\ref{subsec:app_discuss}), as they start with more weak responses that offer greater refinement potential.
Refined responses are generally longer (Table~\ref{tab:ARG_refine}), reflecting added detail for improvement. While length adherence slightly decreases after refinement in Setting~8 (except DeepSeek, which remains perfect), response plan controllability is largely preserved (Table~\ref{tab:ARG}). Across Settings~8 and~9, Qwen3 achieves the highest quality metrics. Overall, the proposed evaluation-guided refinement substantially improves response quality while largely maintaining controllability, demonstrating practical value for iterative ARG systems. 

\subsubsection{Discourse Analysis}
\label{subsubsec:discuss_5}
To address \textbf{\textit{RQ8: How do LLM-generated responses differ from authentic human responses in discourse?}}, we analyze tone--stance profiles in Figure~\ref{fig:discourse_tone_stance} and transition patterns in Figure~\ref{fig:discourse_rel_pos}.
Figure~\ref{fig:discourse_tone_stance} (§\ref{subsec:app_discuss}) shows that both human-authored responses and those generated by all LLMs across settings predominantly exhibit cooperative and socially friendly tones, with cooperative stances (\textit{\%Coop}) being the most prevalent and the combined proportion of cooperative and social stances (\textit{\%Coop}+\textit{\%Soc}) exceeding 0.5 (i.e., 50\%) in most cases. Both human and LLM-generated responses exhibit limited use of defensive language, with defensive stances (\textit{\%Defe}) consistently being the least frequent and remaining below 0.07. In addition, both humans and LLMs frequently employ hedge strategies, with hedging (\textit{\%Hed}) being the second most common stance and exceeding 0.20 in most settings.
However, when conditioned solely on review comments (Setting~1), LLMs tend to generate overly polite responses, with \textit{\%Coop}+\textit{\%Soc} ranging from 0.61 to 0.69, compared to 0.53 for human responses, while employing hedging less frequently than humans (0.16--0.21 vs.\ 0.28). When explicit author inputs are provided, the proportions of cooperative and social stances decrease substantially toward human levels, while the use of hedging increases markedly. These results highlight the effectiveness of incorporating author inputs in producing responses that are more human-like in tone--stance and less unnecessarily overpolite.
Figure~\ref{fig:discourse_rel_pos} (§\ref{subsec:app_discuss}) shows that human-authored responses primarily employ social stances (e.g., thanking reviewers) at early positions and rarely conclude with defensive language. The remaining three stance types are distributed more evenly across relative positions.
Llama-3.3 and DeepSeek exhibit similar patterns, concentrating social language at early positions. In contrast, Phi-4, Qwen3, and GPT-4o place social language at both the beginning and the end of responses, and these patterns remain consistent across different generation settings. Additionally, LLM responses show a relatively high concentration of the Other (\textit{Oth}) category at the final position, which typically corresponds to summary statements concluding the response.

\subsection{Case Studies}
\label{subsec:case_study}
To further demonstrate the analytical utility of \textit{REspEval} in capturing cross-dimensional interactions and trade-offs with detailed outcomes, we present two case studies (§\ref{subsec:app_case_studies}) on infrequent but informative cases where richer inputs do not help. In Figure \ref{fig:app_case_study_1}, added paragraph context introduces irrelevant details that obscure the core answer, degrading \textit{Targeting}, \textit{Specificity}, and \textit{Convincingness}. In Figure \ref{fig:app_case_study_2}, richer inputs cause omission of key details from core edit sentences, reducing both \textit{Specificity} and \textit{ICR}. 
These studies illustrate how our fine-grained, multi-dimensional evaluation surfaces subtle trade-offs that coarser evaluation would miss. 

\section{Conclusion}
\label{sec:conclusion}

We introduced \textit{REspGen}, an author-in-the-loop ARG framework with explicit author input, controllable generation, and evaluation-guided refinement; \textit{Re$^3$Align}, the first dataset enabling this new formulation; and \textit{REspEval}, a comprehensive evaluation suite spanning controllability, input utilization, discourse, and response quality.
Experiments reveal several key insights, including the necessity and effectiveness of author-in-the-loop ARG; the benefits of richer input contexts for improved factual grounding alongside a dilution of focus on core information; trade-offs between controllability and quality, especially under length limits; and the effectiveness of evaluation-guided refinement in improving response quality while preserving controllability.
Our dataset, generation framework, and evaluation tools provide foundational resources for future research in NLP for peer review, controllable generation, and human--AI collaboration.

\section*{Limitations}
This study has several limitations that should be considered when interpreting the results. From a data perspective, our study is restricted to English-language scientific publications, reflecting the limited availability of openly licensed source data. Examining the transferability of our findings to additional languages, domains, and application settings remains an important direction for future work and can be supported by our publicly released annotation models and analysis and evaluation tools.

From a modeling perspective, the implementations and empirical results presented in this study are intended to illustrate the proposed task, generation, and evaluation frameworks. Their primary purpose is to establish technical feasibility and to lay the groundwork for the development of future NLP systems for collaborative author response writing that integrate human expertise and intent with AI assistance. Consequently, the provided implementations have inherent limitations. For example, our approach selectively employs state-of-the-art LLMs and does not systematically evaluate alternative architectures, fine-tuning-based methods, or smaller models. A comprehensive exploration of modeling approaches for the proposed task lies beyond the scope of this work and is left for future research, which can build on the publicly released dataset.

\section*{Ethical Considerations}
All source data used in this work are publicly available under Creative Commons licenses (CC BY 4.0 and CC BY-NC 4.0). Data collection in the original sources followed ethical guidelines, and our dataset construction and redistribution adhere to the original licensing terms. Our \textit{Re$^3$Align} dataset is released under a CC BY-NC 4.0 license. 
Human annotation was conducted by experienced researchers who participated voluntarily and without financial compensation. Annotators were informed of the study’s purpose and provided consent for the use and publication of their annotations. This study does not involve the collection or processing of personal or sensitive information. To protect privacy, author, reviewer, and annotator identities have been excluded from the analysis and data release.

\section*{Acknowledgements}
This work is part of the InterText initiative\footnote{\url{https://intertext.ukp-lab.de/}} at the UKP Lab. This work has been funded by the European Union (ERC, InterText, 101054961). Views and opinions expressed are however those of the author(s) only and do not necessarily reflect those of the European Union or the European Research Council. Neither the European Union nor the granting authority can be held responsible for them. 

It has been also co-funded by the German Research Foundation (DFG) as part of the PEER project (grant GU 798/28-1) and the LOEWE Distinguished Chair “Ubiquitous Knowledge Processing”, LOEWE initiative, Hesse, Germany (Grant Number: LOEWE/4a//519/05/00.002(0002)/81). 

We thank Dr. Hiba Arnaout, Sheng Lu, and Dr. Federico Marcuzzi for their valuable feedback and suggestions on a draft of this paper. We would also like to express our gratitude to the members of the SIG InterText at the UKP Lab for their insightful discussions throughout the project.

\bibliography{custom}

\appendix

\begin{figure*}[]
  \centering
  \includegraphics[width=0.86\textwidth]{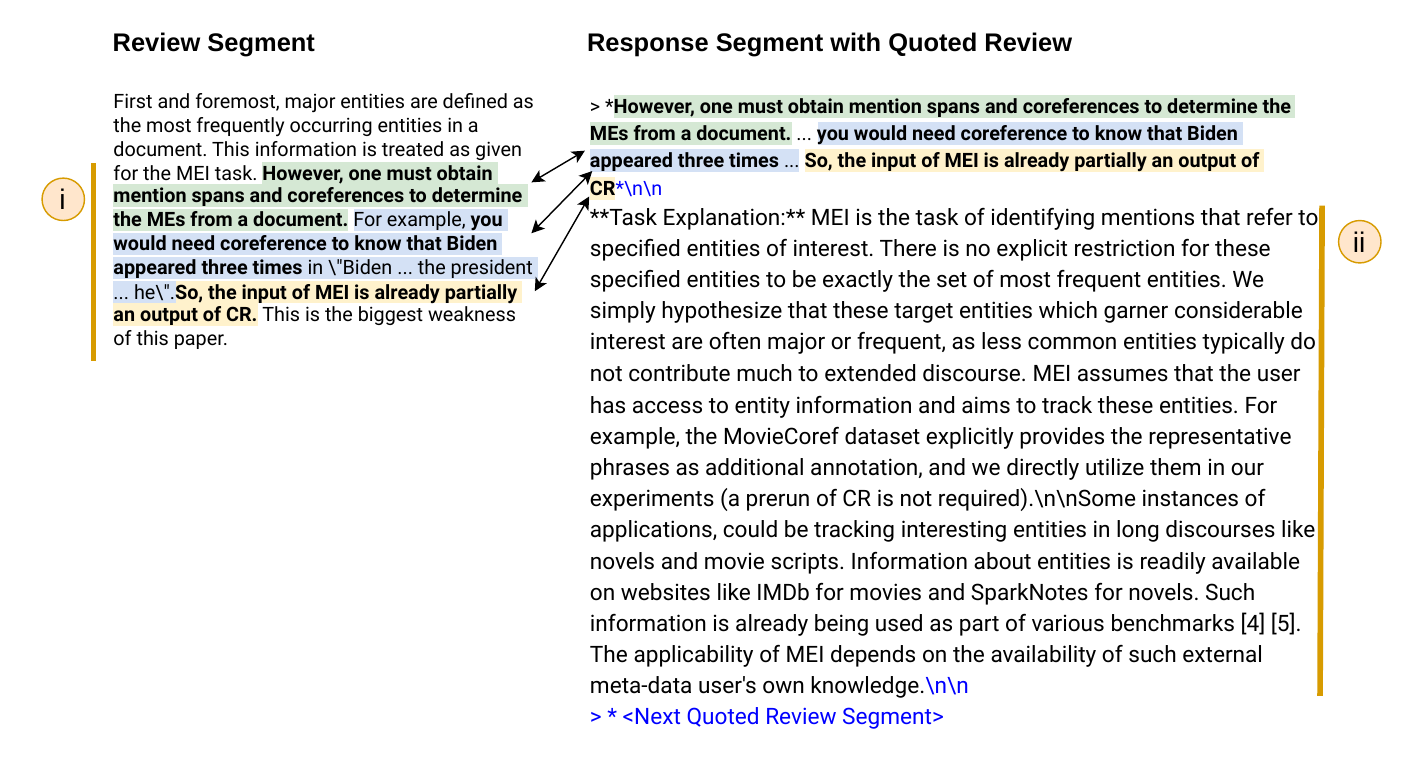}
  \caption{An illustrative example of segment-level review-response pair matching. Given a reviewer–author exchange $(C_k, A_k)$ for reviewer $k$, where $C_k$ is the full review text from reviewer $k$ and $A_k$ is the full author response, we show an aligned review–response segment pair $p^{k}_{mn} = p(c^{k}_m, a^{k}_n)$ with $c^{k}_m \in C_k$ and $a^{k}_n \in A_k$. $c^{k}_m$ is shown as (i) and $a^{k}_n$ as (ii) (see further notation in §\ref{subsec:triplet_alignment} and \ref{subsec:app_re3_align}).}
  \label{fig:app_pair_example}
\end{figure*}

\begin{figure*}[]
  \centering
  \includegraphics[width=0.88\textwidth]{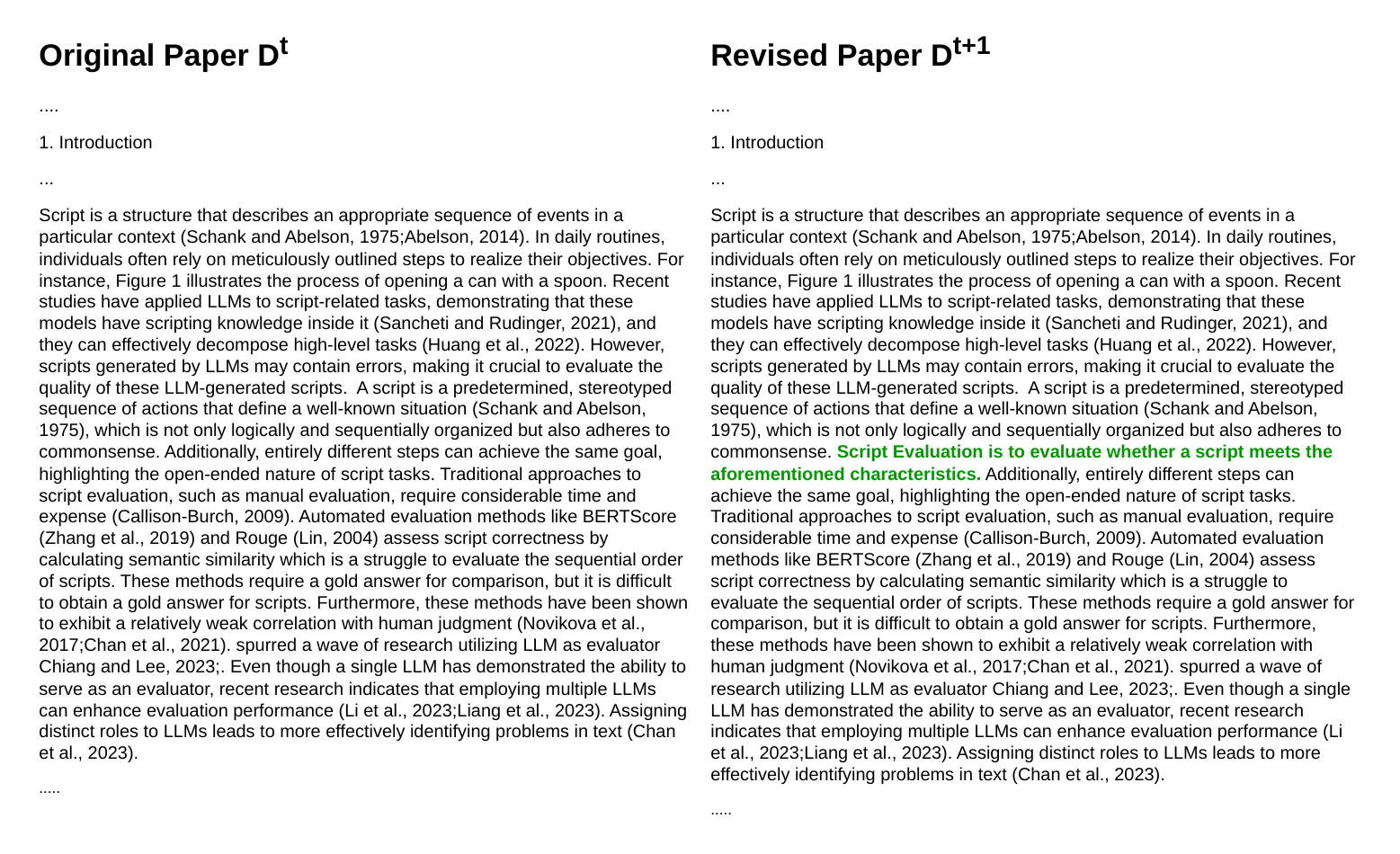}
  \caption{An illustrative example of an aligned sentence-level edit. Given original paper $D^{t}$ and revised paper $D^{t+1}$, with sentences denoted $x^{t}_j \in D^{t}$ and $x^{t+1}_i \in D^{t+1}$. We show an aligned edit $e_{ij} = e(x^{t+1}_i, x^{t}_j)$, where $x^{t+1}_i$ (highlighted in green in $D^{t+1}$) is an added sentence and $x^{t}_j$ is null, as it is a pure addition. The section location and paragraph context of each edit are preserved in the dataset.}
  \label{fig:app_edit_example}
\end{figure*}

\section{Dataset Construction}
\label{sec:app_data_construction}

\subsection{Preprocessing}
\label{subsec:app_prepro}
We preprocess each paper by first grouping its submission versions (v1 and v2), peer reviews, and author responses under a unified paper identifier and preserving all associated metadata if present (e.g., timestamps, reviewer scores, decision information). We then convert every document into the intertextual graph (ITG) representation \cite{f1000rd}, which encodes hierarchical structure and text order through node–edge relationships. Following \citet{re3}, we augment each ITG with sentence-level nodes produced via an assembled sentence-segmentation model, ensuring consistent granularity across documents and facilitating downstream alignment between review comments, responses, and revision edits at the sentence and segment levels.

\subsection{Review-Response Matching Algorithm}
\label{subsec:app_match_algorithm}

Given a review sentence $s_1$ and a response sentence $s_2$, we treat them as a match if any of the following conditions holds: (1) $s_1$ contains $s_2$; (2) $s_2$ contains $s_1$; (3) their SBERT similarity \cite{sbert} exceeds a threshold $t_0$; or (4) their partial-string fuzzy-matching score\footnote{\texttt{fuzzywuzzy.partial\_ratio}} exceeds a threshold $t_1$. The thresholds $t_0$ and $t_1$ were selected based on a pilot study over 20 examples, with the optimal configuration determined to be $t_0 = 85$ and $t_1 = 85$.
Figure \ref{fig:app_pair_example} presents an illustrative example of a matched review–response segment pair identified through the longest contiguous sentence-matching span.

After constructing initial review–response segment pairs, we apply several quality filtering steps to remove problematic cases. Specifically, we discard response segments that are too short (fewer than 2 sentences), which typically arise from noisy quotation matching or incomplete segmentation, and segments that are excessively long (more than 15 sentences), which usually indicate structural inconsistencies such as merged replies or missing quotation boundaries. After filtering, we retain 2,108 review–response segment pairs from EMNLP24 and 13,963 from PeerJ, providing a high-quality basis for subsequent alignment with revision edits.

\subsection{Re$^3$ Triplet Alignment}
\label{subsec:app_re3_align}
The task of Re$^3$ triplet alignment is to determine whether an annotated edit 
$e_{ij} \in E$ is related to a given review--response segment pair 
$p^{k}_{mn} \in P$. We use a two-way alignment strategy: 
(i) aligning the review segment $c^{k}_m$ with each sentence edit 
$e(x^{t+1}_i, x^{t}_j)$ via a function set $\text{CE}$, and 
(ii) aligning the response segment $a^{k}_n$ with the same edit via a 
function set $\text{AE}$. Each function set combines a fine-tuned 
state-of-the-art LLM classifier to capture semantic relations 
(i.e., $\text{CE}_{\text{llm}}$, $\text{AE}_{\text{llm}}$), together with a 
lightweight similarity-based component (i.e., $\text{CE}_{\text{sim}}$, 
$\text{AE}_{\text{sim}}$) for efficiency.

To maintain matching granularity, we split each review and response 
segment into sentences, denoted by $c^{k}_{mp} \in c^{k}_m$ and 
$a^{k}_{nq} \in a^{k}_n$. For review comment--edit alignment (CE), we enumerate all textual pairs $(s_1 = c^{k}_{mp}, s_2 = x^{t}_j +\,\text{``\textbackslash n''} +\, x^{t+1}_i)$.
For author response--edit alignment (AE), we analogously consider all pairs $(s_1 = a^{k}_{nq}, s_2 = x^{t}_j + \text{``\textbackslash n''} + x^{t+1}_i)$

\noindent\textbf{Similarity Matching.} 
Given a review or response sentence ($s_1$) and the combined sentence edit representation ($s_2$), 
$\text{CE}_{\text{sim}}$ and $\text{AE}_{\text{sim}}$ identify a pair as a positive match if \emph{all} of the following conditions hold: 
(i) their partial-string fuzzy matching score is $\geq 60$, 
(ii) their SBERT similarity is $\geq 20$, and 
(iii) their bigram overlap score is $\geq 10$. 
These conditions and thresholds were optimized in a pilot study to ensure high precision. 
All positive matches are aggregated into the set of aligned sentence-level edits for the segment. 
This lightweight component serves to efficiently identify edits with high lexical or semantic similarity to the review or response text.

\noindent\textbf{LLM Classifier.} Edits related to a review--response pair do not necessarily exhibit high surface similarity to the review or response text. To capture such cases and reduce missed alignments, we fine-tune two LLM classifiers, $\text{CE}_{\text{llm}}$ and $\text{AE}_{\text{llm}}$, following prior work \cite{aries,llm-classifier, re3}.  
\citet{llm-classifier} show that LLMs can be fine-tuned using their SeqC approach to achieve state-of-the-art performance on several classification tasks, including edit intent classification, a closely related task requiring paired inputs and fine-grained understanding of sentence edits.  
Following this approach, we concatenate $s_1$ and $s_2$ as the model input and fine-tune a set of base LLMs to perform binary classification of positive vs.\ negative alignment. The base models include Llama 2-13B \cite{llama2} and Llama 3-8B \cite{llama3}, the top-performing models identified by \citet{llm-classifier}, as well as the newly released Llama 3.2-3B model.\footnote{\url{https://ai.meta.com/blog/llama-3-2-connect-2024-vision-edge-mobile-devices/}}

We use existing human-annotated data available from prior work. For fine-tuning the review comment–edit alignment classifier ($\text{CE}_{\text{llm}}$), we combine 466 positive sentence-level samples from Re3-Sci \cite{re3} with paragraph-level samples from ARIES \cite{aries} that we decompose into 213 sentence-level instances, and generate 2,737 negative samples from Re3-Sci. The final dataset is split into train/validation/test sets (1,989/372/1,055). For the author response–edit alignment classifier ($\text{AE}_{\text{llm}}$), we use 1,364 positive samples from Re3-Sci and create 4,092 negative samples, split into train/validation/test sets (3,819/818/819). %

Table~\ref{tab:app_llm_classifier} reports the performance of the fine-tuned LLMs, from which we select the best-performing classifiers for annotation. We use the fine-tuned Llama 2–13B model as $\text{CE}_{\text{llm}}$ (96.3\% accuracy, 83.4 F1) and the fine-tuned Llama 3–8B–Instruct model as $\text{AE}_{\text{llm}}$ (93.7\% accuracy, 92.0 F1). Similar to the similarity-based approach, we enumerate all pairs of $(s_1, s_2)$, and aggregate all samples receiving positive alignment predictions into the set of aligned edits for the segment.

Finally, all aligned edits identified by $\text{CE}_{\text{sim}}$, $\text{CE}_{\text{llm}}$, $\text{AE}_{\text{sim}}$, and $\text{AE}_{\text{llm}}$ are aggregated into the set $[e_{\text{align}}]$, which is then used to construct each triplet sample $
t^{k}_{mn} = \bigl(c^{k}_m,\, a^{k}_n,\, [e_{\text{align}}]\bigr)
$. 

\begin{table}[ht]
\tabcolsep=0.06cm
\fontsize{9}{9}
\selectfont
\tabcolsep=0.09cm
\renewcommand{\arraystretch}{1.2} %
\begin{subtable}[ht]{0.48\textwidth}
       \centering
        \begin{tabular}{lll} \toprule
       Base LLM&Accuracy &F1 \\ \hline
       \begin{tabular}[c]{@{}l@{}}\textbf{Llama2-13b} \end{tabular}& \textbf{96.3} & \textbf{83.4} \\ 
       \begin{tabular}[c]{@{}l@{}}Llama2-13b-chat \end{tabular}& 94.9 & 79.6 \\
       \begin{tabular}[c]{@{}l@{}}Llama3-8b\end{tabular}&94.4 & 74.1\\
       \begin{tabular}[c]{@{}l@{}}Llama3-8b-instruct\end{tabular}&94.8 & 76.3\\
       \begin{tabular}[c]{@{}l@{}}Llama3.2-3b\end{tabular} & 96.1 & 80.8  \\
        \begin{tabular}[c]{@{}l@{}}Llama3.2-3b-chat\end{tabular} & 95.6 & 80.9  \\ \bottomrule
       \end{tabular}
       \caption{LLM classifier performance on review comment-edit alignment. The best-performing classifier (bolded) is used as $\text{CE}_{\text{llm}}$.}
       \label{tab:ce_llm}
\end{subtable}
\begin{subtable}[ht]{0.48\textwidth}
     \centering
        \begin{tabular}{lll} \toprule
       Base LLM&Accuracy &F1 \\ \hline
       \begin{tabular}[c]{@{}l@{}}Llama2-13b \end{tabular}& 92.3 & 90.5  \\
       \begin{tabular}[c]{@{}l@{}}Llama2-13b-chat \end{tabular}& 93.3 & 91.5 \\ 
       \begin{tabular}[c]{@{}l@{}}Llama3-8b \end{tabular}& 93.1 & 91.8 \\
       \begin{tabular}[c]{@{}l@{}}\textbf{Llama3-8b-instruct}\end{tabular}& \textbf{93.7} & \textbf{92.0}\\
       \begin{tabular}[c]{@{}l@{}}Llama3.2-3b\end{tabular} & 92.9 & 91.1  \\
        \begin{tabular}[c]{@{}l@{}}Llama3.2-3b-chat\end{tabular} & 93.0 & 90.9  \\ \bottomrule
       \end{tabular}
       \caption{LLM classifier performance on author response-edit alignment. The best-performing classifier (bolded) is used as $\text{AE}_{\text{llm}}$.
       }
       \label{tab:ae_llm}
\end{subtable}
\caption{Fine-tuned LLM classifier performance.}
\label{tab:app_llm_classifier}
\end{table}

\begin{table*}[ht]
\tabcolsep=0.06cm
\fontsize{8}{8}
\selectfont
\tabcolsep=0.09cm
\renewcommand{\arraystretch}{1.2}%
\begin{tabular}[t]{lll|ll }
 \multicolumn{2}{l}{DISAPERE \cite{disapere}} && \multicolumn{2}{l}{\textbf{REspGen (ours)}}\\
 Label & Definition && Label & Definition \\ \hline
 Evaluative & A subjective judgement of an aspect of the paper &&
 Criticism & A subjective judgement of an aspect of the paper \\ \hline
 Request & A request for information or change in regards to the paper && Request & A request for change in regards to the paper \\
 &  && Question & A request for information that requires an explicit answer \\ 
\end{tabular}
\caption{Review item types.}
\label{tab:app_item_types}
\end{table*}

\begin{table*}[ht]
\tabcolsep=0.06cm
\fontsize{8}{8}
\selectfont
\tabcolsep=0.09cm
\renewcommand{\arraystretch}{1.2}%
\begin{tabular}[t]{ll|ll|ll}
 & & & & & \\[-3pt] \toprule
 Stance Class && Response Action Label && Definition\\\hline
 Cooperative&&answer question&&answer a question \\ 
 &&task has been done&&claim that a requested task has been completed \\
&&task will be done in next version&&claim that a requested task will be completed in resubmission\\
&&accept for future work&&express approval for a suggestion, but for future work \\
&&concede criticism&&accept a criticism \\ \hline
Defensive&&refute question&&reject the validity of a question \\ 
 &&reject criticism&&reject the validity of a criticism \\
&&contradict assertion&&contradict a statement presented as a fact \\
&&reject request&&reject a request from a reviewer \\ \hline
Hedge&&mitigate importance of the question&&mitigate the importance of a question \\ 
 &&mitigate criticism&&mitigate the importance of a criticism\\ \hline
Social&&social&&non-substantive social text \\ \hline
Other (NonArg)&&follow-up question&&clarification question addressed to the reviewer \\ 
 &&structure&&text used to organize sections of the response \\
&&summarize&&summary of the response text \\
&&other&&all other sentences \\ \bottomrule
\end{tabular}
\caption{Response action labels and the corresponding stance classes. The labels are largely adopted from Table 3 of DISAPERE \cite{disapere}, with minimal adjustments to align with the item types used in our framework.}
\label{tab:app_response_actions}
\end{table*}

\begin{figure*}[t]
  \centering
  \includegraphics[width=0.82\textwidth]{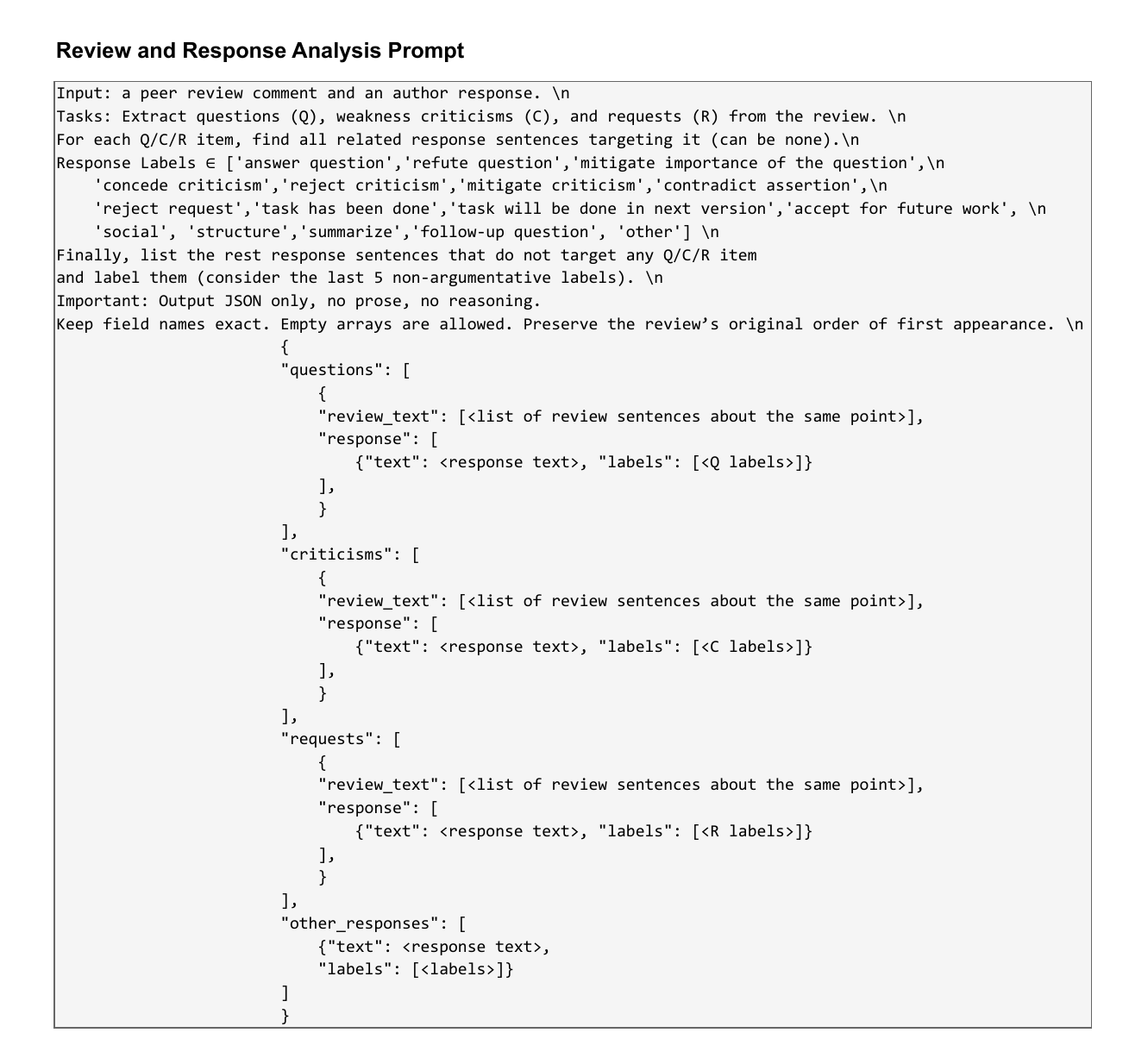}
  \caption{Optimized prompt to itemize review segments and label response actions. }
  \label{fig:app_review_response_labels_prompt}
\end{figure*}

\begin{figure*}[ht]
  \centering
  \includegraphics[width=0.998\textwidth]{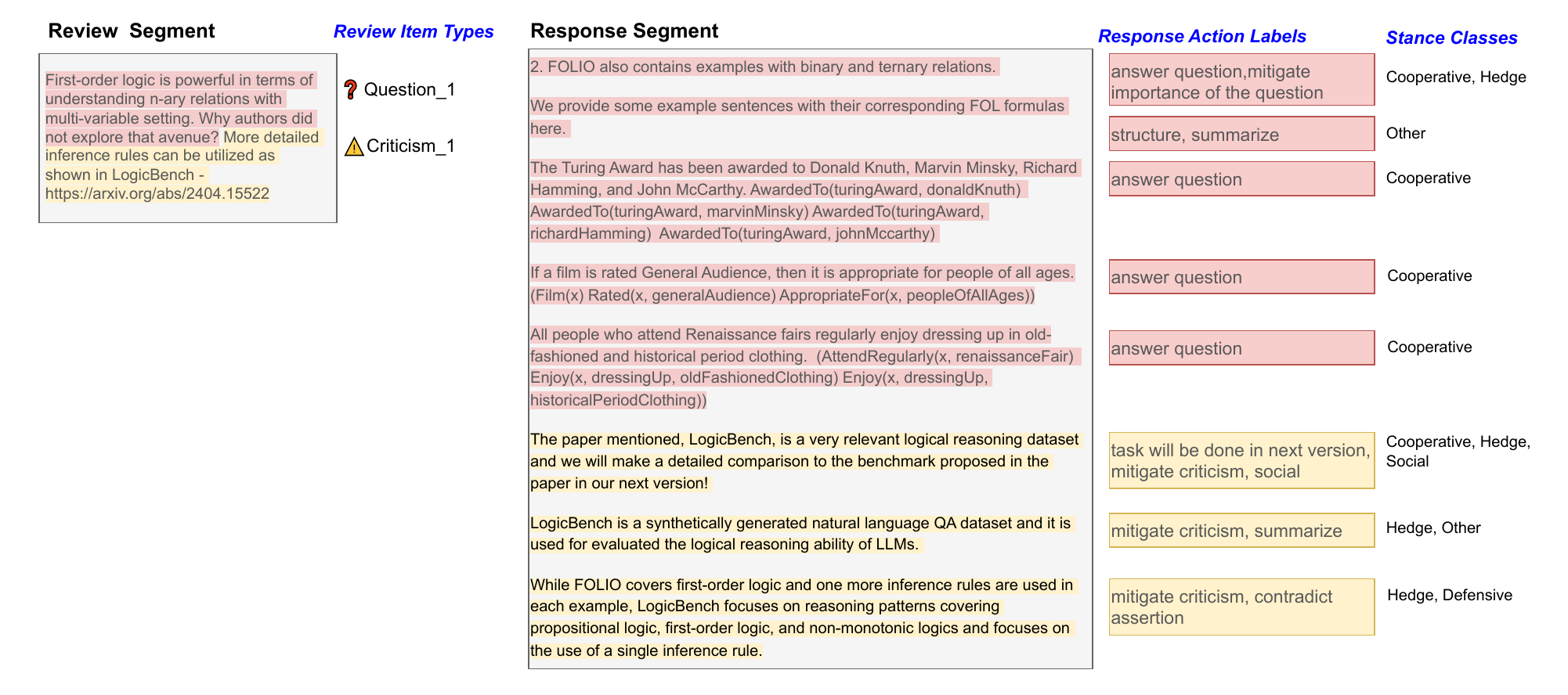}
  \caption{An illustrative example of annotated review items, response action labels, and the stance classes.}
  \label{fig:app_review_response_labels_example}
\end{figure*}

\section{REspGen}
\label{sec:app_respgen}
\subsection{Item-Based Response Planning}
\label{subsec:app_resp_plan}
Table~\ref{tab:app_item_types} provides definitions of our review item types and their mappings to the taxonomy of \citet{disapere}. Following their empirical findings that author responses most directly and explicitly address review actions of type \textit{Request} and \textit{Evaluative}, we rename \textit{Evaluative} as \textit{Criticism} and split \textit{Request} into two categories: \textit{Question} (requests for information) and \textit{Request} (requests for changes).
Table~\ref{tab:app_response_actions} presents our response action taxonomy and the corresponding stance classes, largely adapted from Table 3 of \citet{disapere} with minor adjustments to align with our item types.

Figure~\ref{fig:app_review_response_labels_prompt} shows the optimized GPT-5 prompts for review and response analysis. Given a pair of review and response segments, the model identifies and categorizes review items, extracts the corresponding response spans to each review item, and assigns response action labels. The output is a structured JSON, with example results illustrated in Figure~\ref{fig:app_review_response_labels_example}. Human verification indicates that GPT-5 consistently preserves the JSON format, and analyzing the review and response jointly produces better results than separate analysis, likely due to the additional cross-segment context enabling more reliable linking and reasoning.
\subsection{Input Component Configuration}
\label{subsec:app_input_cond}
Inspired by prior work demonstrating that retrieval-augmented generation substantially enhances LLM outputs by grounding generations in domain-relevant external knowledge \citep{shi2026cognitive}, we adopt a two-stage retrieval pipeline to supply the generator with precise scientific context.
Specifically, we retrieve the most relevant paragraphs from the original submission using a hybrid retrieval–reranking pipeline. Each paragraph is prepended with its corresponding section title. Given a review segment as the query, we apply a two-stage retrieval procedure:
(1) Hybrid first-stage retrieval combining a sparse BM25 retriever \citep{bm25} with a dense retriever built on science-tuned SPECTER2 embeddings \citep{specter2}. BM25 captures exact lexical overlap, while SPECTER2 captures semantic similarity between scientific texts; we combine them to improve recall and robustness, especially when reviewer terminology differs from the paper wording.
Scores from these two retrievers are fused using reciprocal rank fusion \citep{rrf}, yielding a robust initial candidate set.
(2) Reranking of the top candidates using the BAAI/bge-reranker-v2-m3 cross-encoder \citep{bge}, which provides fine-grained semantic relevance scores.
The final top-$k$ paragraphs (five by default) constitute the retrieved paper context (\textit{\textbf{v1}}). 
This module enriches the generator with topic-relevant scientific background while remaining lightweight and domain-agnostic, and is easily applicable across review–response scenarios.

\section{REspEval}
\label{sec:app_respeval}
\begin{figure*}[ht]
  \centering
  \includegraphics[width=0.8\textwidth]{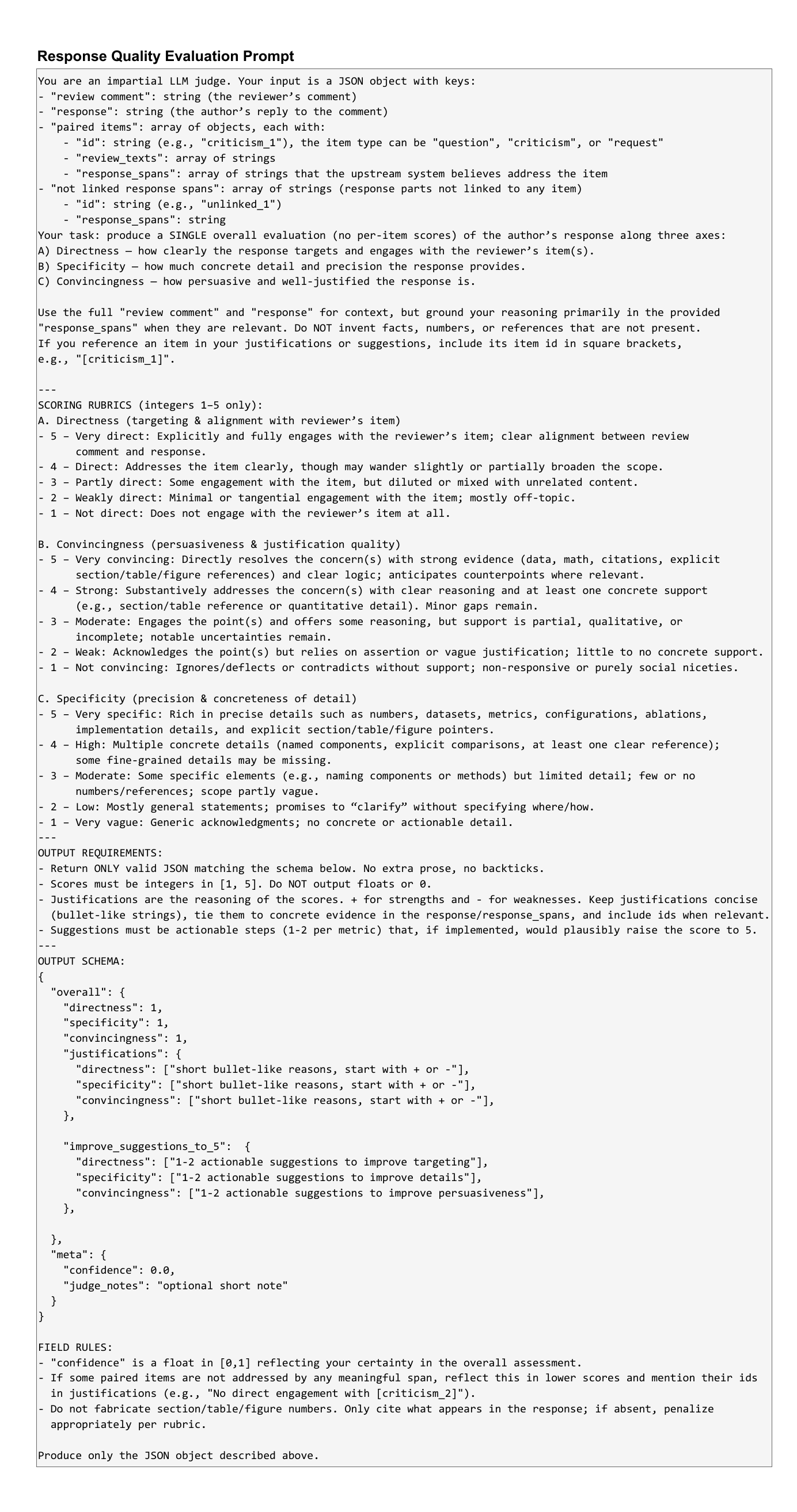}
  \caption{Optimized prompt to evaluate response quality in terms of targeting (directness), specificity, and convincingness. }
  \label{fig:app_quality_eval_prompt}
\end{figure*}

\begin{figure*}[ht]
  \centering
  \includegraphics[width=0.79\textwidth]{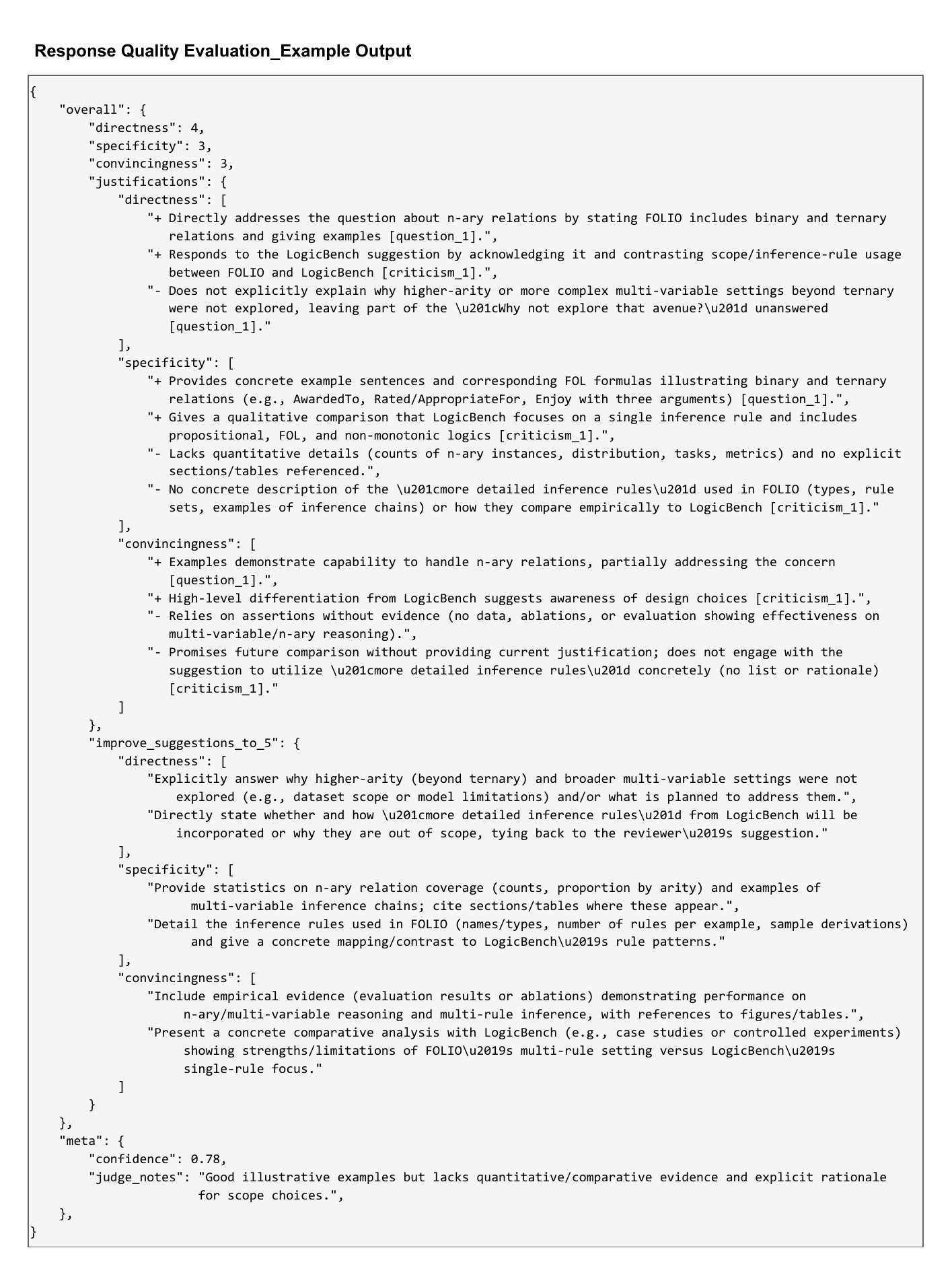}
  \caption{An illustrative output example of response quality evaluation, including scores, justifications and refinement suggestions per metric.}
  \label{fig:app_quality_eval_output}
\end{figure*}

\subsection{Quality Evaluation Procedure}
\label{subsec:app_respeval_qual_proce}
Given the review segment, response, and alignments of review items and response spans, we prompt GPT-5 to assign 5-point scores for targeting, specificity, and convincingness, and to provide both justifications and refinement suggestions for each criterion. 
The evaluation prompt (Figure~\ref{fig:app_quality_eval_prompt}) specifies explicit scoring rubrics. The evaluation output (Figure~\ref{fig:app_quality_eval_output}) includes (i) the three scores (ii) justifications per metric, expressed as concise, evidence-grounded bullet points that separate strengths(+) and weaknesses(-) and reference item IDs when relevant; and (iii) suggestions per metric, consisting of one or two actionable advices that, if implemented, would plausibly raise the score to five.

\subsection{Quality Evaluation Verification}
\label{subsec:app_respeval_qual_veri}
We conduct comprehensive studies to assess the consistency, robustness, interpretability, and reliability of the evaluation procedure.

\begin{table}[t]
\tabcolsep=0.06cm
\fontsize{8}{8}
\selectfont
\tabcolsep=0.09cm
\renewcommand{\arraystretch}{1.2}%
\begin{tabular}[t]{l|l|l|l|l|l }
 & & & & & \\[-3pt] \toprule
 & mean std & median std 
&p90 std & max std & ICC \\\hline
Targeting  &.07 & 0  &.58 &.58  &.94 \\ 
Specificity  & .09 & 0 & .58 &.58 &.93 \\ 
Convincingness & .05 &0 & 0 &.58 &.95 \\ \bottomrule
\end{tabular}
\caption{Consistency verification. Reported are the mean, median, 90th-percentile, and maximum standard deviation, as well as the Intra-class Correlation Coefficients (ICC), computed across repeated evaluation runs for the three scoring dimensions.}
\label{tab:app_qual_consistent}
\end{table}

\paragraph{Consistency.}
To assess consistency, we run the evaluation three times on the 48 experimental samples (§\ref{sec:results_discussion}) and report standard-deviation–based (std) statistics together with Intra-class Correlation Coefficients (ICC, \citealp{icc}). As shown in Table~\ref{tab:app_qual_consistent}, both the median and mean std are zero, indicating that, on average, the model assigns identical scores across runs. For more than half of the samples, all three runs yield exactly the same score. Even at the 90th percentile, the std corresponds to at most a 1-point difference, and no sample exhibits a spread larger than one point (max. std.\ = 0.58). ICC values further demonstrate excellent reliability \cite{icc-expl} across all three evaluation axes. Taken together, these results show that the evaluation procedure is highly stable, with nearly all samples receiving identical or near-identical scores across repeated runs.

\captionsetup[subfigure]{labelformat=simple, labelsep=space}
\renewcommand\thesubfigure{(\alph{subfigure})}
\begin{figure*}[t]
    \centering
    \begin{subfigure}{0.315\textwidth}
        \centering
        \includegraphics[width=\linewidth]{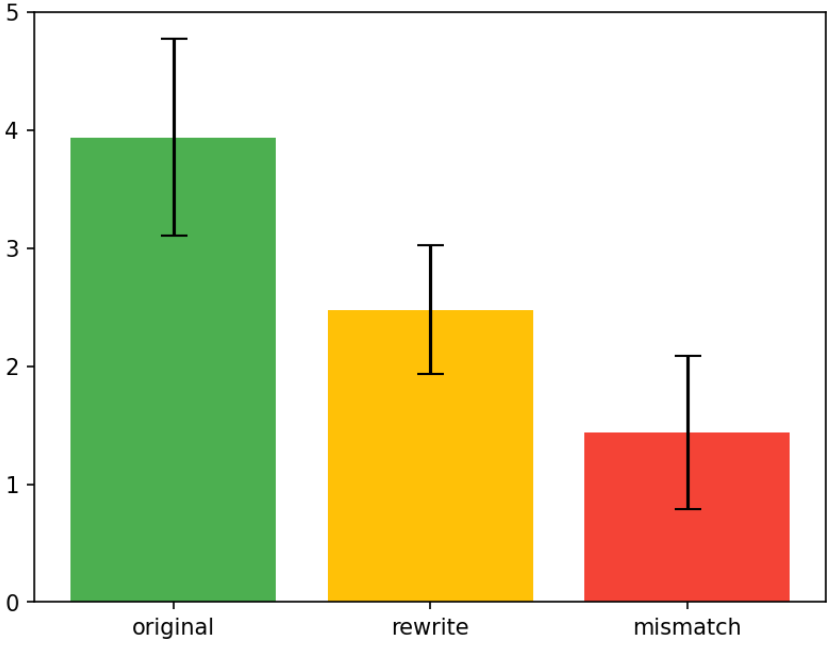}
        \caption{Targeting}
        \label{fig:robust_targ}
    \end{subfigure}
    \hfill
    \begin{subfigure}{0.32\textwidth}
        \centering
        \includegraphics[width=\linewidth]{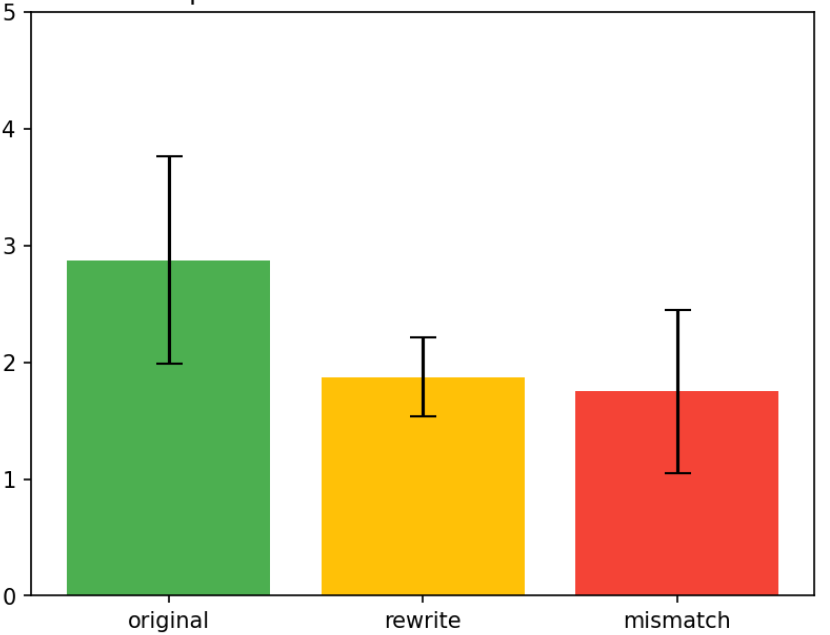}
        \caption{Specificity}
        \label{fig:robust_spec}
    \end{subfigure}
    \hfill
    \begin{subfigure}{0.32\textwidth}
        \centering
        \includegraphics[width=\linewidth]{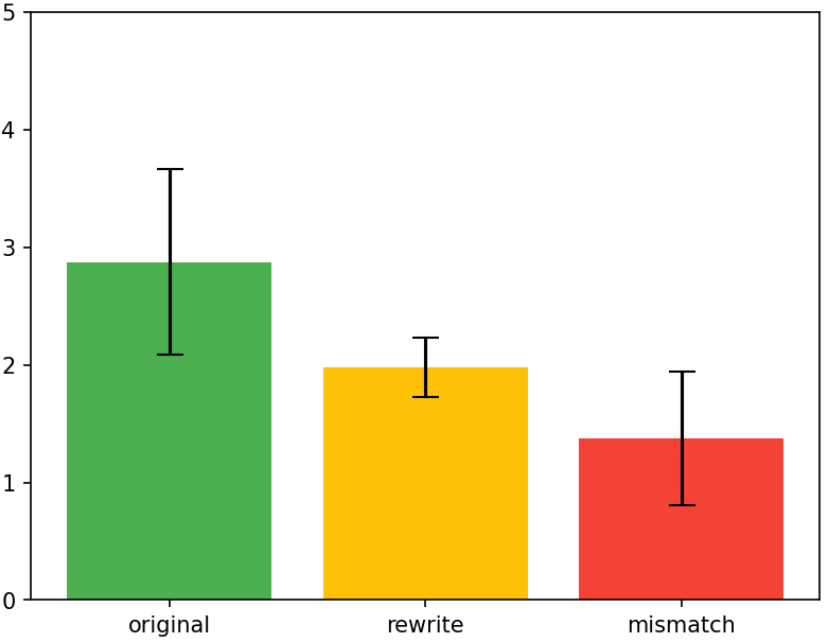}
        \caption{Convincingness}
        \label{fig:roubust_conv}
    \end{subfigure}
    \caption{Robustness verification. Mean scores and corresponding standard deviation error bars are presented for each scoring dimension across the three test conditions: original (green), rewritten (yellow), and mismatched (red).}
    \label{fig:app_qual_robust}
\end{figure*}

\captionsetup[subfigure]{labelformat=simple, labelsep=space}
\renewcommand\thesubfigure{(\alph{subfigure})}
\begin{figure*}[t]
    \centering
    \begin{subfigure}{0.86\textwidth}
        \centering
        \includegraphics[width=\linewidth]{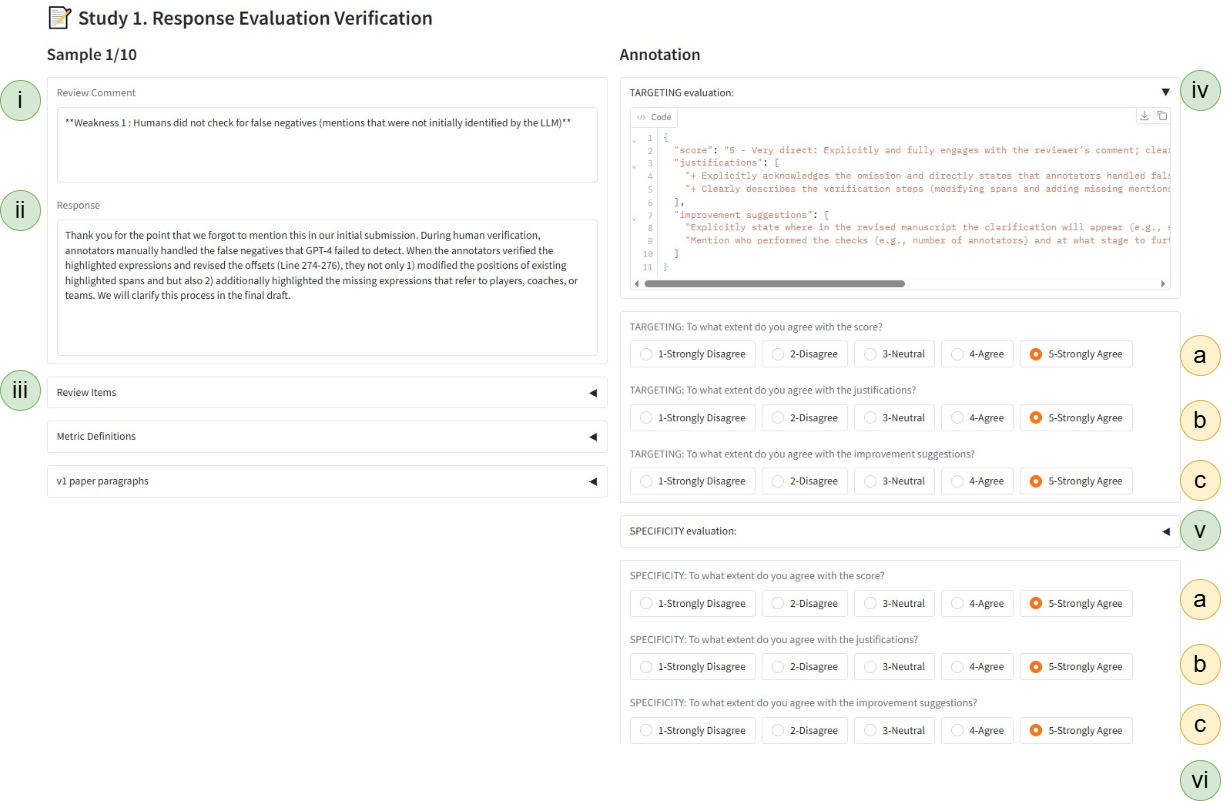}
        \caption{Human Study 1. Response Evaluation Verification.
Shown on the right are: (i) a review segment, (ii) the corresponding response, and (iii) optional contextual materials available to annotators, including extracted review items, metric definitions and scoring rubrics, and the top five relevant paragraphs retrieved from the original paper for additional topic-specific information. On the left, annotators read three evaluation blocks, Targeting, Specificity, and Convincingness (iv–vi), each containing a score, justifications, and refinement suggestions. For each block, annotators indicate the extent to which they agree with (a) the score, (b) the justifications, and (c) the suggestions on a 5-point scale: 1 = strongly disagree, 2 = disagree, 3 = neutral, 4 = agree, 5 = strongly agree.}
        \label{fig:human_study1}
    \end{subfigure}
    \hfill
    \begin{subfigure}{0.86\textwidth}
        \centering
        \includegraphics[width=\linewidth]{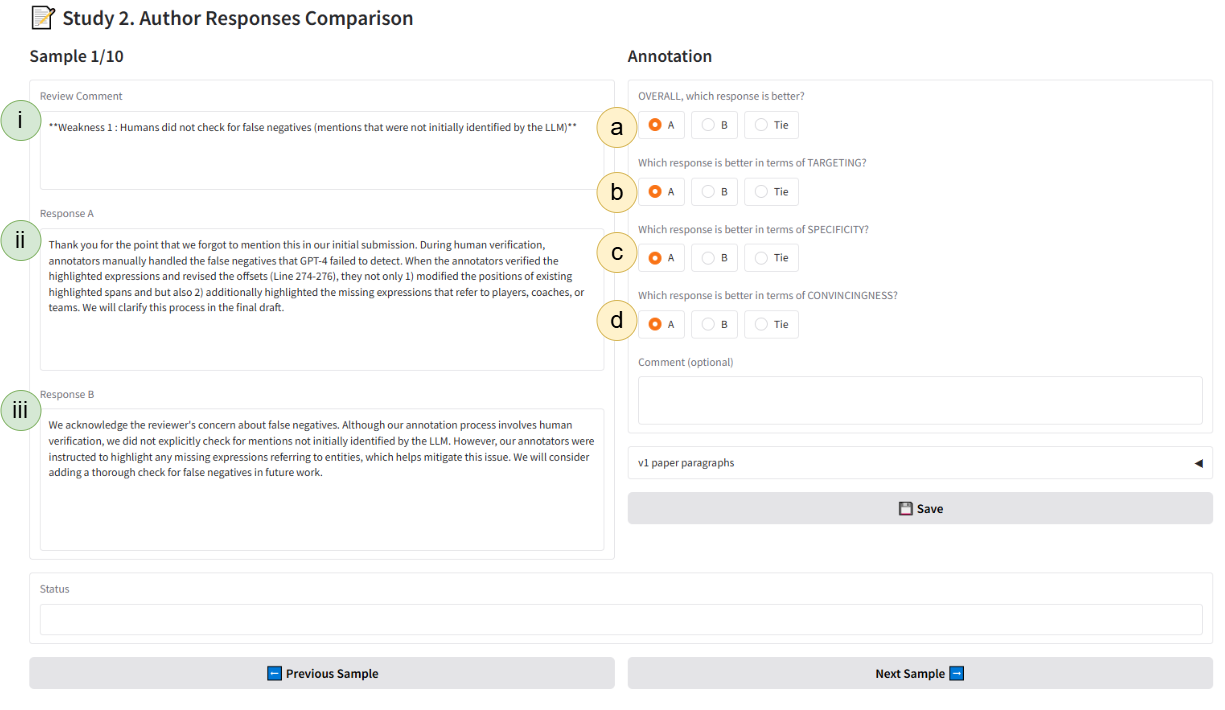}
        \caption{Human Study 2. Author Response Comparison. Shown on the right are: (i) a review segment, (ii) a corresponding Response A, and (iii) another corresponding Response B. On the left, annotators indicate which response is better (or tie) with respect to (b) Targeting, (c) Specificity, and (d) Convincingness, as well as which response wins overall (a).}
        \label{fig:human_study2}
    \end{subfigure}
    \caption{Annotation interface used in the human studies.}
    \label{fig:app_qual_human_study_interface}
\end{figure*}

\begin{table}[t]
\tabcolsep=0.06cm
\fontsize{8}{8}
\selectfont
\tabcolsep=0.09cm
\renewcommand{\arraystretch}{1.2}%
\begin{tabular}[t]{l|l|l|l|l|l}
& & &&  \\[-3pt]\toprule
 \#involvements & 0 & 1-3& 4-6& 7-10&>10  \\\hline
\quad \quad(i) as author &0\% &16.7\%&33.3\%&25.0\%&25.0\%  \\ 
\quad \quad(ii) as reviewer &8.3\%&25.0\%&16.7\%&16.7\%&33.3\% \\ 
\bottomrule
\end{tabular}
\caption{Peer-review experience and expertise of the 12 human annotators.
Each annotator had substantial peer-review experience, indicated by the number of full review cycles they had participated in as (i) an author or co-author and (ii) a reviewer.
}
\label{tab:app_human_annotator}
\end{table}

\begin{table}[t]
\tabcolsep=0.06cm
\fontsize{8}{8}
\selectfont
\tabcolsep=0.09cm
\renewcommand{\arraystretch}{1.1}%
\begin{tabular}[t]{l|l|l|l }
 & & &  \\[-3pt] \toprule
 & \%Agree & \begin{tabular}{l}
      \%Not\\Disagree
 \end{tabular} & \begin{tabular}{l}
      Mean\\Agreement\\ Rating
 \end{tabular} \\\hline
Targeting &&& \\ 
\quad \quad \quad \quad- score &86.7\% & 95.3\% &4.38 \\ 
\quad \quad\quad \quad- justifications &88.6\% & 97.2\% &4.38 \\ 
\quad \quad\quad \quad- suggestions &81.0\% & 97.2\% &4.22 \\ \hline
Specificity &&& \\ 
\quad \quad\quad \quad- score &85.2\% & 95.2\% &4.18 \\ 
\quad \quad\quad \quad- justifications &84.8\% & 97.2\% &4.17 \\ 
\quad \quad\quad \quad- suggestions &85.7\% & 97.2\% &4.18 \\ \hline
Convincingness &&& \\ 
\quad \quad\quad \quad- score &85.7\% & 96.2\% &4.23 \\ 
\quad \quad\quad \quad- justifications &83.8\% & 97.2\% &4.21 \\ 
\quad \quad\quad \quad- suggestions &80.0\% & 97.2\% &4.18 \\ 
\bottomrule
\end{tabular}
\caption{Results from Human Study 1: Response Evaluation Verification.
Scale: 1 = strongly disagree, 2 = disagree, 3 = neutral, 4 = agree, 5 = strongly agree.
Reported are the percentages of human agreement (ratings of 4–5), non-disagreement (ratings of 3–5), and the mean agreement ratings for quality scores, justifications, and suggestions across the three scoring dimensions.}
\label{tab:app_human_study1}
\end{table}

\begin{table}[t]
\tabcolsep=0.06cm
\fontsize{8}{8}
\selectfont
\tabcolsep=0.09cm
\renewcommand{\arraystretch}{1.2}%
\begin{tabular}[t]{l|l|l }
 & &   \\[-3pt] \toprule
 & \%Agreement & Krippendorff’s $\alpha$  \\\hline
Overall &91.4\% & .87 \\ 
Targeting &83.8\% &.81 \\ 
Specificity &91.9\% &.87 \\ 
Convincingness &90.0\% &.89 \\ 
\bottomrule
\end{tabular}
\caption{Results from Human Study 2: Author Response Comparison.
Reported are the percentages of human–LLM agreement on win/loss decisions across the three quality dimensions and overall, and the inter-annotator agreement measured by Krippendorff’s $\alpha$.
}
\label{tab:app_human_study2}
\end{table}

\paragraph{Robustness.}
To verify robustness, we evaluate the experimental samples under three conditions: original, mismatched (paired with a random review), and rewritten by GPT-5 to be less target, specific and convincing. As expected, scores dropped sharply from the original condition to the mismatch condition, with the rewrites showing intermediate degradation (Figure \ref{fig:app_qual_robust}). 
For example, Targeting declined by 1.46 points under rewrites with a large Cliff’s~$\delta$~\cite{cliff} of 0.81, and by 2.50 points under mismatches with an even larger effect ($\delta = 0.96$).
Specificity and Convincingness also decreased significantly, with large effect sizes ($\delta$) ranging from 0.63–0.84. All reductions compared to the original group are statistically significant ($p < 10^{-9}$, paired t-tests \cite{ttest}). These results demonstrate that the scoring is robust to perturbations, which reliably distinguishes genuine responses from degraded or irrelevant ones.

\paragraph{Interpretability and Reliability.}
To assess the interpretability and reliability of our evaluation procedure and its alignment with human judgments, we conduct two human studies with 12 experienced researchers. The annotator pool includes 2 postdoctoral researchers, 8 PhD students, 1 research intern, and 1 professor, all of whom have participated in multiple full peer-review cycles as authors and/or reviewers (Table~\ref{tab:app_human_annotator}), demonstrating strong domain expertise for the task. Both studies were conducted after a joint review of the annotation guidelines and preparatory discussions.

In \textbf{Study~1} (Figure~\ref{fig:human_study1}), annotators rate their agreement (1–5) with the GPT-5’s scores, justifications, and refinement suggestions for \textit{Targeting}, \textit{Specificity}, and \textit{Convincingness}.
In \textbf{Study~2} (Figure~\ref{fig:human_study2}), annotators compare two responses to the same review and judge which is better (or a tie) on the three quality dimensions and overall. 
From these two studies, we collect 1,365 human judges. 

Results from \textbf{Study~1} (Table~\ref{tab:app_human_study1}) show high human agreement with the GPT-5’s scores ($\geq$85.2\%), justifications ($\geq$83.8\%), and suggestions ($\geq$80\%). Across all three aspects and all three quality dimensions, mean agreement ratings exceed 4.17 (on a 1–5 scale, where 4 = agree and 5 = strongly agree), and disagreement rates remain below 5\%. These findings indicate that the GPT-5’s scores and suggestions are highly reliable and that its justifications are generally well-aligned with human interpretability.
\textbf{Study 2} (Table~\ref{tab:app_human_study2}) further demonstrates strong human–LLM agreement on win/loss decisions (83.8\%–91.9\%) and substantial \cite{Landis1977TheMO} inter-annotator agreement (Krippendorff’s $\alpha$ = 0.81–0.89).

\section{Experiments}
\label{sec:app_res_dis}

\subsection{Experimental Setup}
\label{subsec:app_exp_details}
Table~\ref{tab:app_exp_settings} presents detailed descriptions and prompts for each of the nine experimental settings. Experiments with Llama-3.3, Qwen-3, and Phi-4 were conducted on a  NVIDIA A100 GPU with 80GB memory. 

\subsection{Results and Discussion}
\label{subsec:app_discuss}
\begin{figure}[t]
\centering
\begin{tabular}{lcc}
\hline
 & \fontsize{8}{8}\selectfont \textit{8.+Refine\_Cont.\textcircled{\tiny 2}}
 & \fontsize{8}{8}\selectfont\textit{9.+Refine\_Cont.\textcircled{\tiny 3}} \\ \hline
\fontsize{8}{8}\selectfont
  \begin{minipage}[l]{0.12\linewidth}  %
    \raggedright\fontsize{8}{9}\selectfont
    \includegraphics[height=1.4em]{fig_model_logo/microsoft_logo.png}\\
    Phi-4
  \end{minipage}
  &
  \begin{minipage}[c]{0.31\linewidth}  %
    \centering
    \includegraphics[width=\linewidth]{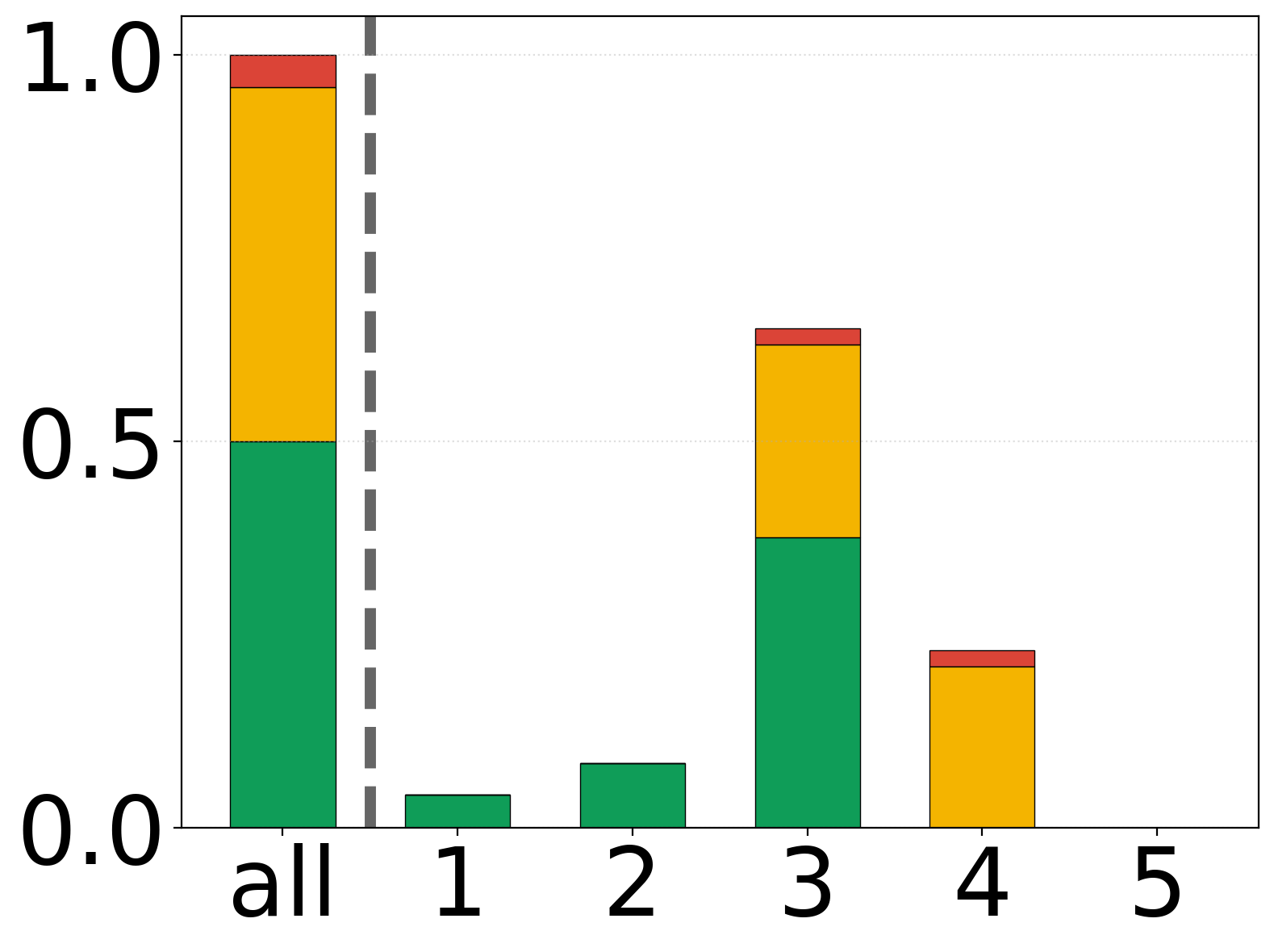}
  \end{minipage}
  &
  \begin{minipage}[c]{0.31\linewidth}  %
    \centering
    \includegraphics[width=\linewidth]{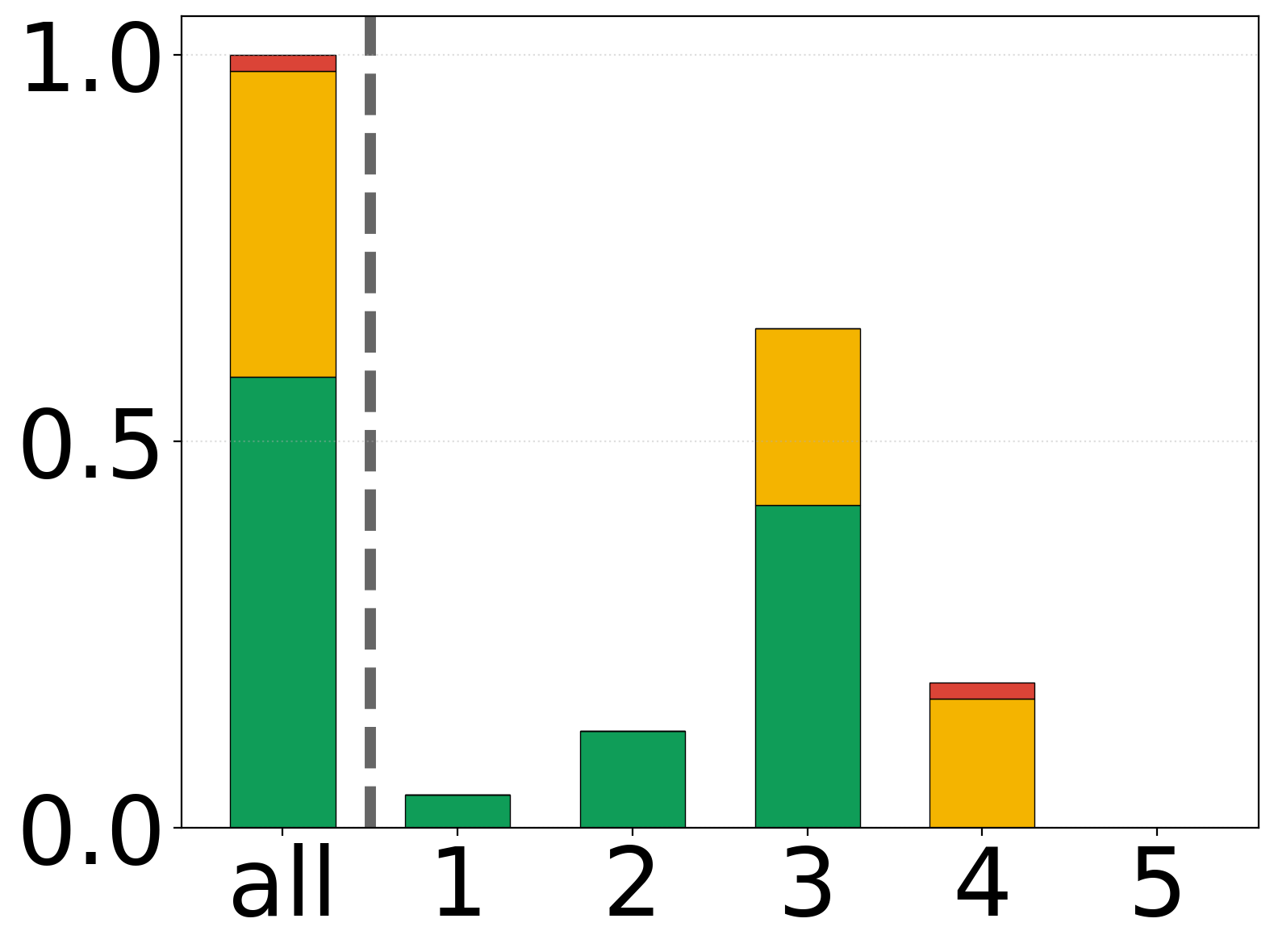}
  \end{minipage}
  \\[0.5em]
\fontsize{8}{8}\selectfont
  \begin{minipage}[l]{0.12\linewidth}  
    \raggedright\fontsize{8}{9}\selectfont
    \includegraphics[height=1.4em]{fig_model_logo/qwen_logo.png}\\
    Qwen3
  \end{minipage}
  &
  \begin{minipage}[c]{0.31\linewidth}  %
    \centering
    \includegraphics[width=\linewidth]{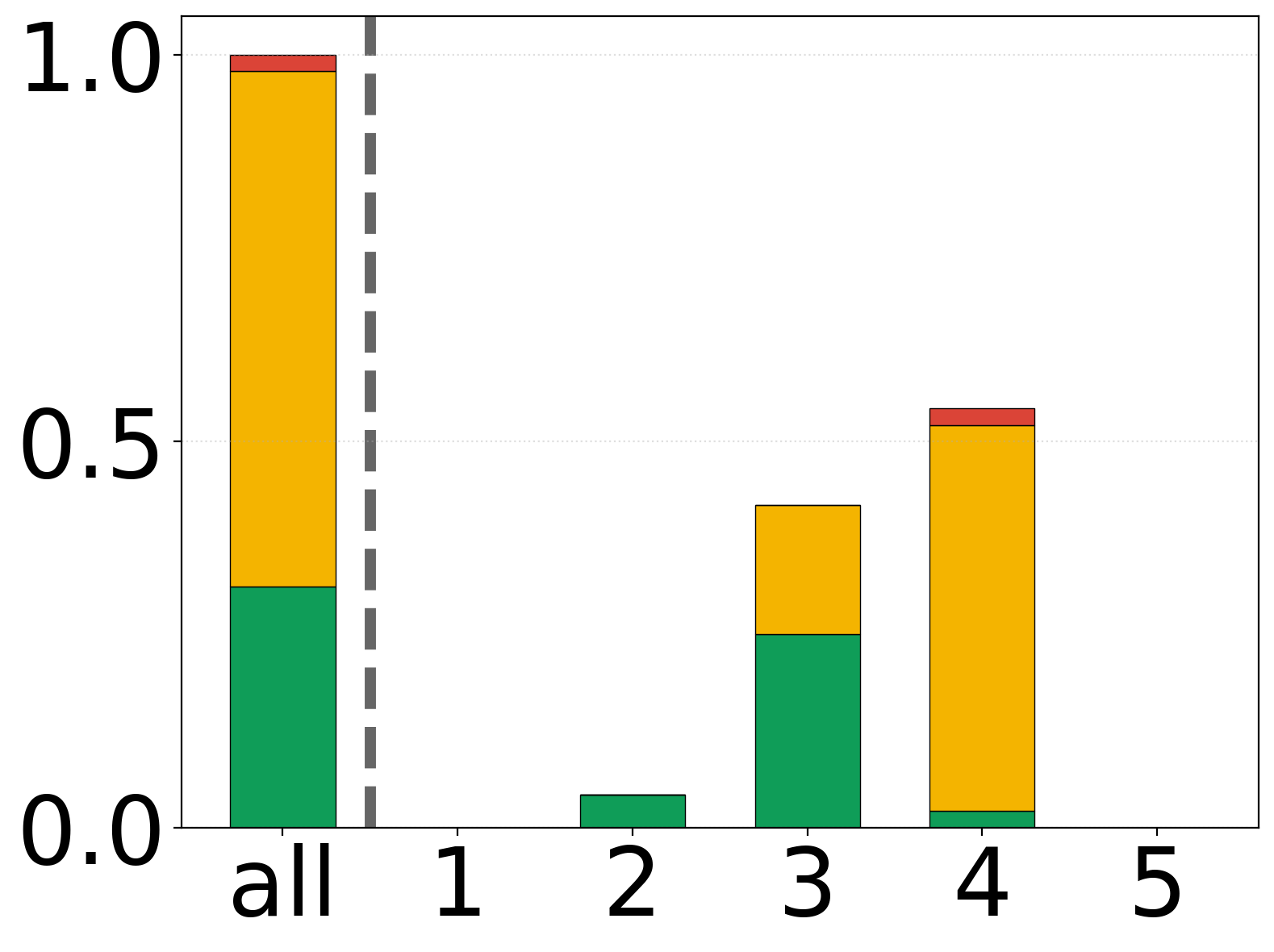}
  \end{minipage}
  &
  \begin{minipage}[c]{0.31\linewidth}  %
    \centering
    \includegraphics[width=\linewidth]{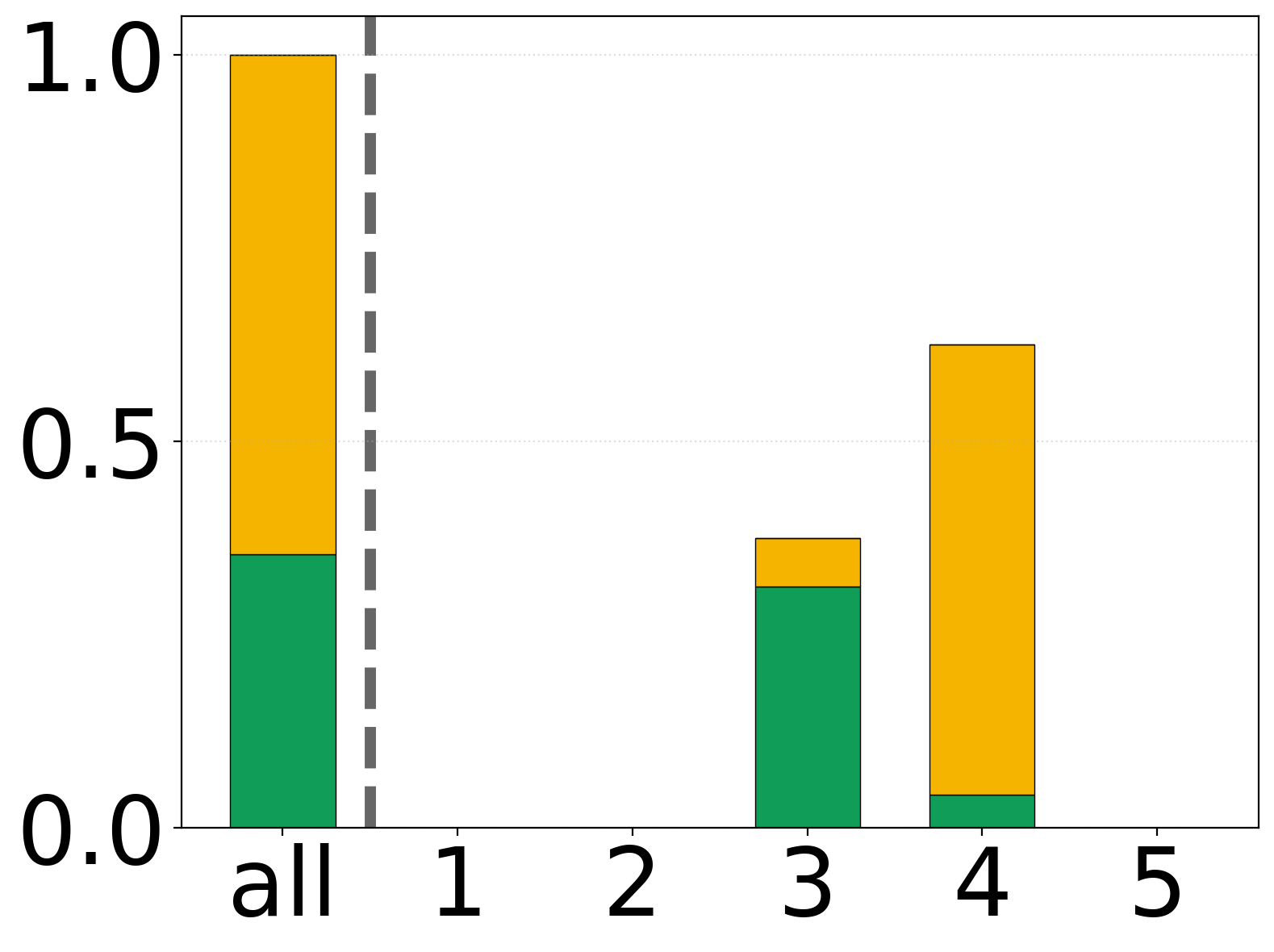}
  \end{minipage}
  \\[0.5em]
\fontsize{8}{8}\selectfont
  \begin{minipage}[l]{0.12\linewidth}  
    \raggedright\fontsize{8}{9}\selectfont
    \includegraphics[height=1.4em]{fig_model_logo/llama_logo.png}\\
    Llama-3.3
  \end{minipage}
  &
  \begin{minipage}[c]{0.31\linewidth}  %
    \centering
    \includegraphics[width=\linewidth]{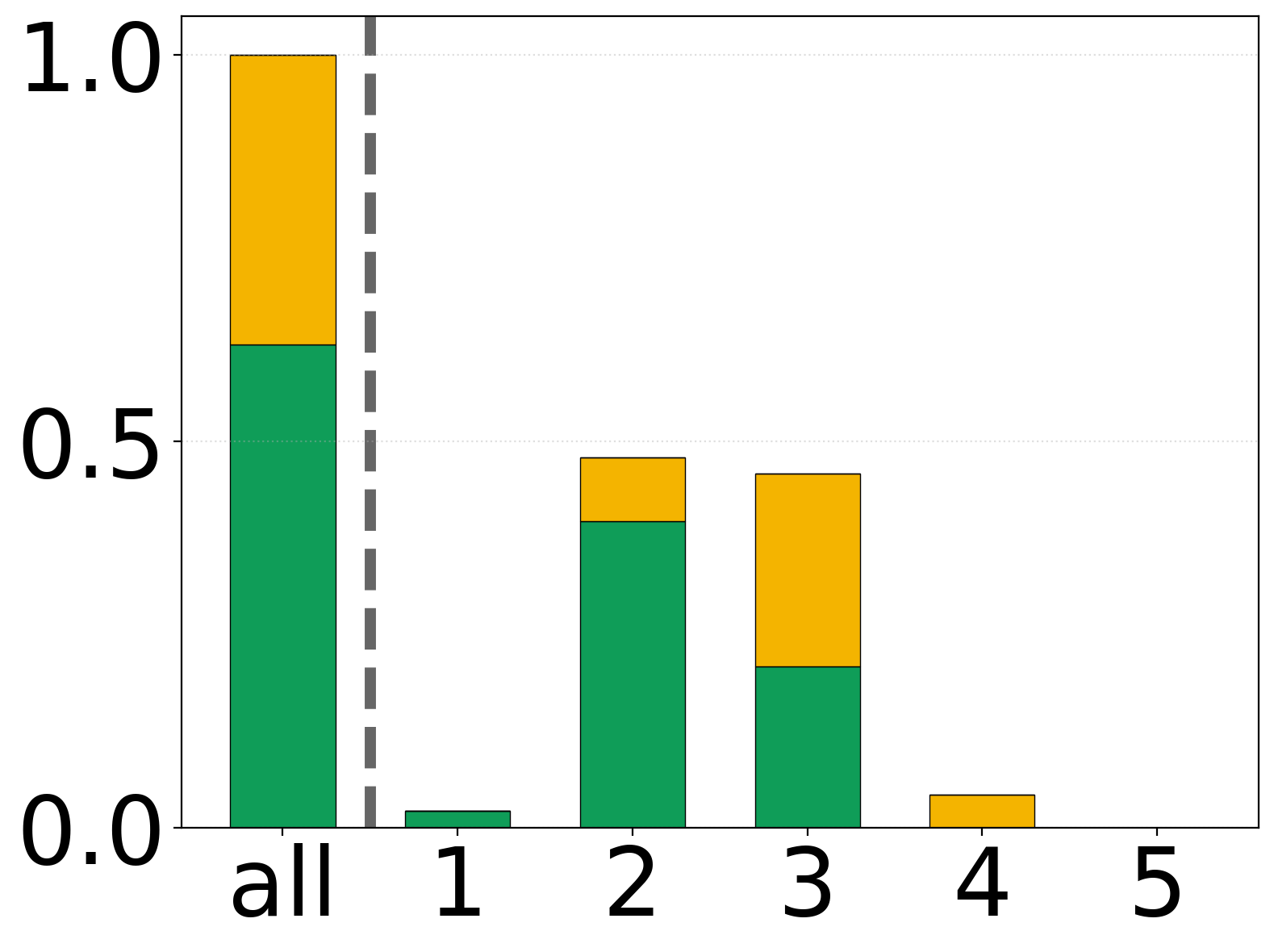}
  \end{minipage}
  &
  \begin{minipage}[c]{0.31\linewidth}  %
    \centering
    \includegraphics[width=\linewidth]{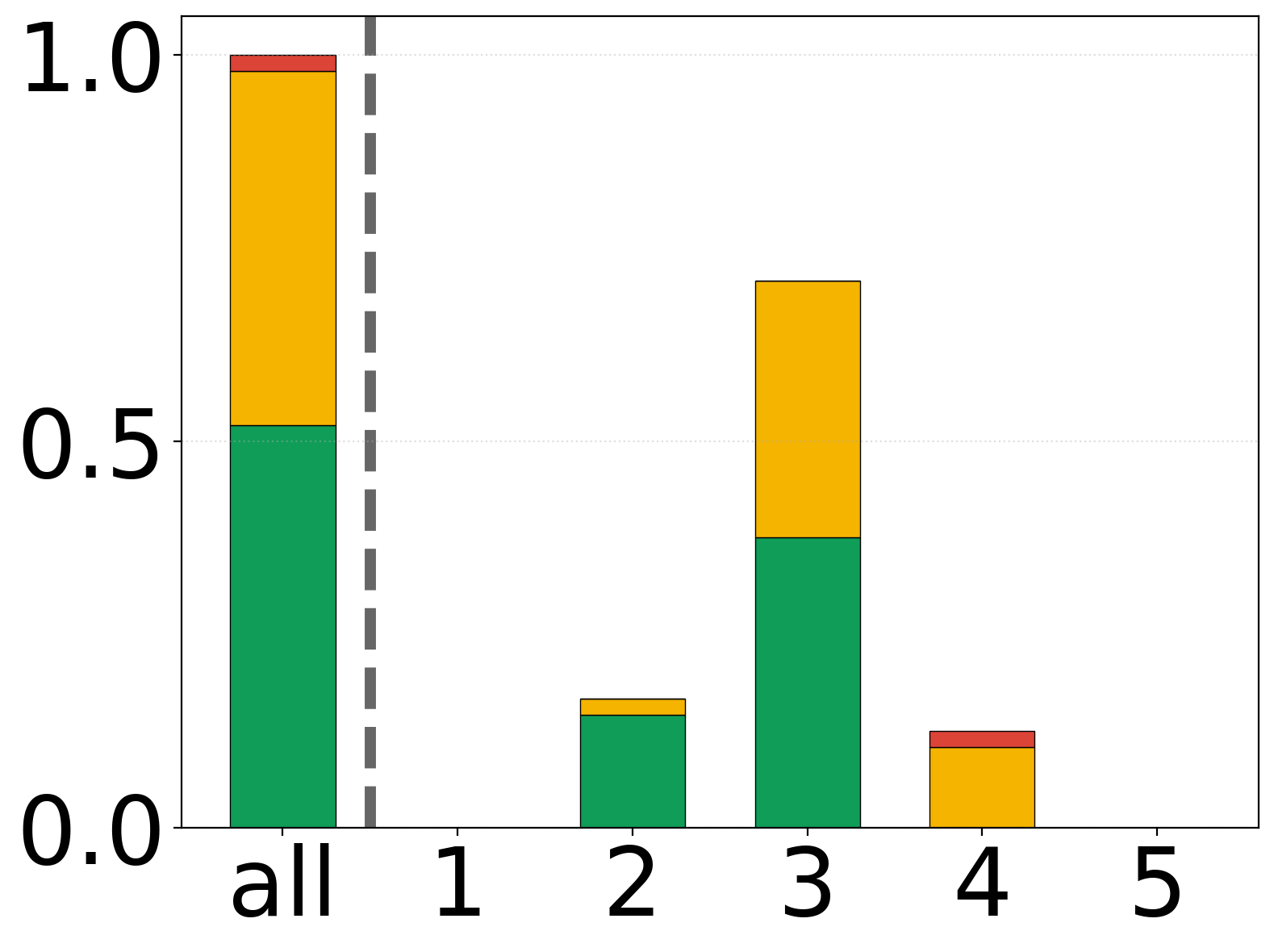}
  \end{minipage}
  \\[0.5em]

\fontsize{8}{8}\selectfont
  \begin{minipage}[l]{0.12\linewidth}  
    \raggedright\fontsize{8}{9}\selectfont
    \includegraphics[height=1.4em]{fig_model_logo/deepseek_logo.png}\\
    DeepSeek
  \end{minipage}
  &
  \begin{minipage}[c]{0.31\linewidth}  %
    \centering
    \includegraphics[width=\linewidth]{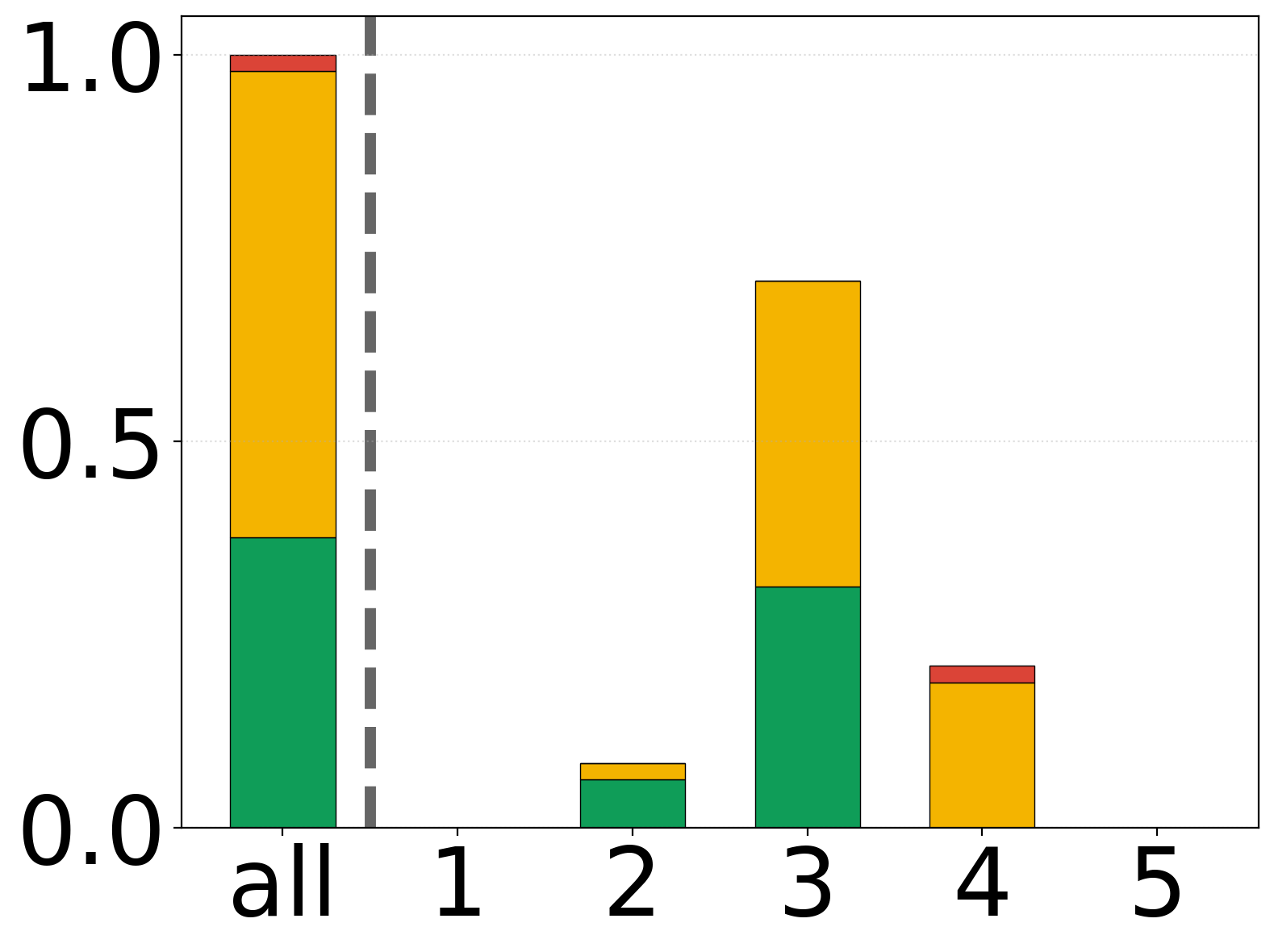}
  \end{minipage}
  &
  \begin{minipage}[c]{0.31\linewidth}  %
    \centering
    \includegraphics[width=\linewidth]{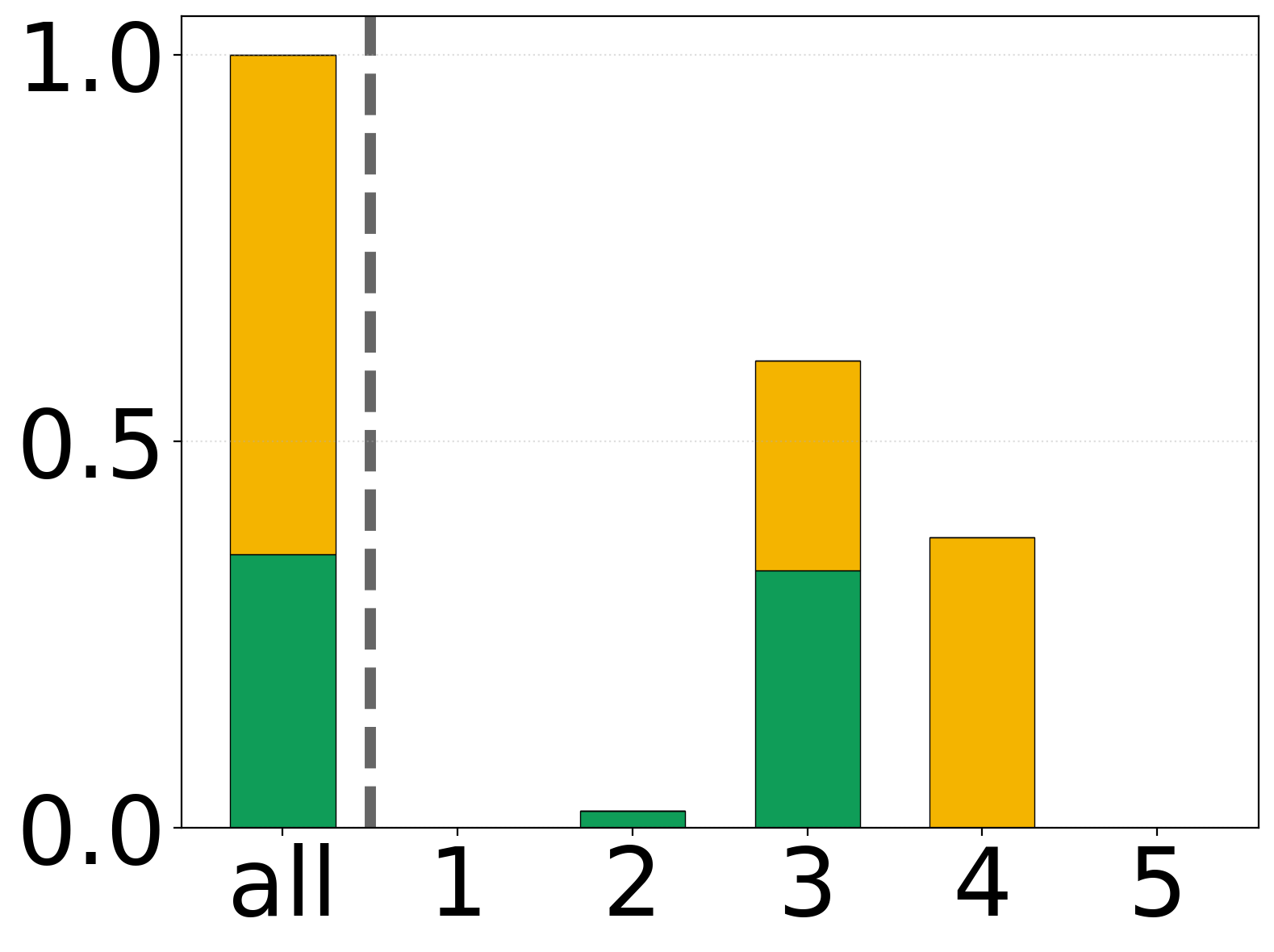}
  \end{minipage}
  \\[0.5em]

\fontsize{8}{8}\selectfont
  \begin{minipage}[l]{0.12\linewidth}  
    \raggedright\fontsize{8}{9}\selectfont
    \includegraphics[height=1.4em]{fig_model_logo/gpt_logo.jpg}\\
    GPT-4o
  \end{minipage}
  &
  \begin{minipage}[c]{0.31\linewidth}  %
    \centering
    \includegraphics[width=\linewidth]{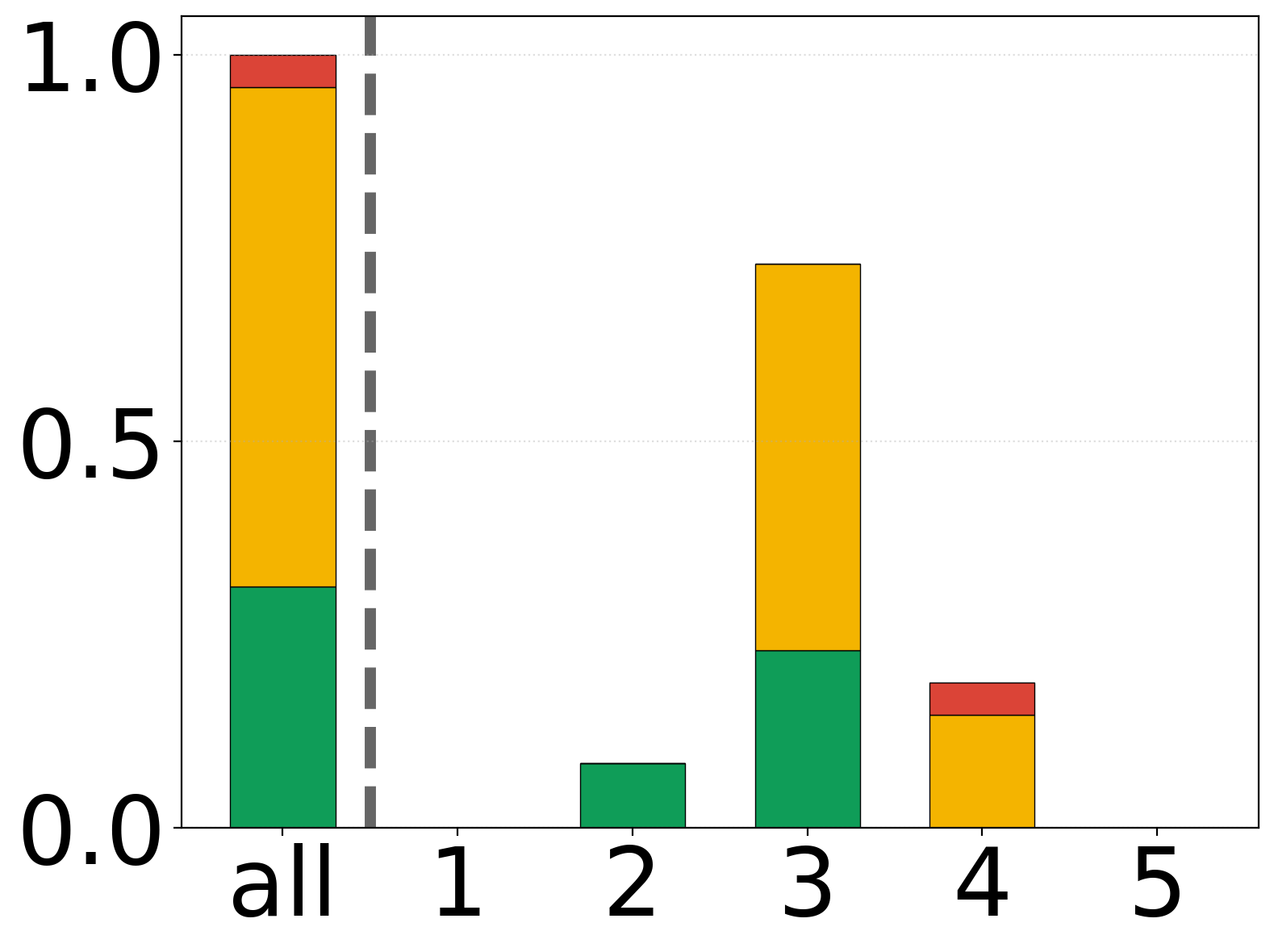}
  \end{minipage}
  &
  \begin{minipage}[c]{0.31\linewidth}  %
    \centering
    \includegraphics[width=\linewidth]{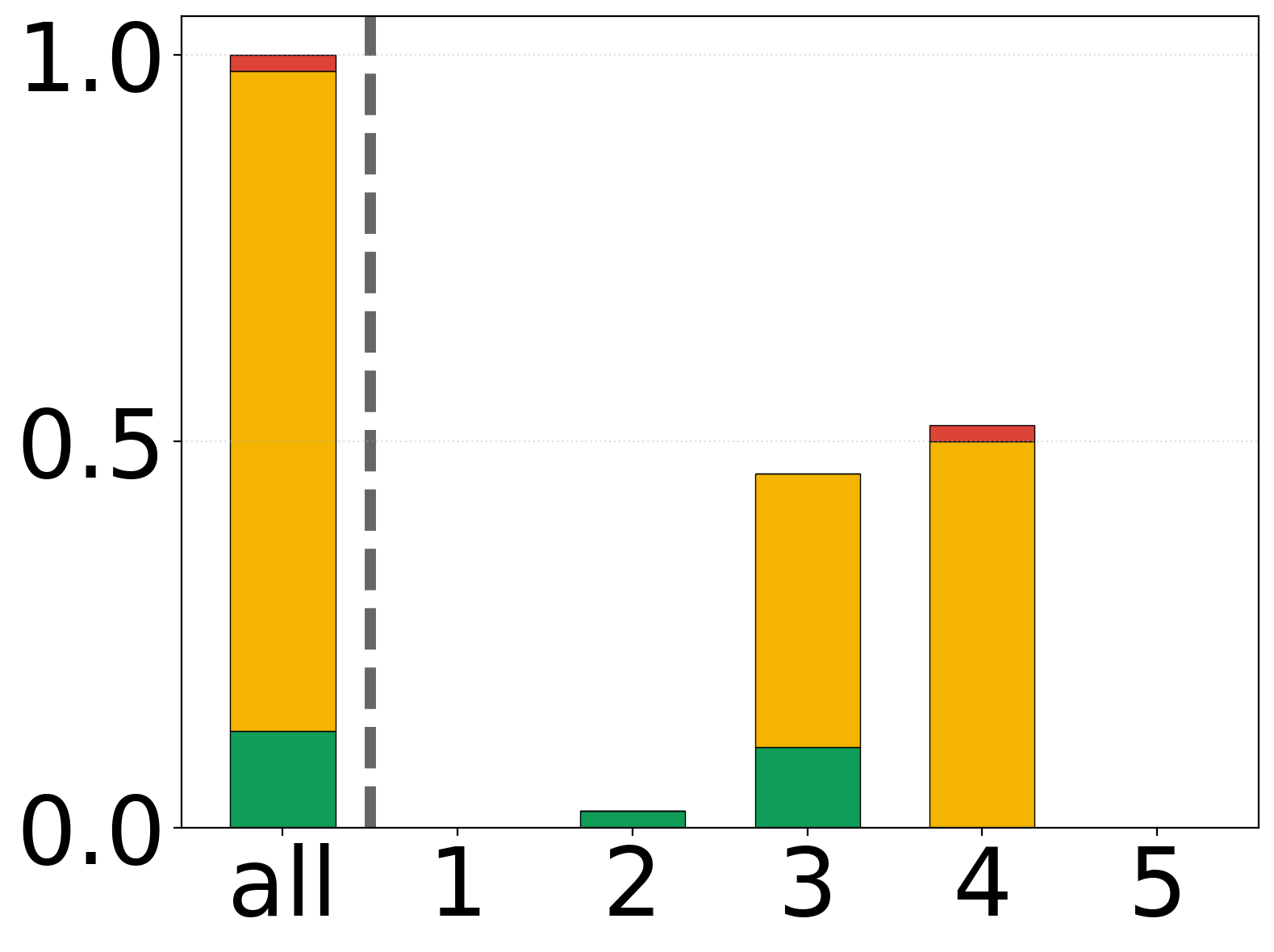}
  \end{minipage}
\\
\hline
\end{tabular}
\caption{Changes in \textit{Convincingness} after refinement across five LLMs. Colors indicate increase (green), no change (yellow), or decrease (red); the first bar shows overall proportions, followed by distributions by initial score.}
\label{fig:refine_conv}
\end{figure}

Table~\ref{tab:ARG} and Table~\ref{tab:ARG_TSP} report \textit{REspEval} evaluation results across five LLMs and nine settings, covering all metrics defined in §\ref{sec:respeval}. 
Table~\ref{tab:app_aix_sig} shows a consistent and significant positive impact of author input on response quality across all five LLMs.
Table~\ref{tab:ARG_refine}, Figure~\ref{fig:refine_spec} and Figure~\ref{fig:refine_conv} provide detailed analyses of refinement effectiveness.
Figure~\ref{fig:discourse_tone_stance} and Figure~\ref{fig:discourse_rel_pos} present discourse analyses comparing LLM-generated and authentic human responses in terms of tone--stance profiles and transition patterns, respectively.

\begin{table*}[ht]
\fontsize{8}{8}
\selectfont
\tabcolsep=0.025cm
\renewcommand{\arraystretch}{1.1}%
\begin{tabular}{ll|ll|ll}
\toprule
Exp. Setting && Description && Prompt\\\hline
\begin{tabular}{l} \textbf{1}.\\
 \textit{noAIx}\end{tabular} &&\begin{tabular}{l}ARG conditioned solely on \\the review segment without \\any additional inputs\end{tabular}&&\begin{tabular}{l}
\textcolor{blue}{L1.}You are a research assistant helping authors prepare an author response for a paper under peer review.\\ 
\textcolor{blue}{L2.}You will receive: \\
\textcolor{blue}{L3.} \quad - The reviewer's comment. \\
\textcolor{blue}{L4.}Your task is to write a specific and convincing response addressing the reviewer's comment.\\
\textcolor{blue}{L5.}\quad - The review comment is: <review segment>.\\ 
\textcolor{blue}{L6.}Output the response only. Do not include any other text.
\end{tabular} \\ \hline 
\begin{tabular}{l}\textbf{2}.\\\textit{wAIX}\\\_\textcircled{\tiny 1}\textit{S} \end{tabular}&&\begin{tabular}{l}ARG conditioned on the \\review segment and author \\input provided as a list of \\simple edit strings \end{tabular}&&\begin{tabular}{l}
\textcolor{blue}{L1.}You are a research assistant helping authors prepare an author response for a paper under peer review.\\ 
\textcolor{blue}{L2.}You will receive: \\
\textcolor{blue}{L3.}\quad    - The reviewer's comment. \\
\textcolor{blue}{L4.}\quad     - The author's additional input regarding the comment.\\ 
\textcolor{blue}{L5.}Your task is to write a specific and convincing response addressing the reviewer's comment.\\
\textcolor{blue}{L6.}\quad - The review comment is: <review segment>. \\
\textcolor{blue}{L7.}\quad - Refer to the author input below:  [<edit string>]\\
\textcolor{blue}{L8.}Output the response only. Do not include any other text.
\end{tabular} \\ \hline 
\begin{tabular}{l}\textbf{3}.\\\textit{wAIX}\\\_\textcircled{\tiny 2}\\\textit{+context}\end{tabular}&&\begin{tabular}{l}ARG conditioned on the \\review segment and author\\ input provided as a list of \\edit strings with paragraph 
\\context and section titles \end{tabular}&&\begin{tabular}{l}
Prompt from \textbf{Setting 2} with only modification in : \\\textcolor{blue}{L7.}- Refer to the author input below: [<<edit string> in <paragraph context> in Section <section title>>]
\end{tabular} \\ \hline 
\begin{tabular}{l}\textbf{4}.\\\textit{wAIX}\\\_\textcircled{\tiny 3}\\\textit{+v1}\end{tabular}&&\begin{tabular}{l}\textbf{Setting 3} with additional\\ \textbf{\textit{v1}} paper content, retrieved \\as the five most relevant \\paragraphs from the \\original submission \end{tabular}&&\begin{tabular}{l}
\textcolor{blue}{L1.}You are a research assistant helping authors prepare an author response for a paper under peer review.\\ 
\textcolor{blue}{L2.}You will receive: \\
\textcolor{blue}{L3.}\quad    - The reviewer's comment. \\
\textcolor{blue}{L4.}\quad     - The author's additional input regarding the comment.\\ 
\textcolor{blue}{L5.}Your task is to write a specific and convincing response addressing the reviewer's comment.\\
\textcolor{blue}{L6.}\quad - The review comment is: <review segment>. \\
\textcolor{blue}{L7.}\quad - Here are the top 5 paragraphs retrieved from the original paper: [<retrieved paragraph>] \\
\textcolor{blue}{L8.}\quad - Refer to the author input below: [<<edit string> in <paragraph context> in Section <section title>>]\\
\textcolor{blue}{L9.}Output the response only. Do not include any other text.
\end{tabular} \\ \hline 
\begin{tabular}{l}\textbf{5}.\textit{+Cont.}\\\_\textcircled{\tiny 1}\textit{lenC}\end{tabular}&&\begin{tabular}{l}\textbf{Setting 4} with additional \\generation length control \end{tabular}&&\begin{tabular}{l}
Prompt from \textbf{Setting 4} adds:\\
\textcolor{blue}{L10.}Please limit the response to NO MORE than <\textit{lenC}> words.
\end{tabular} \\ \hline 
\begin{tabular}{l}\textbf{6}.\\\textit{+Cont.}\\\_\textcircled{\tiny 2}\\\textit{lenC}\\\textit{\&planC}\end{tabular}&&\begin{tabular}{l}\textbf{Setting 4} with additional \\generation length control\\ and response plan control \end{tabular}&&\begin{tabular}{l}
\textcolor{blue}{L1.}You are a research assistant helping authors prepare an author response for a paper under peer review.\\ 
\textcolor{blue}{L2.}You will receive: \\
\textcolor{blue}{L3.}\quad    - The reviewer's comment. And extracted items from the review comment, including questions, \\\quad\quad\quad\quad criticisms and requests.  \\
\textcolor{blue}{L4.}\quad     - The author's additional input regarding the comment.\\ 
\textcolor{blue}{L5.}Your task is to write a clear and convincing response addressing the reviewer's comment and the \\items. Make the response coherent, fluent and human-like, without necessarily listing the items. \\
Write a response addressing the review comment and the items based on the given response action plan.\\
\textcolor{blue}{L6.}\quad- The review comment is: <review segment>. \\
\quad \quad\quad \quad - - The items extracted from the review comment are: \\ \quad \quad\quad \quad \quad \quad- - - questions: [<\#<id>: <question>>] \\\quad \quad\quad \quad \quad \quad - - - criticisms: [<\#<id>: <criticism>>]\\\quad \quad\quad \quad \quad \quad - - - requests: [<\#<id>: <request>>] \\
\quad \quad \quad \quad- - The response action plan is: \\ \quad \quad\quad \quad \quad \quad- - - questions: [<\#<id>: <\textit{planC} for question>>] \\\quad \quad\quad \quad \quad \quad - - - criticisms: [<\#<id>: <\textit{planC} for criticism>>]\\\quad \quad\quad \quad \quad \quad - - - requests: [<\#<id>: <\textit{planC} for request>>] \\
\textcolor{blue}{L7.}\quad - Here are the top 5 paragraphs retrieved from the original paper: [<retrieved paragraph>] \\
\textcolor{blue}{L8.}\quad - Refer to the author input below: [<<edit string> in <paragraph context> in Section <section title>>]\\
\textcolor{blue}{L9.}Output the response only. Do not include any other text.\\
\textcolor{blue}{L10.}Please limit the response to NO MORE than <\textit{lenC}> words.
\end{tabular} \\ \hline 
\begin{tabular}{l}\textbf{7}.\textit{+Cont.}\\\_\textcircled{\tiny 3}\textit{planC}\end{tabular}&&\begin{tabular}{l}\textbf{Setting 4} with additional \\response plan control \end{tabular}&&\begin{tabular}{l}
Prompt from \textbf{Setting 6} removes \textcolor{blue}{L10.}
\end{tabular} \\ \hline 
\begin{tabular}{l}\textbf{8}.\\\textit{+Refine}\\\_\textit{Cont.}\textcircled{\tiny 2}\end{tabular}&&\begin{tabular}{l}Refinement of the response \\generated under \textbf{Setting 6},\\ using its quality and \\factuality evaluations\\ accompanied by justifica-\\tions and improvement \\suggestions\end{tabular}&&\begin{tabular}{l}
Prompt from \textbf{Setting 6} adds: \\
\textcolor{blue}{L11.}Note: This is a refinement round to improve the quality of the previous generated response \\\quad\quad based on its evaluation results.\\
\textcolor{blue}{L12.}\quad - The previous response generated is: <previous response from Setting 6>. \\
\textcolor{blue}{L13.}\quad - The overall response scores (directness, specificity and convincingness, 5-point scale) and the \\\quad\quad\quad respective justifications and improvement suggestions: <quality evaluation of previous response>\\
\textcolor{blue}{L14.}\quad - Factuality score: <GFP score\%> of the atomic facts in the previous response are supported by \\\quad\quad\quad the provided inputs.\\
\textcolor{blue}{L15.}TASK: Please revise the previous response based on the review comment, the provided inputs and \\\quad\quad the requirements, as well as the evaluation results above to improve the directness, specificity, \\\quad\quad convincingness and the factuality of the response. Output the revised response only.

\end{tabular} \\ \hline 
\begin{tabular}{l}\textbf{9}.\\\textit{+Refine}\\\_\textit{Cont.}\textcircled{\tiny 3}\end{tabular}&&\begin{tabular}{l}Refinement of the response \\generated under \textbf{Setting 7},\\ using the same evaluation \\ results as \textbf{Setting 8} \end{tabular}&&\begin{tabular}{l}
Prompt from \textbf{Setting 8} removes \textcolor{blue}{L10.}
\end{tabular} \\ \bottomrule
\end{tabular}
\caption{Author response generation with \textit{REspGen}. Shown are the nine experimental settings, their descriptions, and the prompts used. To explicitly capture the model’s awareness of missing information, we also prompt: \textit{Use placeholders like '[author info: <description>]' if you need extra information from the author to address the review comment.} For the EMNLP24 subset, we additionally clarify the review setting by appending the prompt: \textit{This author response is prepared during the rebuttal phase, before submitting any revisions (like in ARR process). You should use the additional author input to address the review comment if they are useful, and may outline future planned changes in the final version if relevant but do not refer to completed revisions}.}
\label{tab:app_exp_settings}
\end{table*}

\begin{table*}[th]
\fontsize{8}{8}
\selectfont
\centering
\renewcommand{\arraystretch}{1.1} 
\tabcolsep=0.04cm
\begin{tabular}[]{lll|lll|ll|lll|llllllll}
\hline
Metric&&&\multicolumn{2}{c}{Basic}&&\multicolumn{1}{c}{Polite.}&&\multicolumn{2}{c}{Meta}&&\multicolumn{7}{c}{Tone-Stance Profile}\\ \hline

&&& RL&BS &&&&
\#word & \%Ph 
&& \%Coop & \%Defe & \%Hed & \%Soc & \%Other & ArgLoad 
\\ \hline

Human
&&& / &/ &&.829&&
\begin{tabular}{l}115\end{tabular}&\begin{tabular}{l}-\end{tabular}
&&\begin{tabular}{l}.454\end{tabular}&\begin{tabular}{l}.048\end{tabular}
&\begin{tabular}{l}.277\end{tabular}&\begin{tabular}{l}.076\end{tabular}
&\begin{tabular}{l}.145\end{tabular}&\begin{tabular}{l}.779\end{tabular}
\\ \hline
\begin{tabular}{l}\includegraphics[height=1.4em]{fig_model_logo/microsoft_logo.png}\hspace{0.5em}\\Phi-4\end{tabular}
&\begin{tabular}{l}\textit{1.noAIx}\\\textit{2.wAIx\_\textcircled{\tiny 1}\textit{S}}\\\textit{3.wAIx\_\textcircled{\tiny 2}\textit{+context}}\\\textit{4.wAIx\_\textcircled{\tiny 3}\textit{+v1}}\\\textit{5.+Cont.\_\textcircled{\tiny 1}\textit{lenC}}\\\textit{6.+Cont.\_\textcircled{\tiny 2}\textit{lenC\&planC}}\\\textit{7.+Cont.\_\textcircled{\tiny 3}\textit{planC}}\\\textit{8.+Refine\_Cont.\textcircled{\tiny 2}}\\\textit{9.+Refine\_Cont.\textcircled{\tiny 3}}\end{tabular}
&
&\begin{tabular}{l}.140\\\textbf{.158}\\.136\\.144\\.144\\.155\\.155\\.133\\.131\end{tabular}
&\begin{tabular}{l}.834\\\textbf{.835}\\.826\\.825\\.761\\.829\\.829\\.820\\.820\end{tabular}&
&\begin{tabular}{l}\textbf{.850}\\.798\\.772\\.761\\.761\\.798\\.798\\.777\\.772\end{tabular}&

&\begin{tabular}{l}161\\127\\\cellcolor{lightgreen}\textbf{428}\\343\\343\\284\\284\\312\\\cellcolor{lightgreen2}368 \end{tabular}
&\begin{tabular}{l}0\\0\\0\\0\\0\\0\\0\\0\\0 \end{tabular}&

&\begin{tabular}{l}\textbf{.517}\\.443\\.409\\.446\\.444\\.445\\.434\\.484\\.451\end{tabular}
&\begin{tabular}{l}.018\\.019\\\cellcolor{lightgreen2}\textbf{.062}\\.037\\.034\\.029\\.026\\.021\\.017\end{tabular}
&\begin{tabular}{l}.194\\.245\\.221\\.199\\.222\\.220\\\textbf{.247}\\.197\\.224\end{tabular}
&\begin{tabular}{l}\textbf{.115}\\.065\\.077\\.034\\.054\\.078\\.077\\.058\\.058\end{tabular}
&\begin{tabular}{l}.136\\.227\\.212\\\cellcolor{lightgreen}\textbf{.284}\\\cellcolor{lightgreen3}.246\\.228\\.216\\.240\\.250\end{tabular}
&\begin{tabular}{l}\textbf{.728}\\.708\\.691\\.682\\.699\\.694\\.708\\.702\\.692\end{tabular}

\\ \hline
\begin{tabular}{l}\includegraphics[height=1.4em]{fig_model_logo/qwen_logo.png}\hspace{0.5em}\\Qwen3\end{tabular}
&\begin{tabular}{l}\textit{1.noAIx}\\\textit{2.wAIx\_\textcircled{\tiny 1}\textit{S}}\\\textit{3.wAIx\_\textcircled{\tiny 2}\textit{+context}}\\\textit{4.wAIx\_\textcircled{\tiny 3}\textit{+v1}}\\\textit{5.+Cont.\_\textcircled{\tiny 1}\textit{lenC}}\\\textit{6.+Cont.\_\textcircled{\tiny 2}\textit{lenC\&planC}}\\\textit{7.+Cont.\_\textcircled{\tiny 3}\textit{planC}}\\\textit{8.+Refine\_Cont.\textcircled{\tiny 2}}\\\textit{9.+Refine\_Cont.\textcircled{\tiny 3}}\end{tabular}
&
&\begin{tabular}{l}.149\\.158\\.173\\.166\\.171\\\textbf{.178}\\.171\\.163\\.146\end{tabular}
&\begin{tabular}{l}.833\\.835\\.840\\.839\\.843\\\textbf{.846}\\.842\\.838\\.829\end{tabular}&
&\begin{tabular}{l}\textbf{.841}\\.798\\.808\\.797\\.807\\.825\\.823\\.797\\.782\end{tabular}&

&\begin{tabular}{l}123\\127\\164\\205\\125\\130\\216\\142\\\textbf{290} \end{tabular}
&\begin{tabular}{l}\cellcolor{lightgreen}\textbf{95.8}\\0\\12.5\\4.2\\25.0\\4.2\\2.1\\2.1\\0\end{tabular}&

&\begin{tabular}{l}.499\\.443\\.492\\.468\\.440\\.462\\.448\\\textbf{.520}\\.508\end{tabular}
&\begin{tabular}{l}.035\\.019\\.045\\.036\\\textbf{.054}\\.040\\.039\\.046\\.028\end{tabular}
&\begin{tabular}{l}.205\\.245\\.237\\\cellcolor{lightgreen}\textbf{.263}\\\cellcolor{lightgreen2}.258\\.230\\.251\\.241\\.207\end{tabular}
&\begin{tabular}{l}\textbf{.110}\\.065\\.095\\.072\\.075\\.082\\.089\\.059\\.065\end{tabular}
&\begin{tabular}{l}.150\\\textbf{.227}\\.131\\.161\\.173\\.187\\.174\\.135\\.192\end{tabular}
&\begin{tabular}{l}.740\\.708\\.773\\.767\\.752\\.732\\.738\\\textbf{.807}\\.743\end{tabular}

\\ \hline
\begin{tabular}{l}\includegraphics[height=1.4em]{fig_model_logo/llama_logo.png}\hspace{0.5em}\\Llama-3.3\end{tabular}
&\begin{tabular}{l}\textit{1.noAIx}\\\textit{2.wAIx\_\textcircled{\tiny 1}\textit{S}}\\\textit{3.wAIx\_\textcircled{\tiny 2}\textit{+context}}\\\textit{4.wAIx\_\textcircled{\tiny 3}\textit{+v1}}\\\textit{5.+Cont.\_\textcircled{\tiny 1}\textit{lenC}}\\\textit{6.+Cont.\_\textcircled{\tiny 2}\textit{lenC\&planC}}\\\textit{7.+Cont.\_\textcircled{\tiny 3}\textit{planC}}\\\textit{8.+Refine\_Cont.\textcircled{\tiny 2}}\\\textit{9.+Refine\_Cont.\textcircled{\tiny 3}}\end{tabular}
&
&\begin{tabular}{l}.175\\.200\\\cellcolor{lightgreen2}.206\\\cellcolor{lightgreen}\textbf{.207}\\.203\\\cellcolor{lightgreen2}.206\\.196\\.198\\.166\end{tabular}
&\begin{tabular}{l}.839\\.847\\.848\\.849\\\cellcolor{lightgreen2}\textbf{.853}\\\cellcolor{lightgreen3}.852\\.848\\.848\\.838\end{tabular}&
&\begin{tabular}{l}\cellcolor{lightgreen3}.873\\\cellcolor{lightgreen2}\textbf{.876}\\.863\\.853\\.843\\.865\\.860\\.830\\.830\end{tabular}&

&\begin{tabular}{l}126\\169\\183\\198\\82\\82\\214\\125\\\textbf{304} \end{tabular}
&\begin{tabular}{l}\cellcolor{lightgreen2}\textbf{83.3}\\6.3\\4.2\\0\\12.5\\4.2\\0\\2.1\\2.1\end{tabular}&

&\begin{tabular}{l}.524\\.466\\.432\\.427\\.529\\\cellcolor{lightgreen2}.567\\.468\\\cellcolor{lightgreen}\textbf{.595}\\.475\end{tabular}
&\begin{tabular}{l}.026\\.025\\.026\\.026\\.034\\\textbf{.050}\\.026\\.027\\.018\end{tabular}
&\begin{tabular}{l}.159\\.208\\.230\\\textbf{.241}\\.216\\.220\\.229\\.198\\.211\end{tabular}
&\begin{tabular}{l}\textbf{.115}\\.098\\.084\\.085\\.075\\.069\\.081\\.047\\.053\end{tabular}
&\begin{tabular}{l}.176\\.203\\.229\\.220\\.147\\.094\\.196\\.132\\\textbf{.243}\end{tabular}
&\begin{tabular}{l}.709\\.699\\.688\\.695\\.779\\\cellcolor{lightgreen2}\textbf{.837}\\.723\\.821\\.704\end{tabular}

\\ \hline
\begin{tabular}{l}\includegraphics[height=1.4em]{fig_model_logo/deepseek_logo.png}\hspace{0.5em}\\DeepSeek\end{tabular}
&\begin{tabular}{l}\textit{1.noAIx}\\\textit{2.wAIx\_\textcircled{\tiny 1}\textit{S}}\\\textit{3.wAIx\_\textcircled{\tiny 2}\textit{+context}}\\\textit{4.wAIx\_\textcircled{\tiny 3}\textit{+v1}}\\\textit{5.+Cont.\_\textcircled{\tiny 1}\textit{lenC}}\\\textit{6.+Cont.\_\textcircled{\tiny 2}\textit{lenC\&planC}}\\\textit{7.+Cont.\_\textcircled{\tiny 3}\textit{planC}}\\\textit{8.+Refine\_Cont.\textcircled{\tiny 2}}\\\textit{9.+Refine\_Cont.\textcircled{\tiny 3}}\end{tabular}
&
&\begin{tabular}{l}.171\\.198\\.199\\.192\\\textbf{.202}\\.201\\.197\\.190\\.185\end{tabular}
&\begin{tabular}{l}.838\\.845\\.844\\.846\\.850\\\cellcolor{lightgreen3}\textbf{.852}\\.850\\.847\\.844\end{tabular}&
&\begin{tabular}{l}\textbf{.870}\\.843\\.855\\.835\\.824\\.863\\.863\\.825\\.831\end{tabular}&

&\begin{tabular}{l}113\\154\\172\\194\\96\\93\\179\\97\\\textbf{194} \end{tabular}
&\begin{tabular}{l}\cellcolor{lightgreen2}\textbf{83.3}\\22.9\\25.0\\14.6\\14.6\\2.1\\2.1\\2.1\\8.3\end{tabular}&

&\begin{tabular}{l}\cellcolor{lightgreen3}\textbf{.565}\\.449\\.429\\.439\\.428\\.459\\.437\\.529\\.528\end{tabular}
&\begin{tabular}{l}.026\\.021\\.018\\.043\\.051\\.050\\.048\\\cellcolor{lightgreen}\textbf{.068}\\.057\end{tabular}
&\begin{tabular}{l}.179\\.219\\.214\\.232\\.222\\\cellcolor{lightgreen}\textbf{.263}\\.245\\\cellcolor{lightgreen3}.253\\.242\end{tabular}
&\begin{tabular}{l}\textbf{.126}\\.095\\.090\\.077\\.081\\.099\\.100\\.087\\.085\end{tabular}
&\begin{tabular}{l}.104\\.218\\\cellcolor{lightgreen2}\textbf{.249}\\.210\\.219\\.129\\.171\\.063\\.088\end{tabular}
&\begin{tabular}{l}.769\\.688\\.661\\.714\\.700\\.772\\.730\\\cellcolor{lightgreen}\textbf{.850}\\\cellcolor{lightgreen3}.827\end{tabular}

\\ \hline
\begin{tabular}{l}\includegraphics[height=1.4em]{fig_model_logo/gpt_logo.jpg}\hspace{0.5em}\\GPT-4o\end{tabular}
&\begin{tabular}{l}\textit{1.noAIx}\\\textit{2.wAIx\_\textcircled{\tiny 1}\textit{S}}\\\textit{3.wAIx\_\textcircled{\tiny 2}\textit{+context}}\\\textit{4.wAIx\_\textcircled{\tiny 3}\textit{+v1}}\\\textit{5.+Cont.\_\textcircled{\tiny 1}\textit{lenC}}\\\textit{6.+Cont.\_\textcircled{\tiny 2}\textit{lenC\&planC}}\\\textit{7.+Cont.\_\textcircled{\tiny 3}\textit{planC}}\\\textit{8.+Refine\_Cont.\textcircled{\tiny 2}}\\\textit{9.+Refine\_Cont.\textcircled{\tiny 3}}\end{tabular}
&
&\begin{tabular}{l}.159\\.172\\.157\\.162\\.200\\\cellcolor{lightgreen3}\textbf{.204}\\.166\\.192\\.155\end{tabular}
&\begin{tabular}{l}.833\\.839\\.836\\.837\\\cellcolor{lightgreen3}.852\\\cellcolor{lightgreen}\textbf{.854}\\.840\\.850\\.837\end{tabular}&
&\begin{tabular}{l}\cellcolor{lightgreen}\textbf{.878}\\.859\\.839\\.829\\.864\\.871\\.848\\.849\\.825\end{tabular}&

&\begin{tabular}{l}247\\265\\311\\339\\158\\156\\336\\163\\\cellcolor{lightgreen3}\textbf{367} \end{tabular}
&\begin{tabular}{l}\cellcolor{lightgreen3}\textbf{54.2}\\14.6\\8.3\\4.2\\0\\0\\0\\16.7\\8.3\end{tabular}&

&\begin{tabular}{l}.499\\.475\\.478\\\textbf{.514}\\.482\\.446\\.473\\.473\\.509\end{tabular}
&\begin{tabular}{l}.016\\.019\\.041\\\cellcolor{lightgreen3}\textbf{.061}\\.011\\.042\\.027\\.024\\.030\end{tabular}
&\begin{tabular}{l}.213\\.186\\.143\\.186\\.211\\.227\\.221\\\textbf{.229}\\.205\end{tabular}
&\begin{tabular}{l}\cellcolor{lightgreen}\textbf{.148}\\.117\\.094\\.107\\.118\\\cellcolor{lightgreen3}.127\\.111\\\cellcolor{lightgreen2}.133\\.079\end{tabular}
&\begin{tabular}{l}.125\\.183\\\textbf{.244}\\.132\\.179\\.159\\.168\\.142\\.178\end{tabular}
&\begin{tabular}{l}.728\\.680\\.662\\\textbf{.761}\\.704\\.715\\.721\\.725\\.744\end{tabular}
\\ \hline
\end{tabular}
\caption[LLMs]{Evaluation results for the ARG task across five LLMs and nine settings.
Reported metrics include basic similarity measures (Rouge-L (RL), BERTScore(BS)), average sentence politeness (calculated using \url{https://huggingface.co/Genius1237/xlm-roberta-large-tydip}), metadata including word count (\#word) and the proportion of samples with placeholders (\%Ph), as well as the Tone-Stance Profile (\textit{\%Coop}, \textit{\%Defe}, \textit{\%Hed}, \textit{\%Soc}, \textit{\%Other}, \textit{ArgLoad}), see §\ref{sec:respeval} for definitions. All scores and proportions are normalized to [0,1]. For each LLM, the largest value per metric is bolded, and the top three values across LLMs are highlighted in green. 
Settings 1-9: no author input (1), rough edit string as author input (2), add paragraph context (3), with paper retrieval (4), plus length control (5), plan control (7), or combined controls (6), refinement on 6 and 7 (8–9). Full descriptions in Table~\ref{tab:app_exp_settings}.}
\label{tab:ARG_TSP}
\end{table*}

\begin{table}[t]
\fontsize{8}{8}
\selectfont
\centering
\renewcommand{\arraystretch}{1.2} 
\tabcolsep=0.05cm
\begin{tabular}[t]{lll|lll|lll|lll}
\hline
\multicolumn{2}{l}{Metric} &&\multicolumn{2}{c}{Targeting}&&\multicolumn{2}{c}{Specificity}&&\multicolumn{2}{c}{Convincingness}\\\hline
\multicolumn{2}{l}{LLM/Setting} &&t-test&Wilco&&t-test&Wilco&&t-test&Wilco
\\ \hline
\begin{tabular}{l}\includegraphics[height=1.4em]{fig_model_logo/microsoft_logo.png}\hspace{0.5em}\\Phi-4\end{tabular}
&\begin{tabular}{r}1 vs. 2\\1 vs. 3\\1 vs. 4\end{tabular}
&&\begin{tabular}{l}\xmark\\\xmark\\\xmark\end{tabular}
&\begin{tabular}{l}\cmark\\\xmark\\\xmark\end{tabular}
&&\begin{tabular}{l}\cmark\\\cmark\\\cmark\end{tabular}
&\begin{tabular}{l}\cmark\\\cmark\\\cmark\end{tabular}
&&\begin{tabular}{l}\cmark\\\cmark\\\cmark\end{tabular}
&\begin{tabular}{l}\cmark\\\cmark\\\cmark\end{tabular}
\\ \hline
\begin{tabular}{l}\includegraphics[height=1.4em]{fig_model_logo/qwen_logo.png}\hspace{0.5em}\\Qwen3\end{tabular}
&\begin{tabular}{r}1 vs. 2\\1 vs. 3\\1 vs. 4\end{tabular}
&&\begin{tabular}{l}\cmark\\\cmark\\\cmark\end{tabular}
&\begin{tabular}{l}\cmark\\\cmark\\\cmark\end{tabular}
&&\begin{tabular}{l}\cmark\\\cmark\\\cmark\end{tabular}
&\begin{tabular}{l}\cmark\\\cmark\\\cmark\end{tabular}
&&\begin{tabular}{l}\cmark\\\cmark\\\cmark\end{tabular}
&\begin{tabular}{l}\cmark\\\cmark\\\cmark\end{tabular}
\\ \hline
\begin{tabular}{l}\includegraphics[height=1.4em]{fig_model_logo/llama_logo.png}\hspace{0.5em}\\Llama-3.3\end{tabular}
&\begin{tabular}{r}1 vs. 2\\1 vs. 3\\1 vs. 4\end{tabular}
&&\begin{tabular}{l}\xmark\\\cmark\\\cmark\end{tabular}
&\begin{tabular}{l}\xmark\\\cmark\\\cmark\end{tabular}
&&\begin{tabular}{l}\cmark\\\cmark\\\cmark\end{tabular}
&\begin{tabular}{l}\cmark\\\cmark\\\cmark\end{tabular}
&&\begin{tabular}{l}\cmark\\\cmark\\\cmark\end{tabular}
&\begin{tabular}{l}\cmark\\\cmark\\\cmark\end{tabular}
\\ \hline
\begin{tabular}{l}\includegraphics[height=1.4em]{fig_model_logo/deepseek_logo.png}\hspace{0.5em}\\DeepSeek\end{tabular}
&\begin{tabular}{r}1 vs. 2\\1 vs. 3\\1 vs. 4\end{tabular}
&&\begin{tabular}{l}\cmark\\\xmark\\\cmark\end{tabular}
&\begin{tabular}{l}\cmark\\\cmark\\\cmark\end{tabular}
&&\begin{tabular}{l}\cmark\\\cmark\\\cmark\end{tabular}
&\begin{tabular}{l}\cmark\\\cmark\\\cmark\end{tabular}
&&\begin{tabular}{l}\cmark\\\cmark\\\cmark\end{tabular}
&\begin{tabular}{l}\cmark\\\cmark\\\cmark\end{tabular}
\\ \hline
\begin{tabular}{l}\includegraphics[height=1.4em]{fig_model_logo/gpt_logo.jpg}\hspace{0.5em}\\GPT-4o\end{tabular}
&\begin{tabular}{r}1 vs. 2\\1 vs. 3\\1 vs. 4\end{tabular}
&&\begin{tabular}{l}\xmark\\\xmark\\\cmark\end{tabular}
&\begin{tabular}{l}\xmark\\\xmark\\\cmark\end{tabular}
&&\begin{tabular}{l}\cmark\\\cmark\\\cmark\end{tabular}
&\begin{tabular}{l}\cmark\\\cmark\\\cmark\end{tabular}
&&\begin{tabular}{l}\cmark\\\cmark\\\cmark\end{tabular}
&\begin{tabular}{l}\cmark\\\cmark\\\cmark\end{tabular}
\\ \hline
\end{tabular}
\caption{Significance tests for quality improvements comparing Setting 1 with Settings 2–4.
Significance is indicated by \cmark\ (p < 0.05) under paired one-sided t-tests (t-test) \cite{ttest} and Wilcoxon tests (Wilco) \cite{wilcoxon}. Author input, regardless of format or level of detail, yields consistent and significant gains in \textit{Specificity} and \textit{Convincingness}. Only \textit{Targeting} improvements are occasionally non-significant, primarily because baseline scores are already high: LLMs can reliably identify and stay focused on the reviewer’s discussion points.
}
\label{tab:app_aix_sig}
\end{table}

\begin{table}[t]
\fontsize{8}{8}
\selectfont
\centering
\renewcommand{\arraystretch}{1.1} 
\tabcolsep=0.08cm
\begin{tabular}[]{lll|llll|llll|llllll|llllll}
\hline
\multicolumn{2}{l}{Metric} &&\multicolumn{3}{c}{}&&\multicolumn{3}{c}{Quality}\\\hline
\multicolumn{2}{l}{LLM/Setting} &&len&len$\uparrow$ &eD
&& Targ&Spec&Conv 
\\ \hline
\begin{tabular}{l}\includegraphics[height=1.4em]{fig_model_logo/microsoft_logo.png}\hspace{0.5em}\\Phi-4\end{tabular}
&\begin{tabular}{r}6.\\8.\\\hline 7.\\9.\end{tabular}
&&\begin{tabular}{l}284\\312\\ 284\\368\end{tabular}
&\begin{tabular}{l}\\95.8\\\\93.8\end{tabular}
&\begin{tabular}{l}\\.714\\\\.722\end{tabular}
&&\begin{tabular}{l}.829\\\cellcolor{lightgreen}.929\\.821\\\cellcolor{lightgreen}.929\end{tabular}
&\begin{tabular}{l}.600\\\cellcolor{lightgreen}.733\\.583\\\cellcolor{lightgreen}.713\end{tabular}
&\begin{tabular}{l}.613\\\cellcolor{lightgreen}.725\\.596\\\cellcolor{lightgreen}.725\end{tabular}
\\ \hline
\begin{tabular}{l}\includegraphics[height=1.4em]{fig_model_logo/qwen_logo.png}\hspace{0.5em}\\Qwen3\end{tabular}
&\begin{tabular}{l}6.\\8.\\\hline 7.\\9.\end{tabular}
&&\begin{tabular}{l}130\\142\\216\\290\end{tabular}
&\begin{tabular}{l}\\85.4\\\\97.9\end{tabular}
&\begin{tabular}{l}\\.570\\\\.549\end{tabular}
&&\begin{tabular}{l}.913\\\cellcolor{lightgreen}.938\\.938\\\cellcolor{lightgreen}.983\end{tabular}
&\begin{tabular}{l}.700\\\cellcolor{lightgreen}.771\\.725\\\cellcolor{lightgreen}.842\end{tabular}
&\begin{tabular}{l}.700\\\cellcolor{lightgreen}.758\\.725\\\cellcolor{lightgreen}.800\end{tabular}
\\ \hline
\begin{tabular}{l}\includegraphics[height=1.4em]{fig_model_logo/llama_logo.png}\hspace{0.5em}\\Llama-3.3\end{tabular}
&\begin{tabular}{l}6.\\8.\\\hline 7.\\9.\end{tabular}
&&\begin{tabular}{l}82\\125\\214\\304\end{tabular}
&\begin{tabular}{l}\\89.6\\\\95.8\end{tabular}
&\begin{tabular}{l}\\.506\\\\.499\end{tabular}
&&\begin{tabular}{l}.804\\\cellcolor{lightgreen}.892\\.850\\\cellcolor{lightgreen}.888\end{tabular}
&\begin{tabular}{l}.467\\\cellcolor{lightgreen}.667\\.575\\\cellcolor{lightgreen}.750\end{tabular}
&\begin{tabular}{l}.504\\\cellcolor{lightgreen}.638\\.592\\\cellcolor{lightgreen}.700\end{tabular}
\\ \hline
\begin{tabular}{l}\includegraphics[height=1.4em]{fig_model_logo/deepseek_logo.png}\hspace{0.5em}\\DeepSeek\end{tabular}
&\begin{tabular}{l}6.\\8.\\\hline 7.\\9.\end{tabular}
&&\begin{tabular}{l}93\\97\\179\\194\end{tabular}
&\begin{tabular}{l}\\66.7\\\\68.8\end{tabular}
&\begin{tabular}{l}\\.542\\\\.547\end{tabular}
&&\begin{tabular}{l}.867\\\cellcolor{lightgreen}.913\\.888\\\cellcolor{lightgreen}.925\end{tabular}
&\begin{tabular}{l}.588\\\cellcolor{lightgreen}.704\\.663\\\cellcolor{lightgreen}.746\end{tabular}
&\begin{tabular}{l}.625\\\cellcolor{lightgreen}.704\\.671\\\cellcolor{lightgreen}.742\end{tabular}
\\ \hline
\begin{tabular}{l}\includegraphics[height=1.4em]{fig_model_logo/gpt_logo.jpg}\hspace{0.5em}\\GPT-4o\end{tabular}
&\begin{tabular}{l}6.\\8.\\\hline 7.\\9.\end{tabular}
&&\begin{tabular}{l}156\\163\\336\\367\end{tabular}
&\begin{tabular}{l}\\79.2\\\\72.9\end{tabular}
&\begin{tabular}{l}\\.409\\\\.413\end{tabular}
&&\begin{tabular}{l}.879\\\cellcolor{lightgreen}.900\\.913\\.925\end{tabular}
&\begin{tabular}{l}.596\\\cellcolor{lightgreen}.675\\.692\\\cellcolor{lightgreen}.721\end{tabular}
&\begin{tabular}{l}.621\\\cellcolor{lightgreen}.675\\.700\\\cellcolor{lightgreen}.721\end{tabular}
\\ \hline
\end{tabular}
\caption[LLMs]{Refinement effects.
Reported are response length before and after refinement (len), the proportion of longer responses after refinement (len$\uparrow$), edit distance (eD), and mean quality scores for \textit{Targeting}, \textit{Specificity}, and \textit{Convincingness}. All quality metrics improve across all LLMs and settings, with statistically significant gains under both paired one-sided t-tests and Wilcoxon signed-rank tests (marked in green). The only non-significant case is \textit{Targeting} in Setting 9 with GPT-4o, as the initial targeting is already strong.}
\label{tab:ARG_refine}
\end{table}

\subsection{Case Studies}
\label{subsec:app_case_studies}
To further demonstrate the analytical utility of \textit{REspEval}, we present two case studies examining cross-dimensional interactions, trade-offs, and detailed evaluation outcomes. We focus on non-trivial cases where richer inputs do not consistently improve response quality and can in fact hurt certain evaluation dimensions. While richer inputs improve response quality in most cases (>92\%), the following examples illustrate informative exceptions that underscore the value of our fine-grained, multi-dimensional evaluation. 

\paragraph{Case Study 1: Richer Inputs Degrade Response Quality Across Targeting, Specificity, and Convincingness.} Figure \ref{fig:app_case_study_1} illustrates a case where the review poses a direct question requiring a focused answer about execution environment and document type. In Setting 2, only the edit sentences (highlighted in green) are used as author input. In Setting 3, the surrounding paragraph context (highlighted in blue) is additionally included, forming a richer contextualized input.

The final two blocks show responses generated by DeepSeek under each setting, with information drawn from respective sources color-coded accordingly. Since the edit sentences alone contain the key information needed to address the review question, adding paragraph context introduces irrelevant details and ambiguity, resulting in a response that is less targeted (4 vs. 2), less specific (3 vs. 1), and less convincing (3 vs. 1).

The quality evaluations further provide justifications grounding each score. For instance, the low \textit{Specificity} score for the Setting 3 response is justified as: \textit{"- Critical details are missing: execution environment is missing, no document description, no sections/tables/figures or configurations [question\_1]."} In contrast, the Setting 2 response receives a higher \textit{Specificity} score justified by: \textit{"+ Provides some concrete elements: Microsoft Word environment, built-in formatting functions, Office Add-ins, standard Word document [question\_1]."}

\paragraph{Case Study 2: Richer Inputs Reduce Specificity and Input Coverage Recall (ICR).} Figure \ref{fig:app_case_study_2} illustrates a case where the review critiques the measurement of ToM problem complexity solely by the number of state changes, requiring careful argumentation. In Setting 2, the edit sentences (highlighted in green) define stateful and stateless complexity, justify their necessity, and introduce operationalization notations. In Setting 3, the paragraph context (highlighted in blue) further provides connections to existing domain terminology and notation context. The final two blocks show GPT-4o responses under each setting, with information sources color-coded accordingly.

While both responses are substantive and well-reasoned, the Setting 2 response covers most core information: defining stateful and stateless complexity, justifying their necessity, and introducing the relevant notations. In Setting 3, the richer context shifts focus and causes omission of the notations, reducing both specificity (4 vs. 3) and ICR (0.83 vs. 0.75). This is reflected in the specificity justifications: the Setting 2 response is credited for \textit{"+ Introduces specific constructs: stateful vs. stateless complexity and their roles; defines germane load as a ratio [criticism\_1]."} and \textit{"+ Provides a formal decomposition of p and notes |p| as number of state changes [criticism\_1]."} whereas the Setting 3 response, despite introducing core conceptual components, is penalized for \textit{"- Lacks quantitative examples, metrics, or references to specific sections/figures beyond a vague mention of adding to "A Comparison with the Cognitive Load Theory" [criticism\_1]."}

\newcounter{settingrow}
\renewcommand\thesubfigure{(\arabic{settingrow}.\alph{subfigure})}

\newcommand{\RowStart}{%
  \stepcounter{settingrow}%
  \setcounter{subfigure}{0}%
}
\setlength{\tabcolsep}{0.1pt}      %
\renewcommand{\arraystretch}{0.0005}  %

\newcolumntype{L}[1]{>{\raggedright\arraybackslash}m{#1}}
\newcolumntype{C}[1]{>{\centering\arraybackslash}m{#1}}

\newcommand{\LabelColA}{0.01\linewidth}   %
\newcommand{\LabelColB}{0\linewidth}   %
\newcommand{\ImgColW}{0.165\linewidth}     %
\setcounter{settingrow}{0}
\begin{figure*}[ht]
\fontsize{10}{10}
\centering
\begin{tabular}{C{\ImgColW} C{\ImgColW} C{\ImgColW} C{\ImgColW} C{\ImgColW}C{\ImgColW}}
\toprule
\makecell{\selectfont\begin{tabular}{l}\includegraphics[height=1.2em]{fig_model_logo/microsoft_logo.png} Phi-4\hspace{0.5em}\end{tabular}}&
\makecell{\selectfont\begin{tabular}{l}\includegraphics[height=1.2em]{fig_model_logo/qwen_logo.png} Qwen3\hspace{0.5em}\end{tabular}} &
\makecell{\selectfont\begin{tabular}{l}\includegraphics[height=1.2em]{fig_model_logo/llama_logo.png} Llama-3.3\hspace{0.5em}\end{tabular}} &
\makecell{\selectfont\begin{tabular}{l}\includegraphics[height=1.2em]{fig_model_logo/deepseek_logo.png} DeepSeek\hspace{0.5em}\end{tabular}
} &
\makecell{\selectfont\begin{tabular}{l}\includegraphics[height=1.2em]{fig_model_logo/gpt_logo.jpg} GPT-4o\hspace{0.5em}\end{tabular}}
&
\makecell{\selectfont\begin{tabular}{l}\includegraphics[height=1.5em]{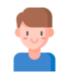} Human\hspace{0.5em}\end{tabular}}\\
\midrule

\RowStart
\subcaptionbox{}{\includegraphics[width=\linewidth]{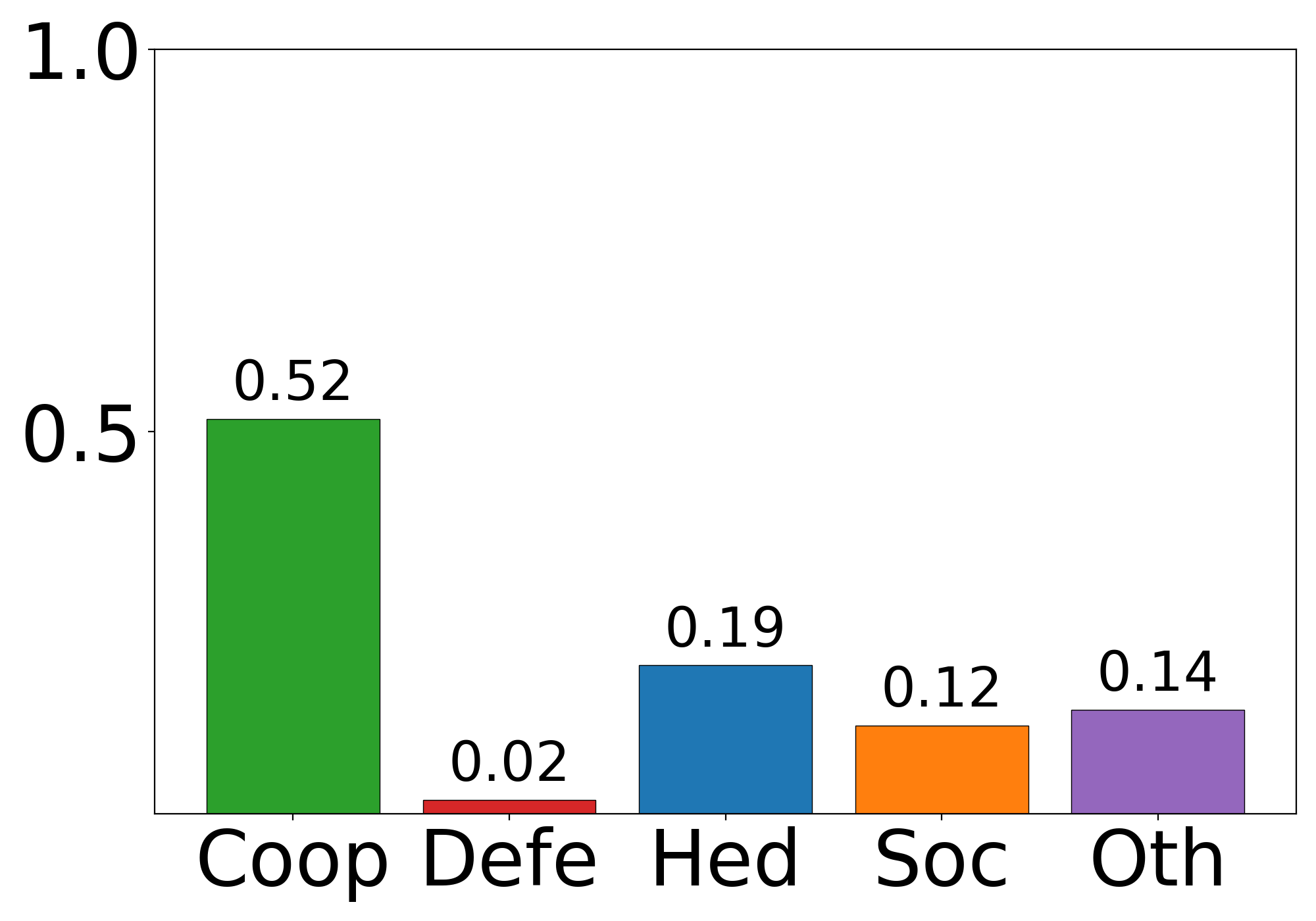}} &
\subcaptionbox{}{\includegraphics[width=\linewidth]{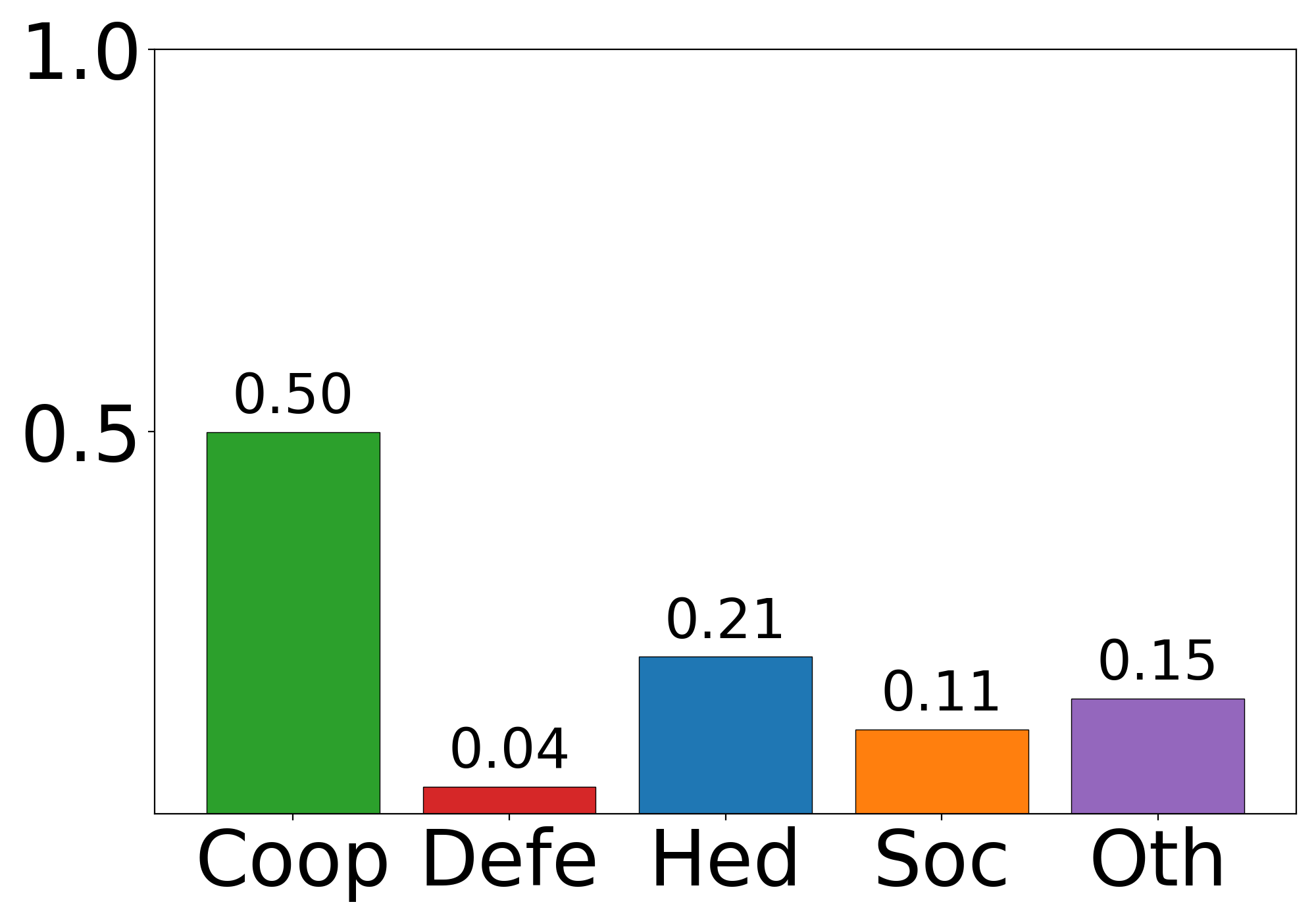}} &
\subcaptionbox{}{\includegraphics[width=\linewidth]{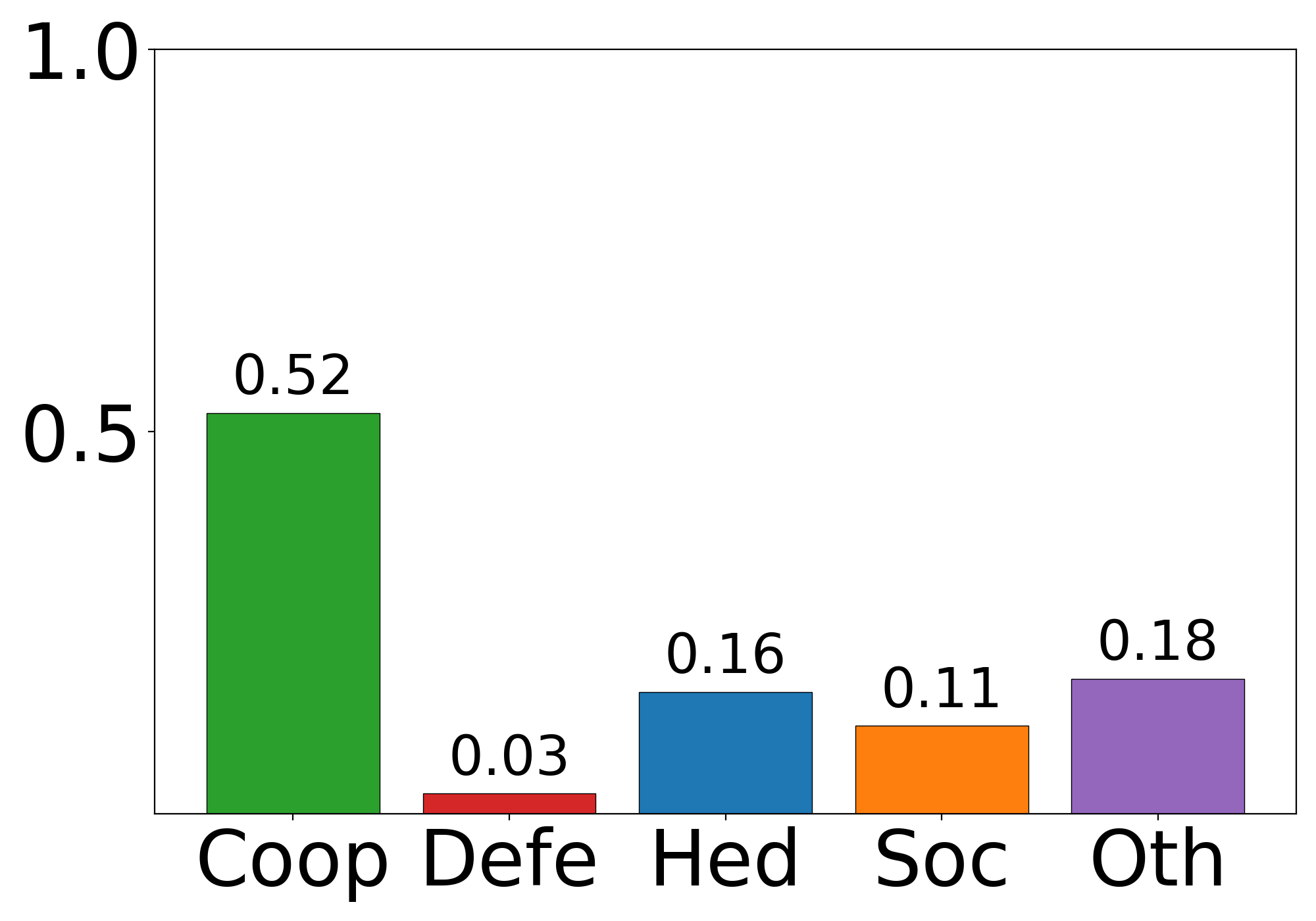}} &
\subcaptionbox{}{\includegraphics[width=\linewidth]{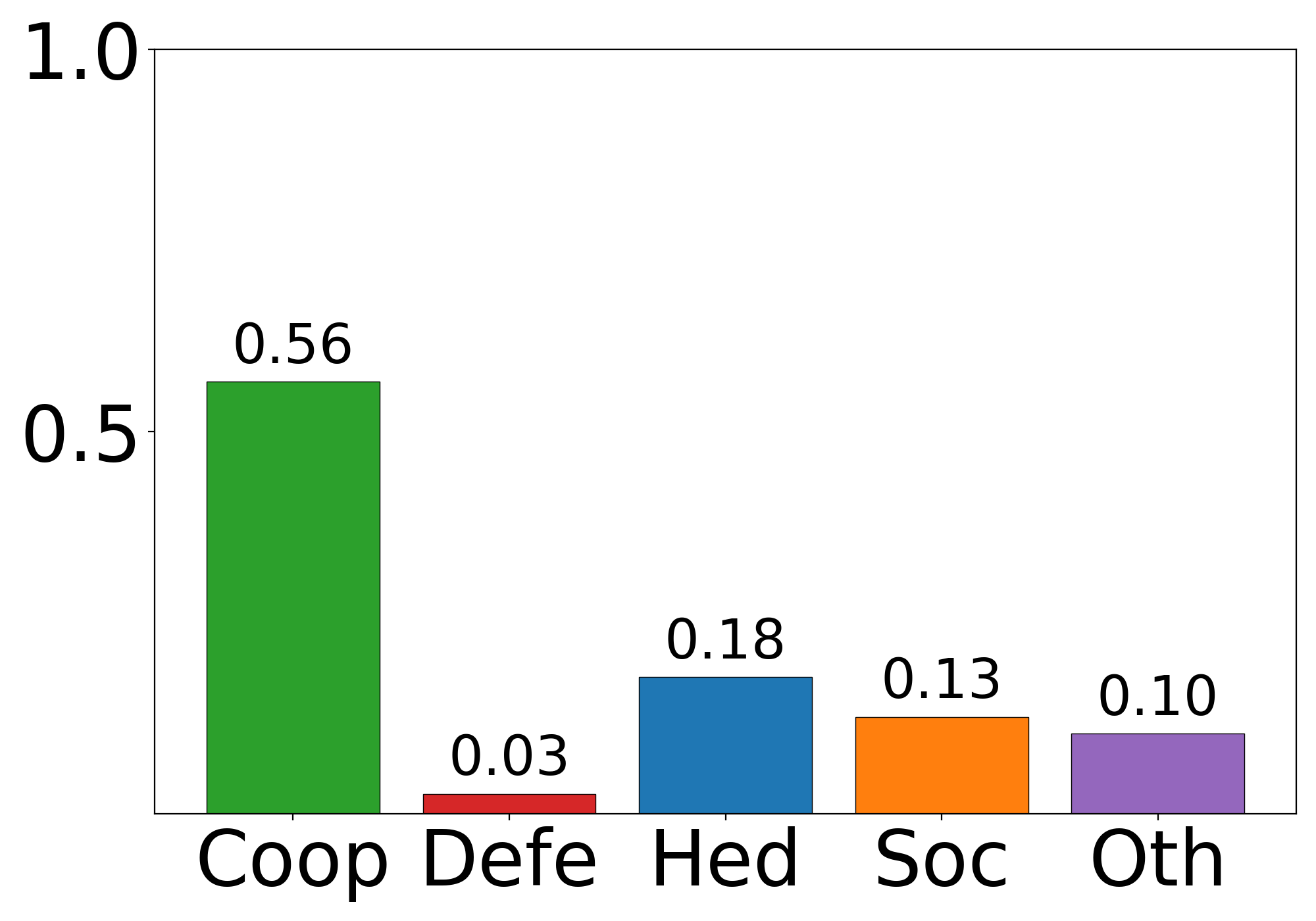}} &
\subcaptionbox{}{\includegraphics[width=\linewidth]{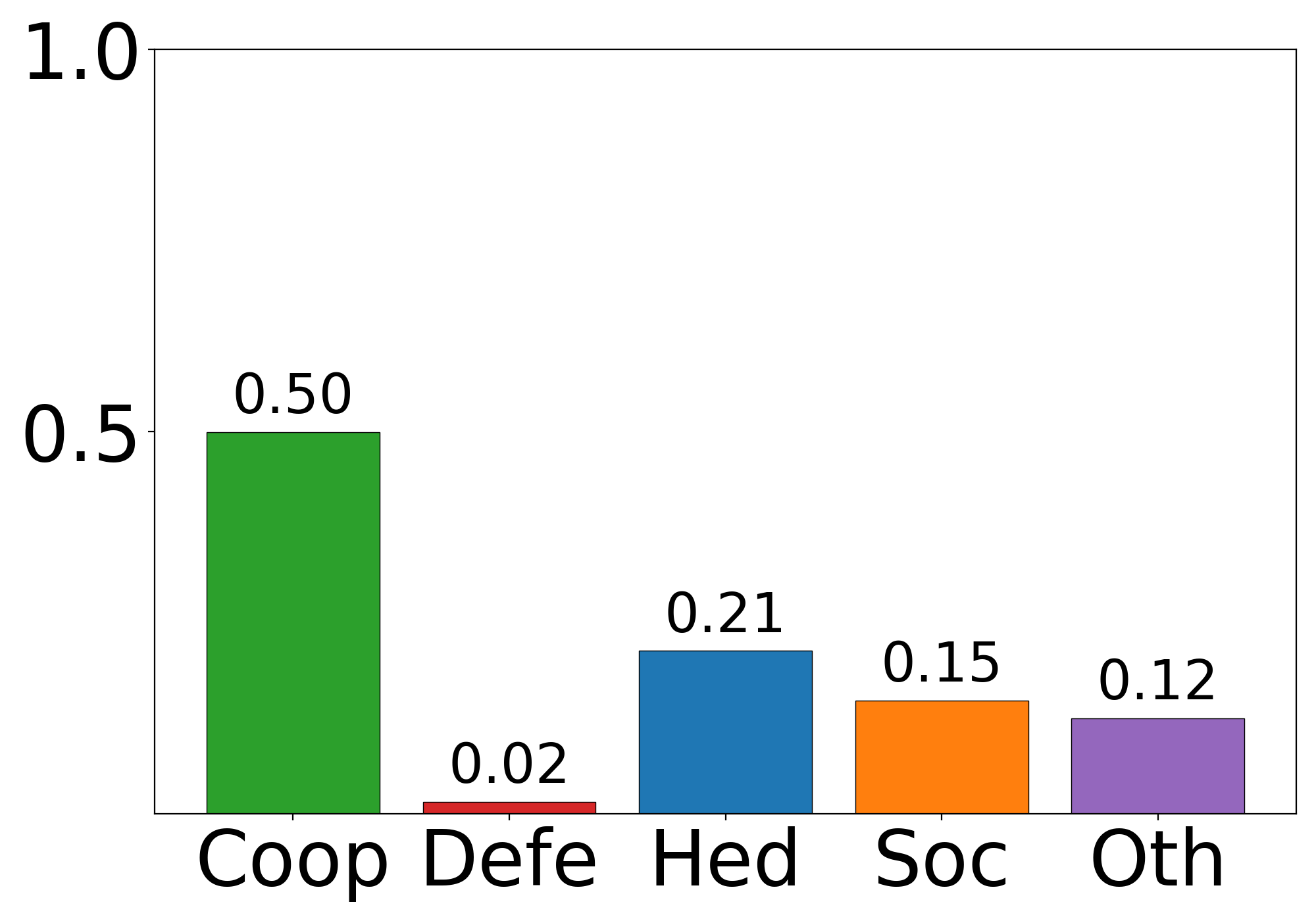}} &
\subcaptionbox{}{\includegraphics[width=\linewidth]{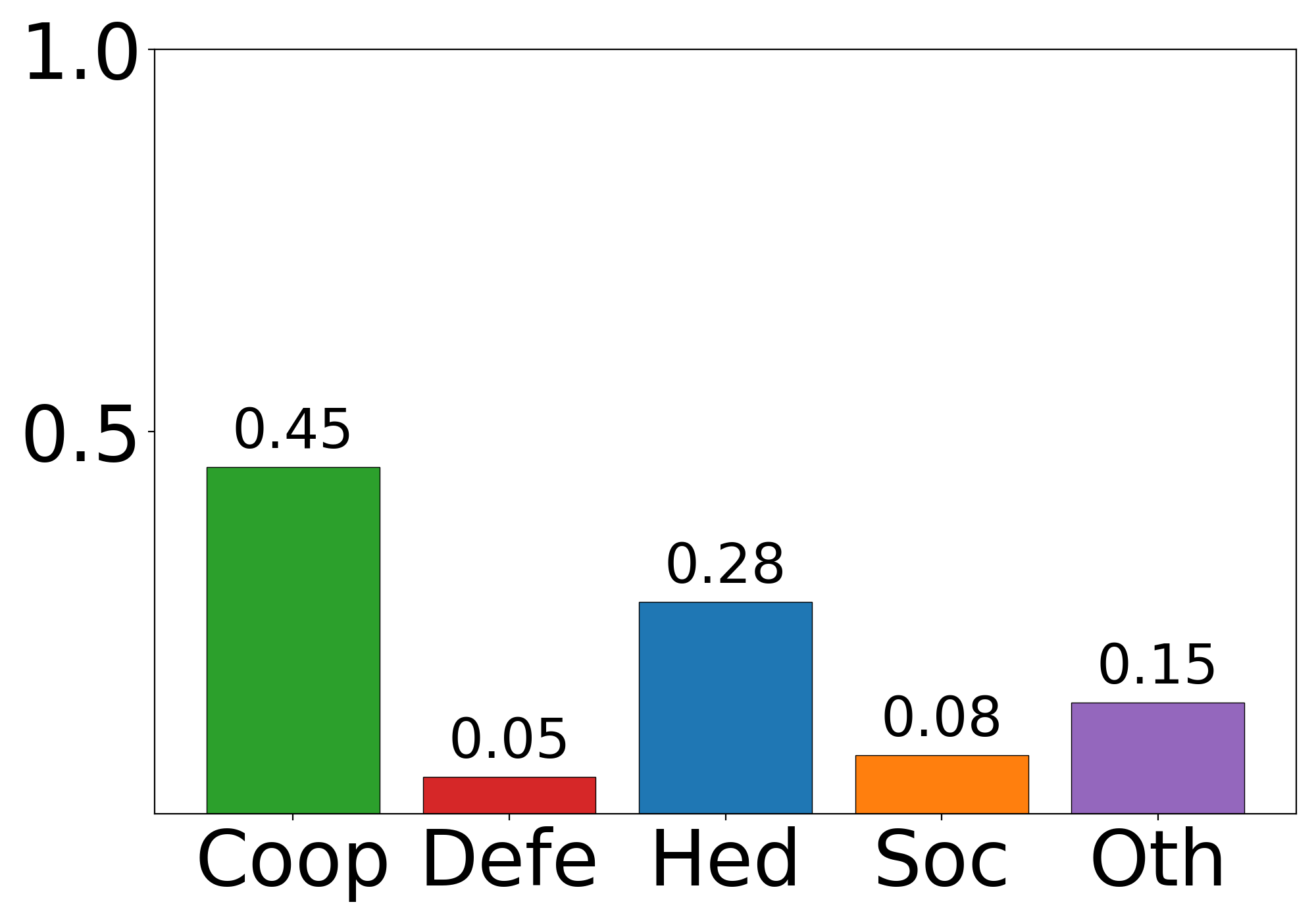}}
\\
\RowStart
\subcaptionbox{}{\includegraphics[width=\linewidth]{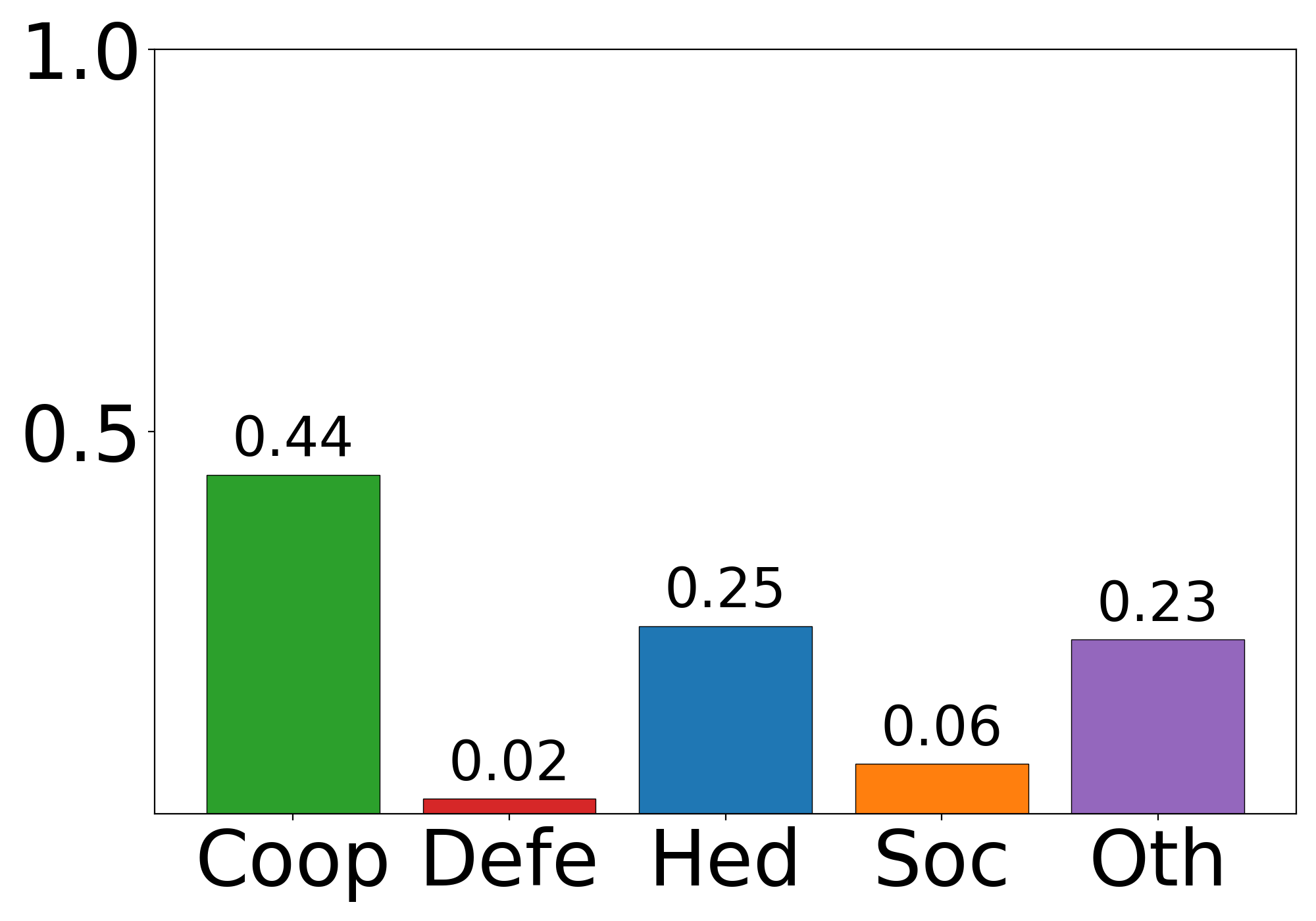}} &
\subcaptionbox{}{\includegraphics[width=\linewidth]{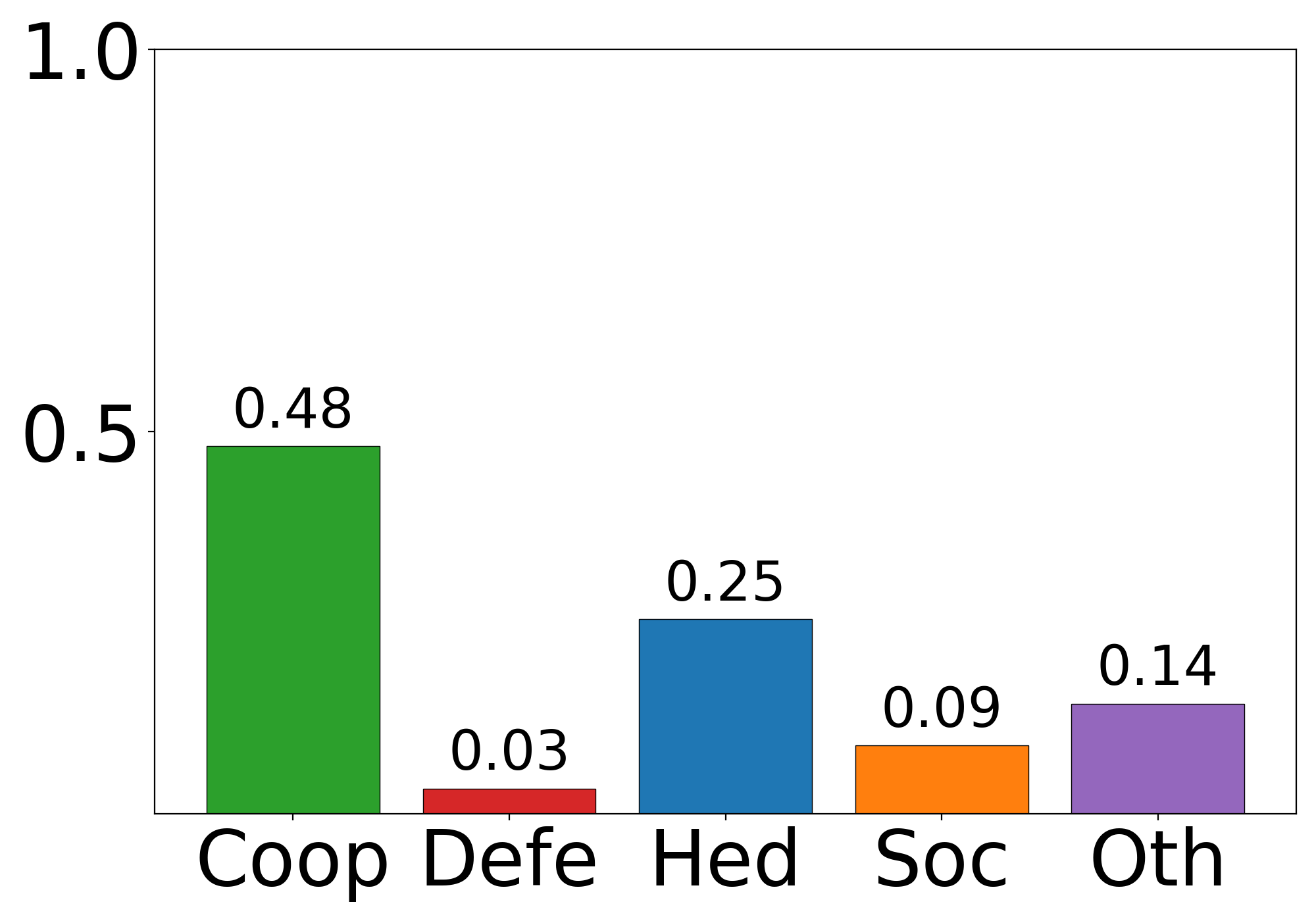}} &
\subcaptionbox{}{\includegraphics[width=\linewidth]{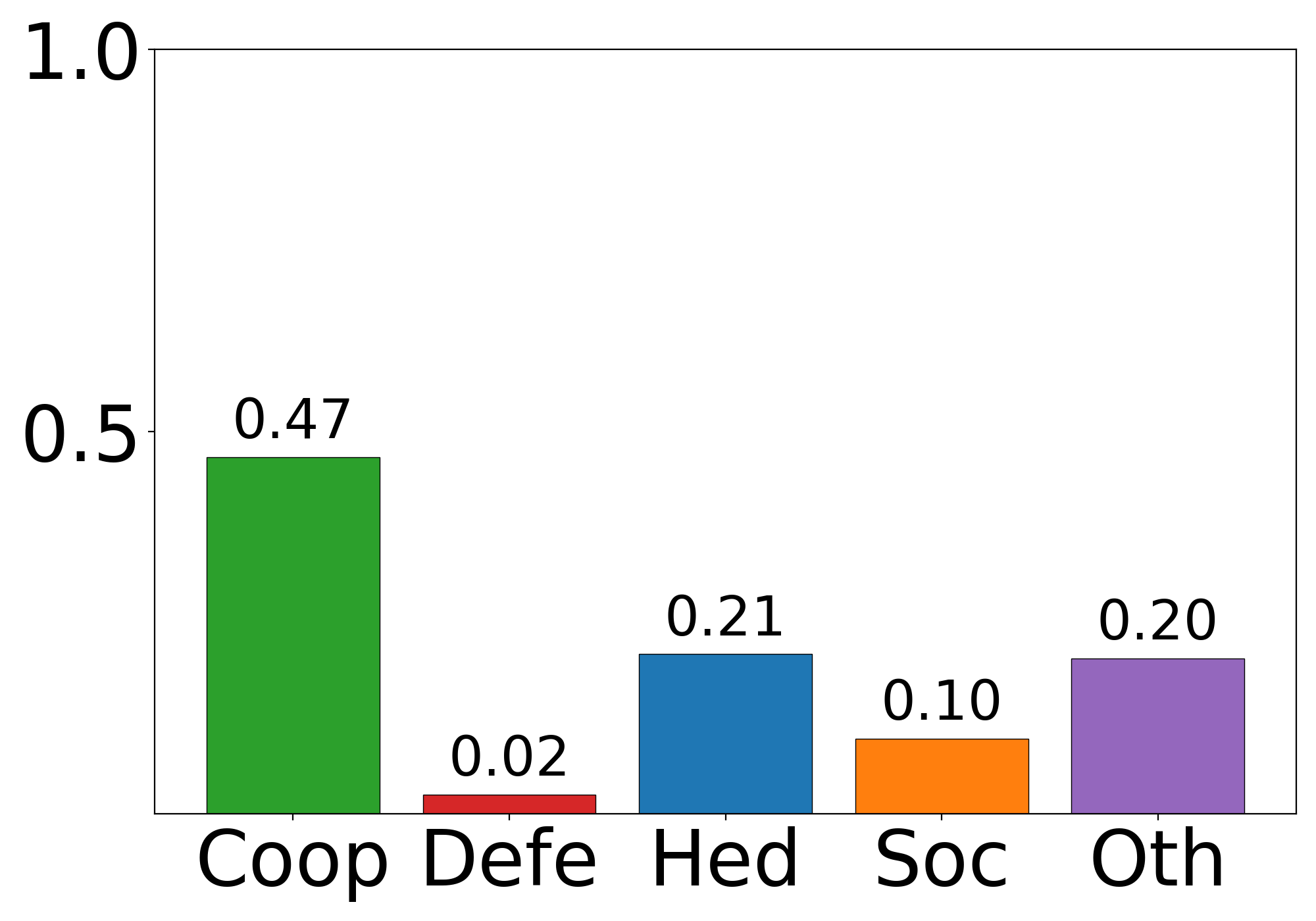}} &
\subcaptionbox{}{\includegraphics[width=\linewidth]{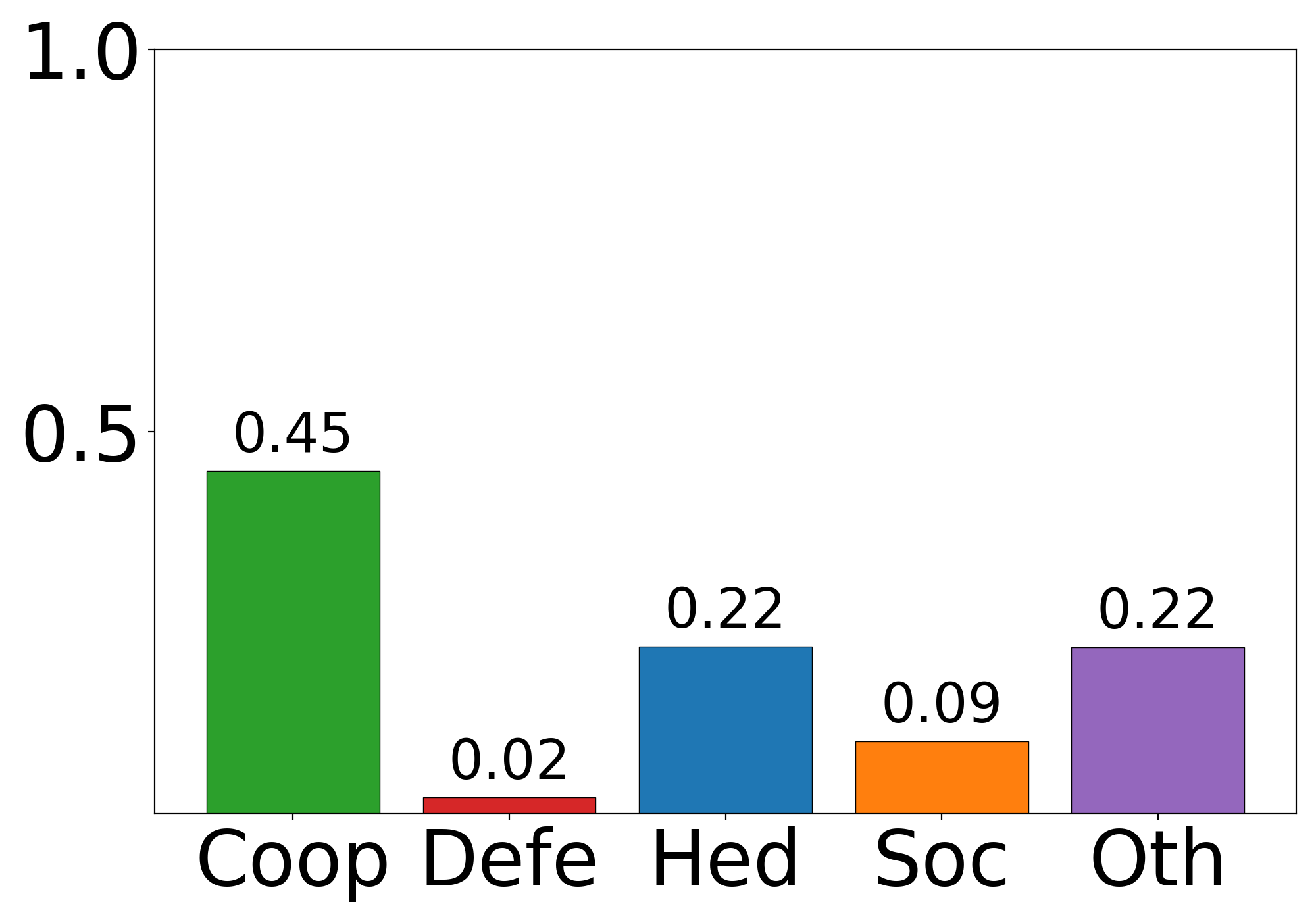}} &
\subcaptionbox{}{\includegraphics[width=\linewidth]{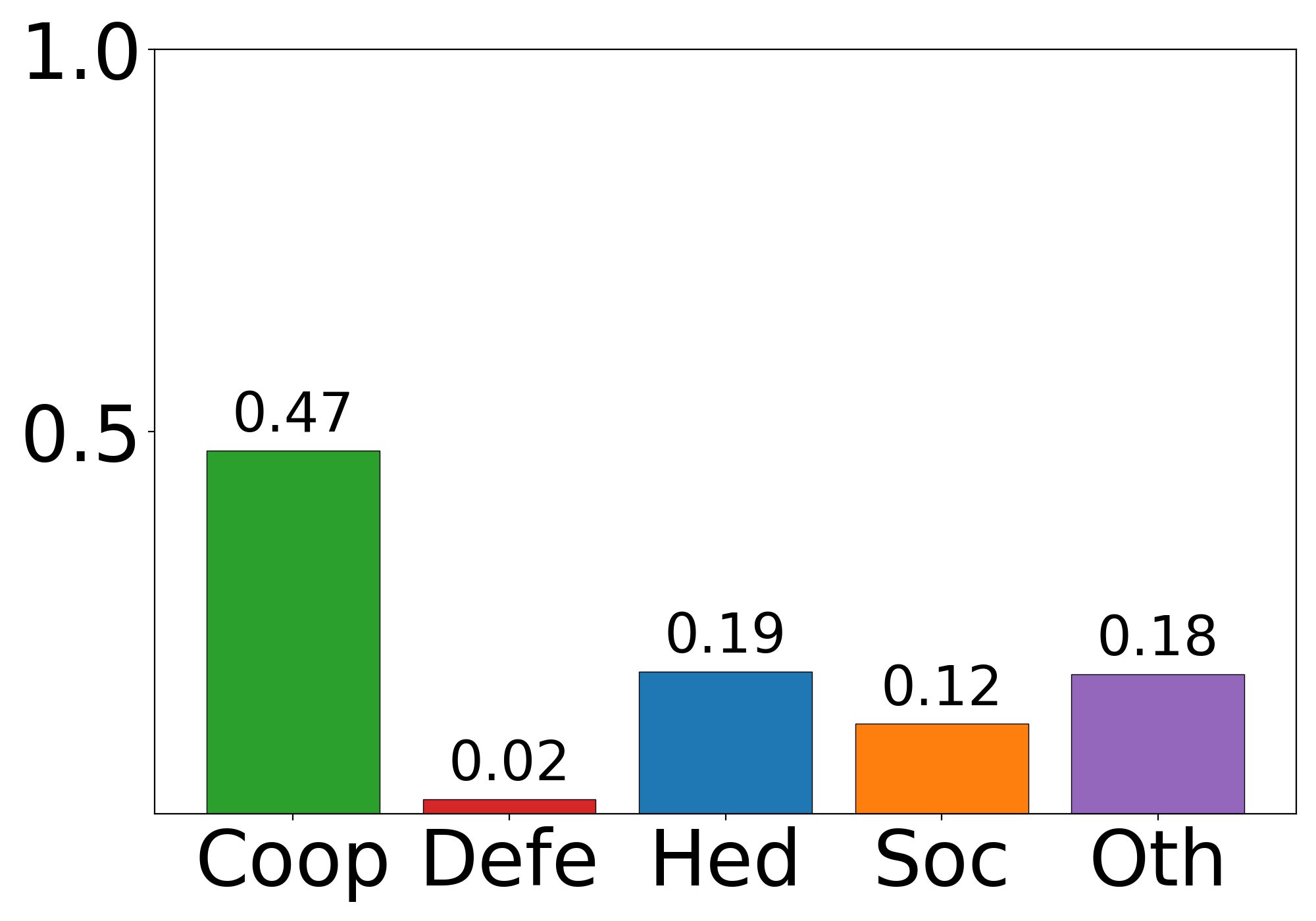}} \\
\RowStart
\subcaptionbox{}{\includegraphics[width=\linewidth]{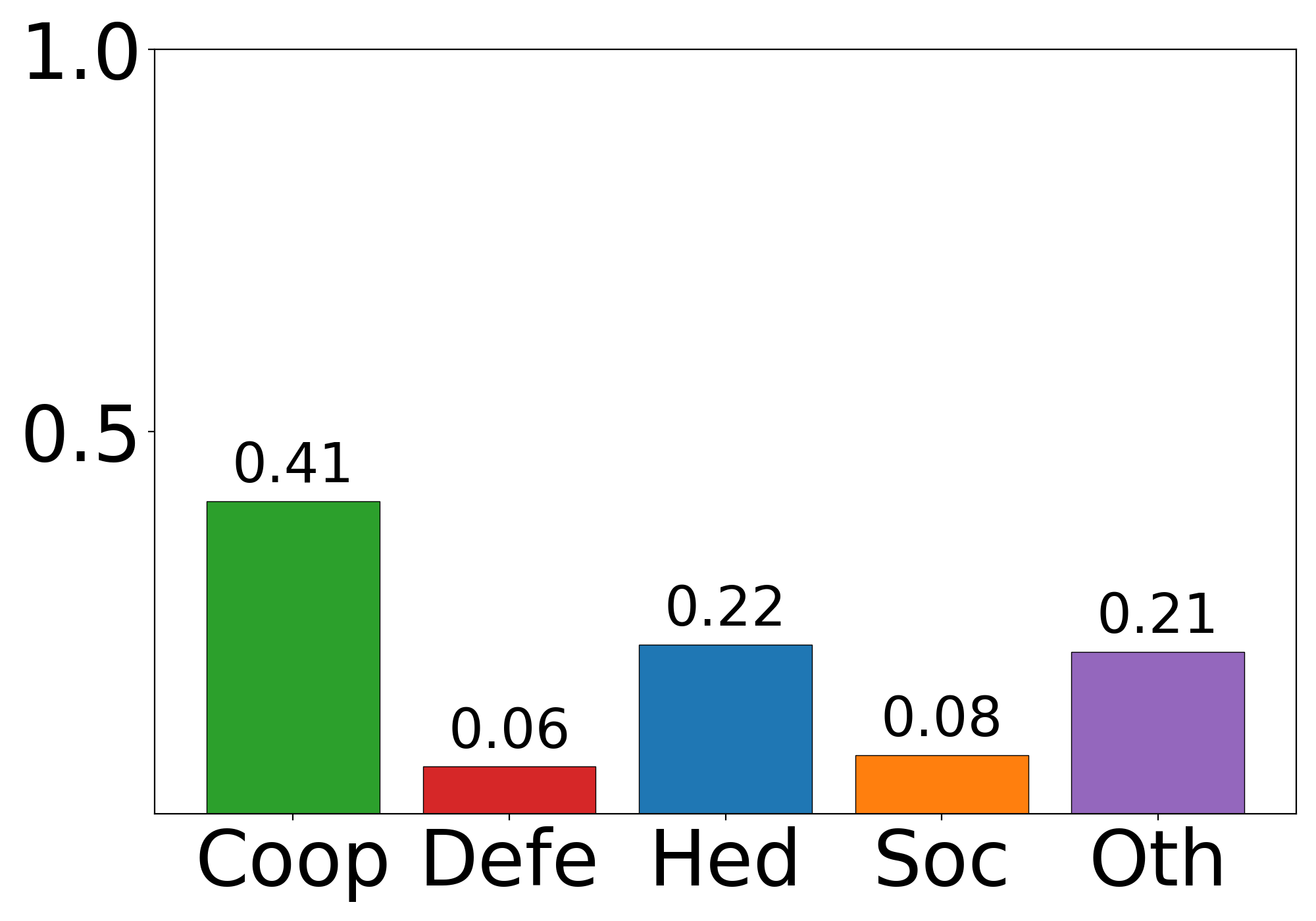}} &
\subcaptionbox{}{\includegraphics[width=\linewidth]{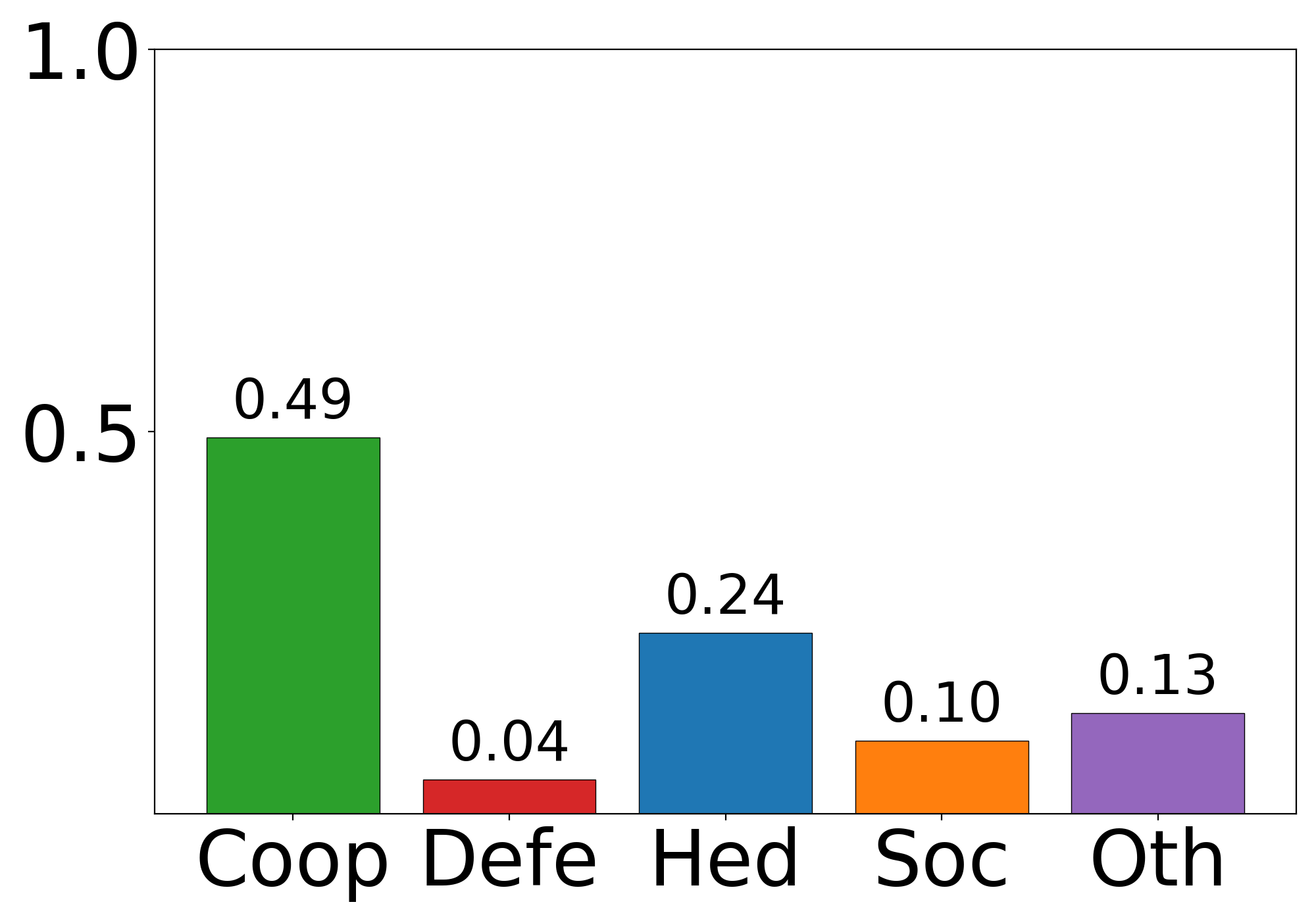}} &
\subcaptionbox{}{\includegraphics[width=\linewidth]{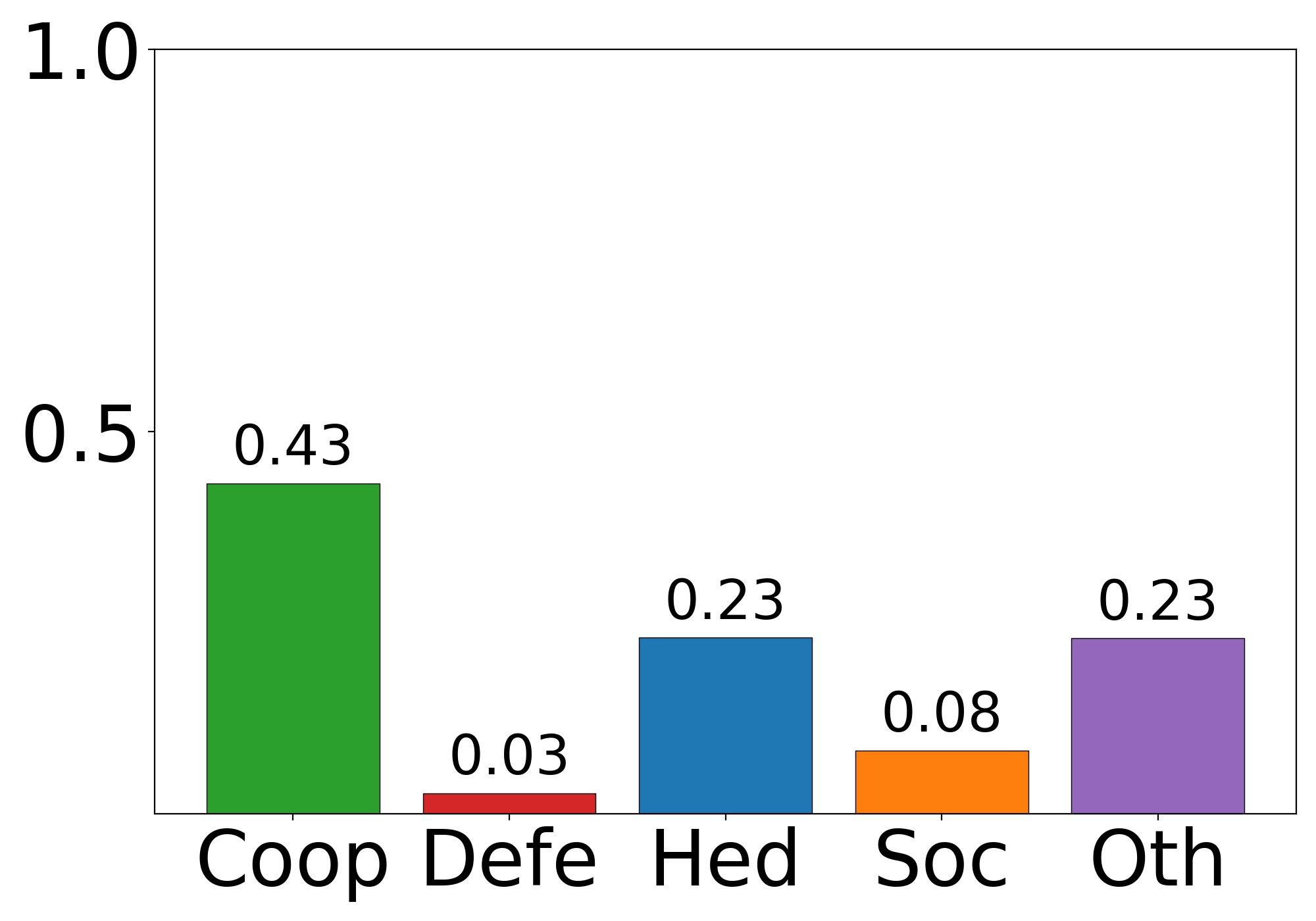}} &
\subcaptionbox{}{\includegraphics[width=\linewidth]{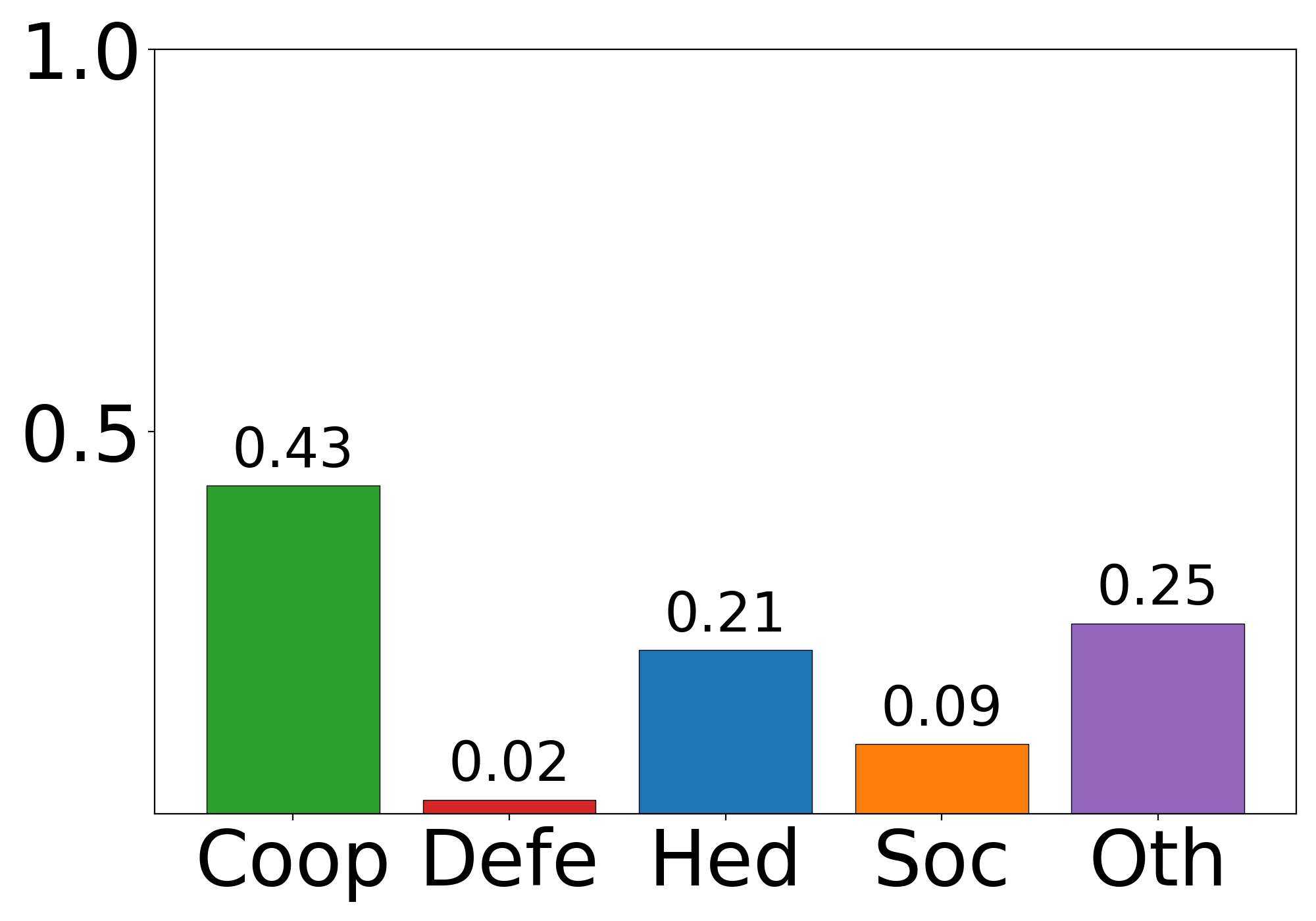}} &
\subcaptionbox{}{\includegraphics[width=\linewidth]{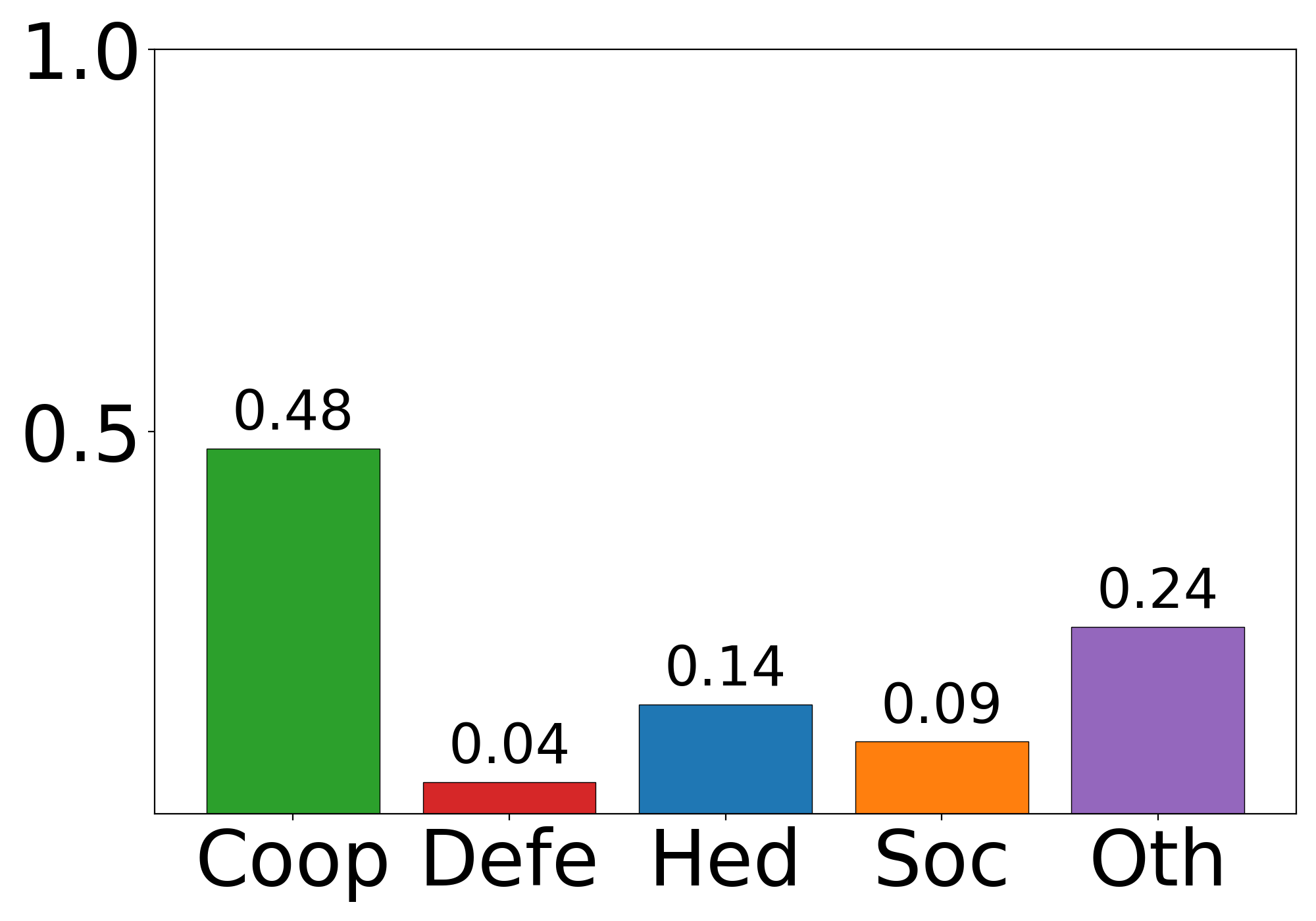}} \\

\RowStart
\subcaptionbox{}{\includegraphics[width=\linewidth]{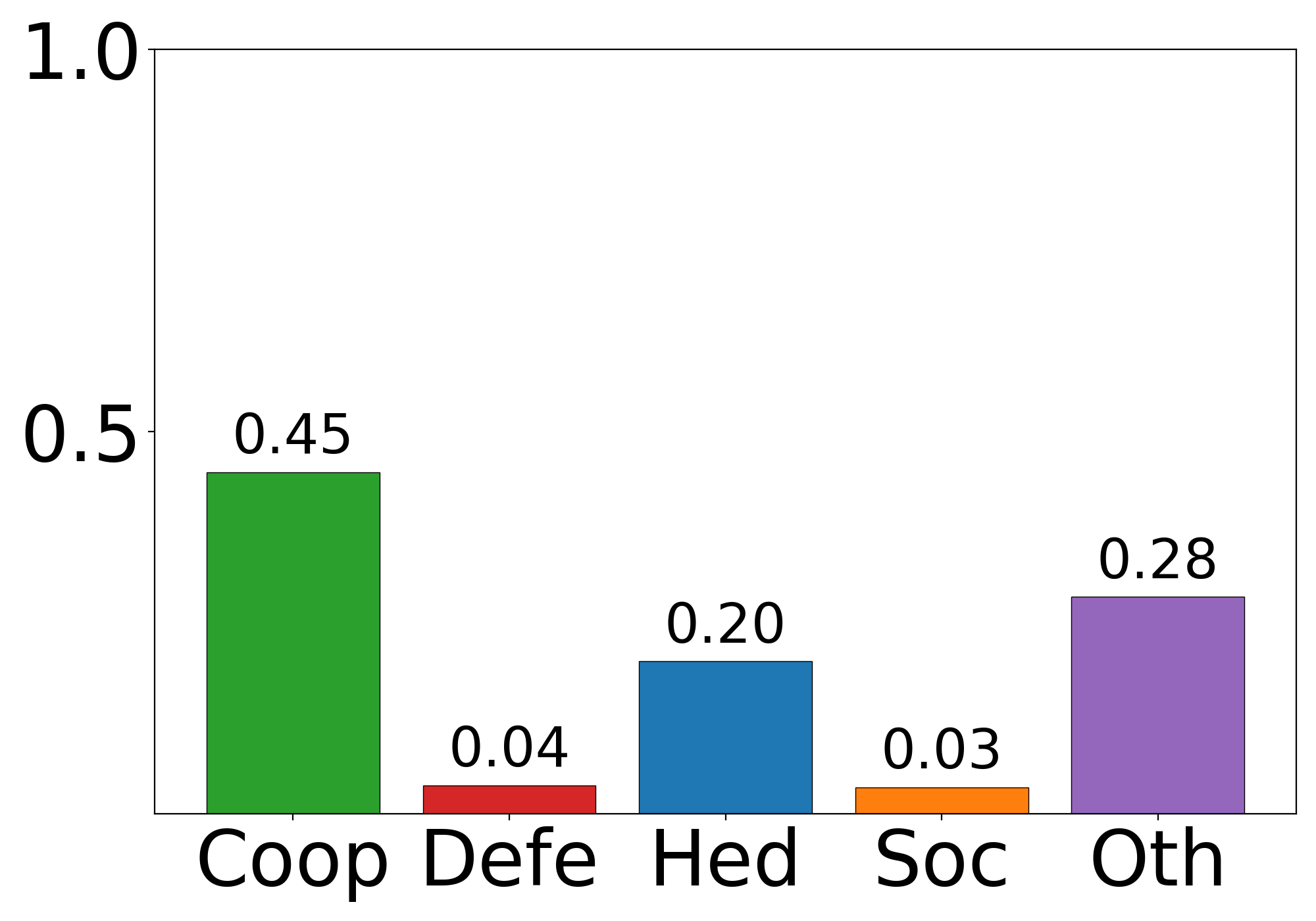}} &
\subcaptionbox{}{\includegraphics[width=\linewidth]{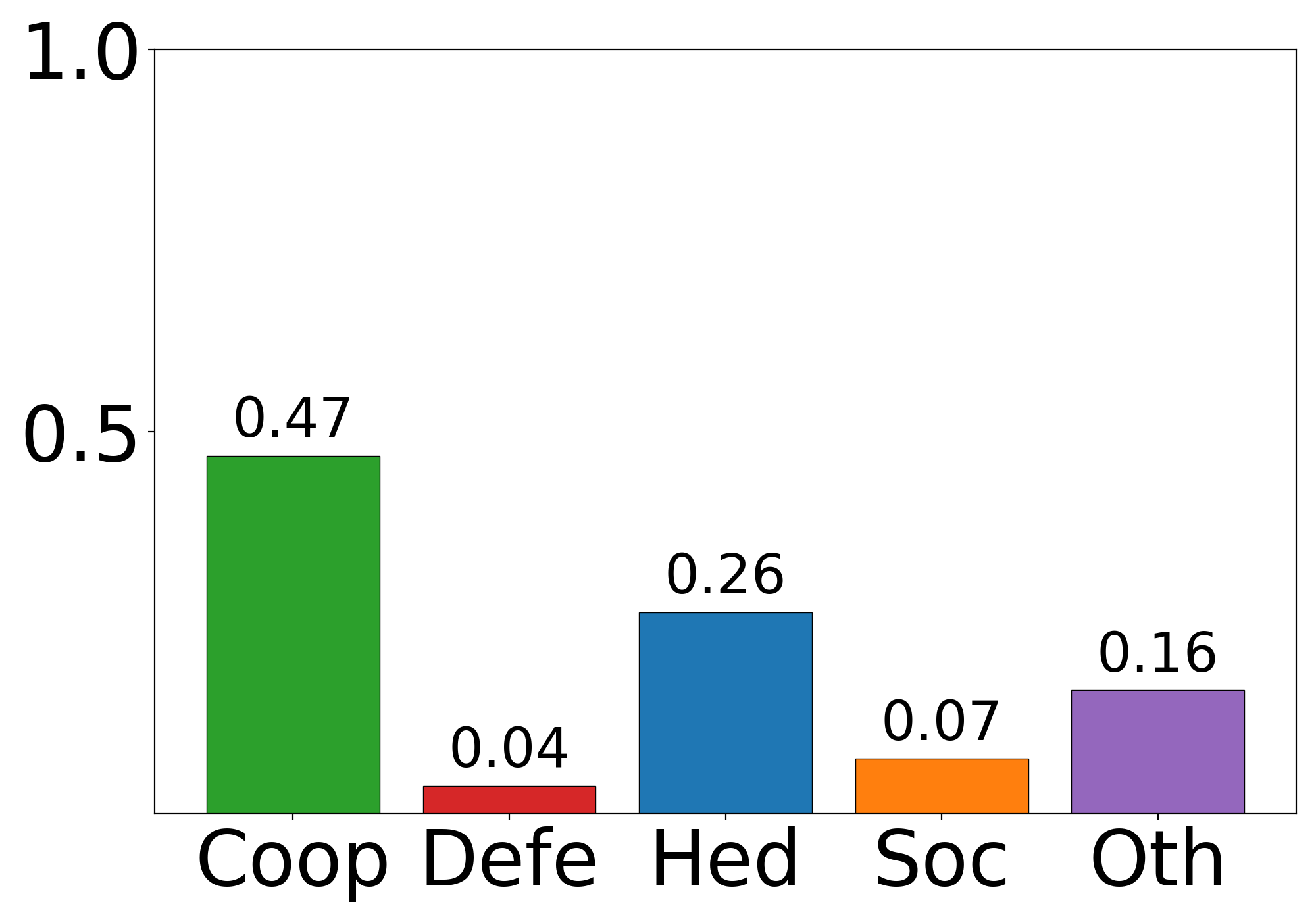}} &
\subcaptionbox{}{\includegraphics[width=\linewidth]{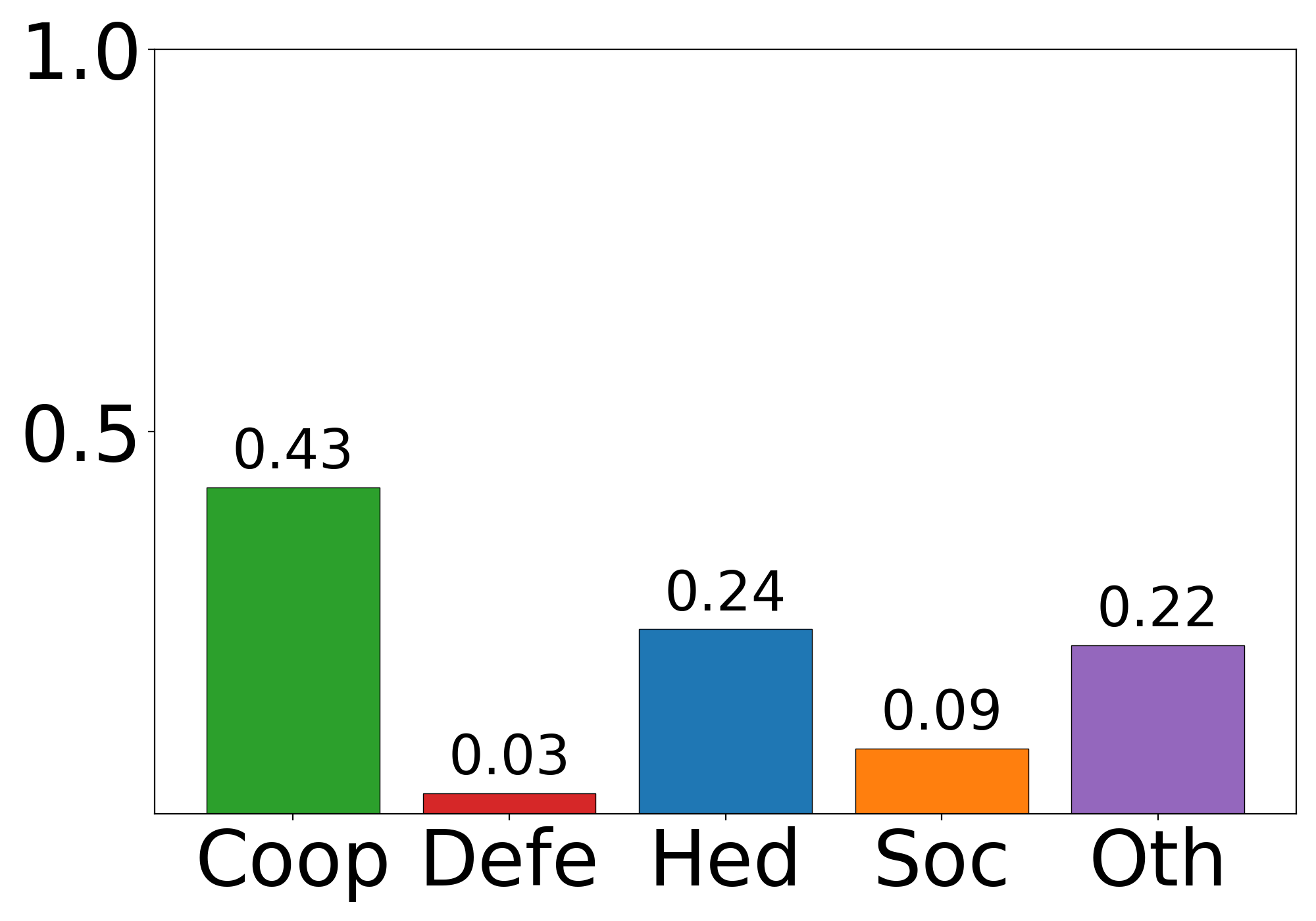}} &
\subcaptionbox{}{\includegraphics[width=\linewidth]{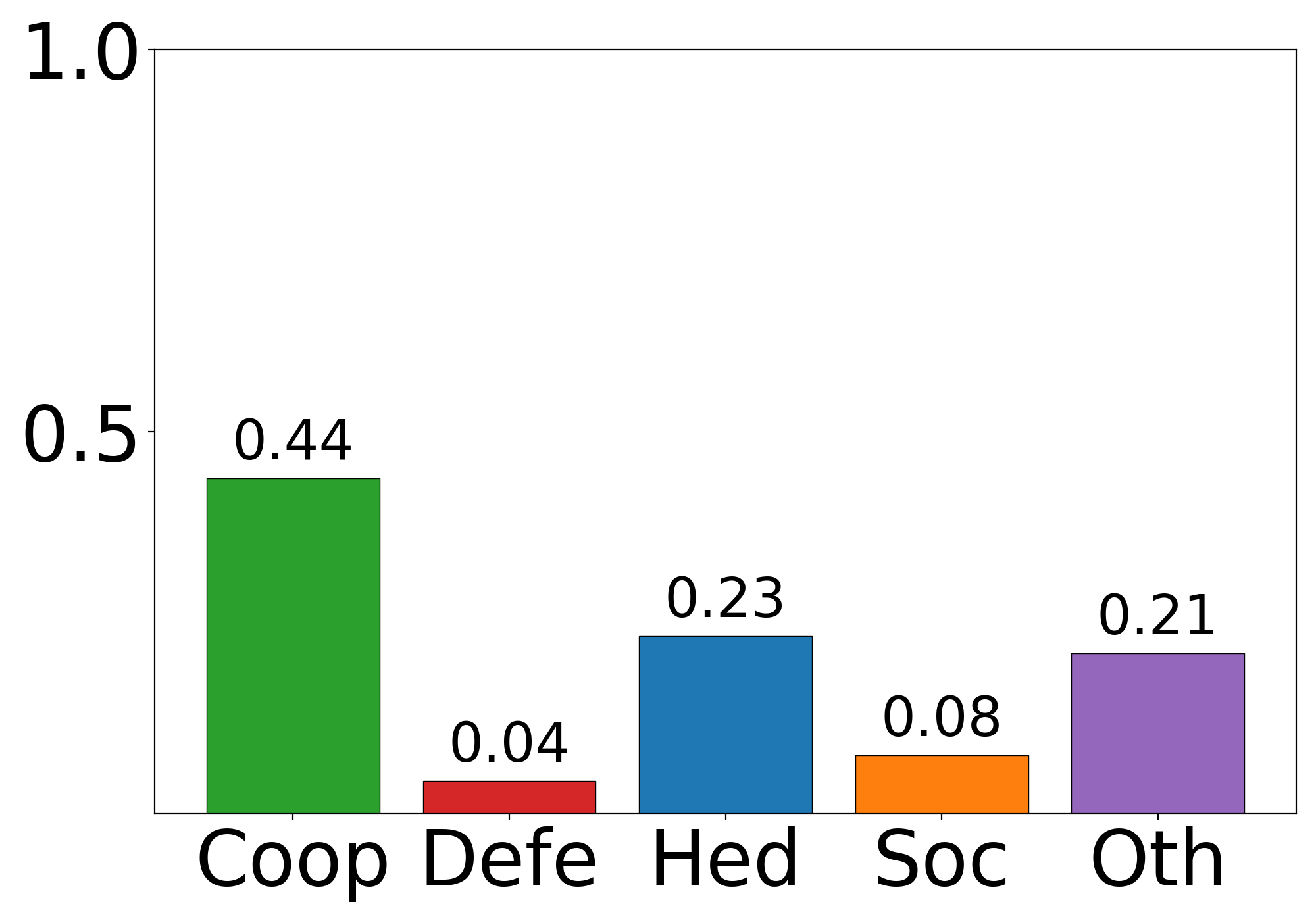}} &
\subcaptionbox{}{\includegraphics[width=\linewidth]{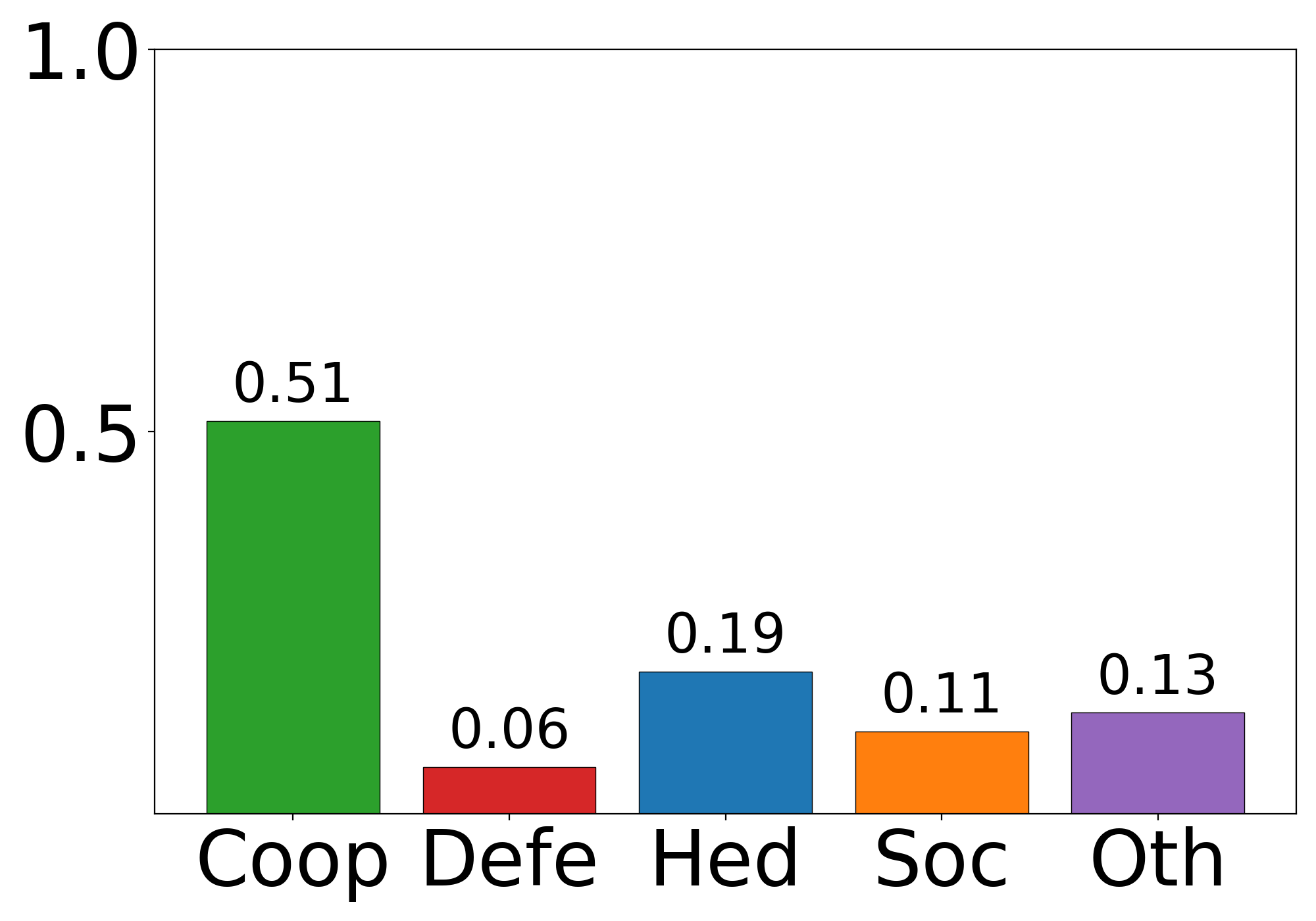}} \\

\RowStart
\subcaptionbox{}{\includegraphics[width=\linewidth]{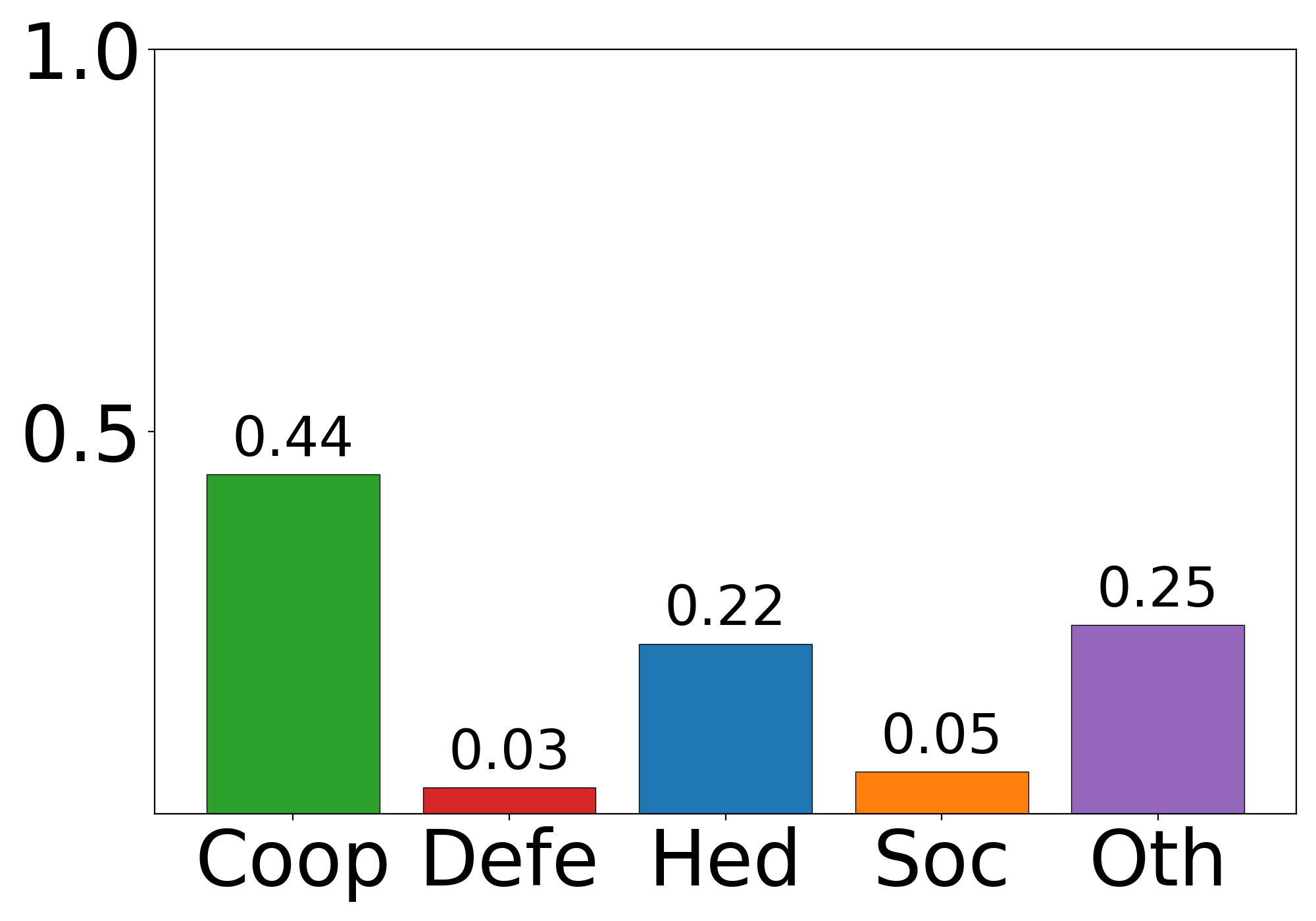}} &
\subcaptionbox{}{\includegraphics[width=\linewidth]{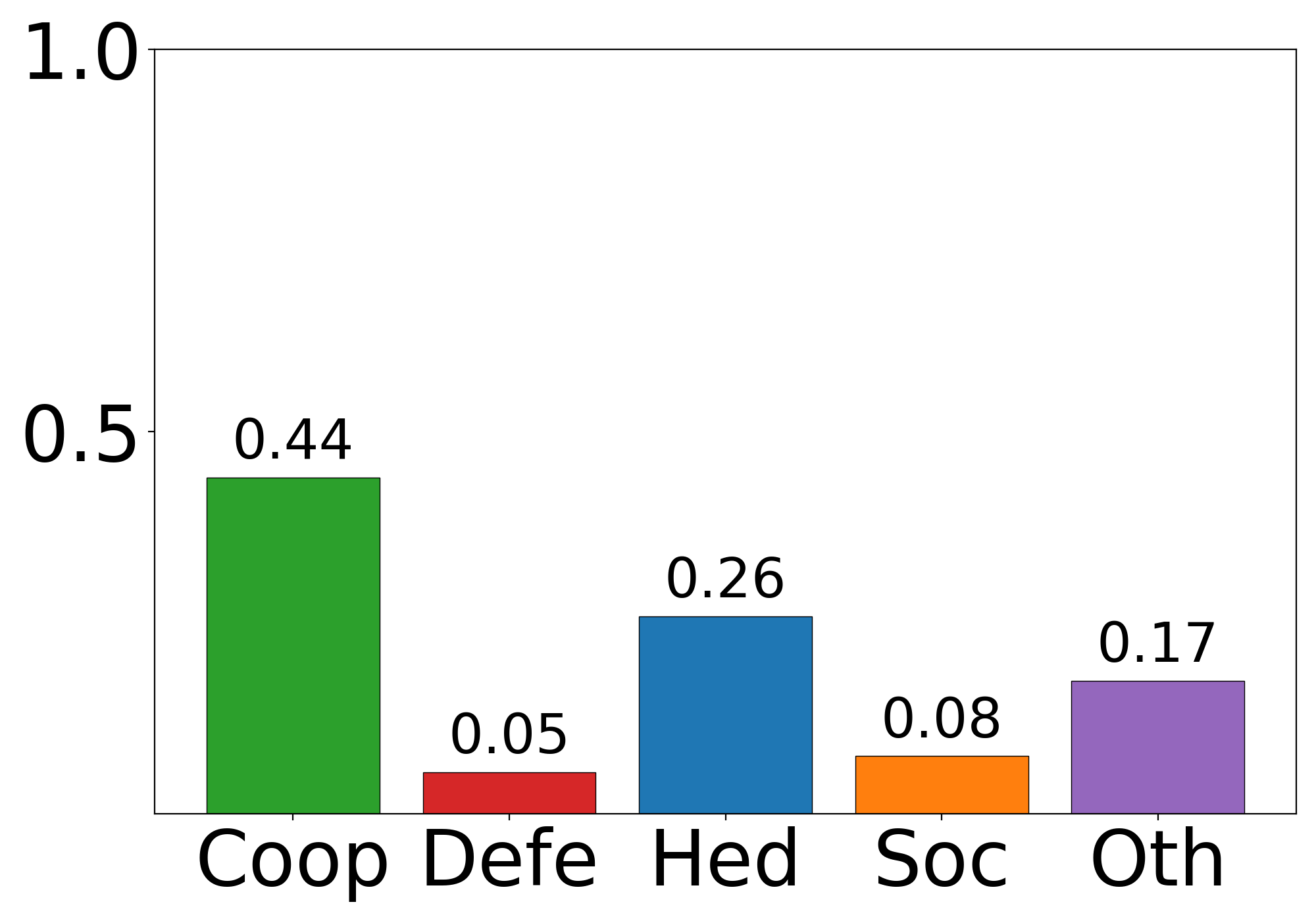}} &
\subcaptionbox{}{\includegraphics[width=\linewidth]{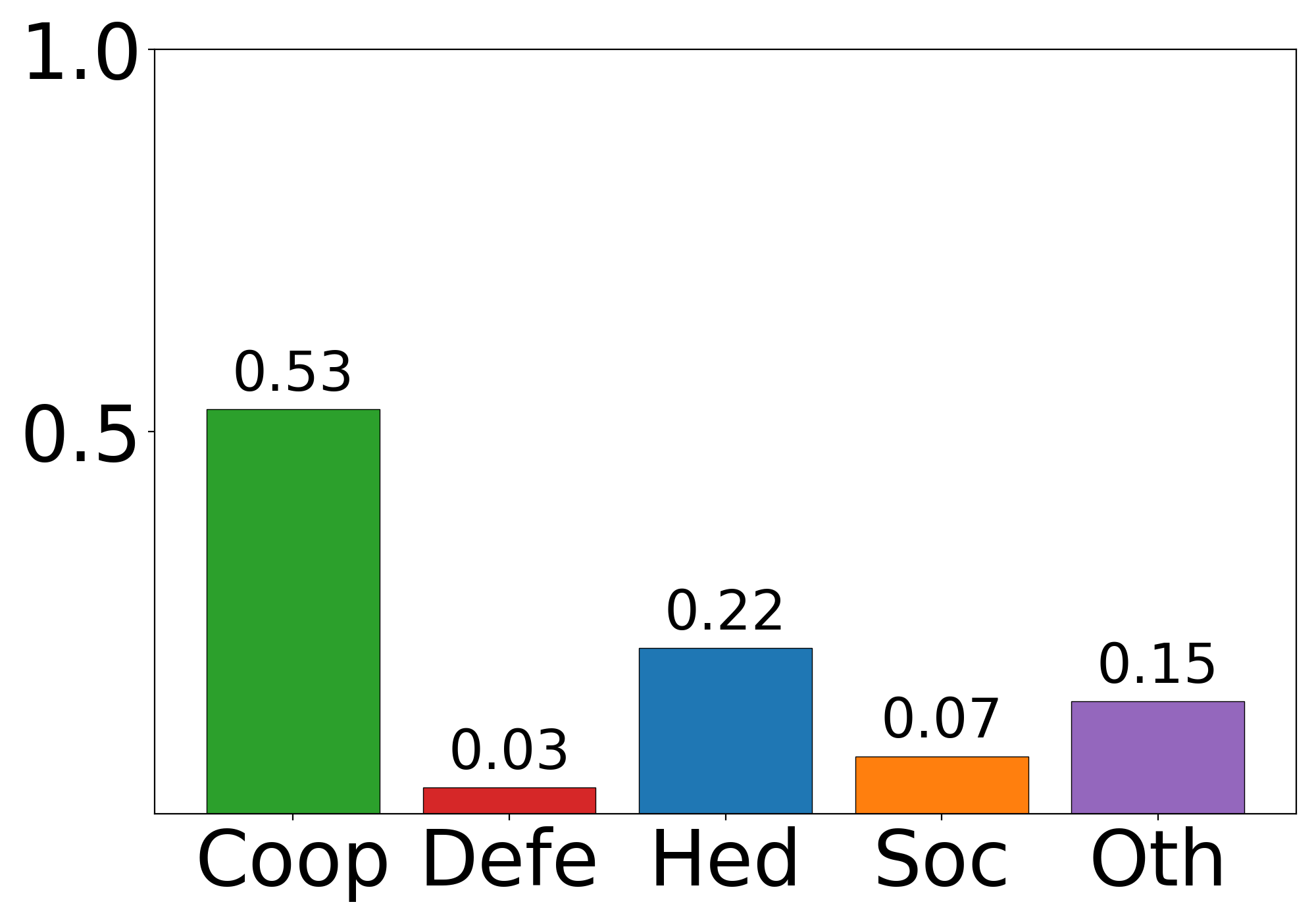}} &
\subcaptionbox{}{\includegraphics[width=\linewidth]{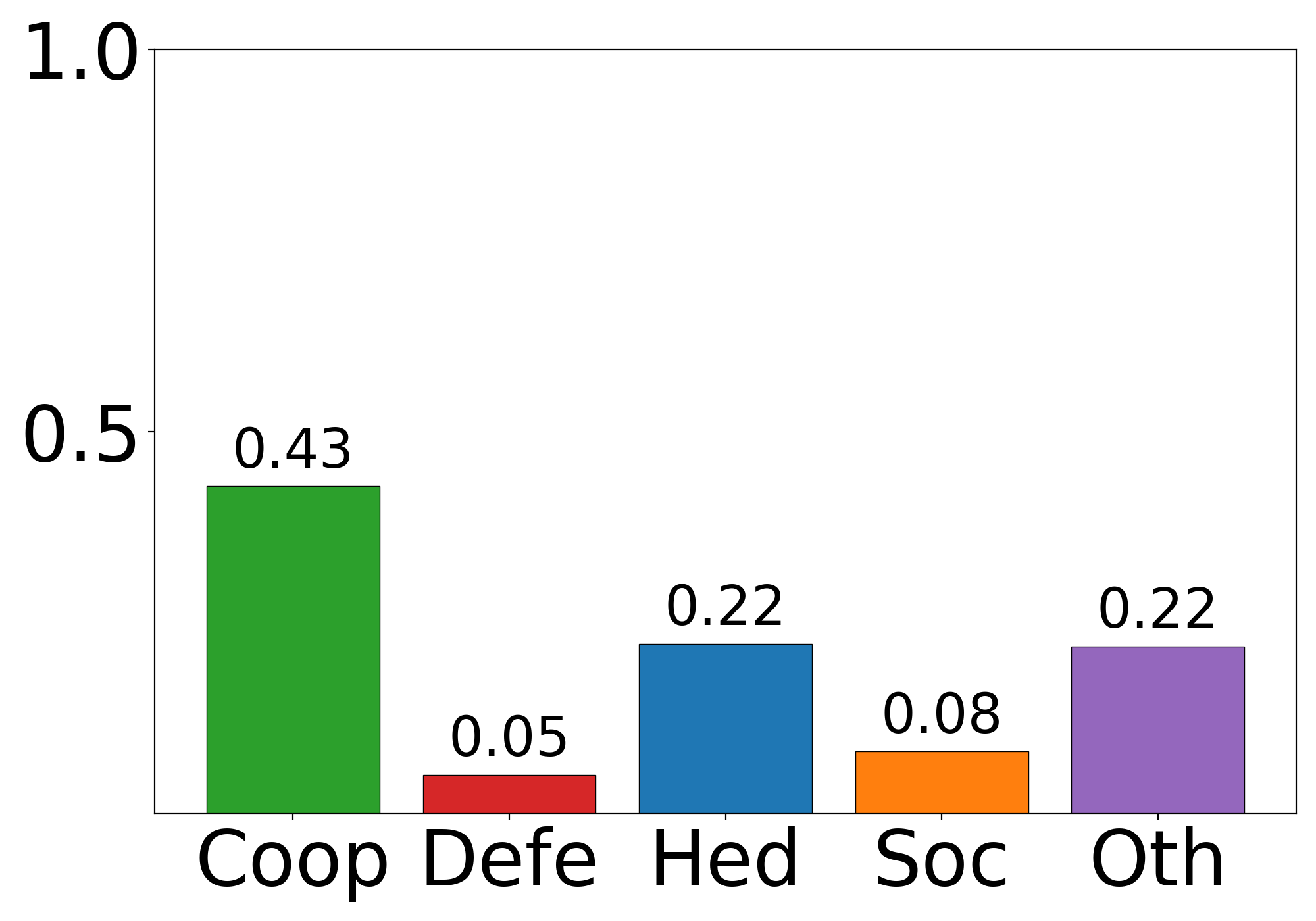}} &
\subcaptionbox{}{\includegraphics[width=\linewidth]{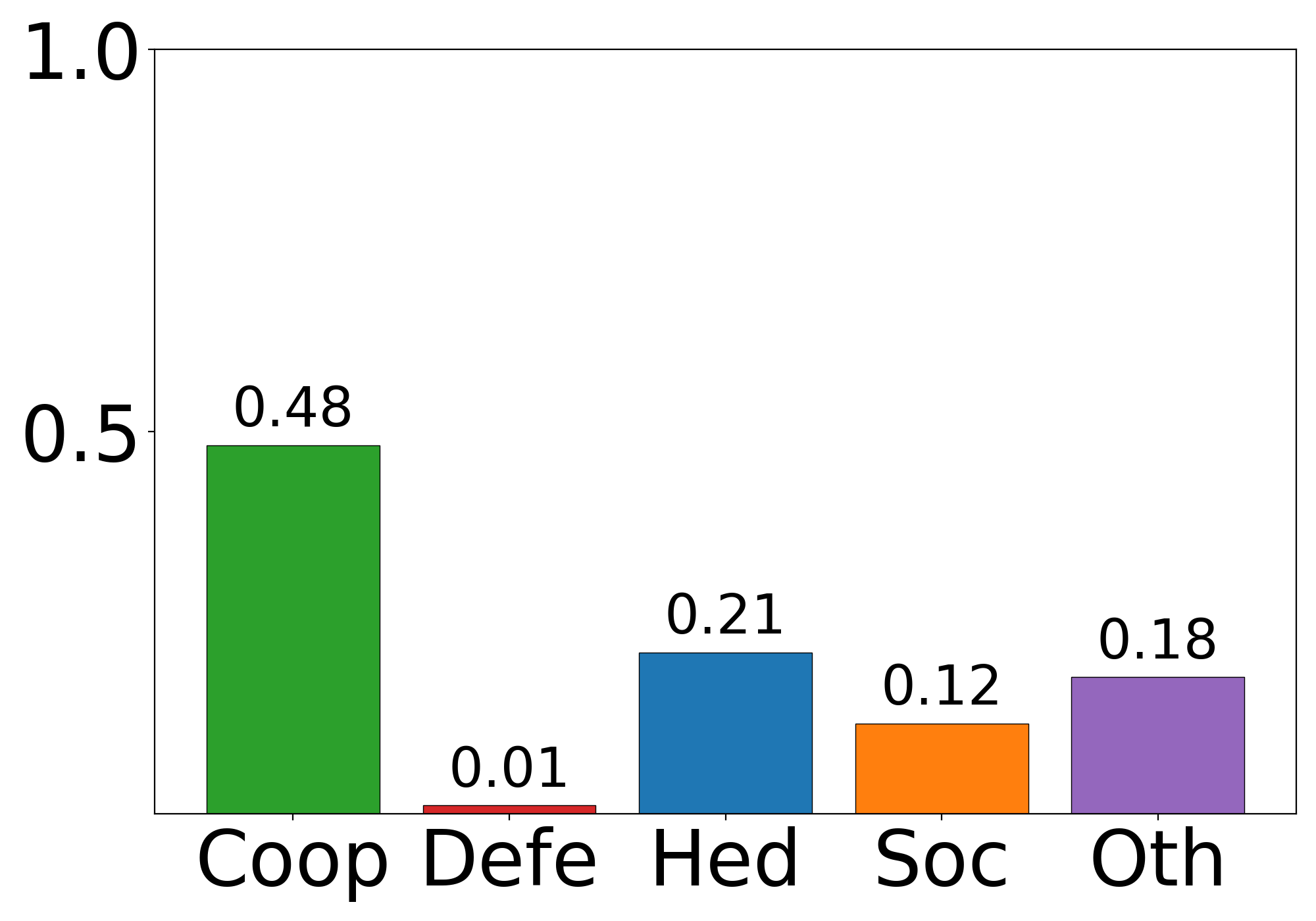}} \\

\RowStart
\subcaptionbox{}{\includegraphics[width=\linewidth]{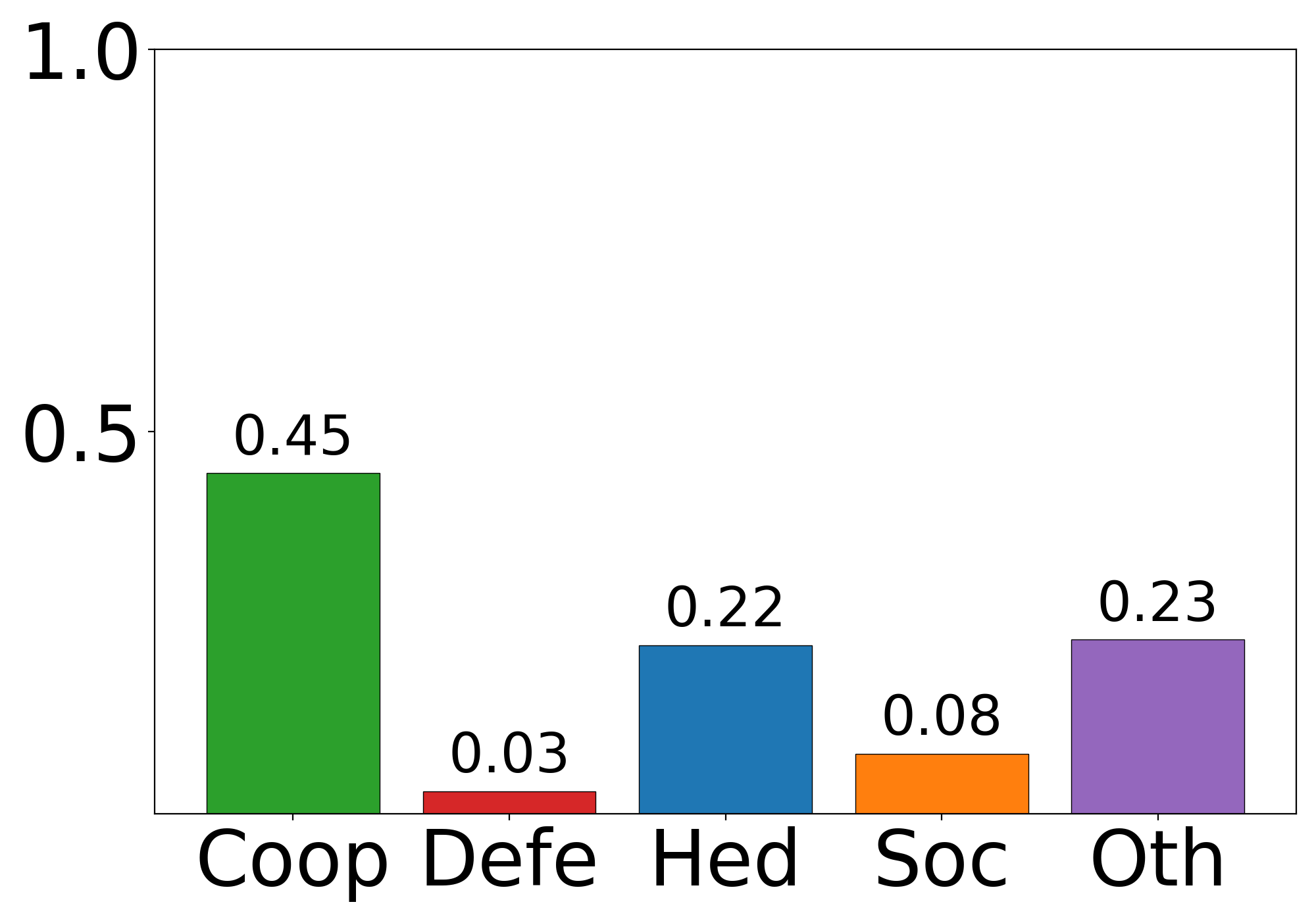}} &
\subcaptionbox{}{\includegraphics[width=\linewidth]{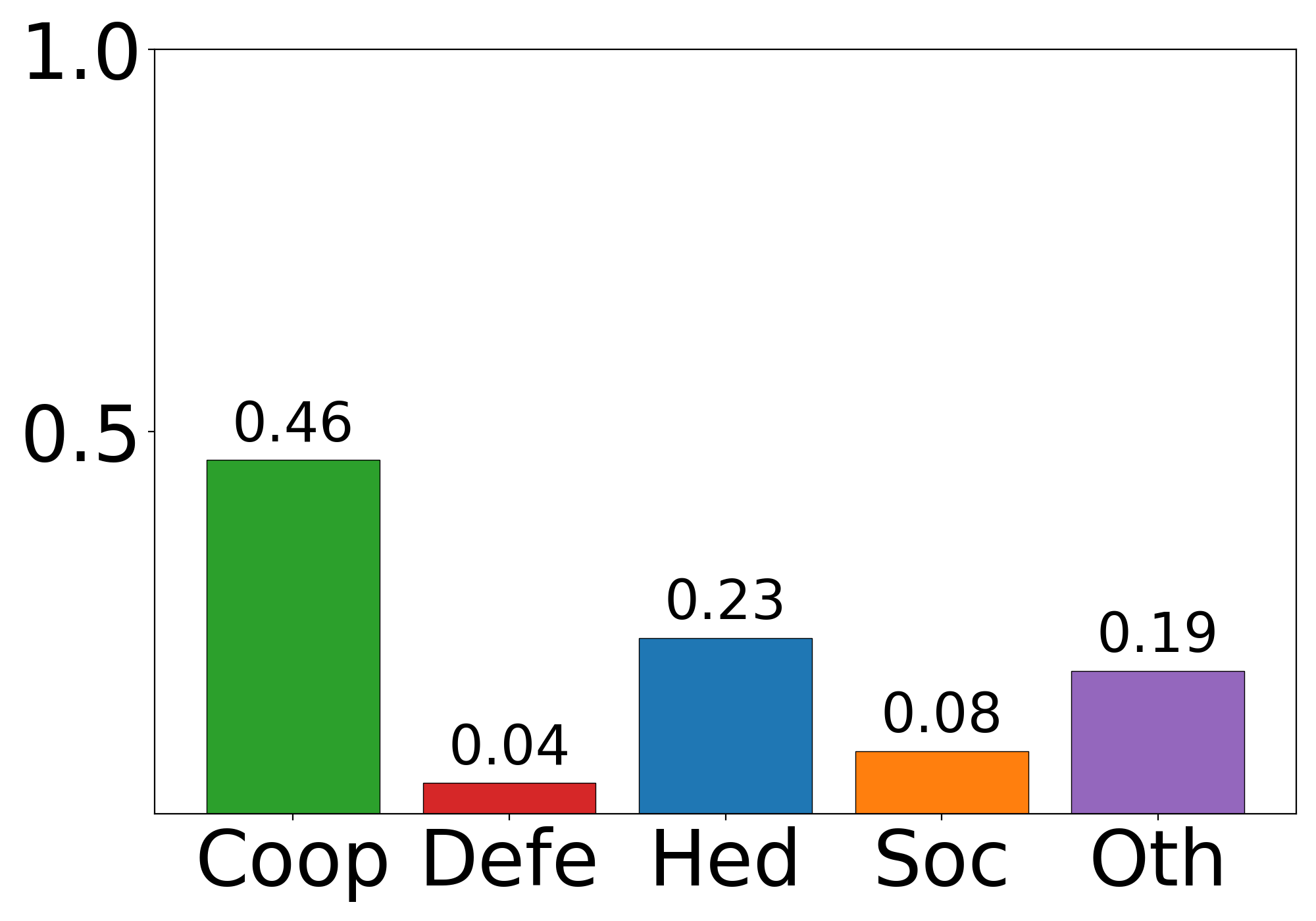}} &
\subcaptionbox{}{\includegraphics[width=\linewidth]{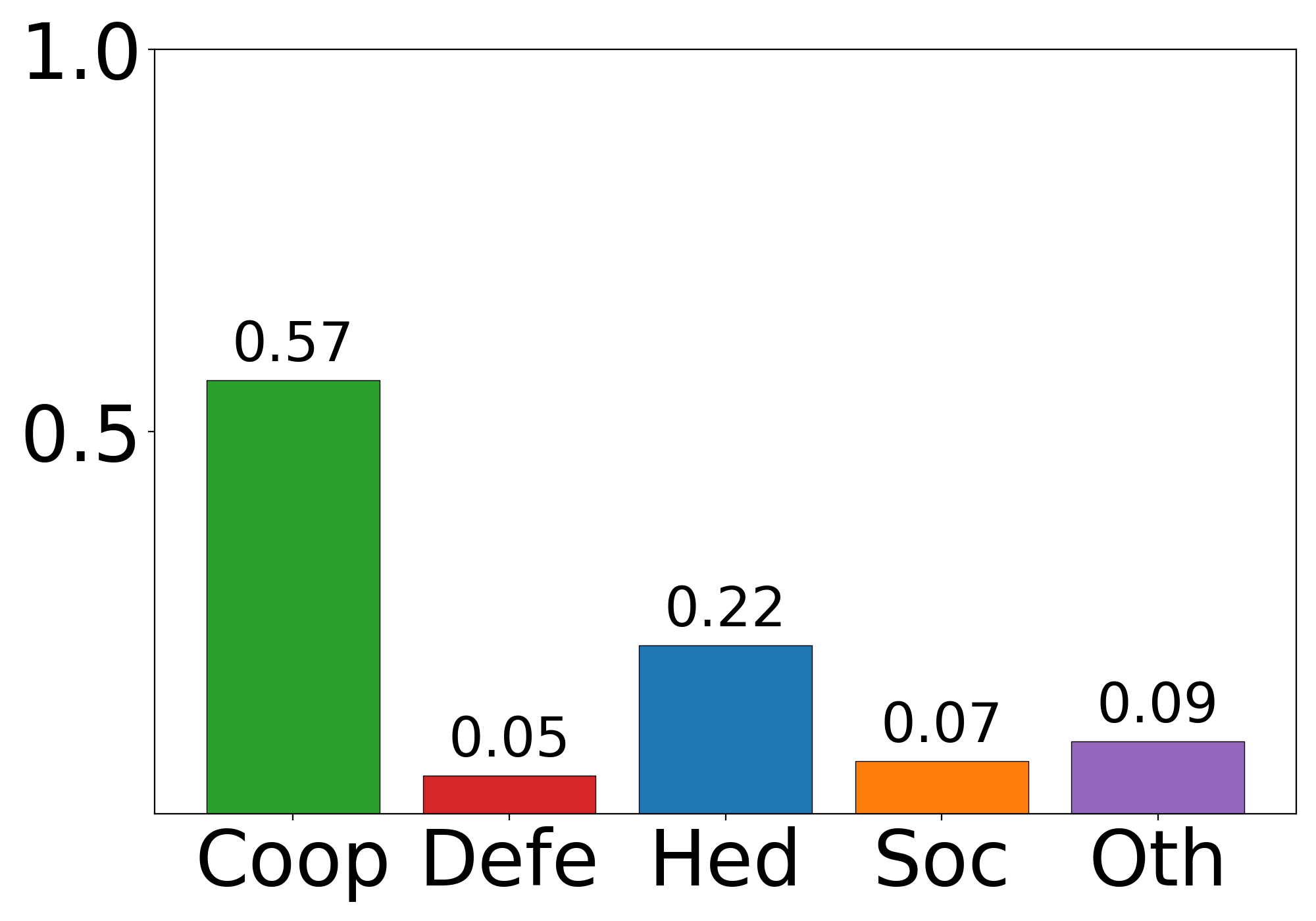}} &
\subcaptionbox{}{\includegraphics[width=\linewidth]{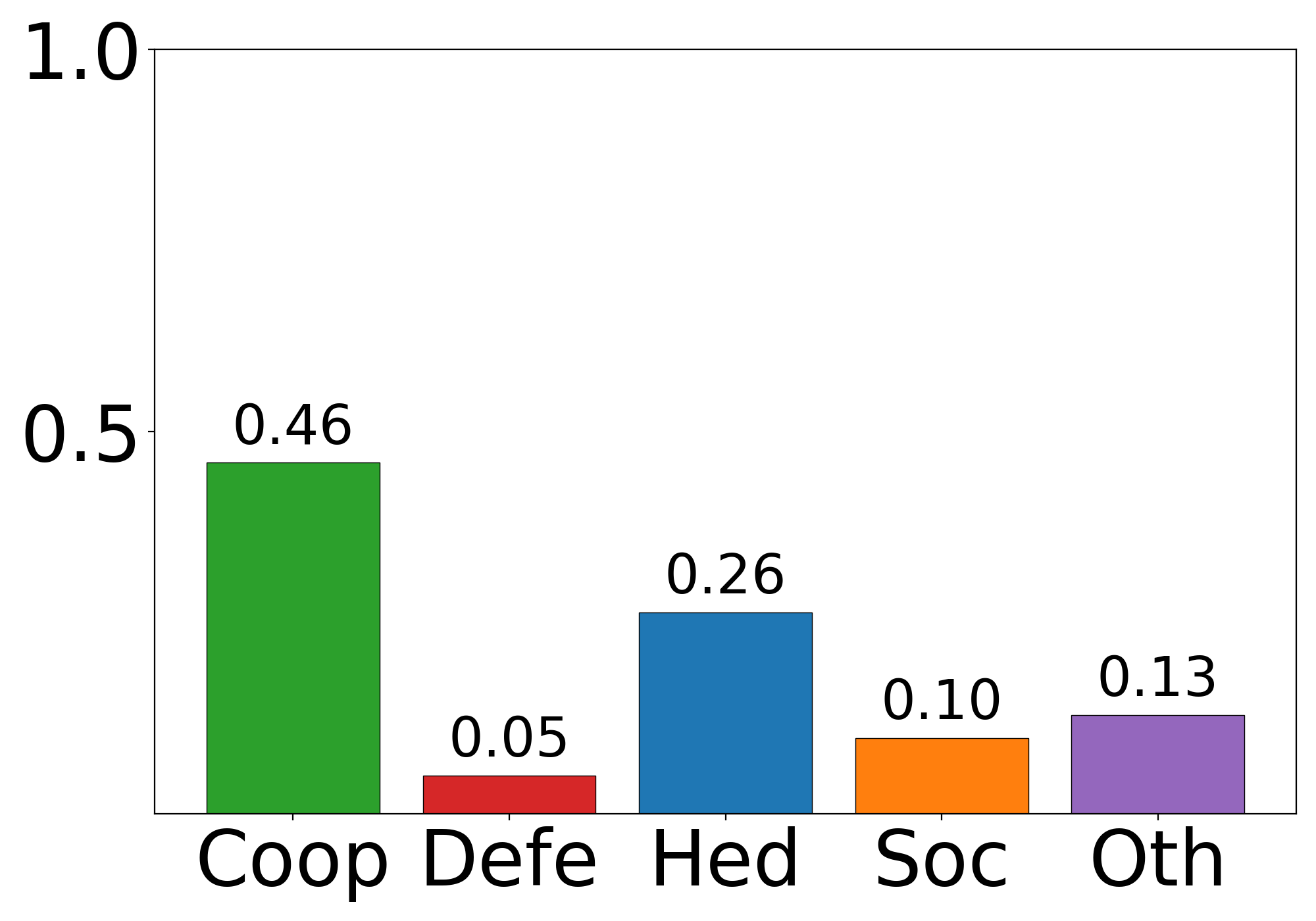}} &
\subcaptionbox{}{\includegraphics[width=\linewidth]{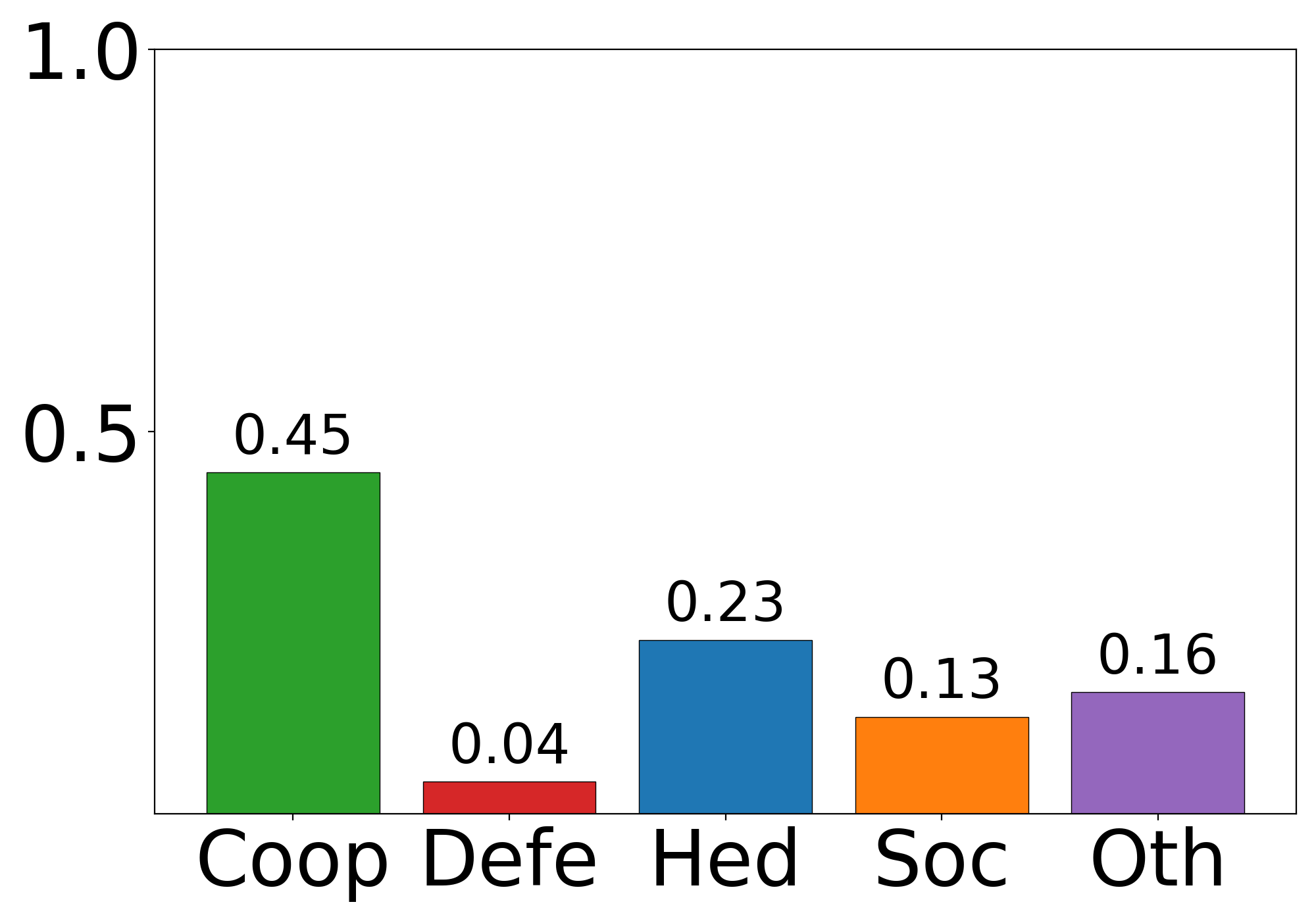}} \\

\RowStart
\subcaptionbox{}{\includegraphics[width=\linewidth]{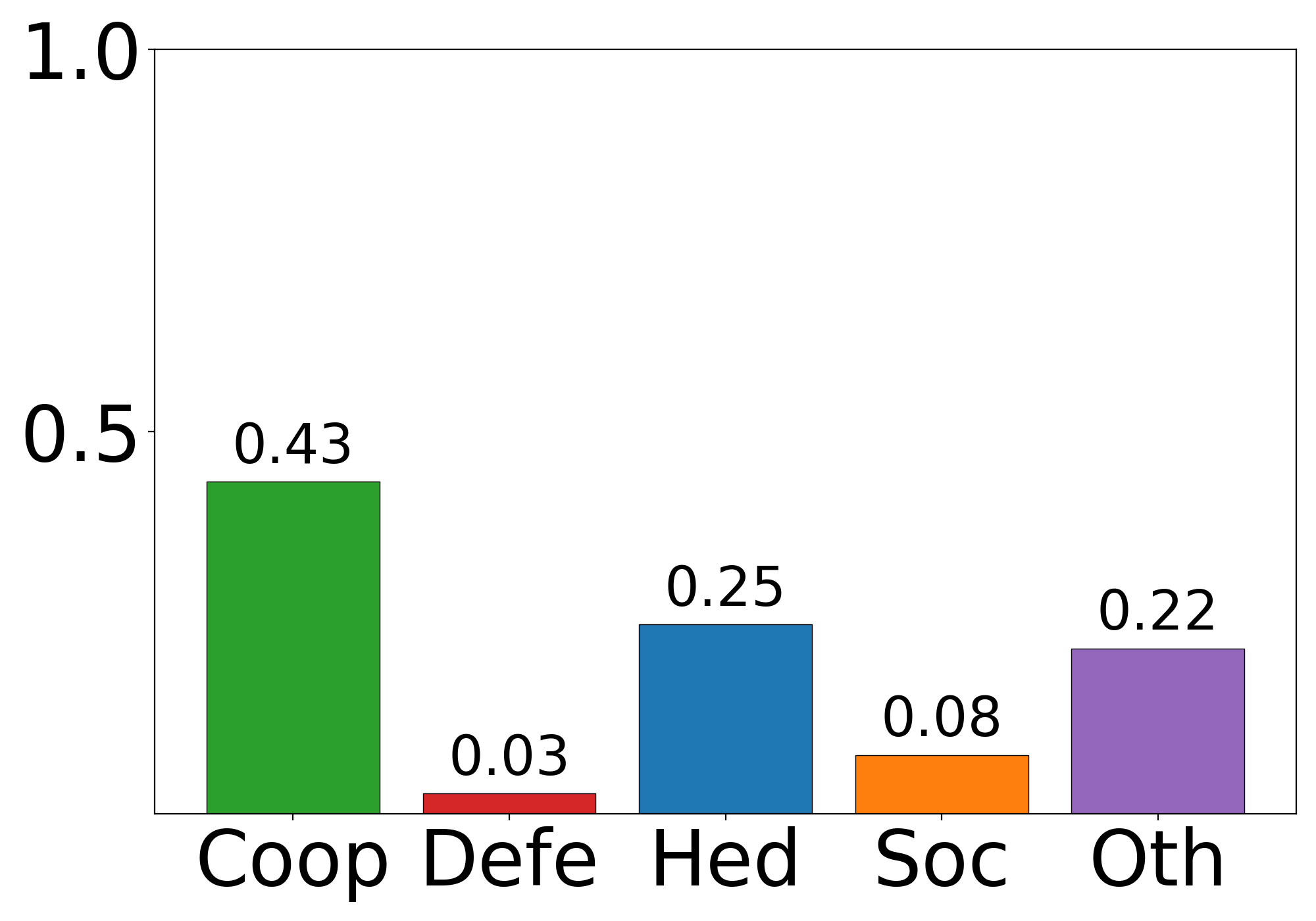}} &
\subcaptionbox{}{\includegraphics[width=\linewidth]{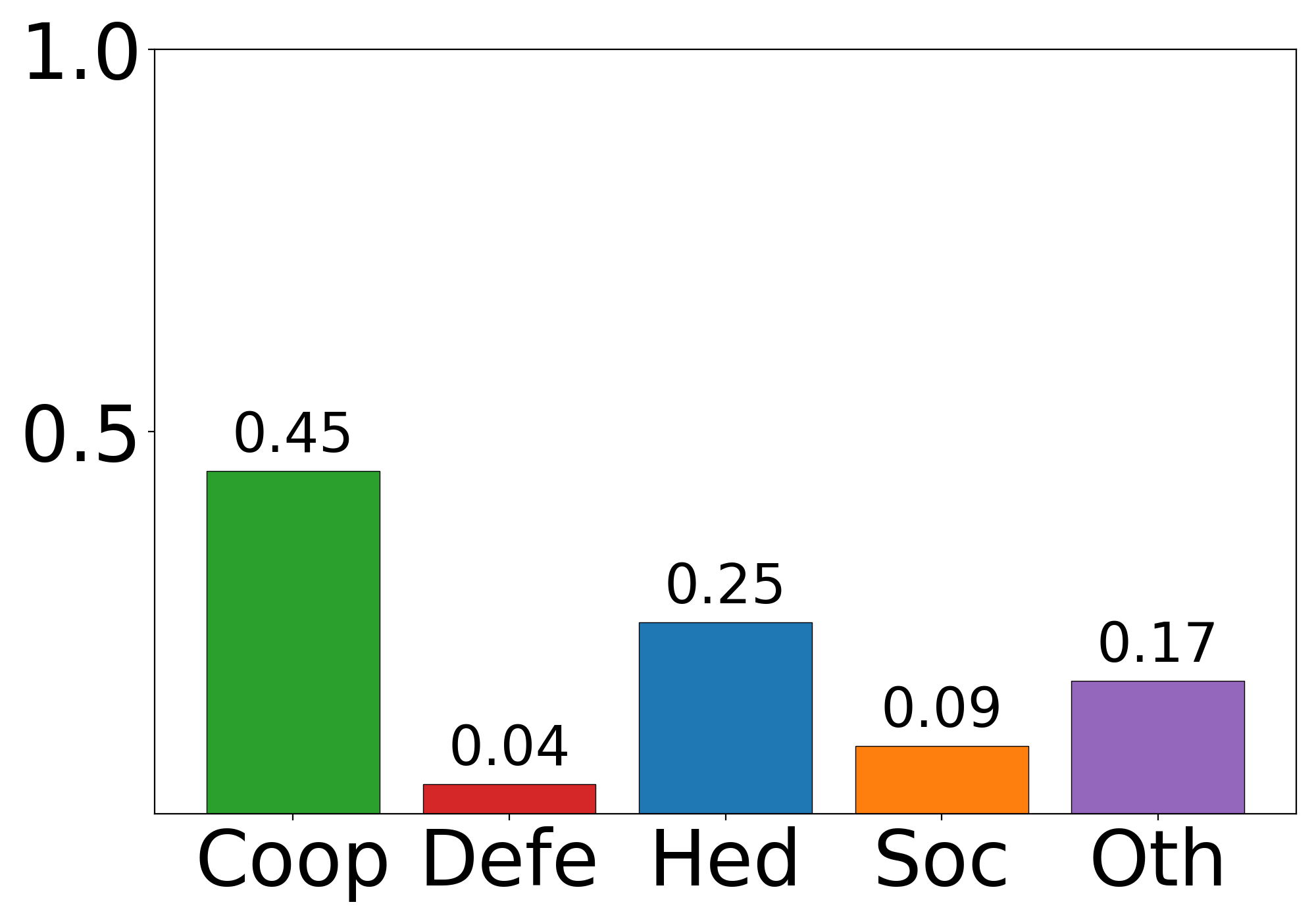}} &
\subcaptionbox{}{\includegraphics[width=\linewidth]{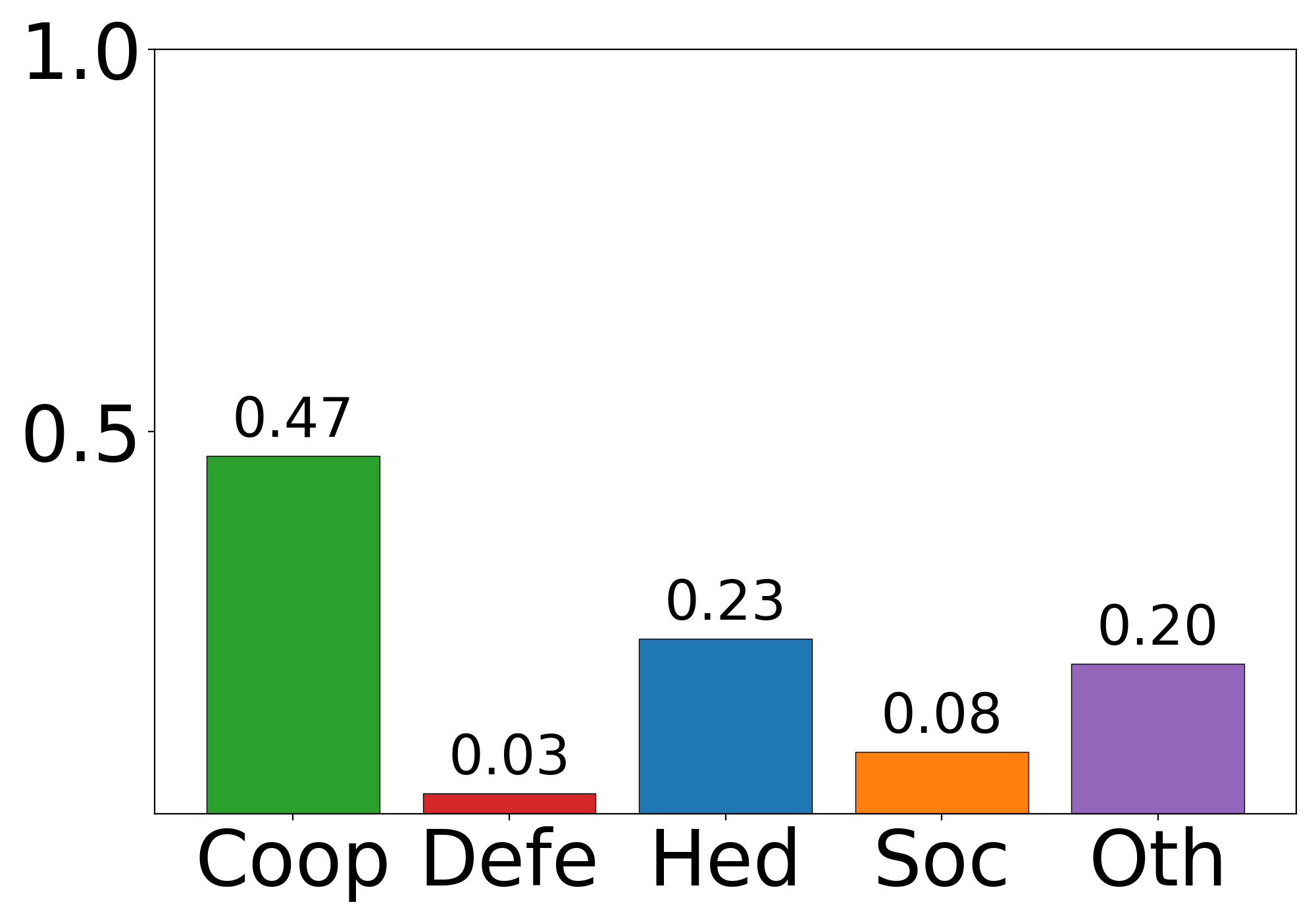}} &
\subcaptionbox{}{\includegraphics[width=\linewidth]{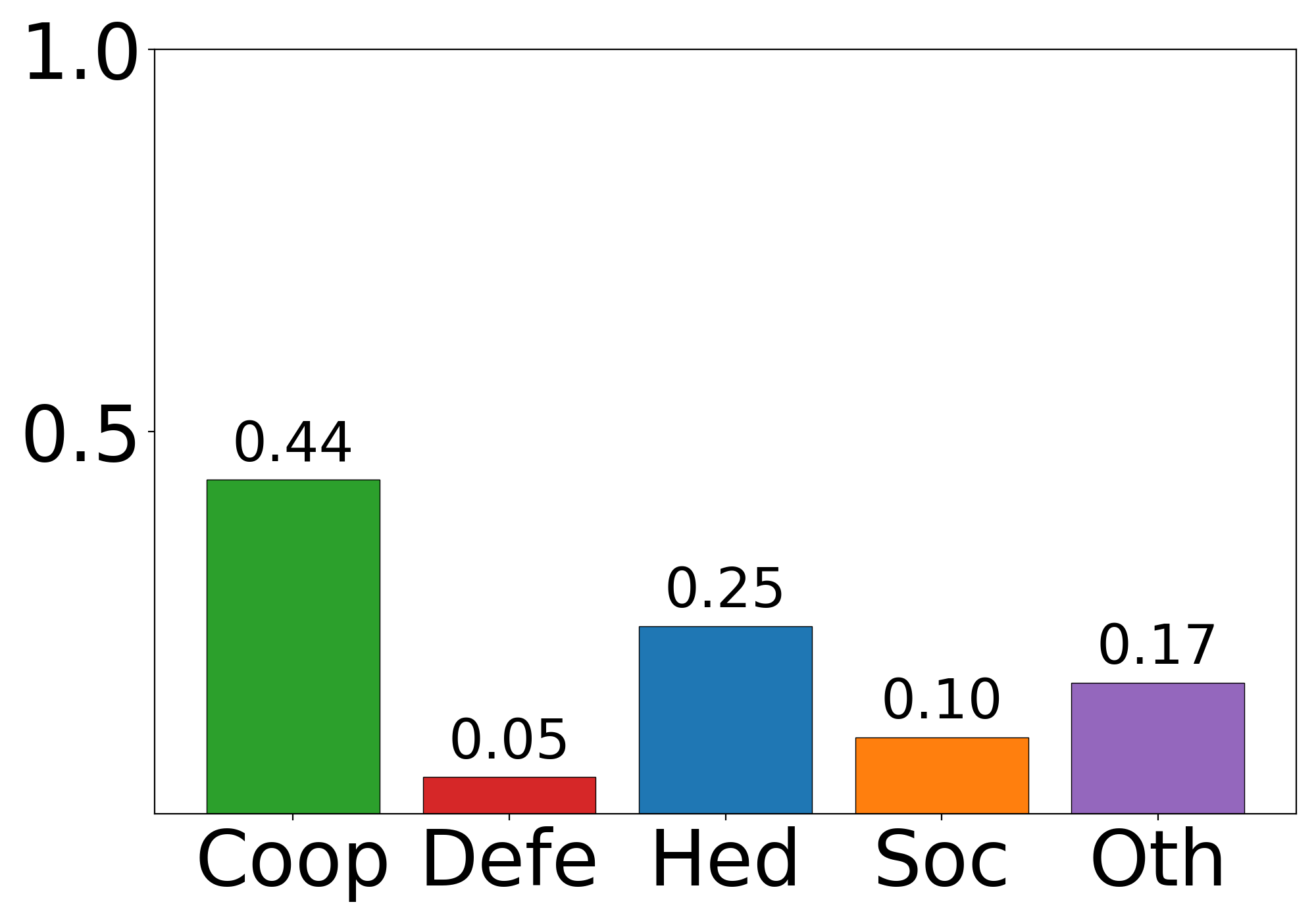}} &
\subcaptionbox{}{\includegraphics[width=\linewidth]{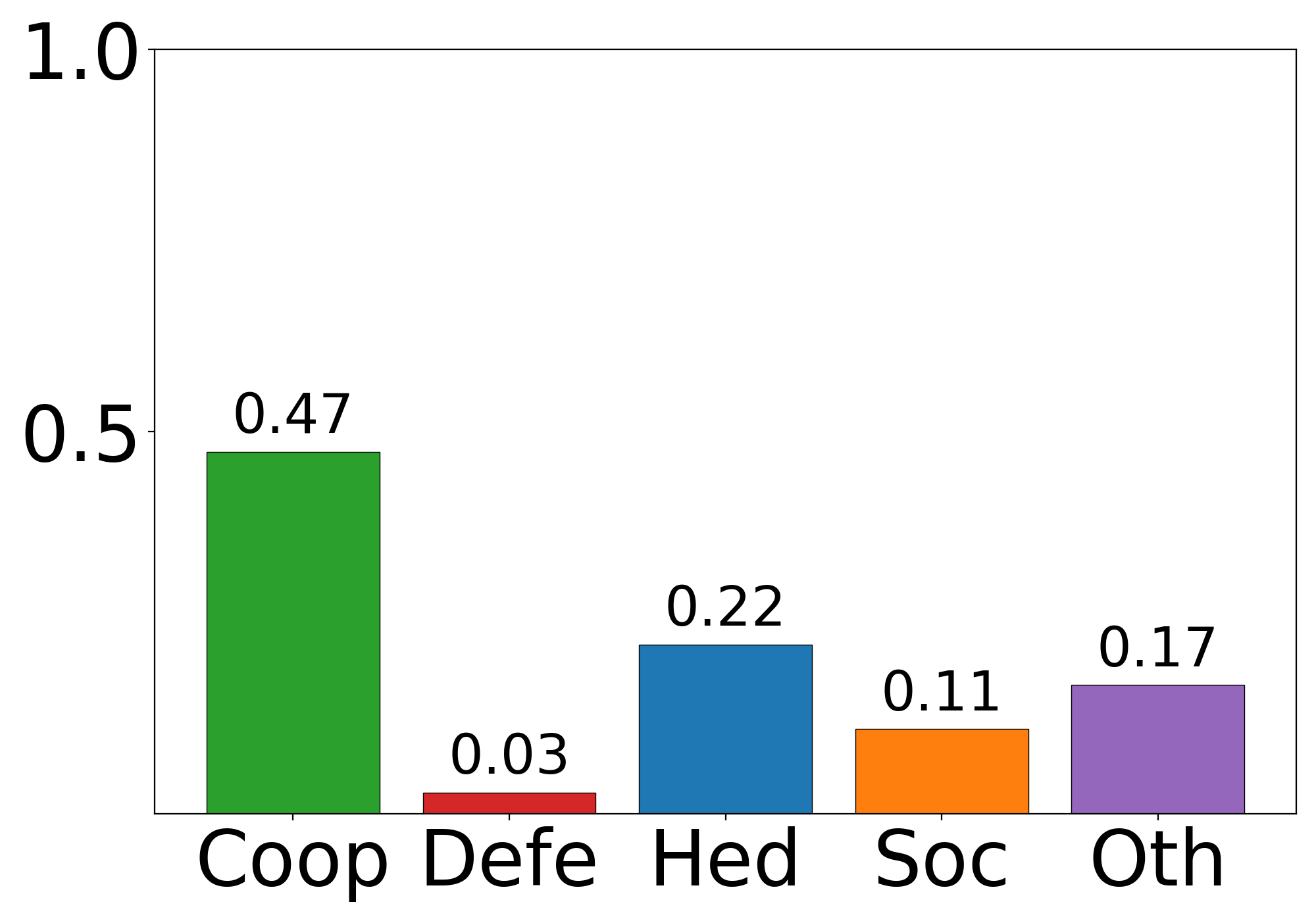}} \\

\RowStart
\subcaptionbox{}{\includegraphics[width=\linewidth]{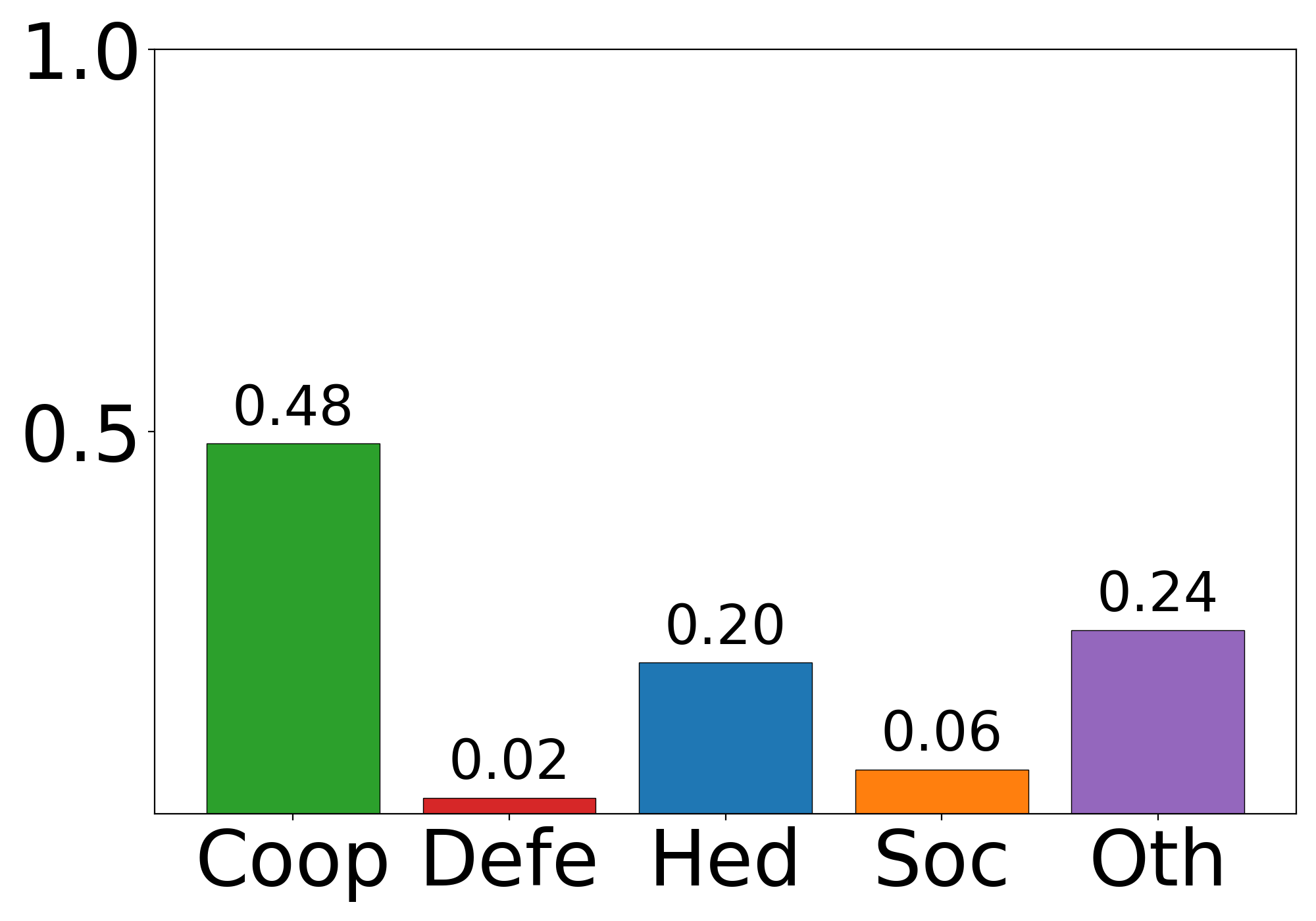}} &
\subcaptionbox{}{\includegraphics[width=\linewidth]{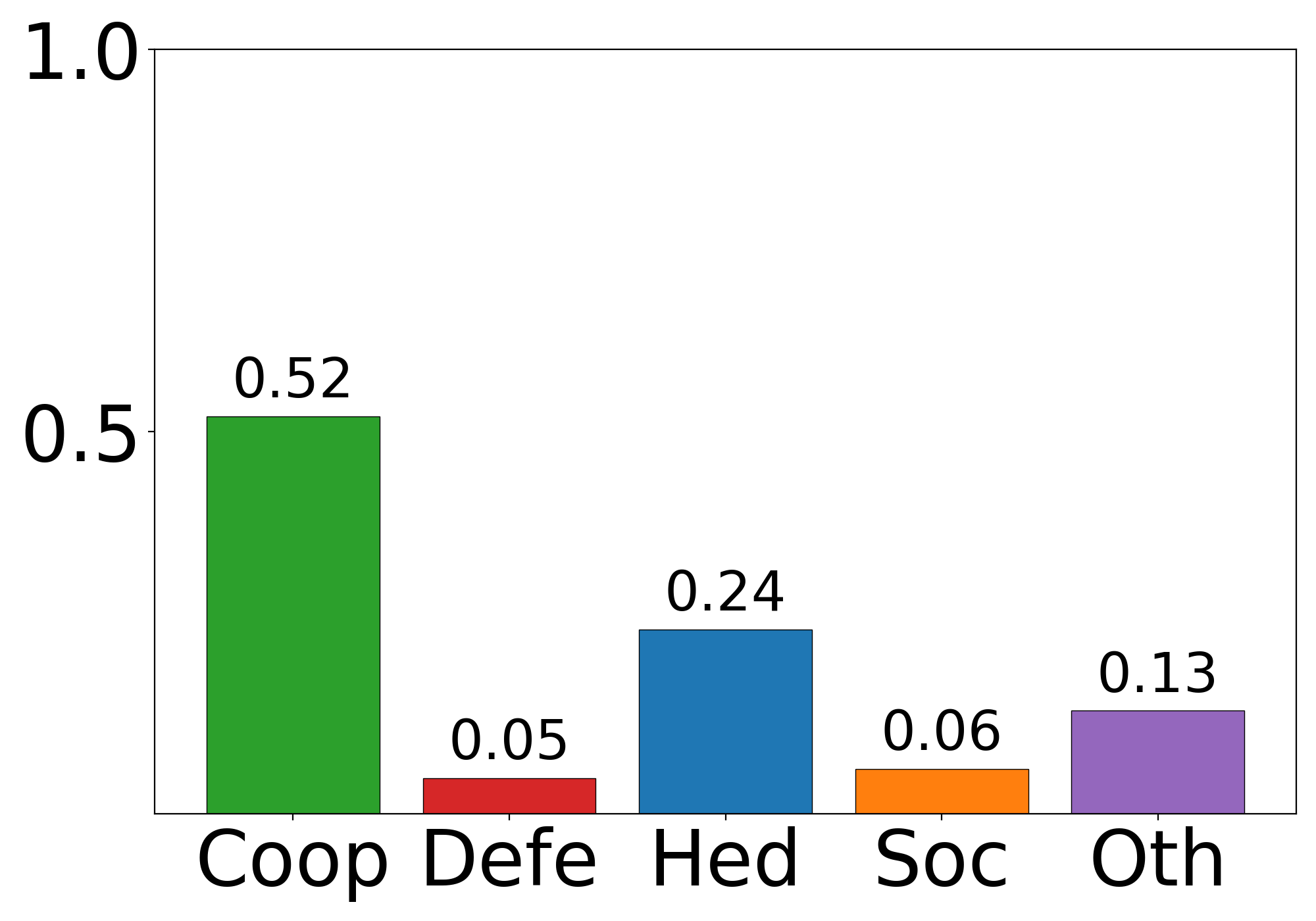}} &
\subcaptionbox{}{\includegraphics[width=\linewidth]{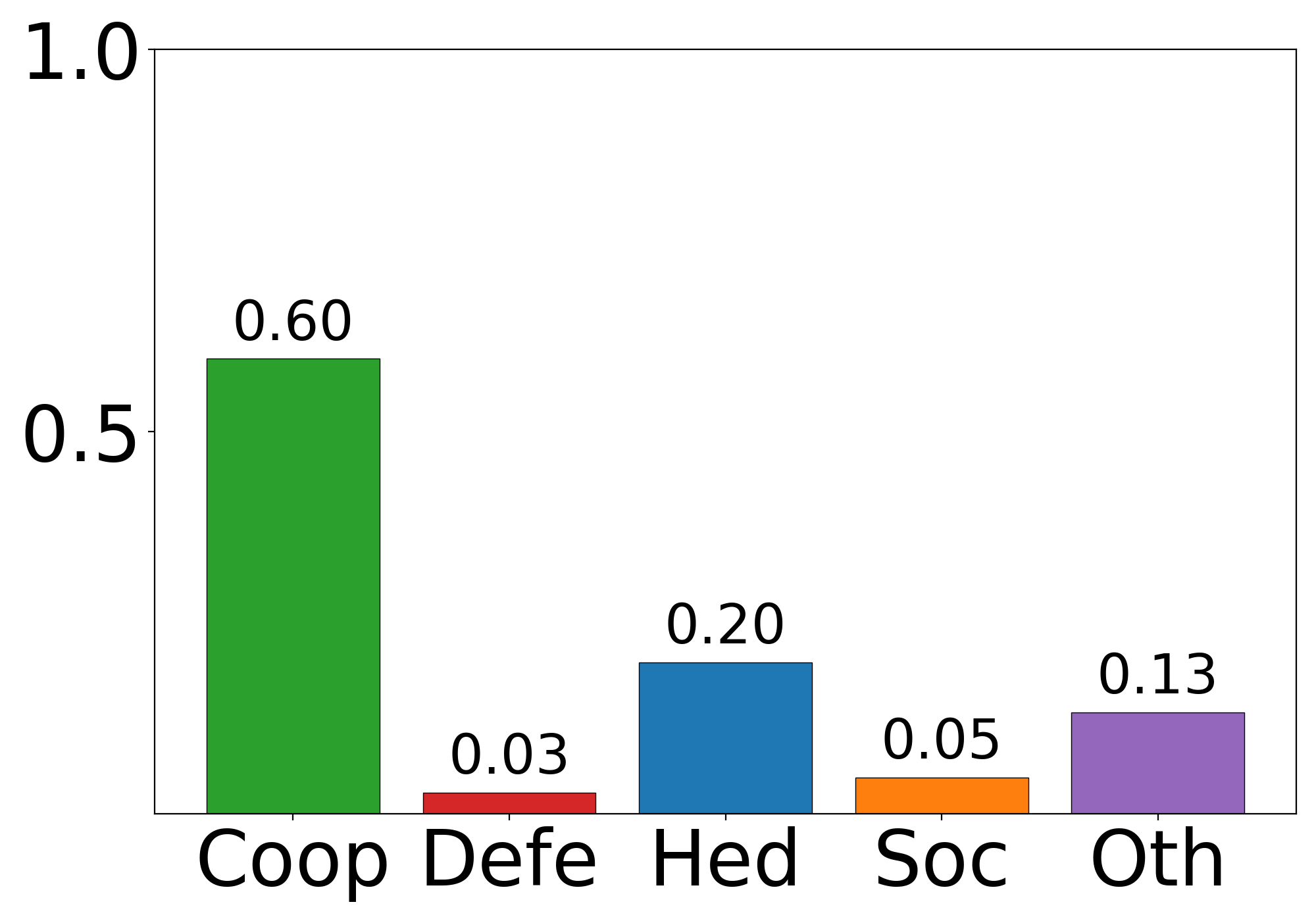}} &
\subcaptionbox{}{\includegraphics[width=\linewidth]{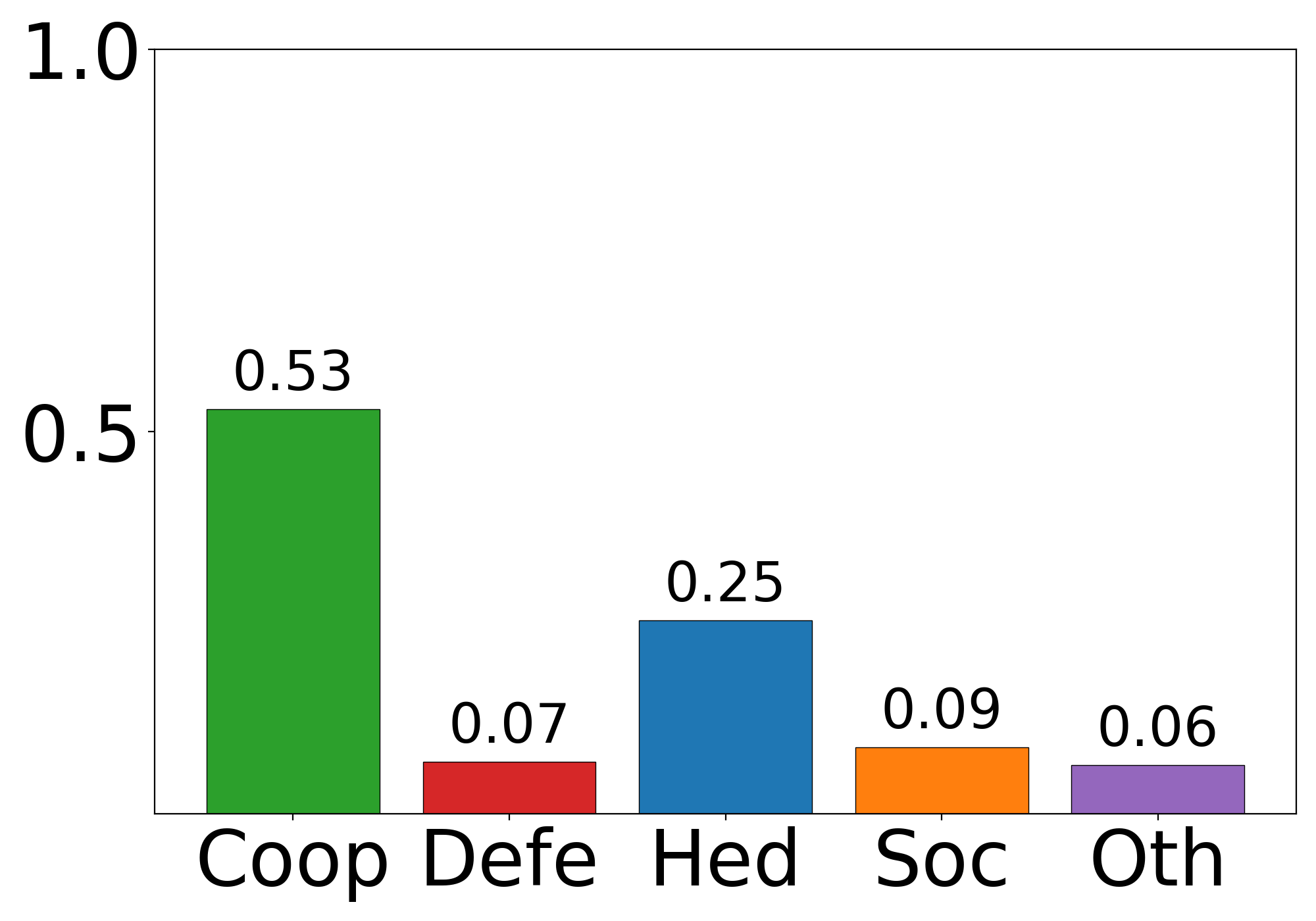}} &
\subcaptionbox{}{\includegraphics[width=\linewidth]{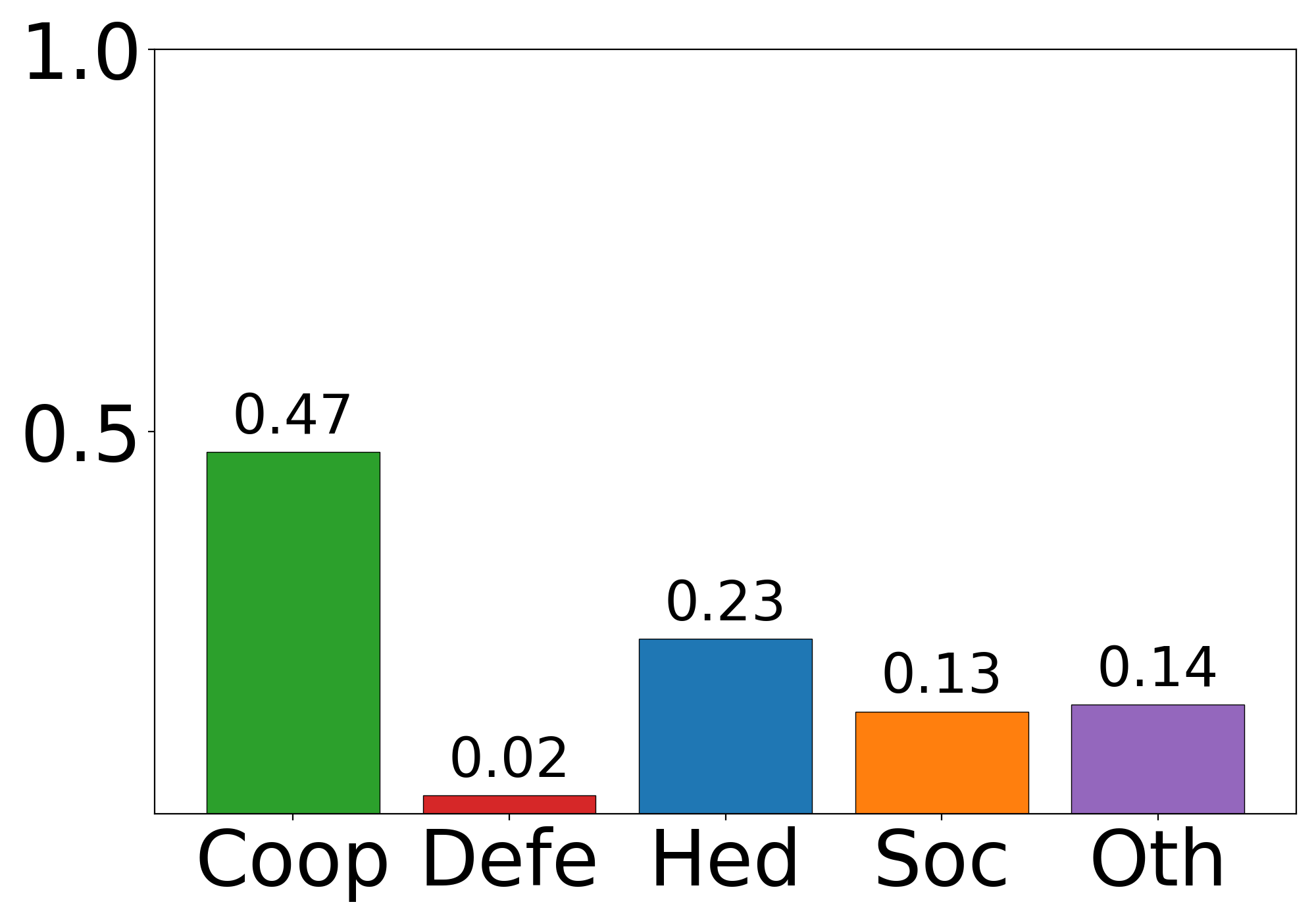}} \\

\RowStart
\subcaptionbox{}{\includegraphics[width=\linewidth]{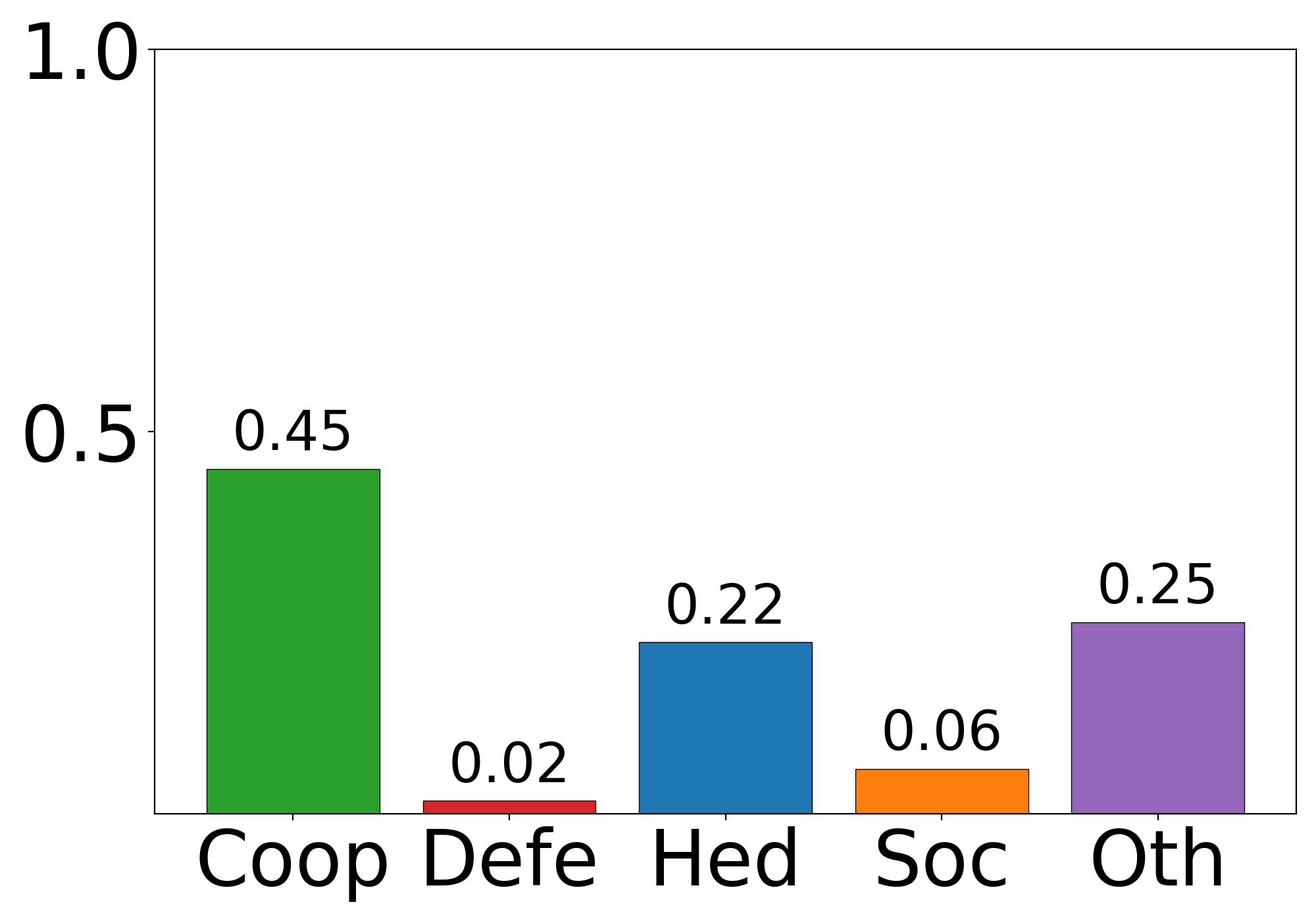}} &
\subcaptionbox{}{\includegraphics[width=\linewidth]{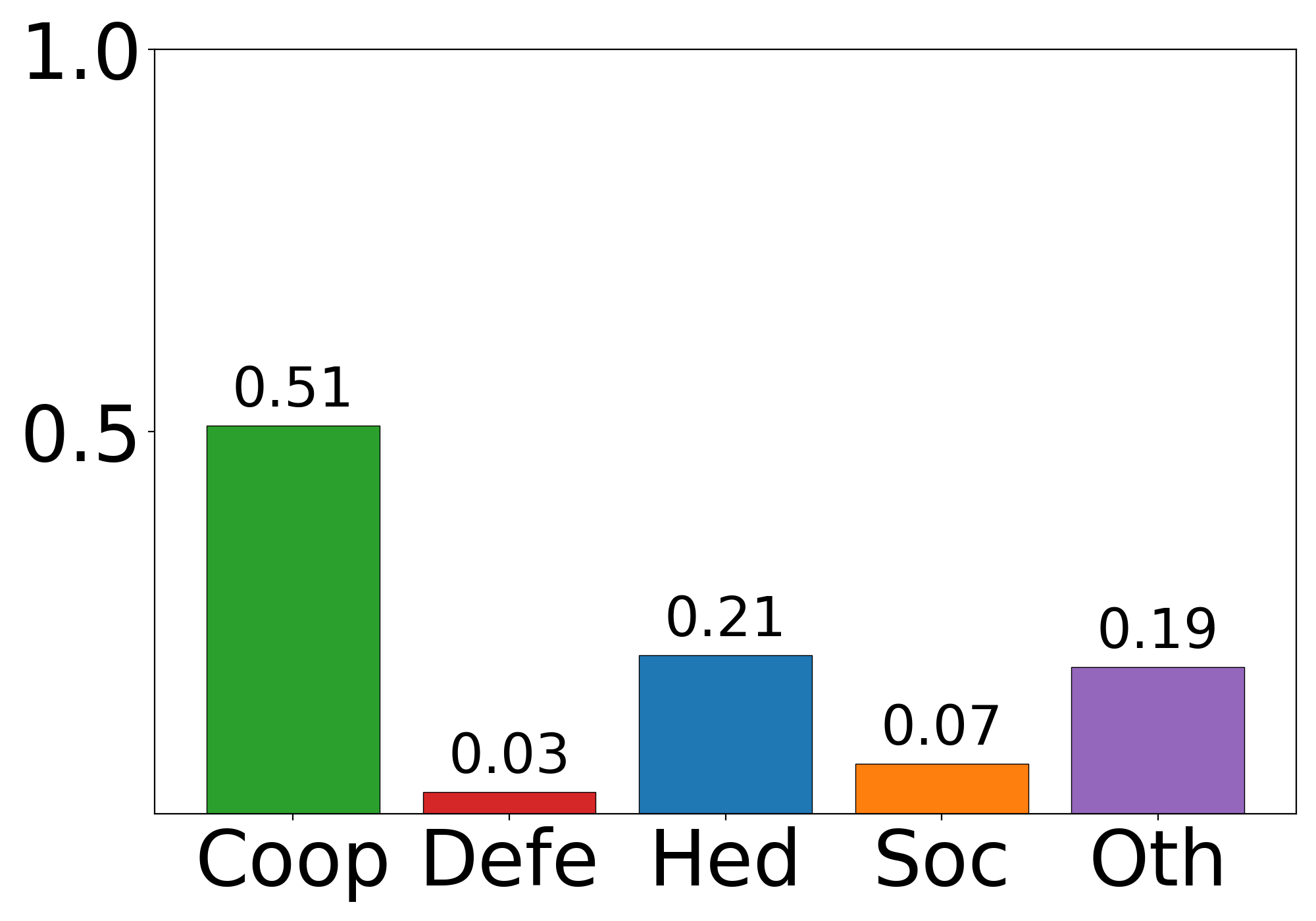}} &
\subcaptionbox{}{\includegraphics[width=\linewidth]{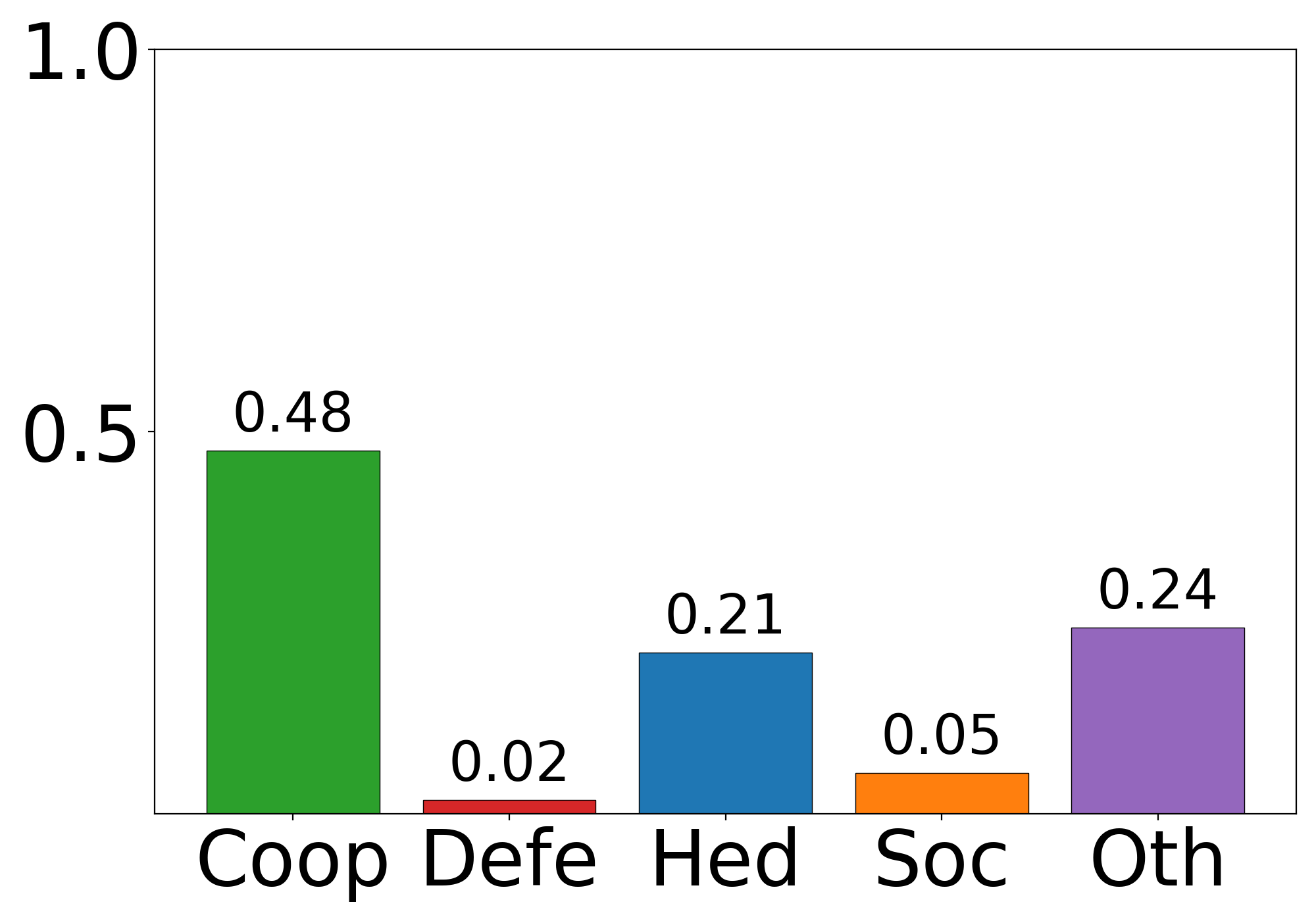}} &
\subcaptionbox{}{\includegraphics[width=\linewidth]{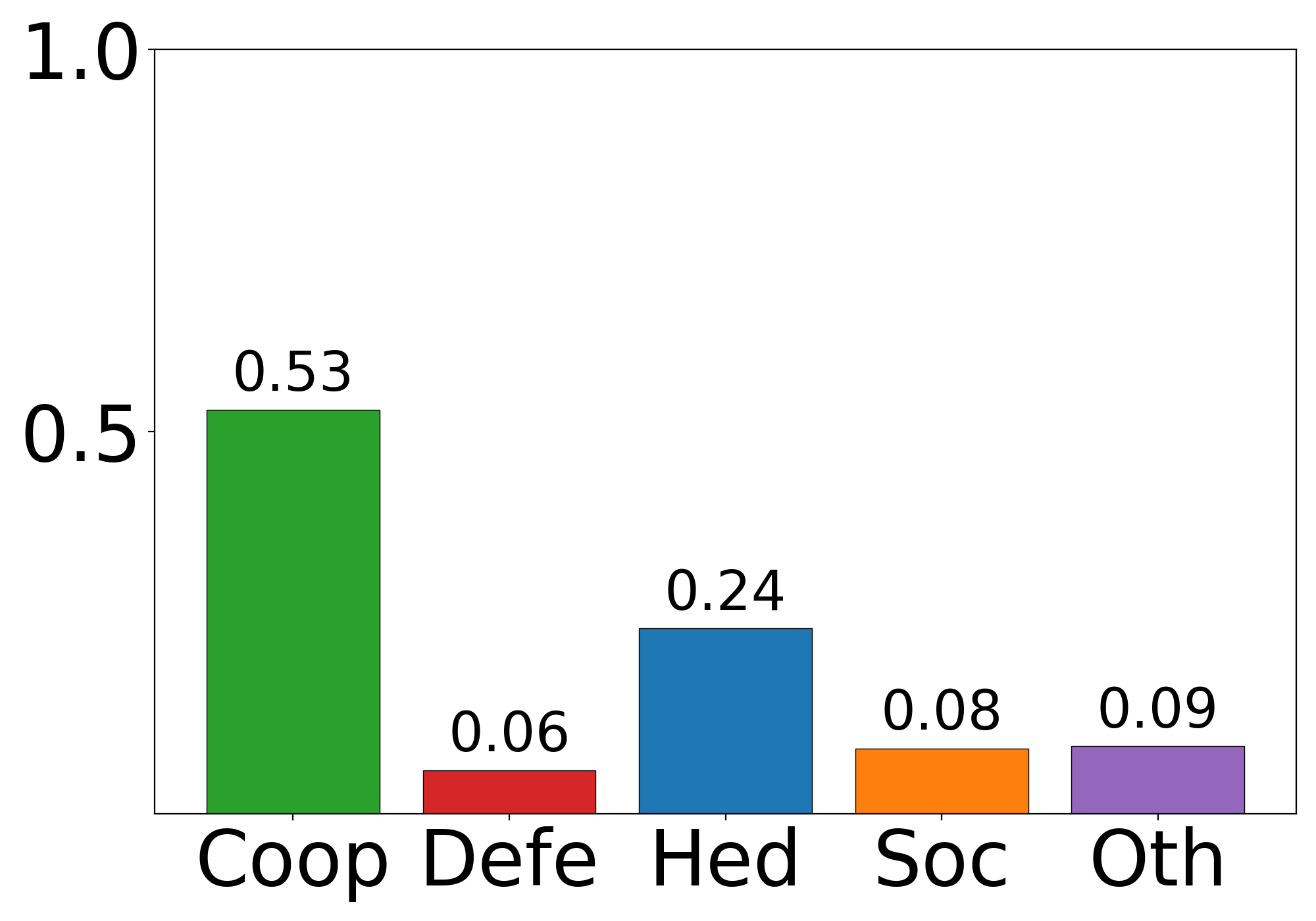}} &
\subcaptionbox{}{\includegraphics[width=\linewidth]{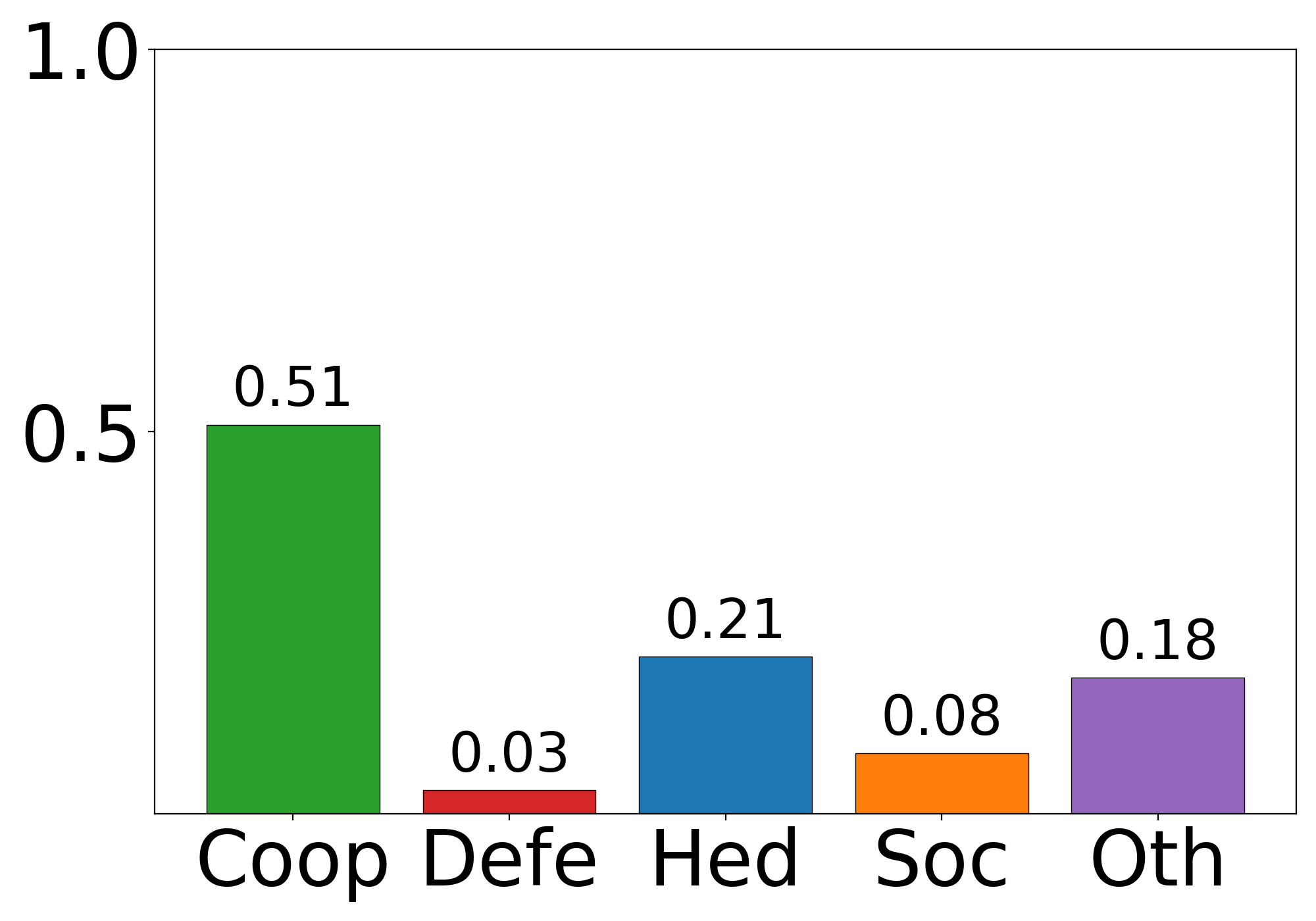}} \\
\bottomrule
\end{tabular}
\caption{Tone--stance profiles of author responses generated by five LLMs (a--e) under nine settings (1--9), together with authentic human responses (1.f). The figure shows average word-weighted percentages (normalized to [0,1]) of cooperative (Coop), defensive (Defe), hedge (Hed), social (Soc), and other (Oth) stances; detailed definitions are provided in §\ref{subsec:respeval_discourse}.}
\label{fig:discourse_tone_stance}
\end{figure*}

\setcounter{settingrow}{0}
\begin{figure*}[ht]
\centering
\fontsize{10}{10}
\begin{tabular}{C{\ImgColW} C{\ImgColW} C{\ImgColW} C{\ImgColW} C{\ImgColW}C{\ImgColW}}
\toprule
\makecell{\selectfont\begin{tabular}{l}\includegraphics[height=1.2em]{fig_model_logo/microsoft_logo.png} Phi-4\hspace{0.5em}\end{tabular}}  &
\makecell{\selectfont\begin{tabular}{l}\includegraphics[height=1.2em]{fig_model_logo/qwen_logo.png} Qwen3\hspace{0.5em}\end{tabular}} &
\makecell{\selectfont\begin{tabular}{l}\includegraphics[height=1.2em]{fig_model_logo/llama_logo.png} Llama-3.3\hspace{0.5em}\end{tabular}} &
\makecell{\selectfont\begin{tabular}{l}\includegraphics[height=1.2em]{fig_model_logo/deepseek_logo.png} DeepSeek\hspace{0.5em}\end{tabular}
} &
\makecell{\selectfont\begin{tabular}{l}\includegraphics[height=1.2em]{fig_model_logo/gpt_logo.jpg} GPT-4o\hspace{0.5em}\end{tabular}}
&
\makecell{\selectfont\begin{tabular}{l}\includegraphics[height=1.5em]{fig_model_logo/human_logo.png} Human\hspace{0.5em}\end{tabular}}\\
\midrule

\RowStart
\subcaptionbox{}{\includegraphics[width=\linewidth]{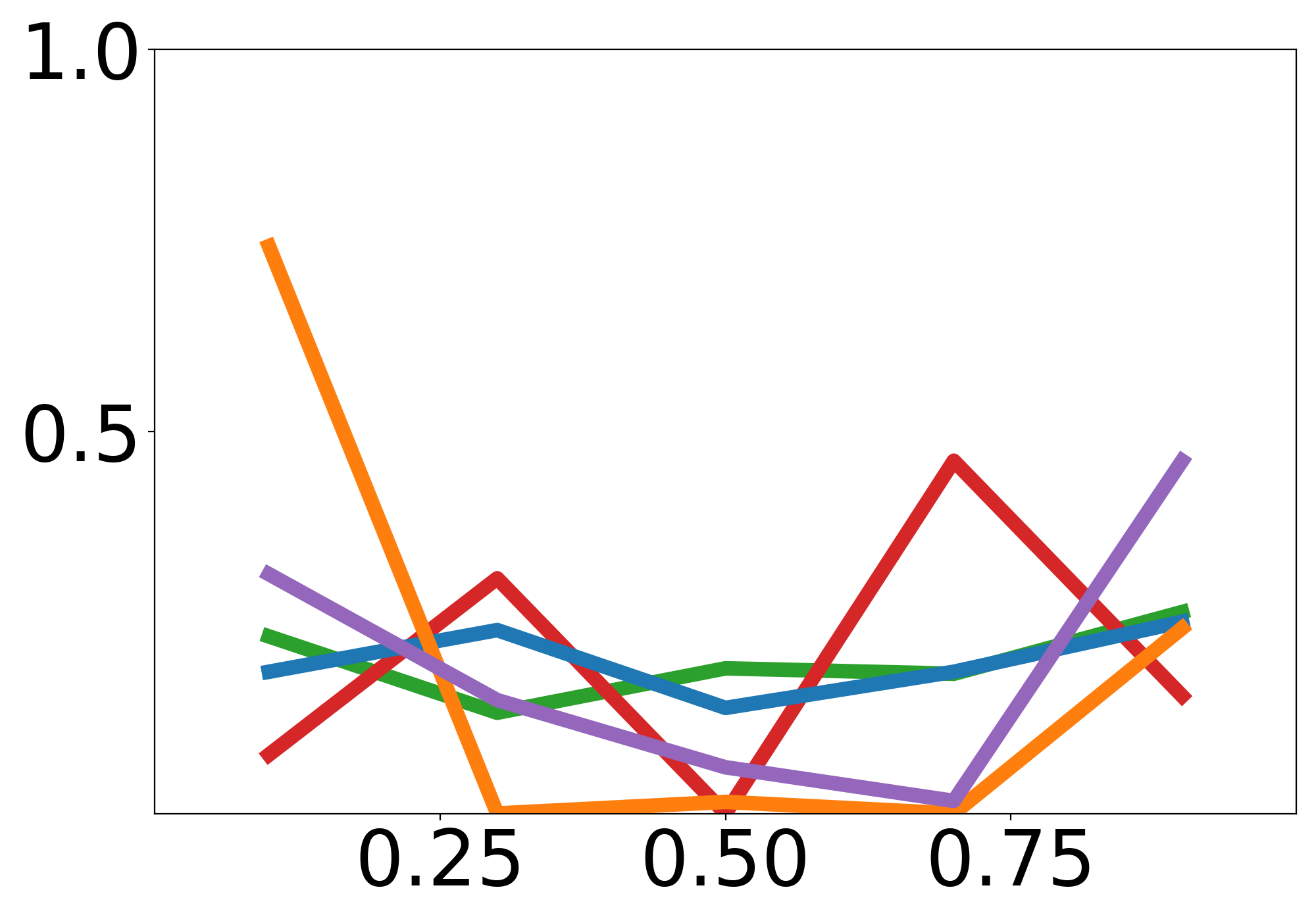}} &
\subcaptionbox{}{\includegraphics[width=\linewidth]{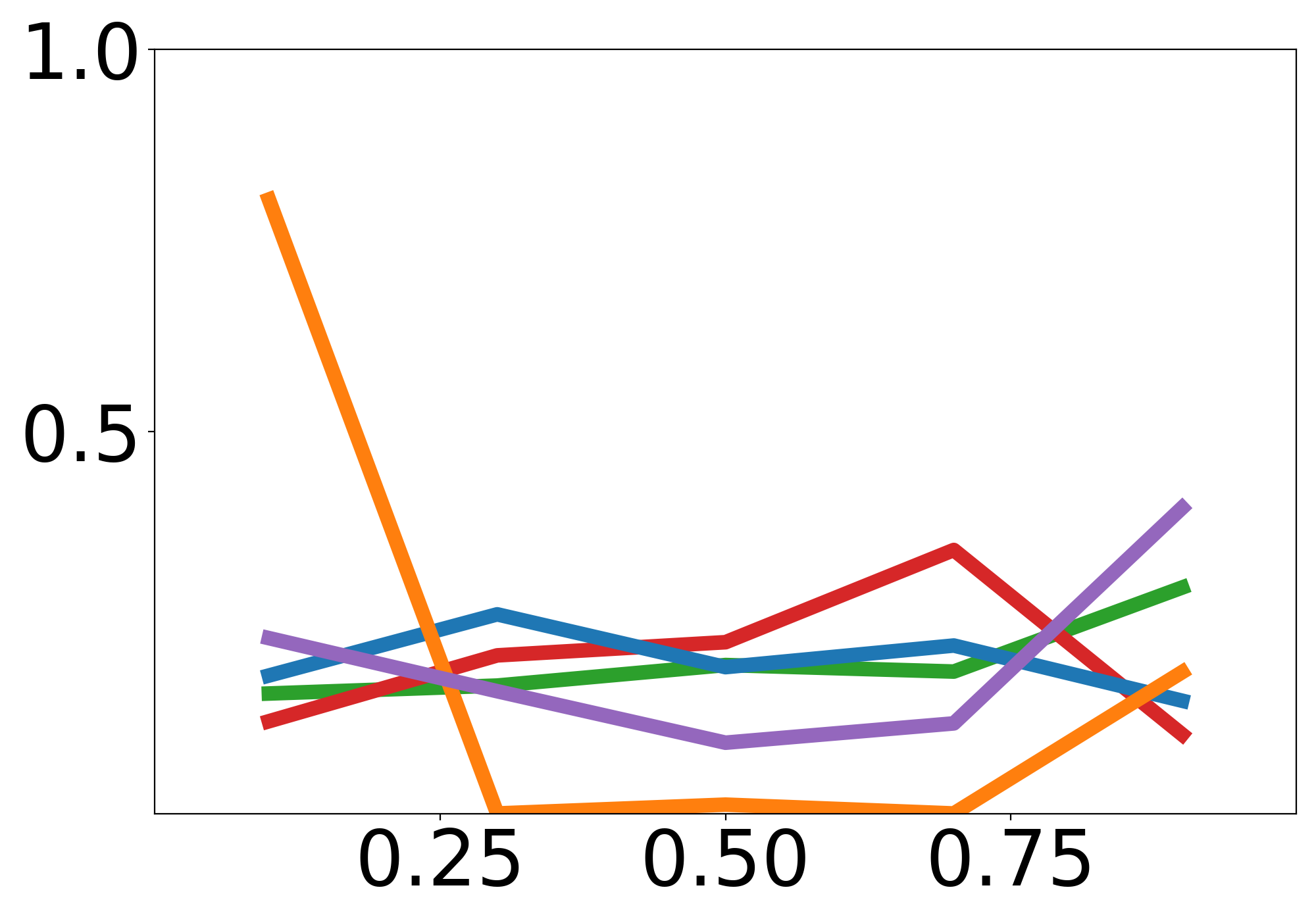}} &
\subcaptionbox{}{\includegraphics[width=\linewidth]{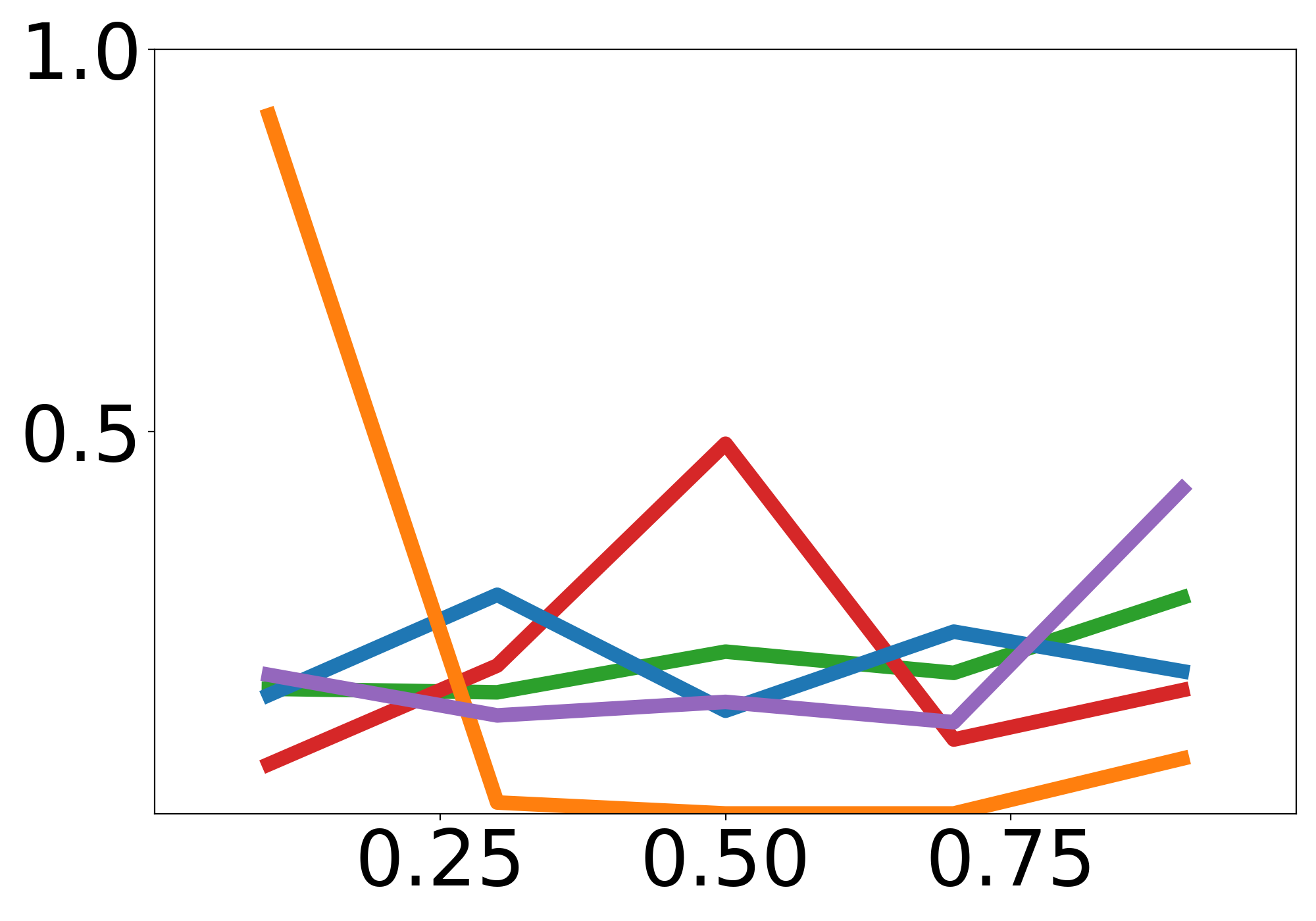}} &
\subcaptionbox{}{\includegraphics[width=\linewidth]{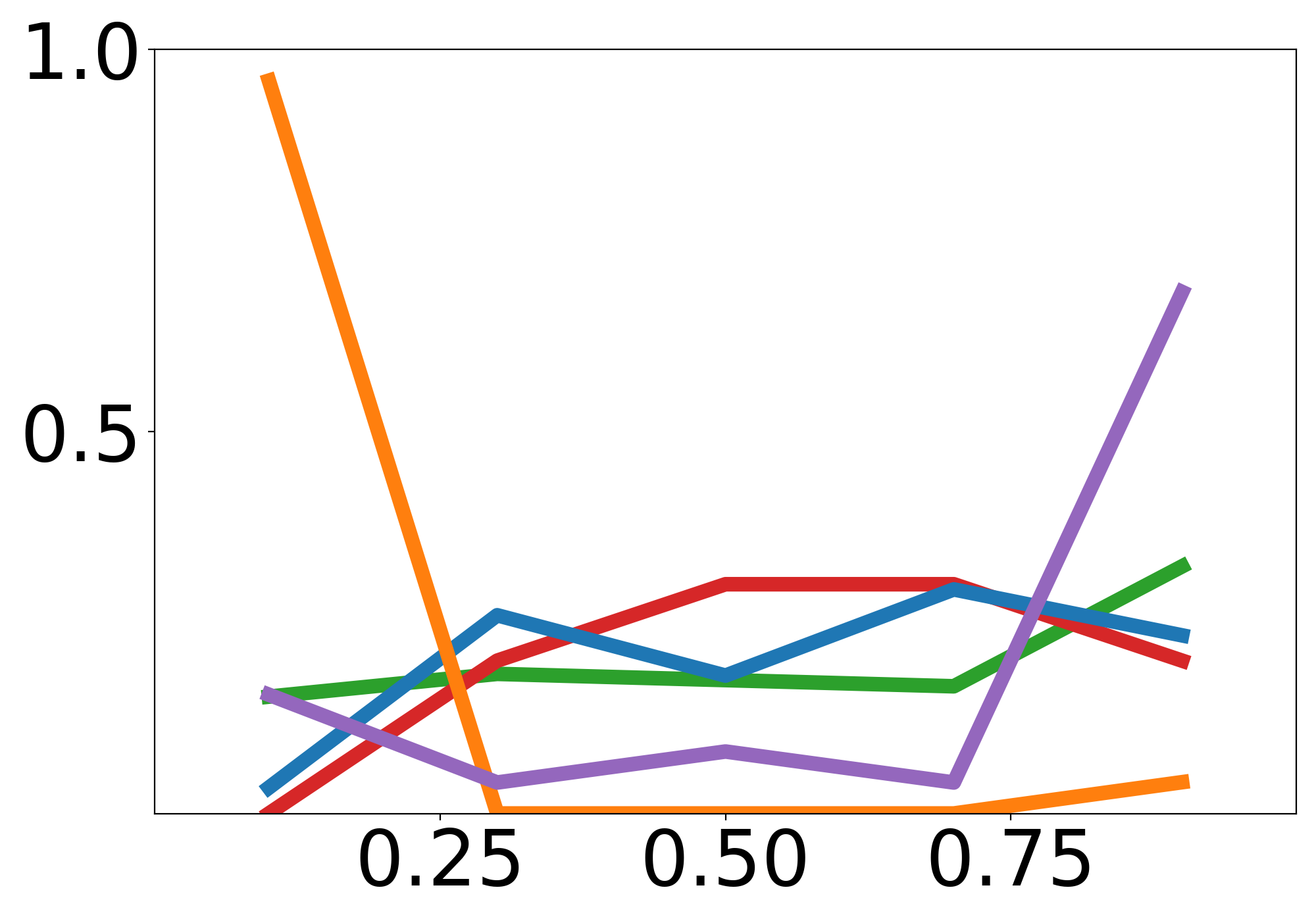}} &
\subcaptionbox{}{\includegraphics[width=\linewidth]{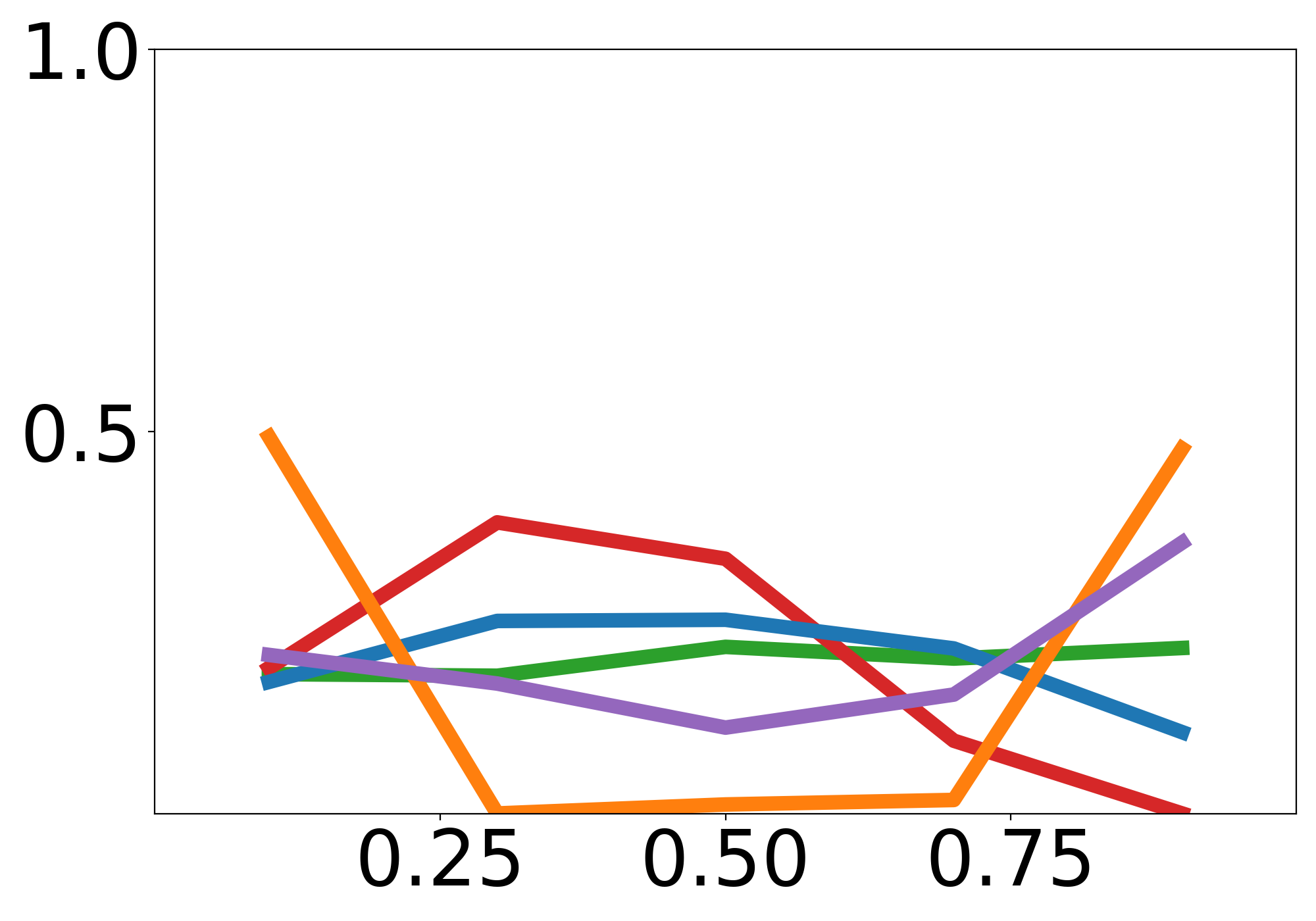}} &
\subcaptionbox{}{\includegraphics[width=\linewidth]{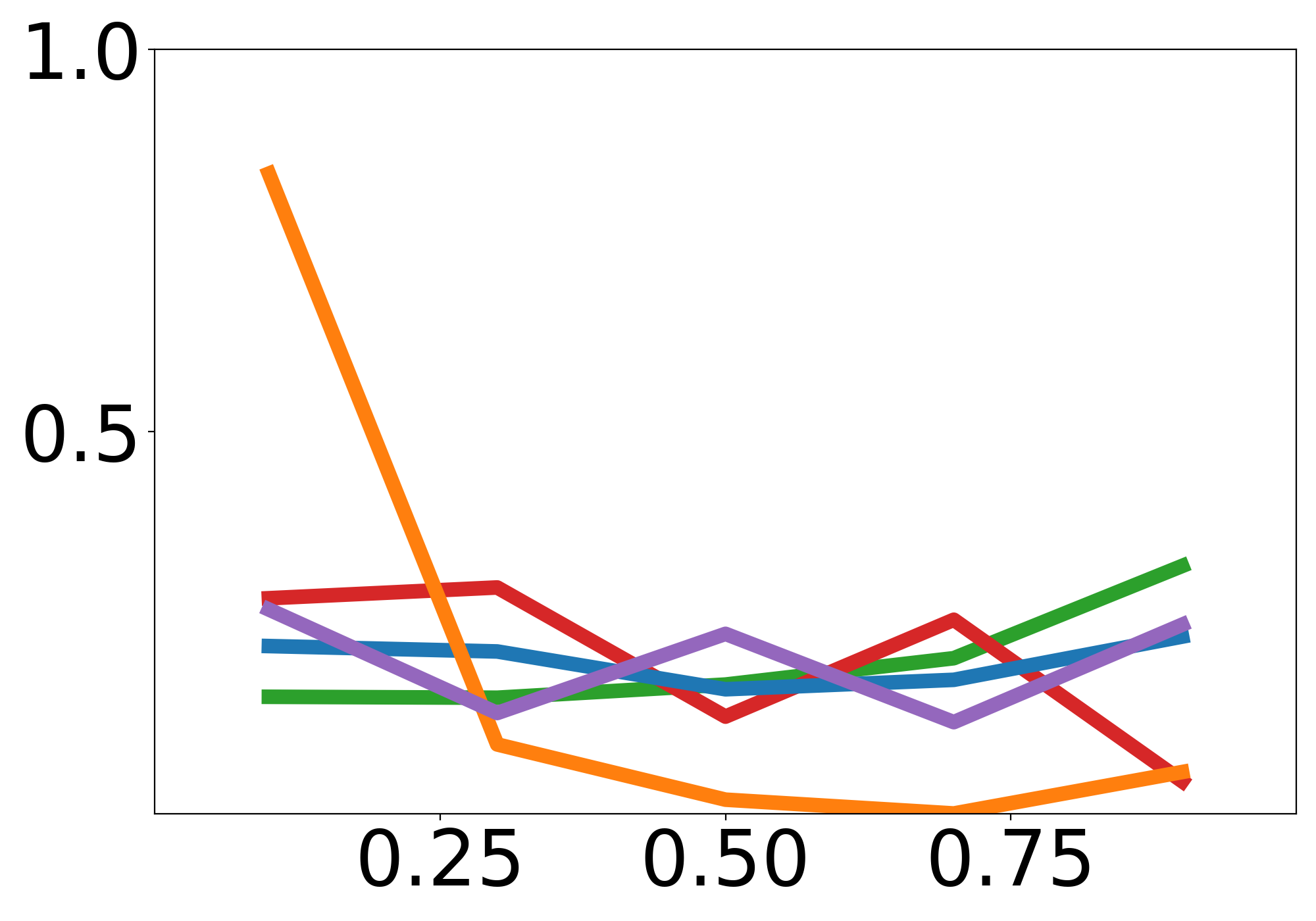}}
\\
\RowStart
\subcaptionbox{}{\includegraphics[width=\linewidth]{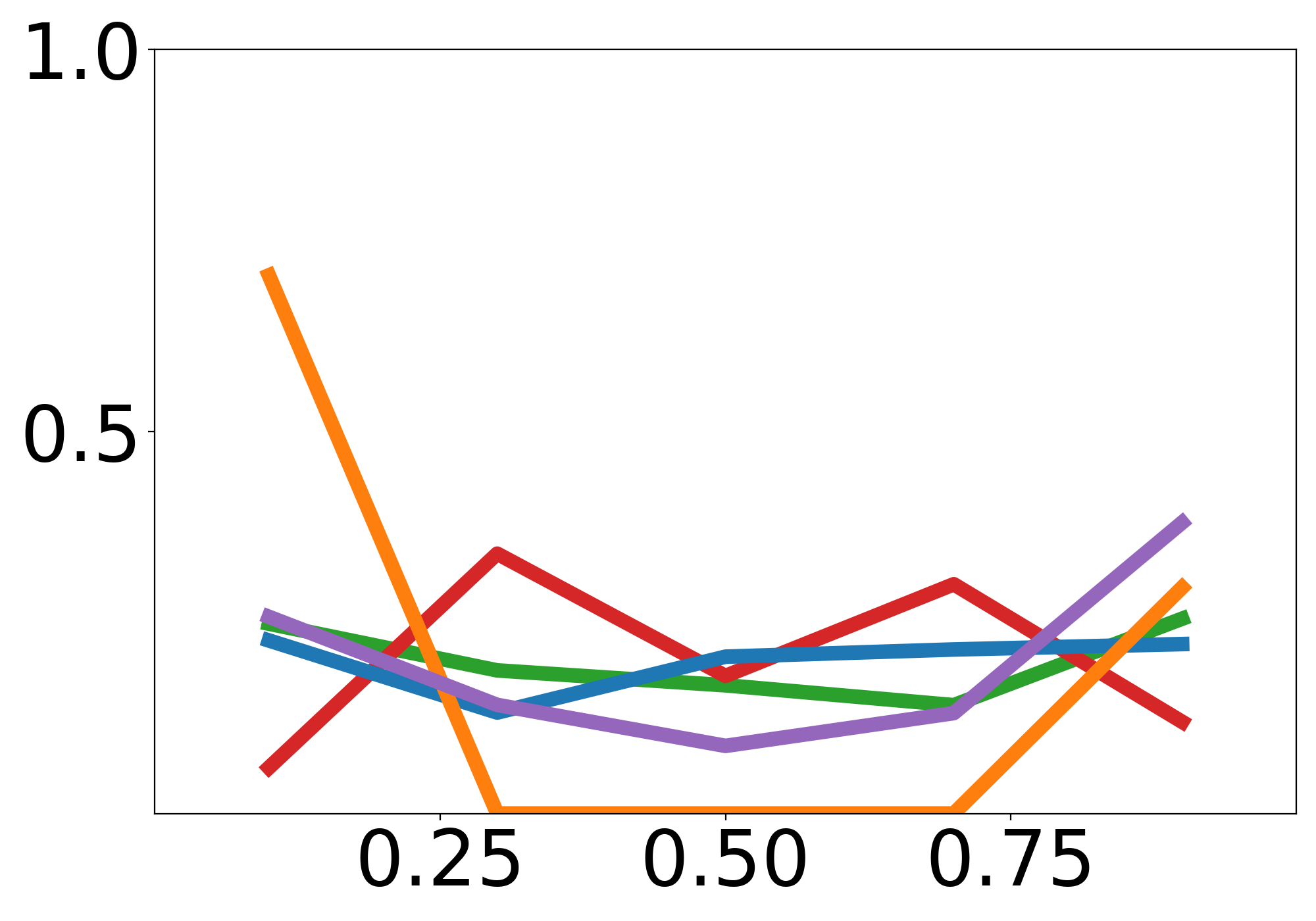}} &
\subcaptionbox{}{\includegraphics[width=\linewidth]{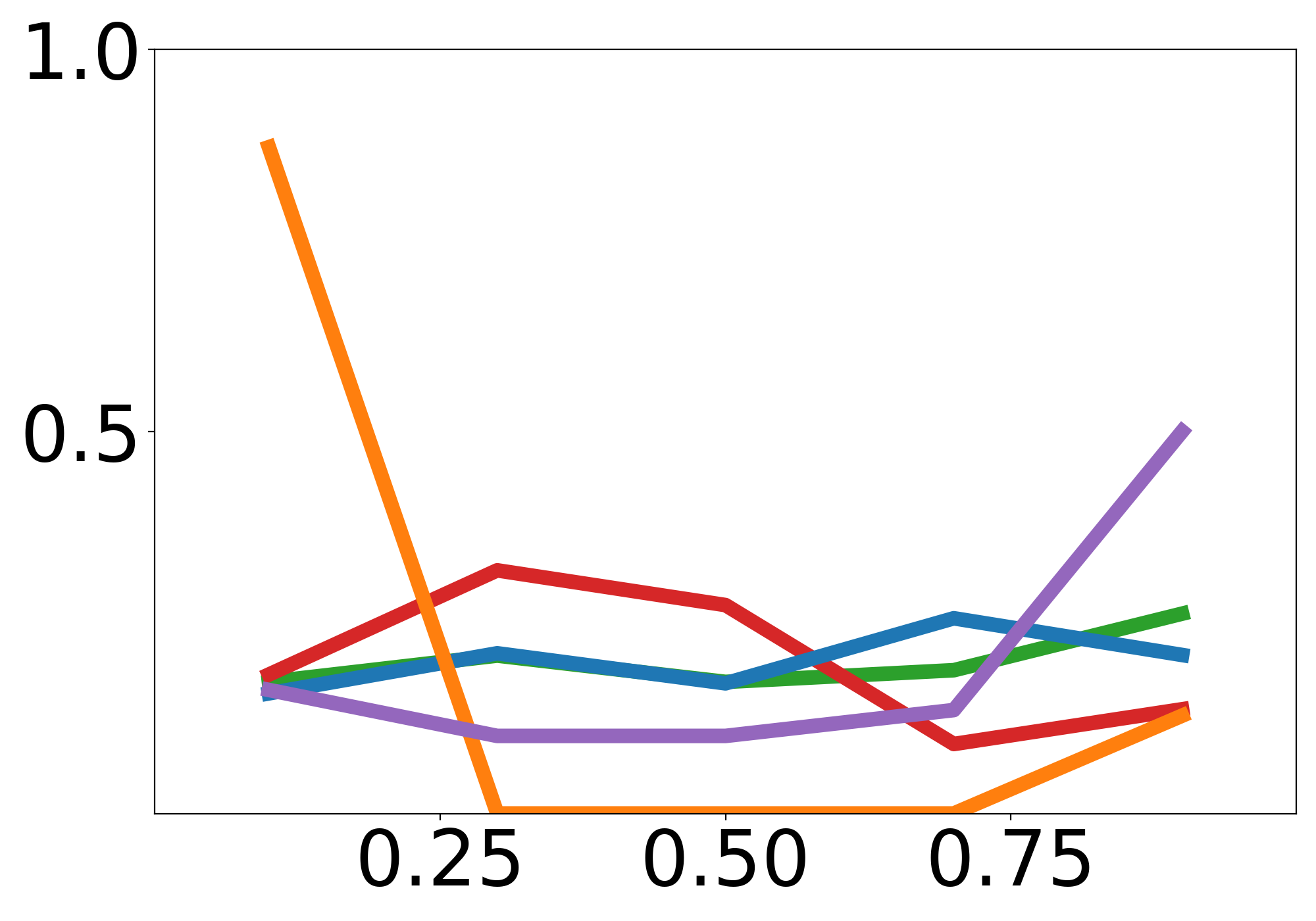}} &
\subcaptionbox{}{\includegraphics[width=\linewidth]{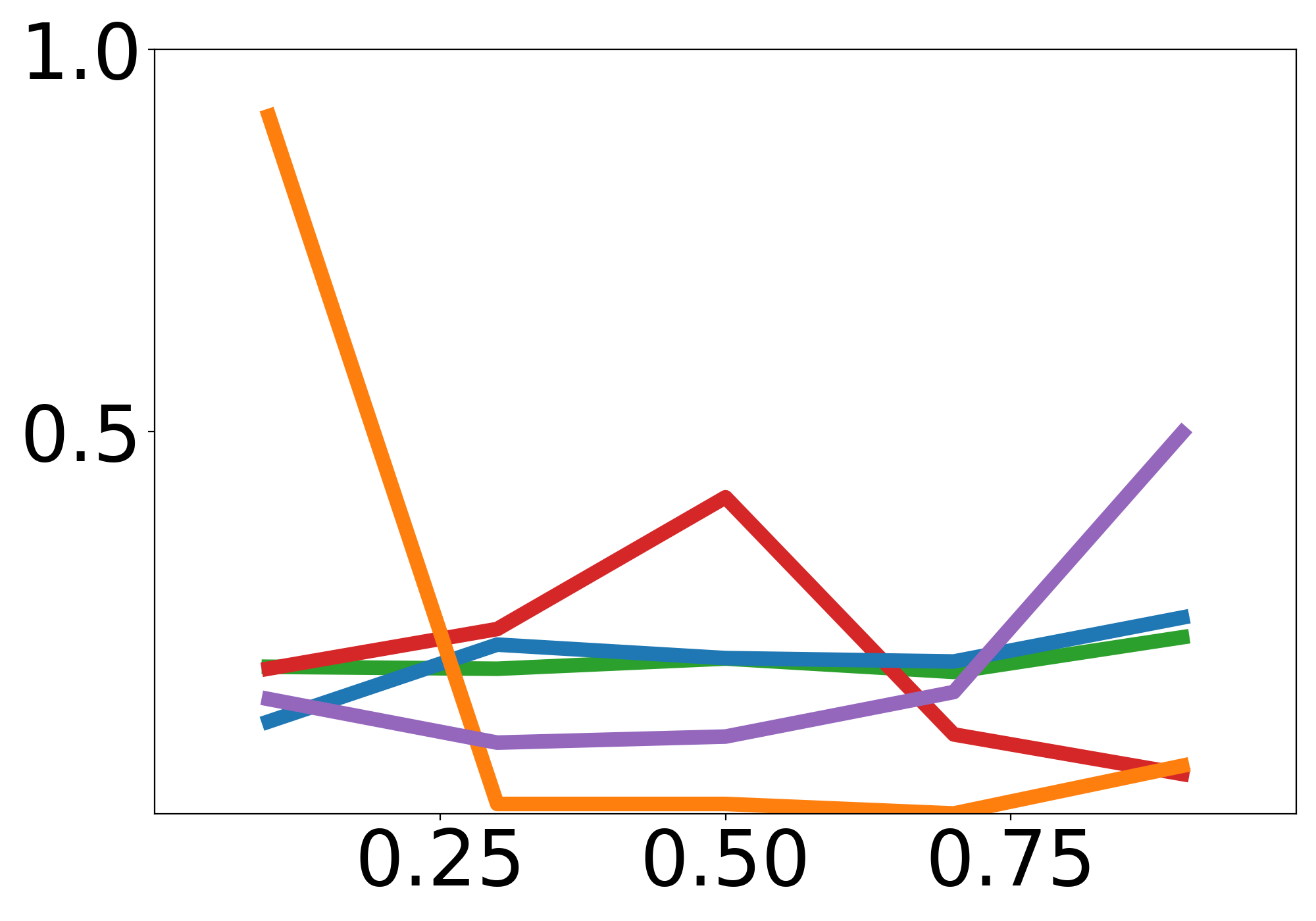}} &
\subcaptionbox{}{\includegraphics[width=\linewidth]{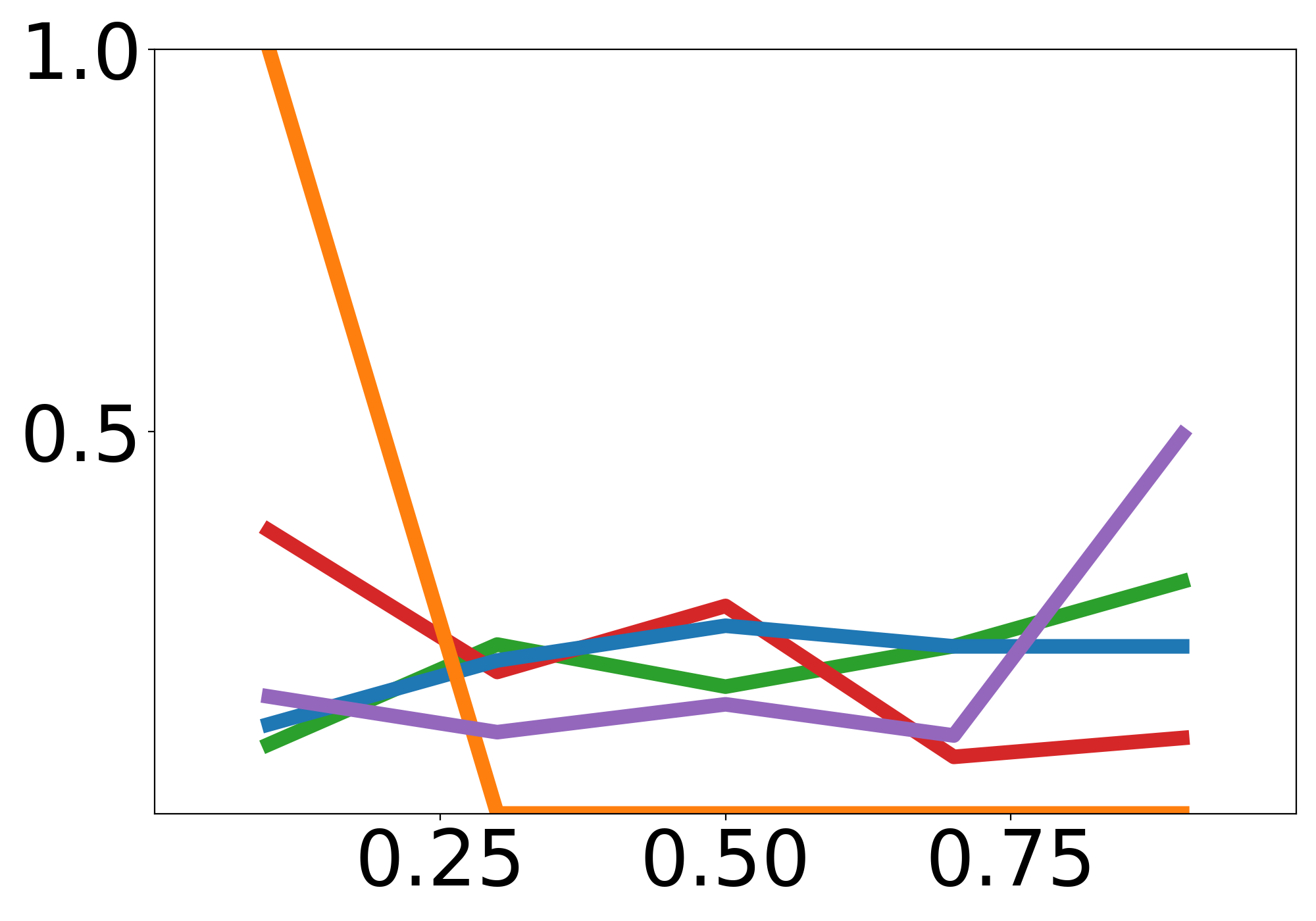}} &
\subcaptionbox{}{\includegraphics[width=\linewidth]{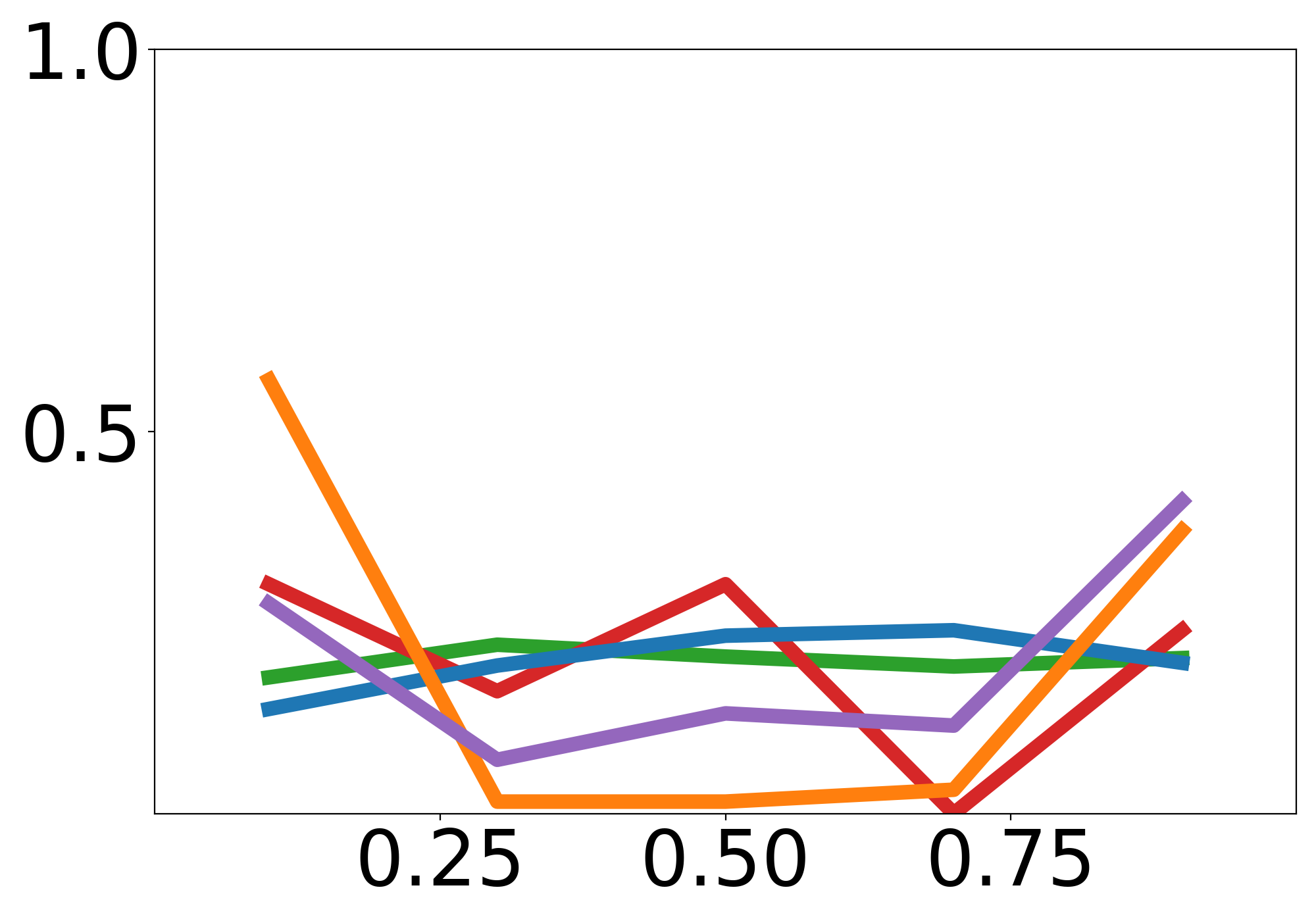}} \\
\RowStart
\subcaptionbox{}{\includegraphics[width=\linewidth]{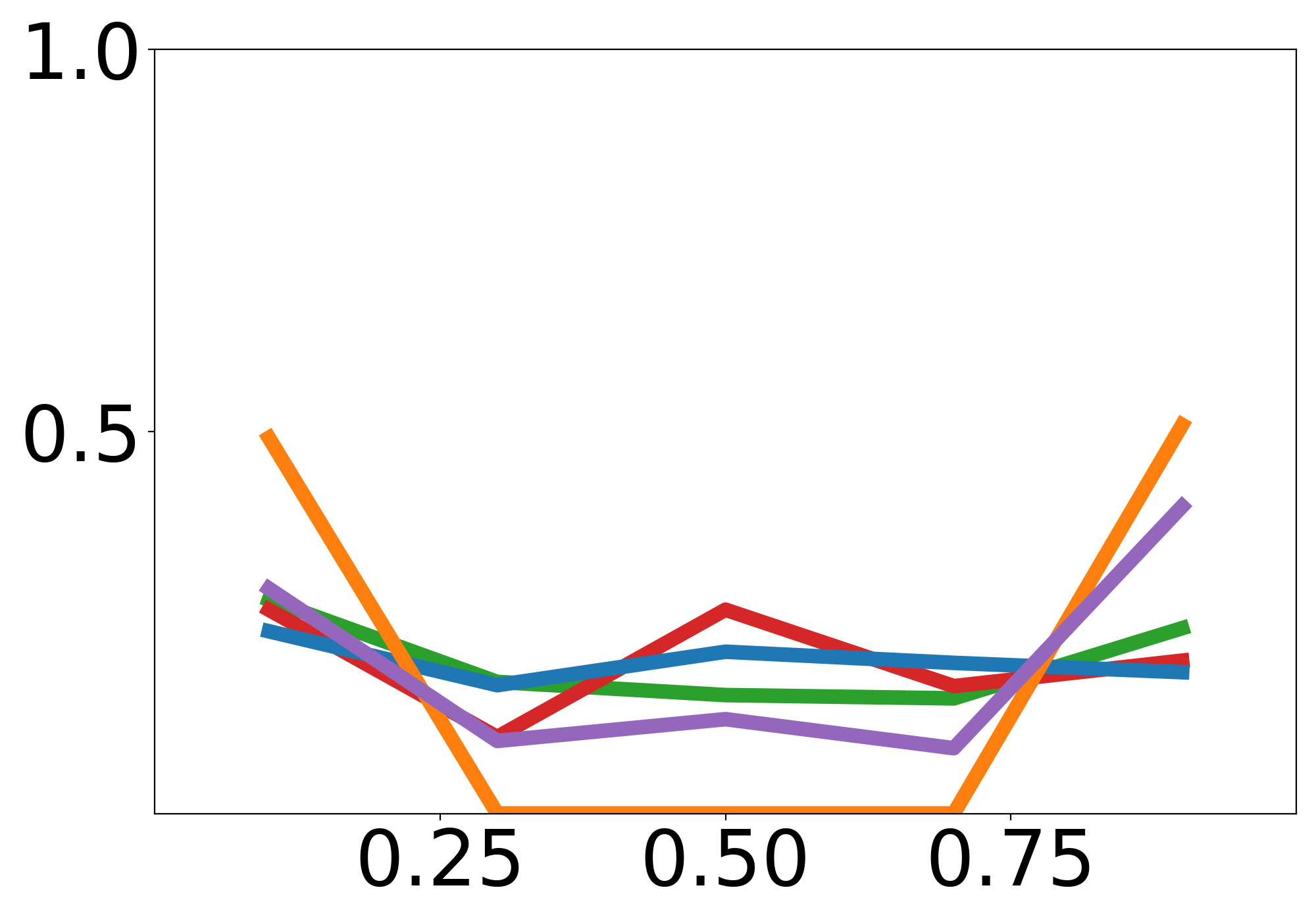}} &
\subcaptionbox{}{\includegraphics[width=\linewidth]{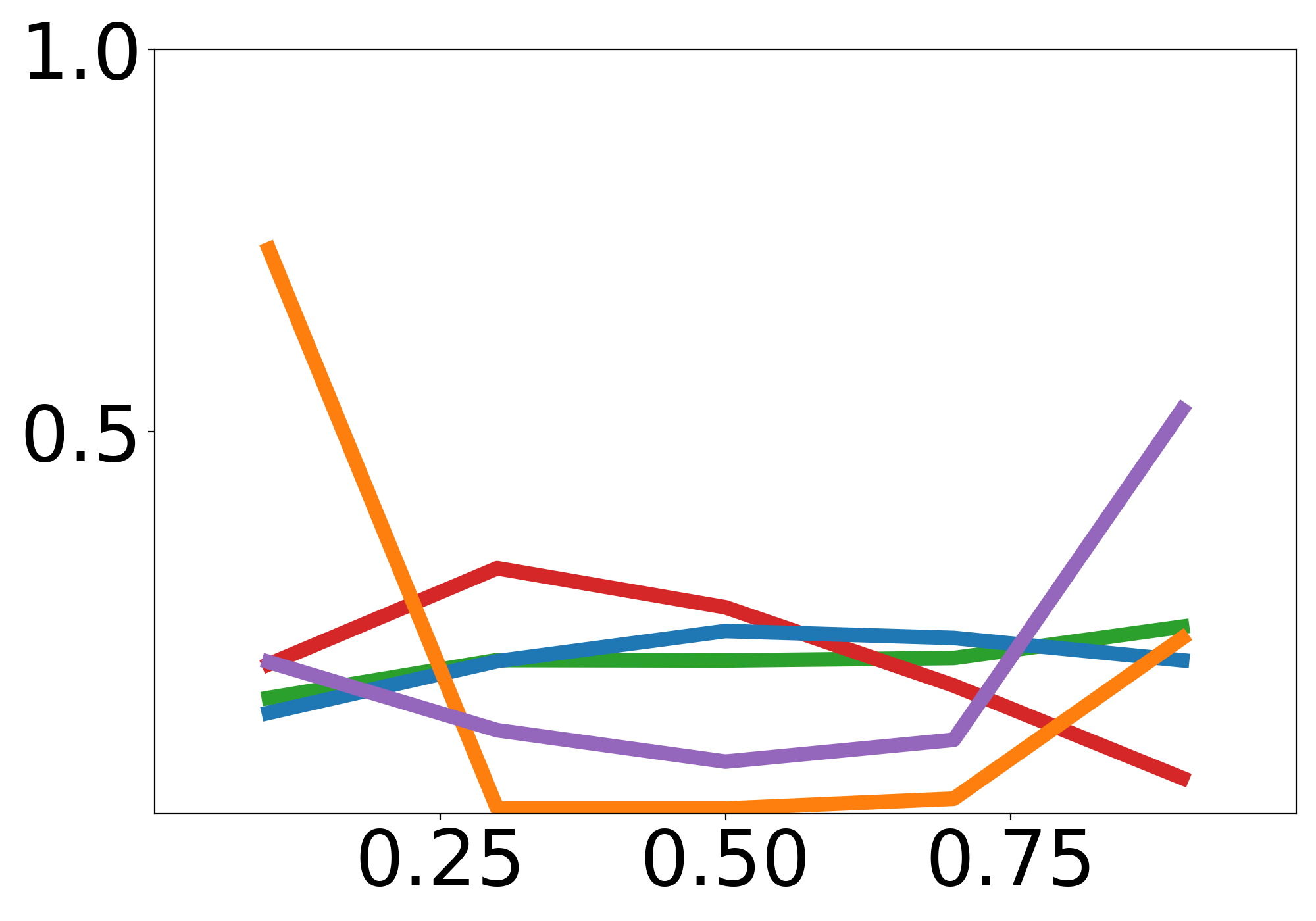}} &
\subcaptionbox{}{\includegraphics[width=\linewidth]{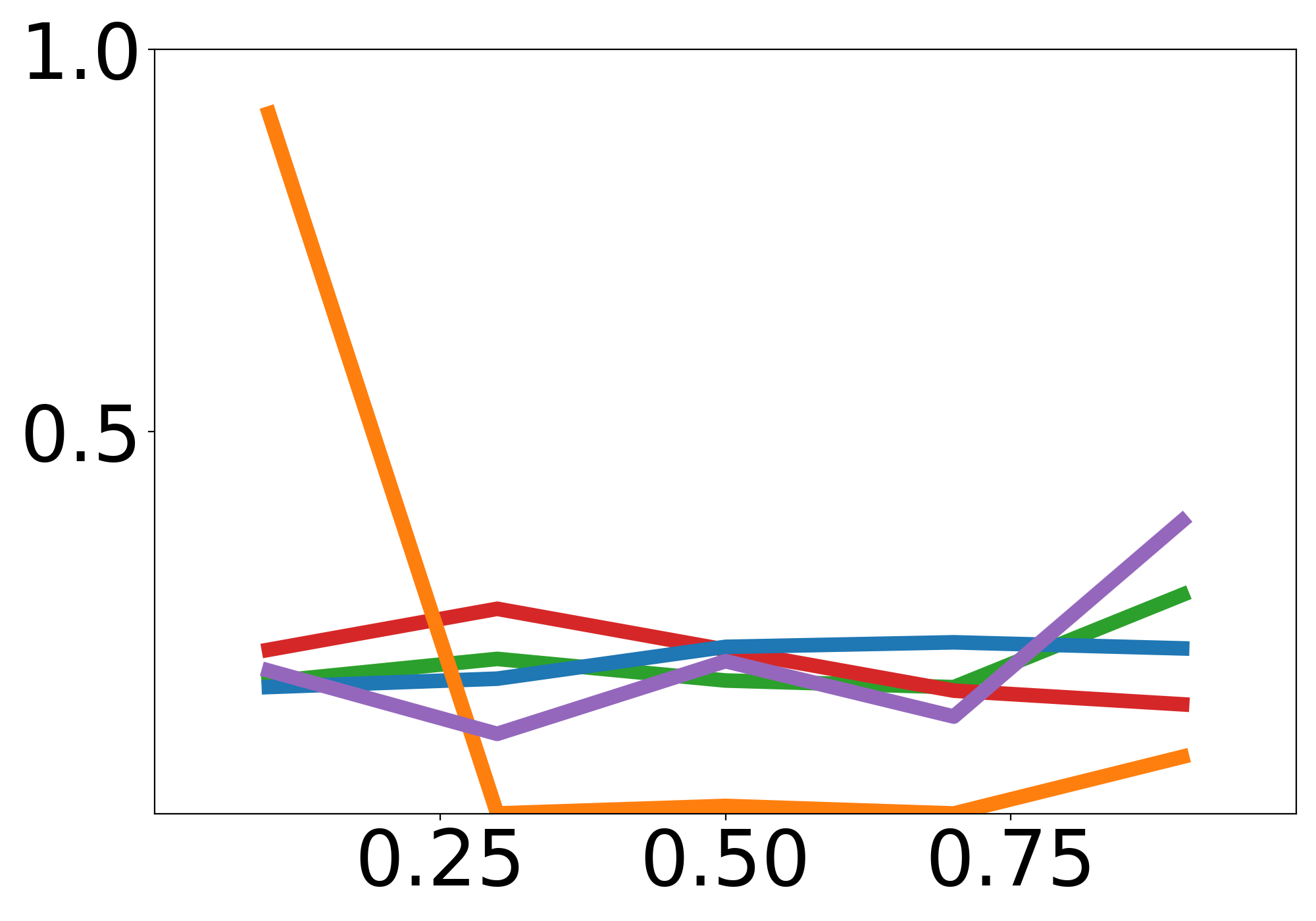}} &
\subcaptionbox{}{\includegraphics[width=\linewidth]{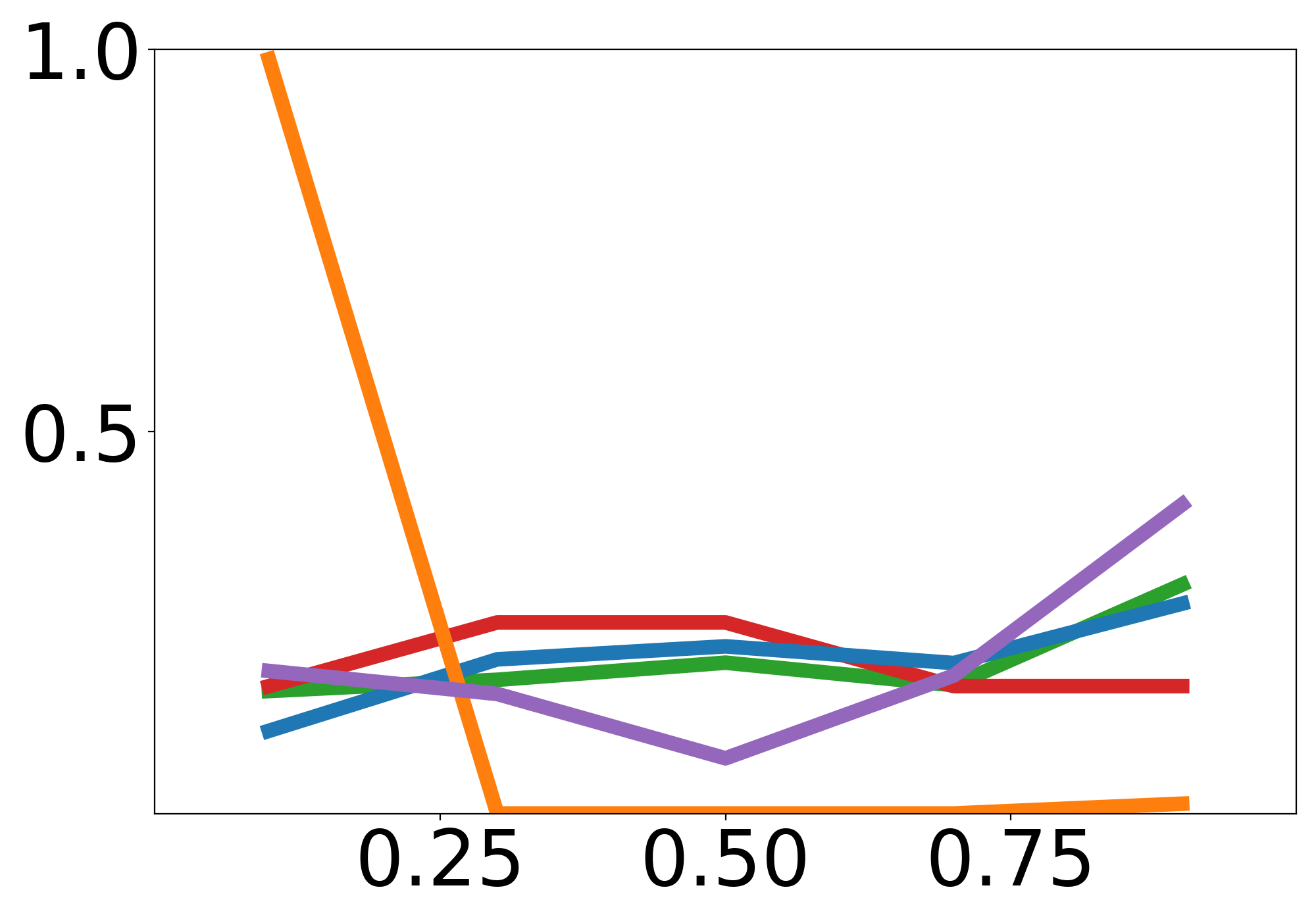}} &
\subcaptionbox{}{\includegraphics[width=\linewidth]{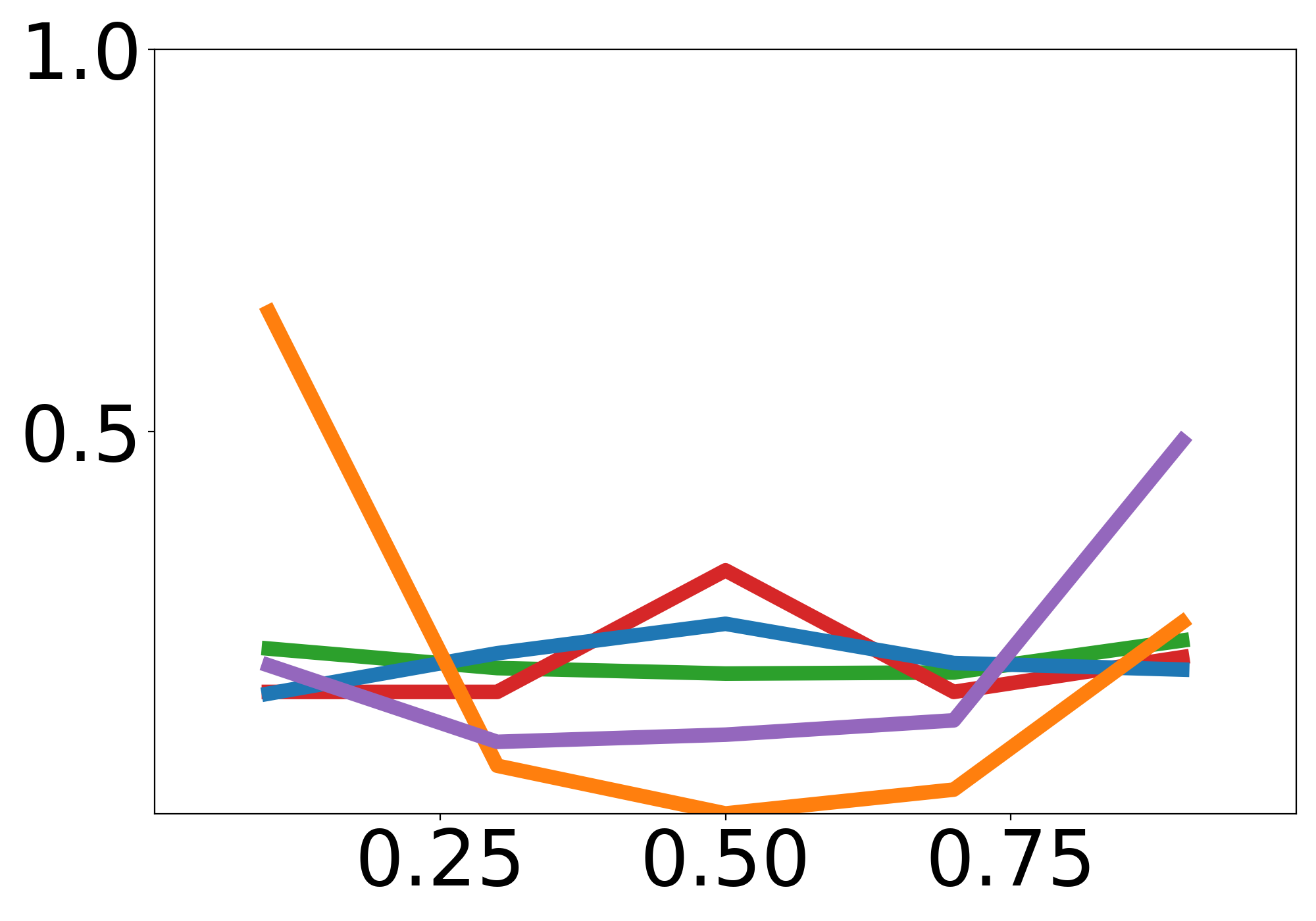}} \\

\RowStart
\subcaptionbox{}{\includegraphics[width=\linewidth]{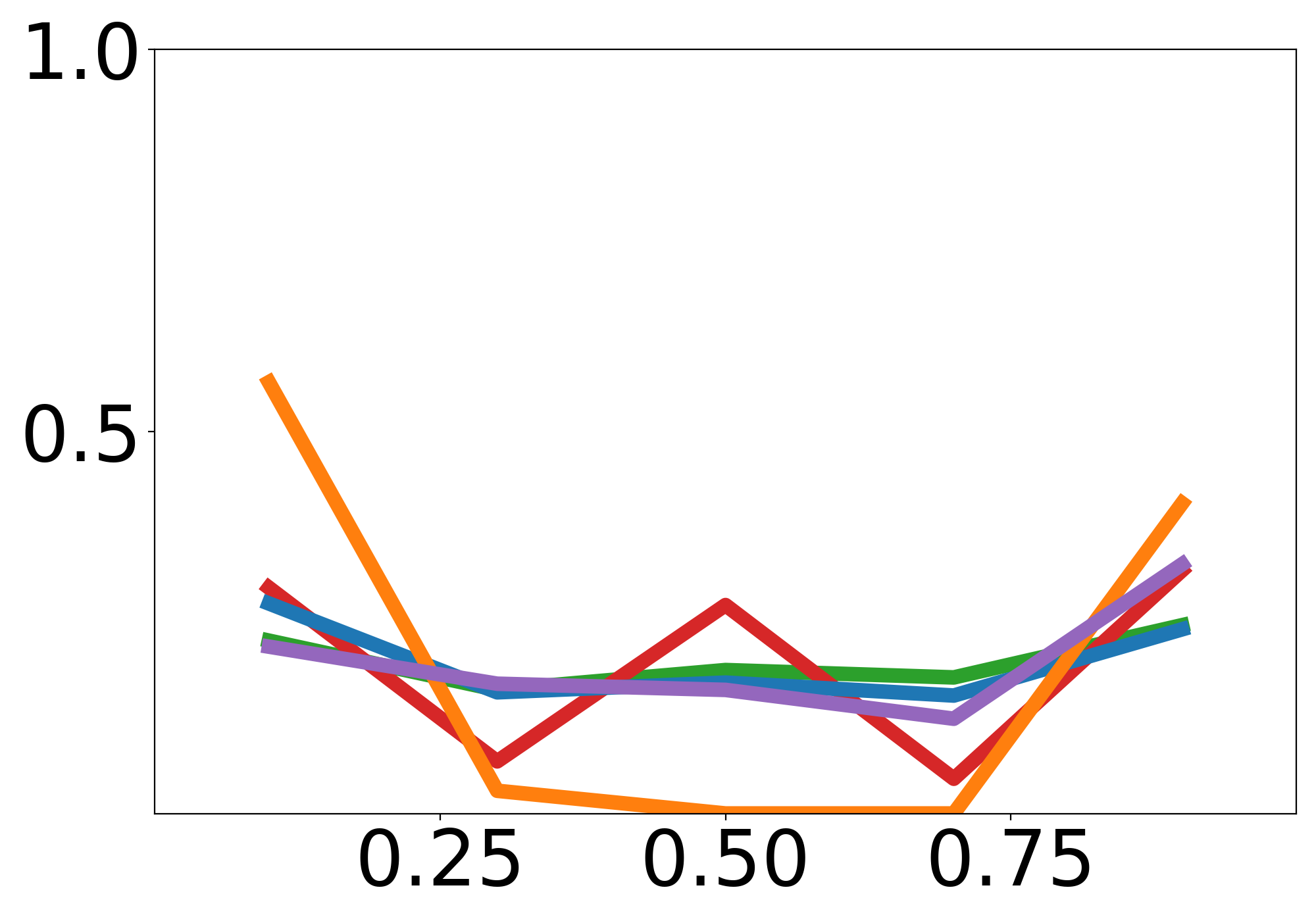}} &
\subcaptionbox{}{\includegraphics[width=\linewidth]{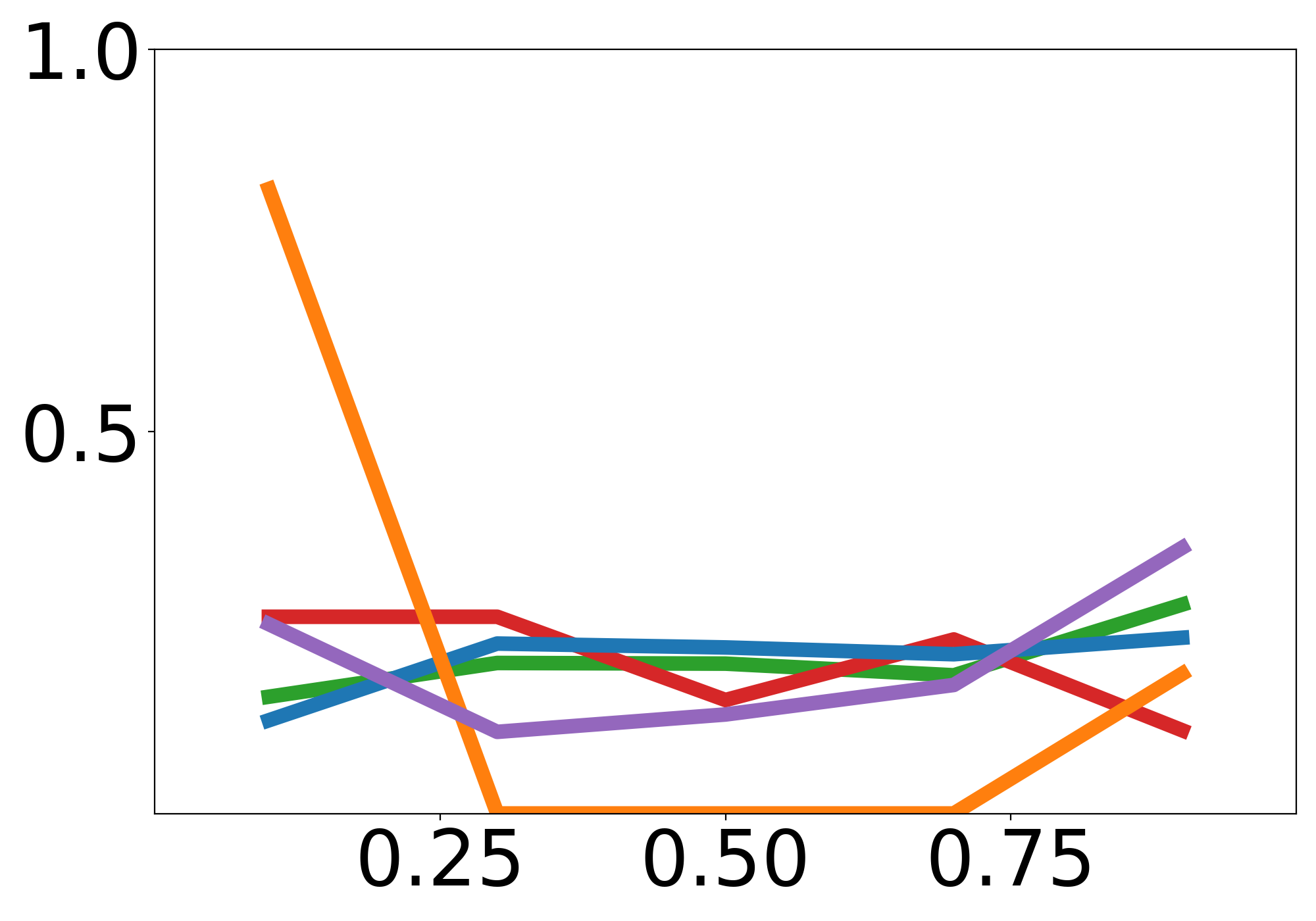}} &
\subcaptionbox{}{\includegraphics[width=\linewidth]{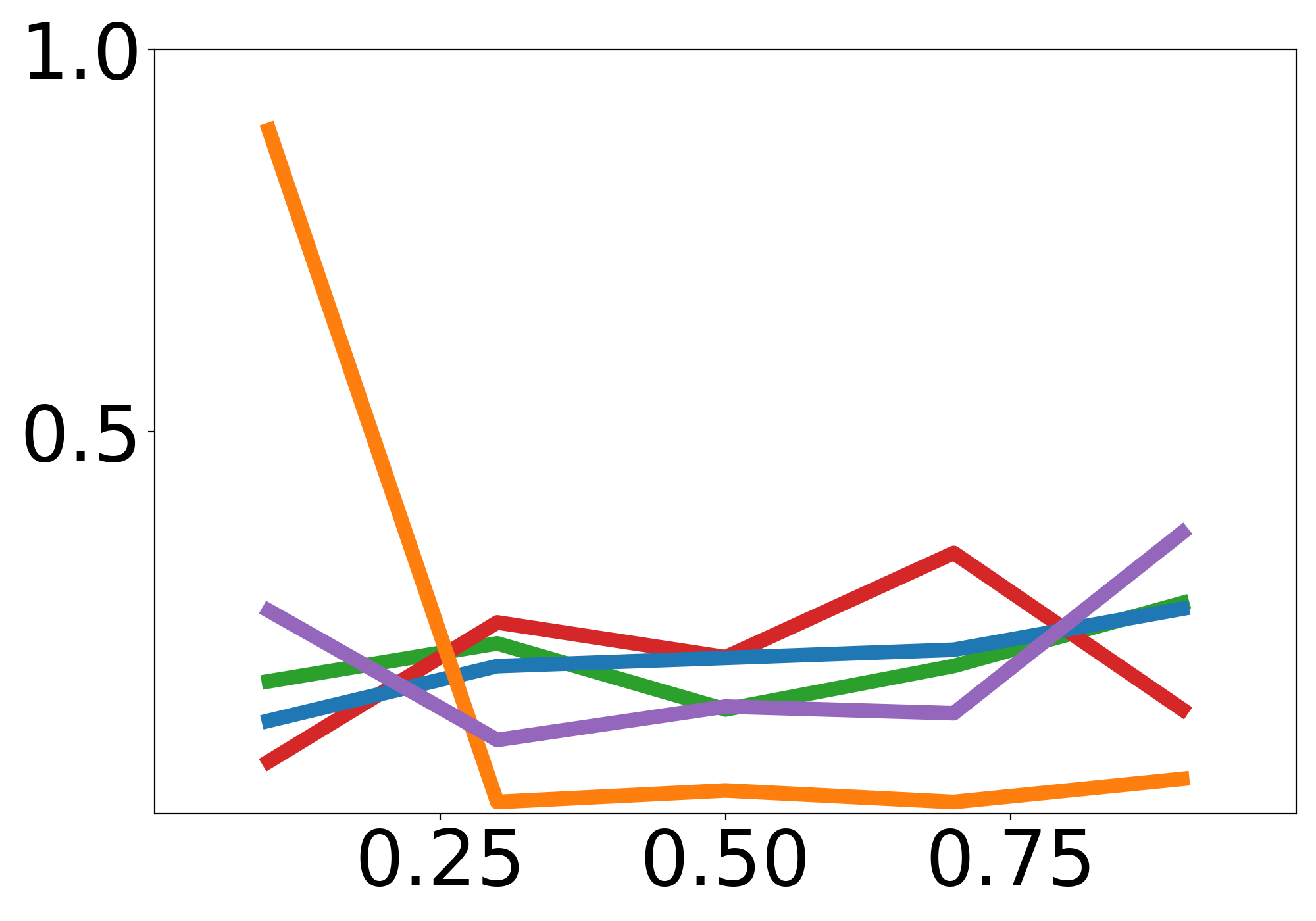}} &
\subcaptionbox{}{\includegraphics[width=\linewidth]{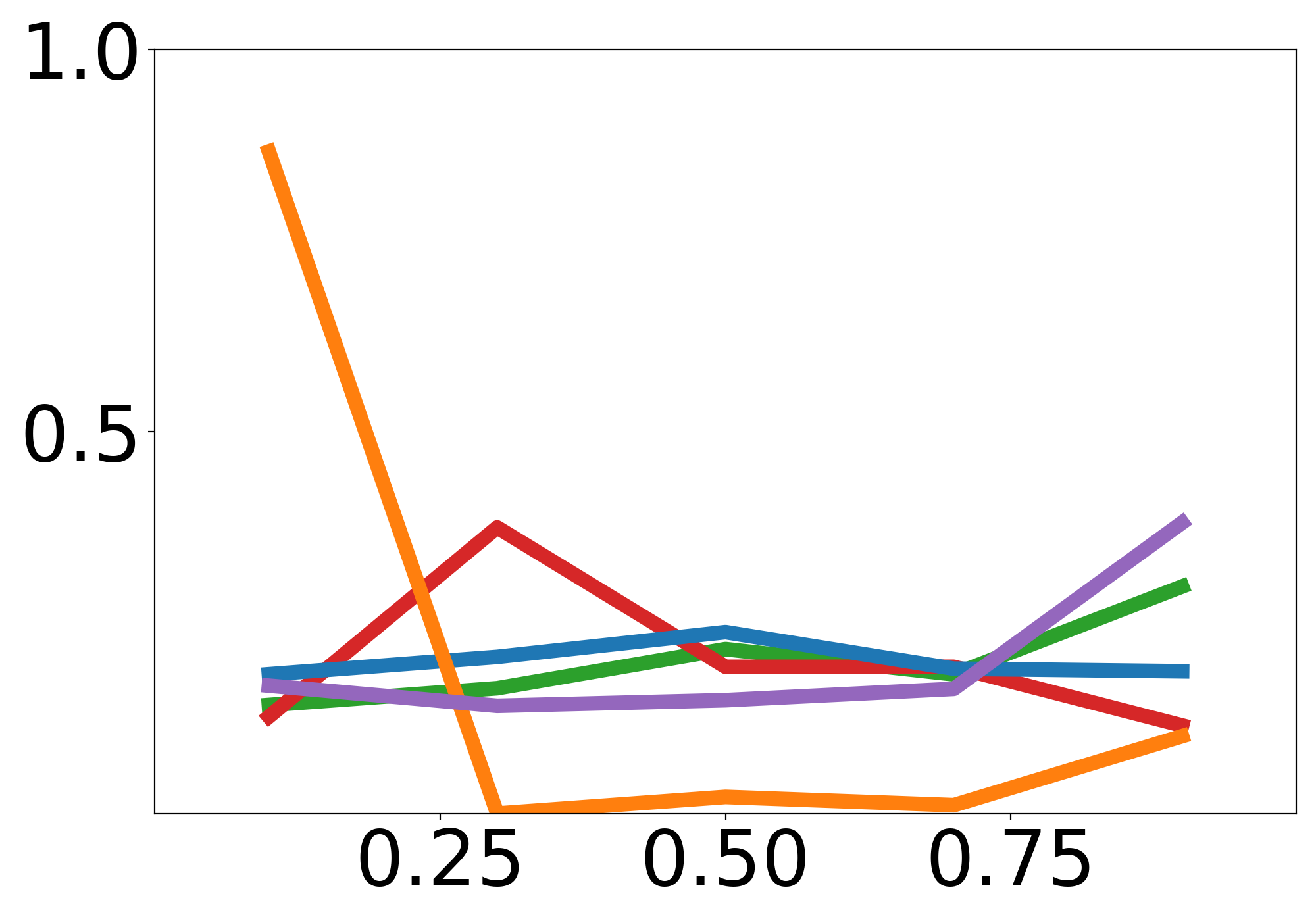}} &
\subcaptionbox{}{\includegraphics[width=\linewidth]{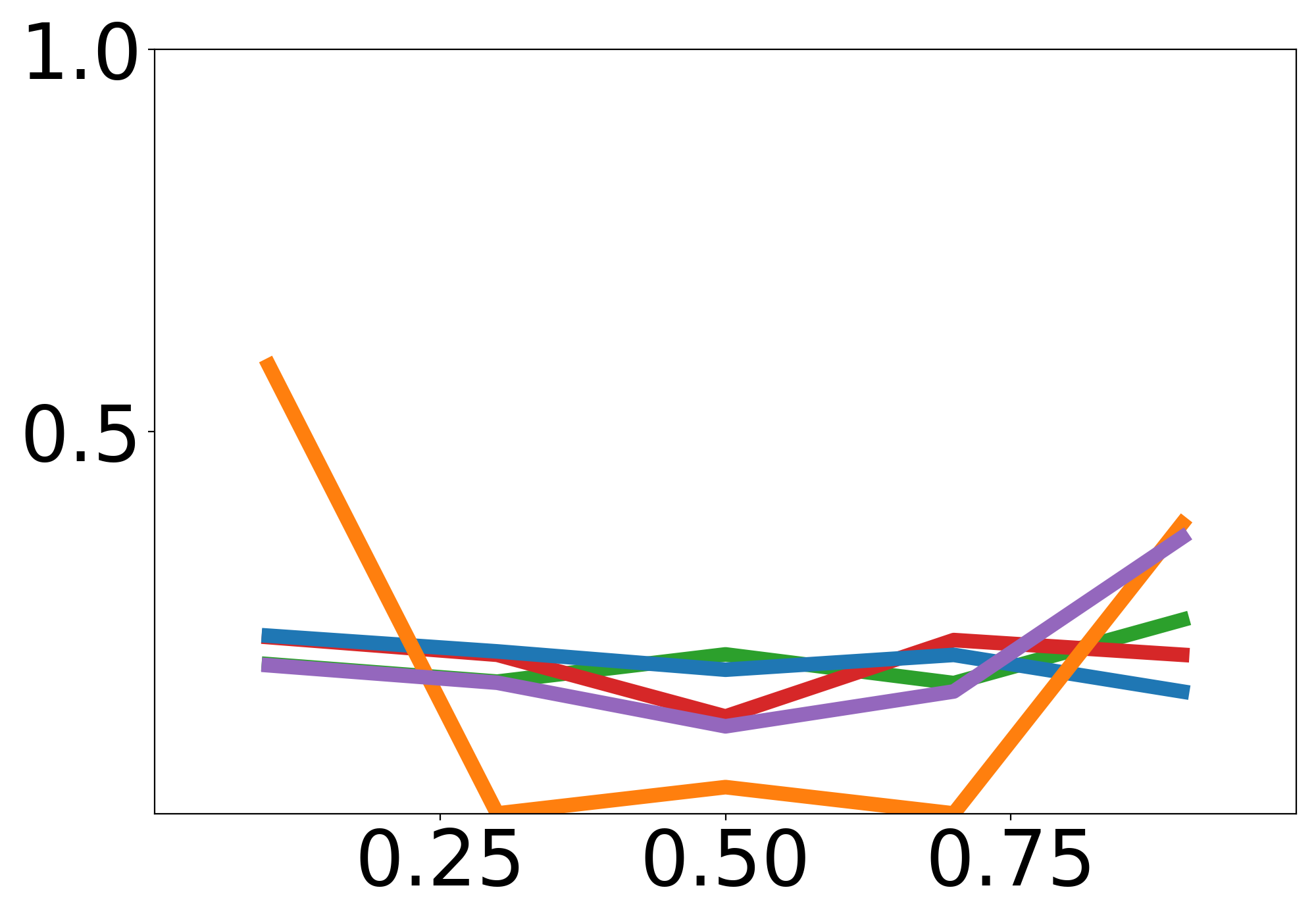}} \\

\RowStart
\subcaptionbox{}{\includegraphics[width=\linewidth]{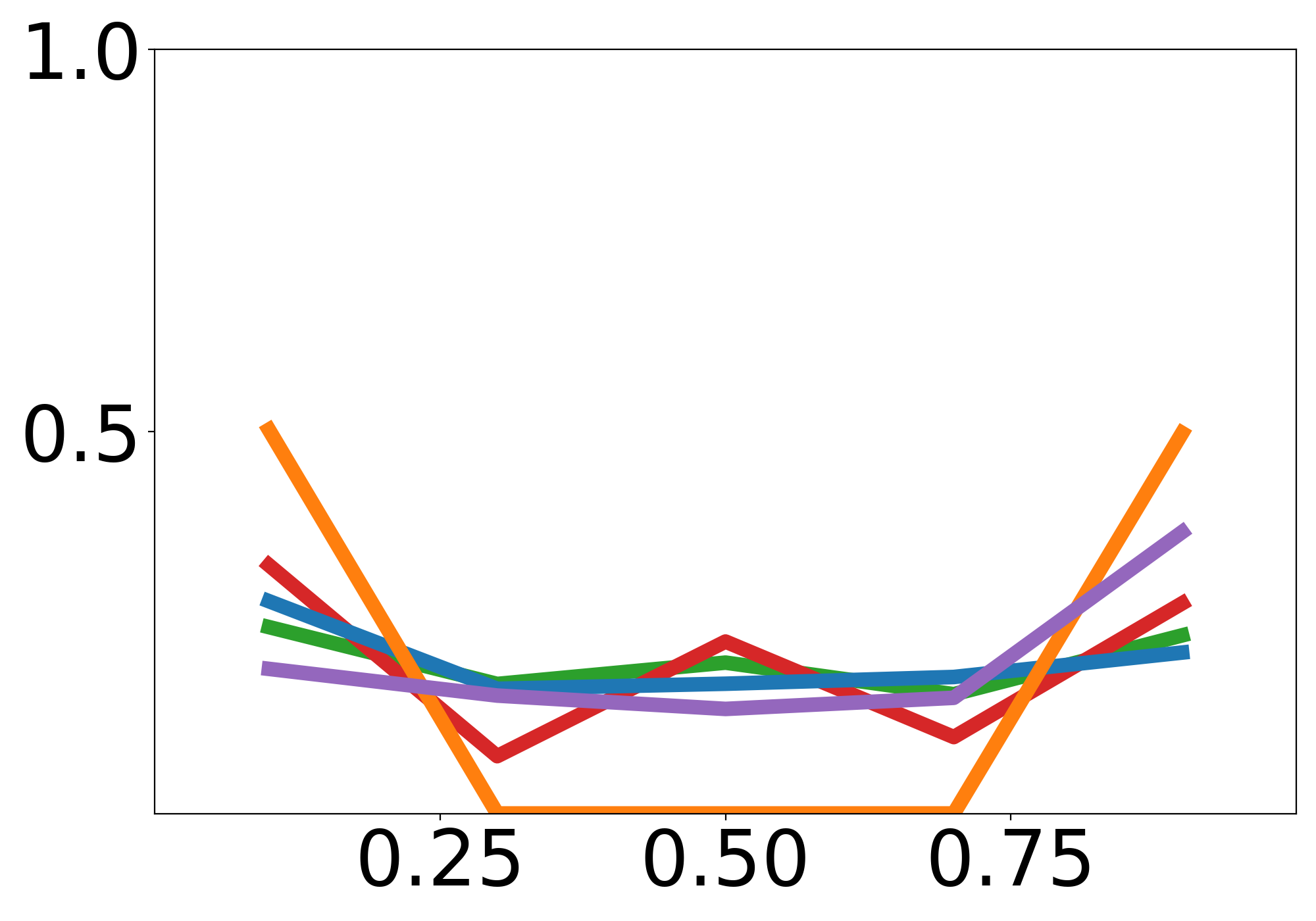}} &
\subcaptionbox{}{\includegraphics[width=\linewidth]{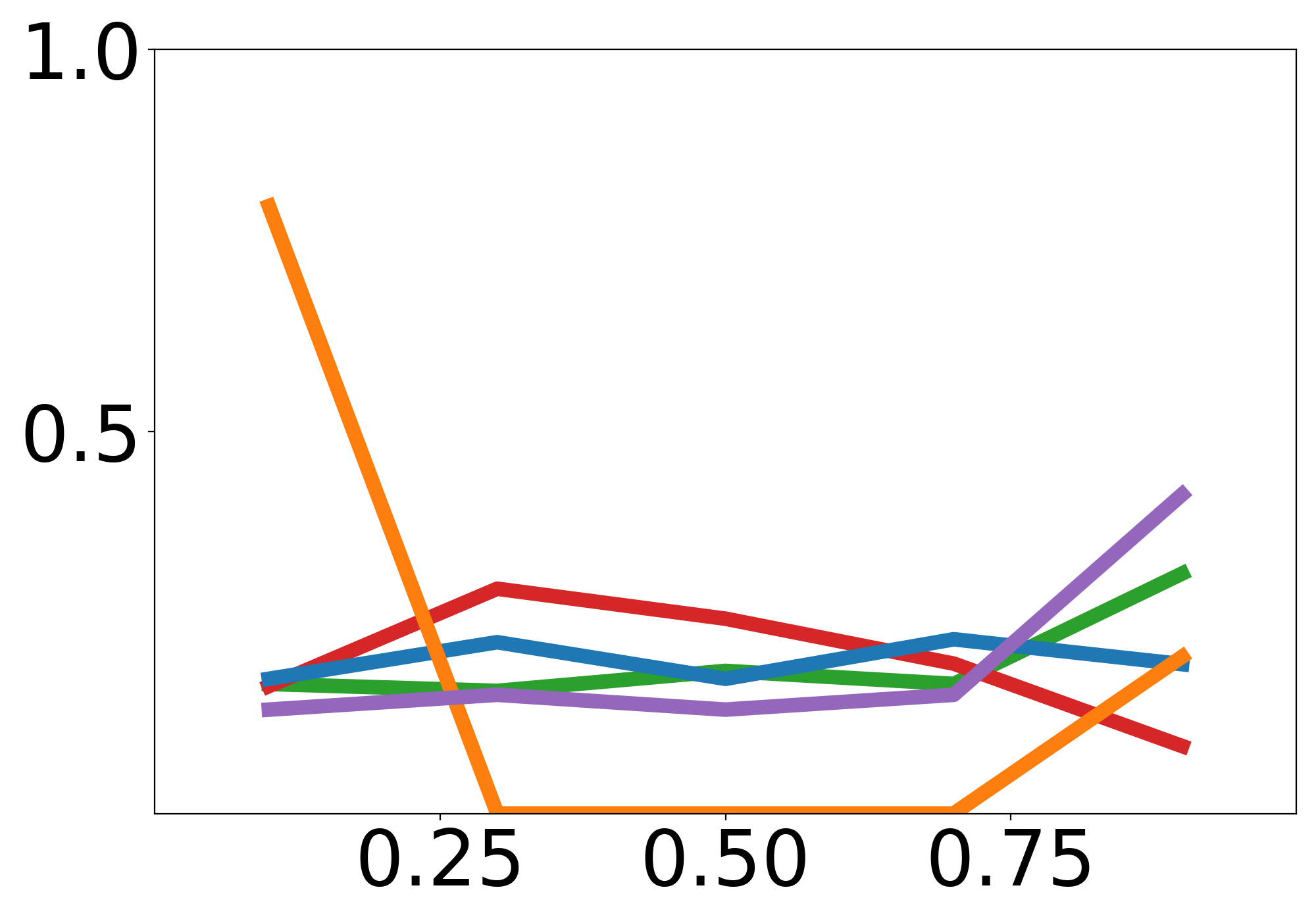}} &
\subcaptionbox{}{\includegraphics[width=\linewidth]{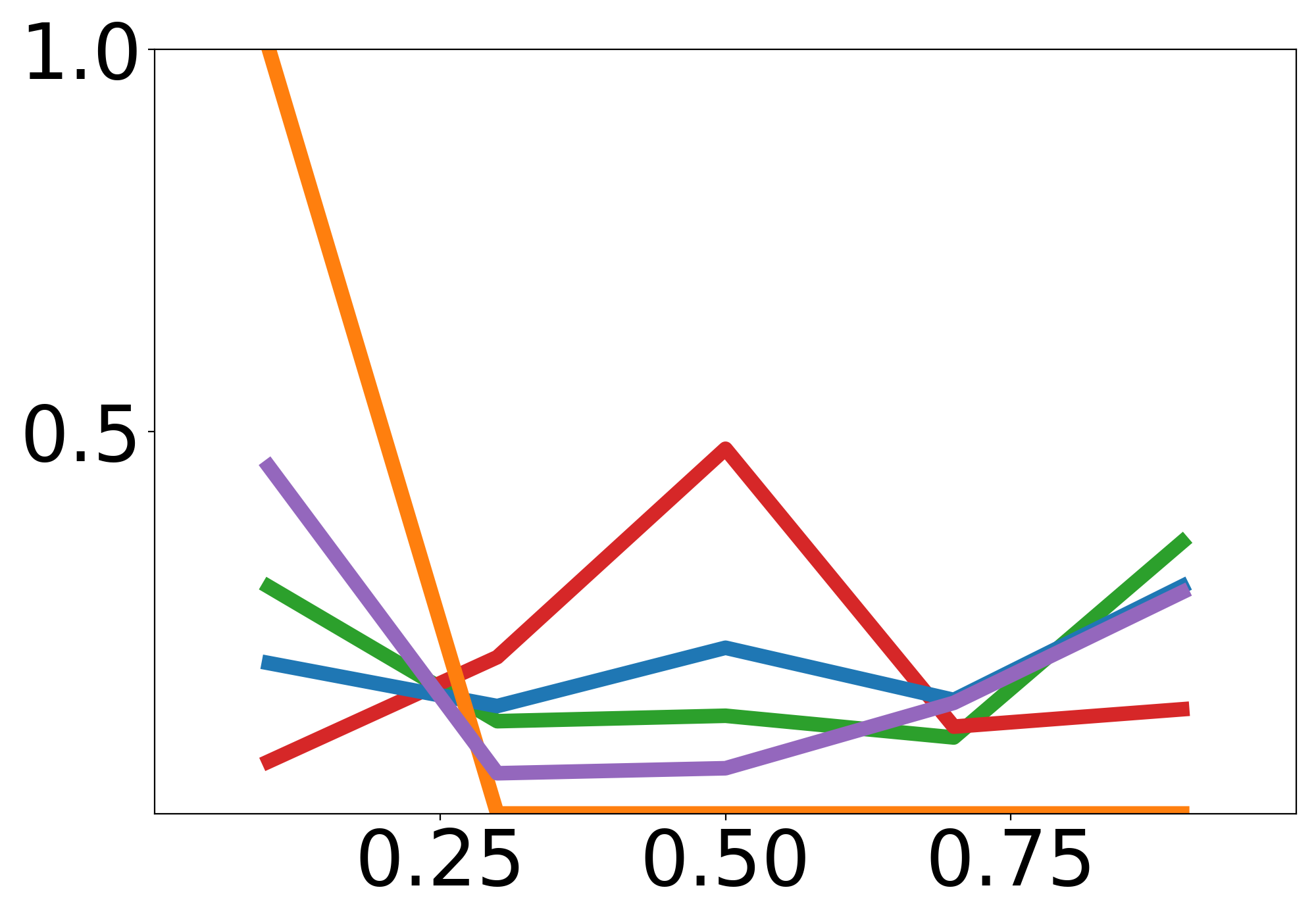}} &
\subcaptionbox{}{\includegraphics[width=\linewidth]{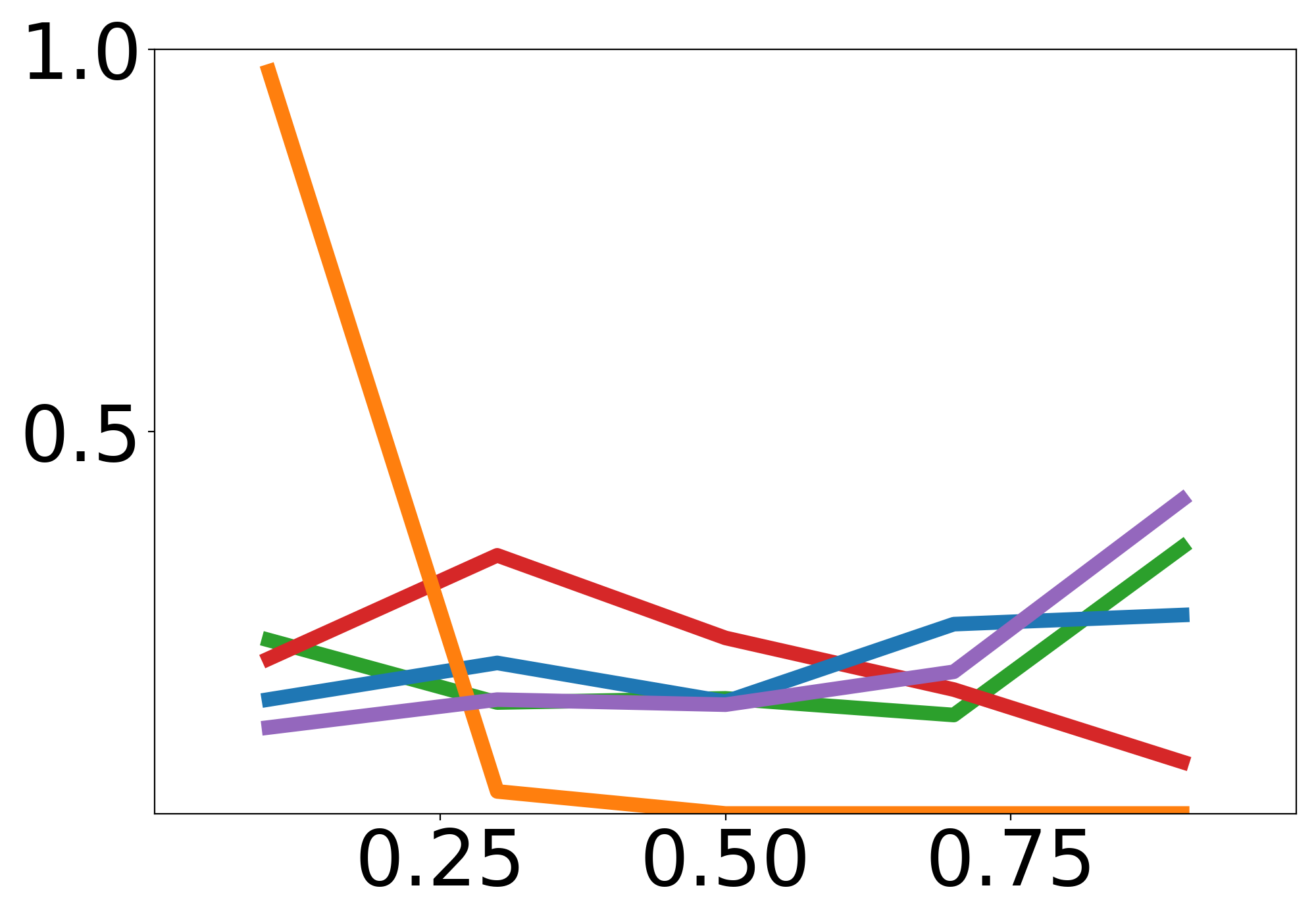}} &
\subcaptionbox{}{\includegraphics[width=\linewidth]{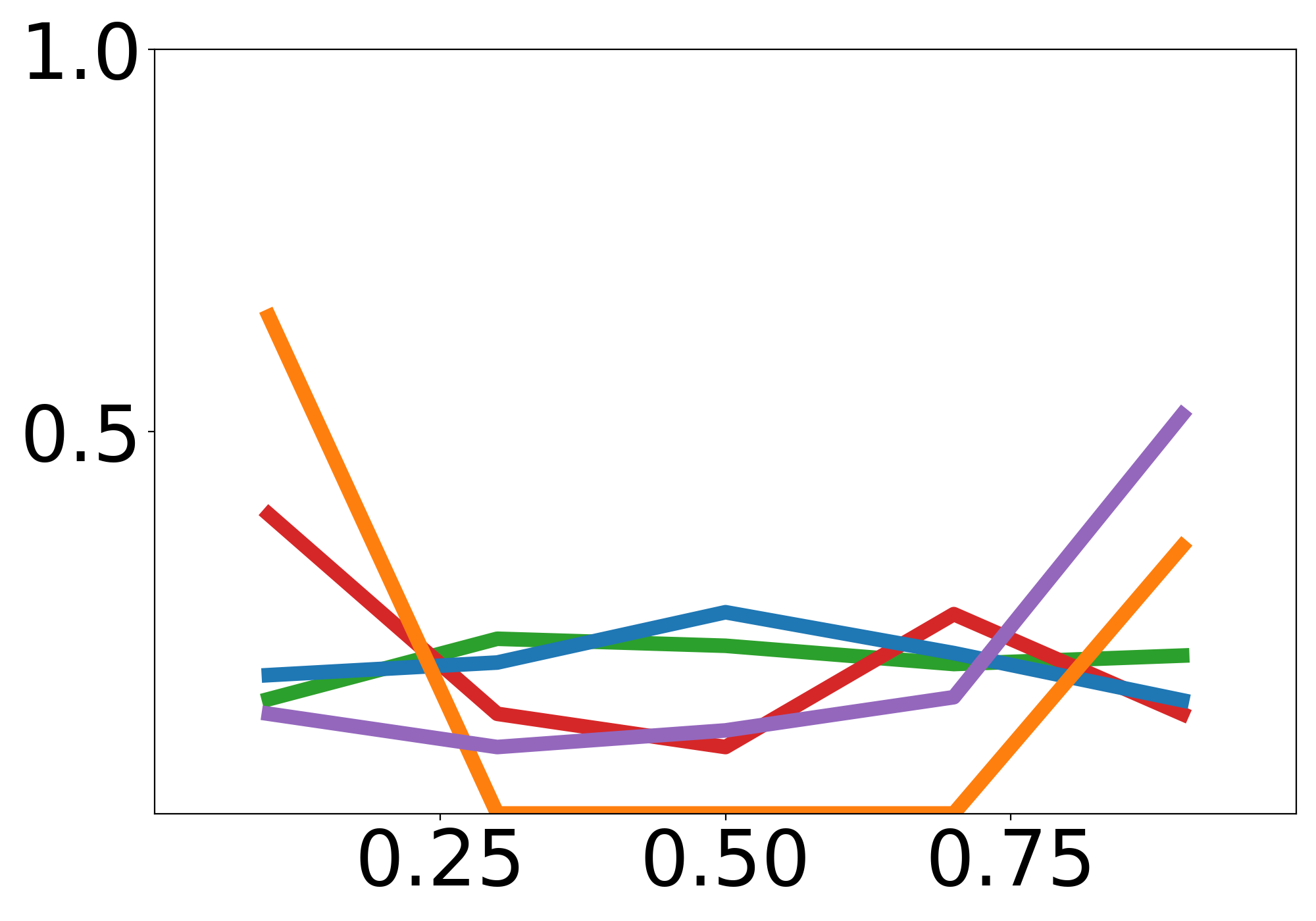}} \\

\RowStart
\subcaptionbox{}{\includegraphics[width=\linewidth]{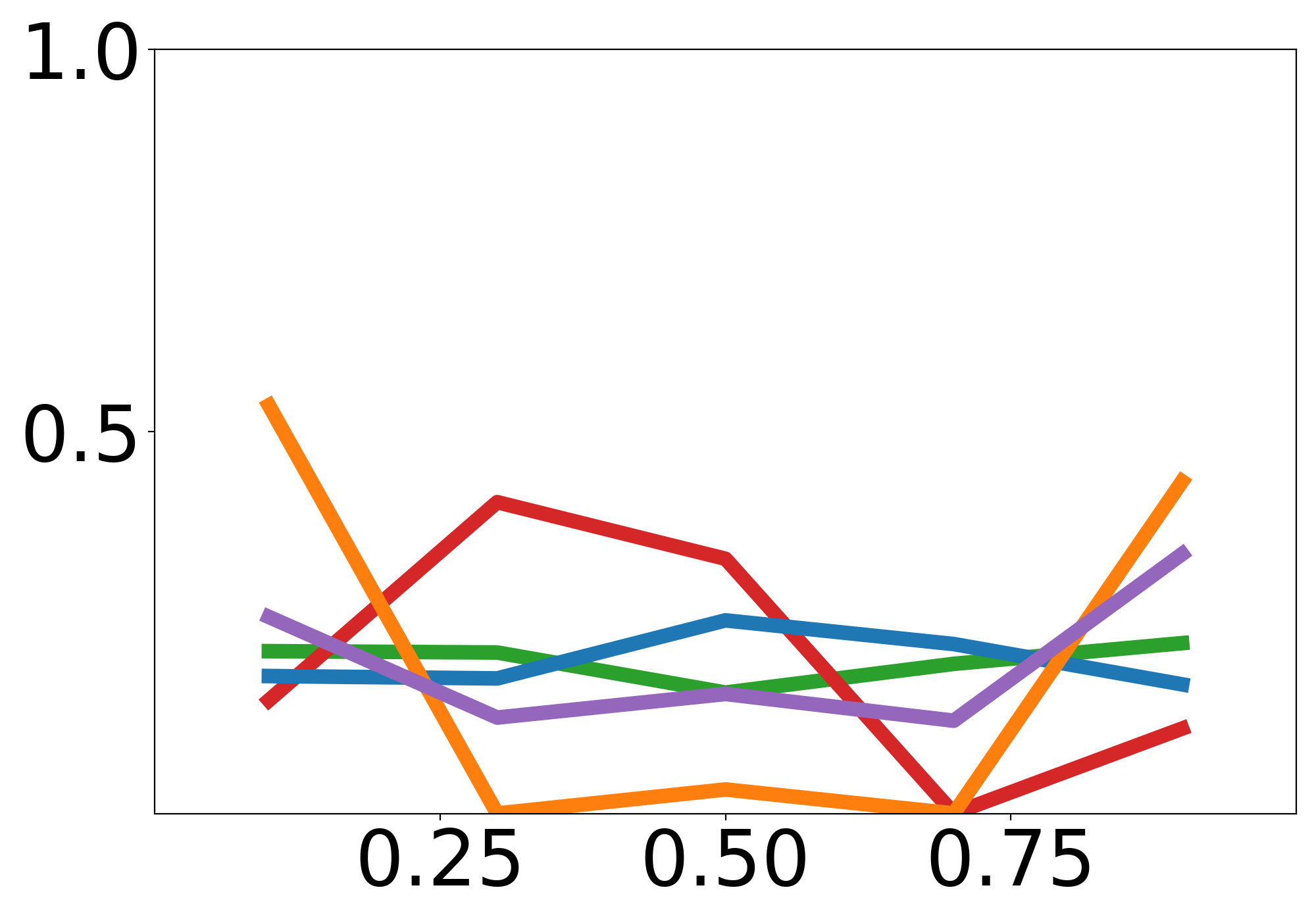}} &
\subcaptionbox{}{\includegraphics[width=\linewidth]{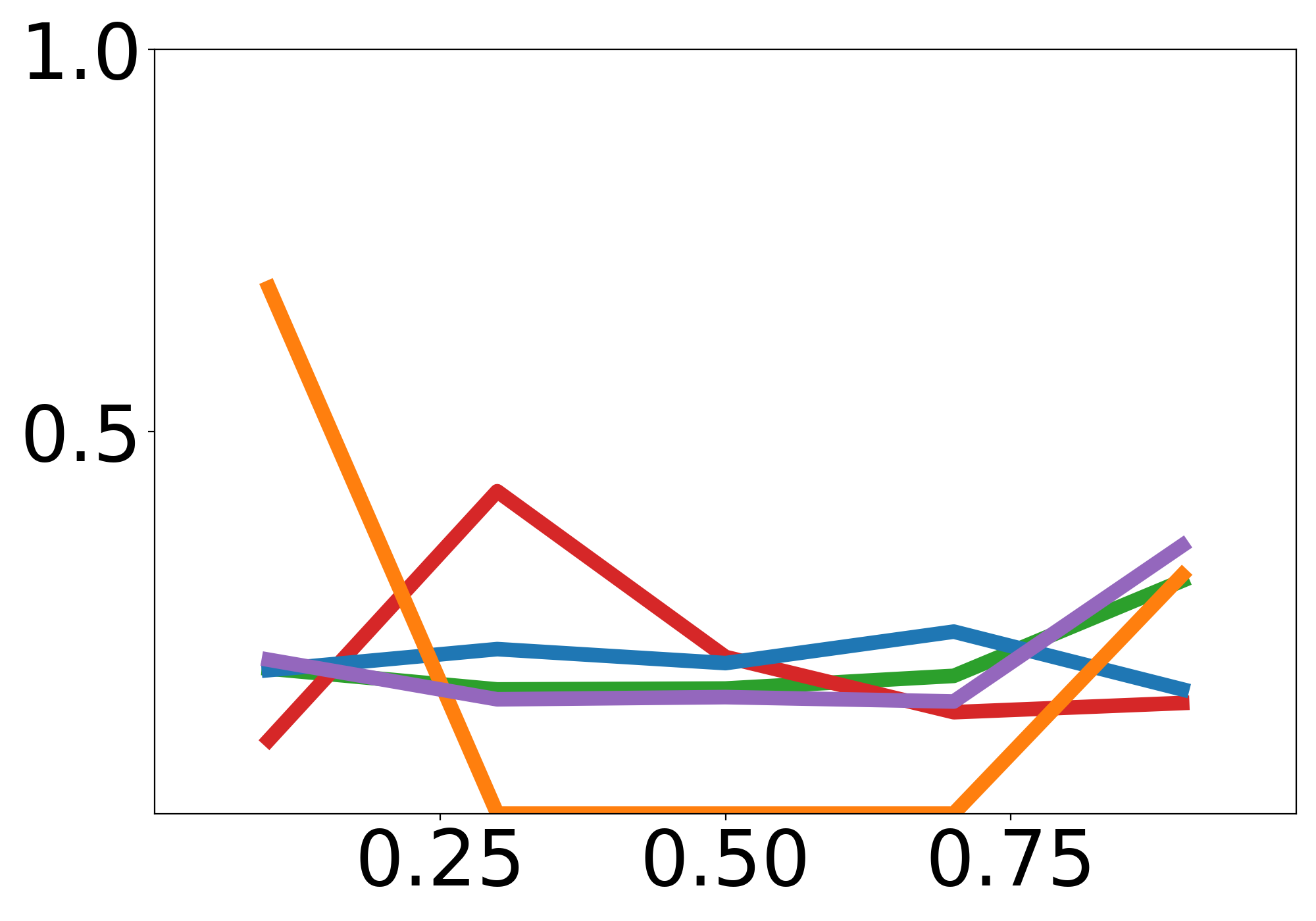}} &
\subcaptionbox{}{\includegraphics[width=\linewidth]{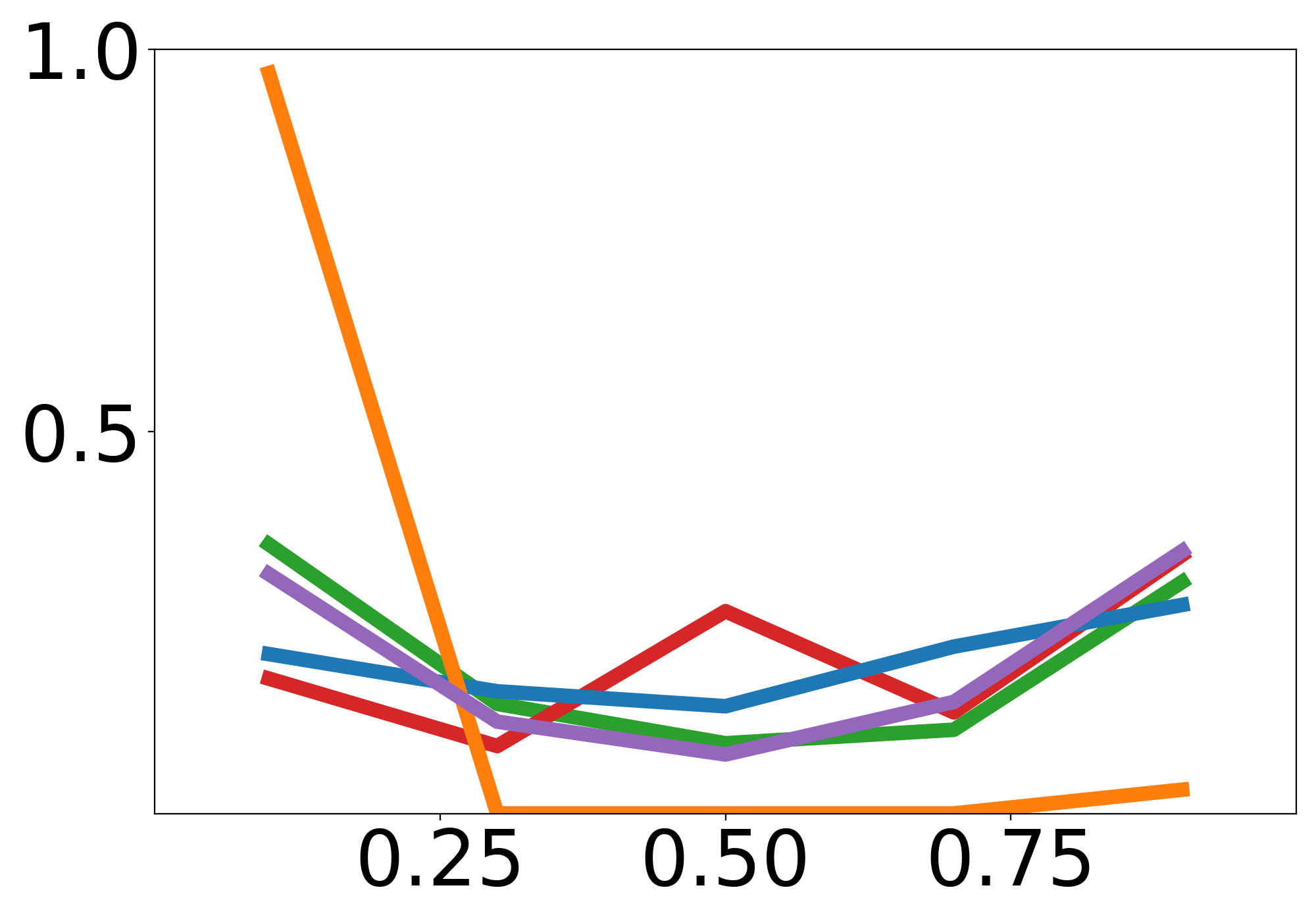}} &
\subcaptionbox{}{\includegraphics[width=\linewidth]{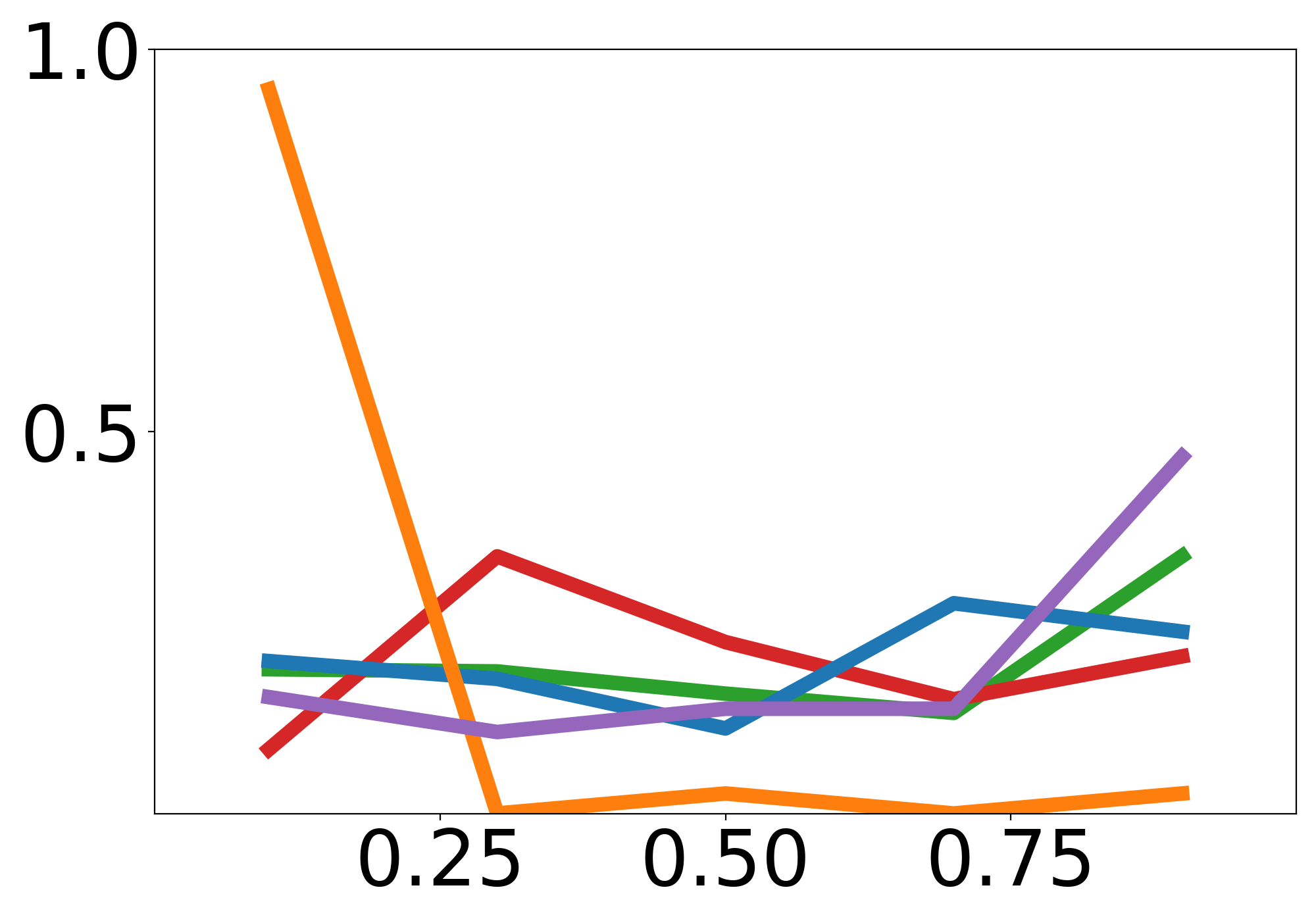}} &
\subcaptionbox{}{\includegraphics[width=\linewidth]{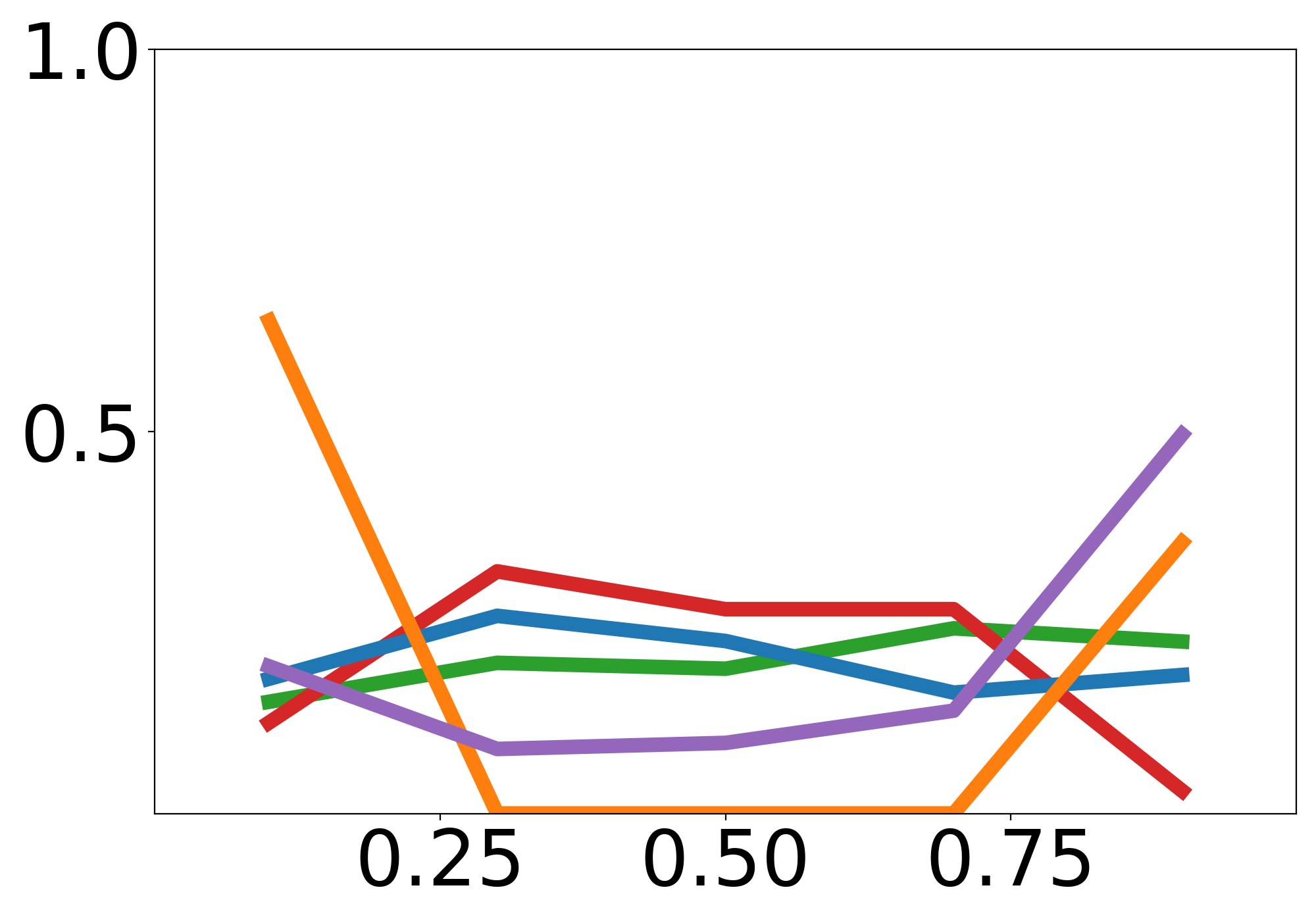}} \\

\RowStart
\subcaptionbox{}{\includegraphics[width=\linewidth]{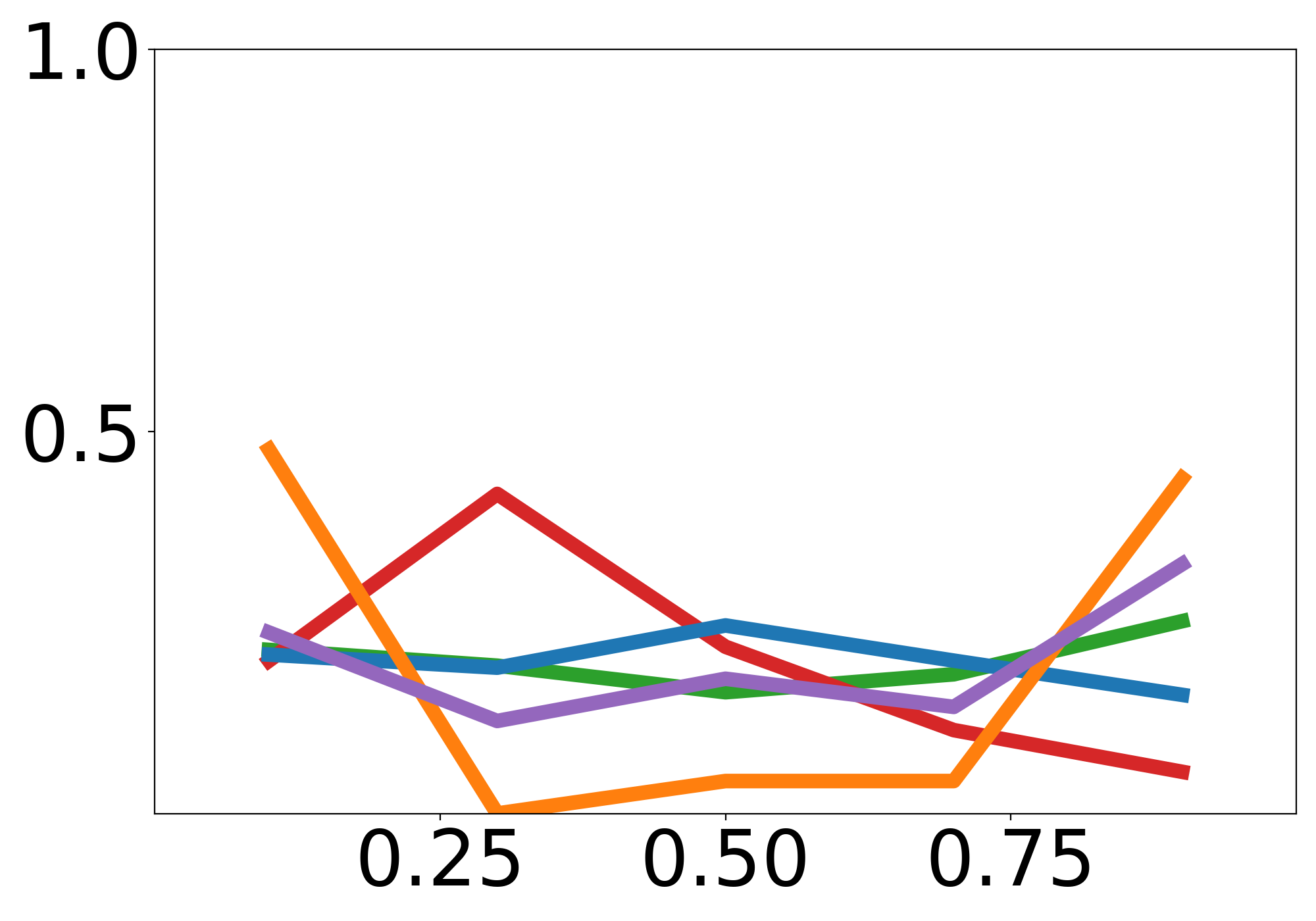}} &
\subcaptionbox{}{\includegraphics[width=\linewidth]{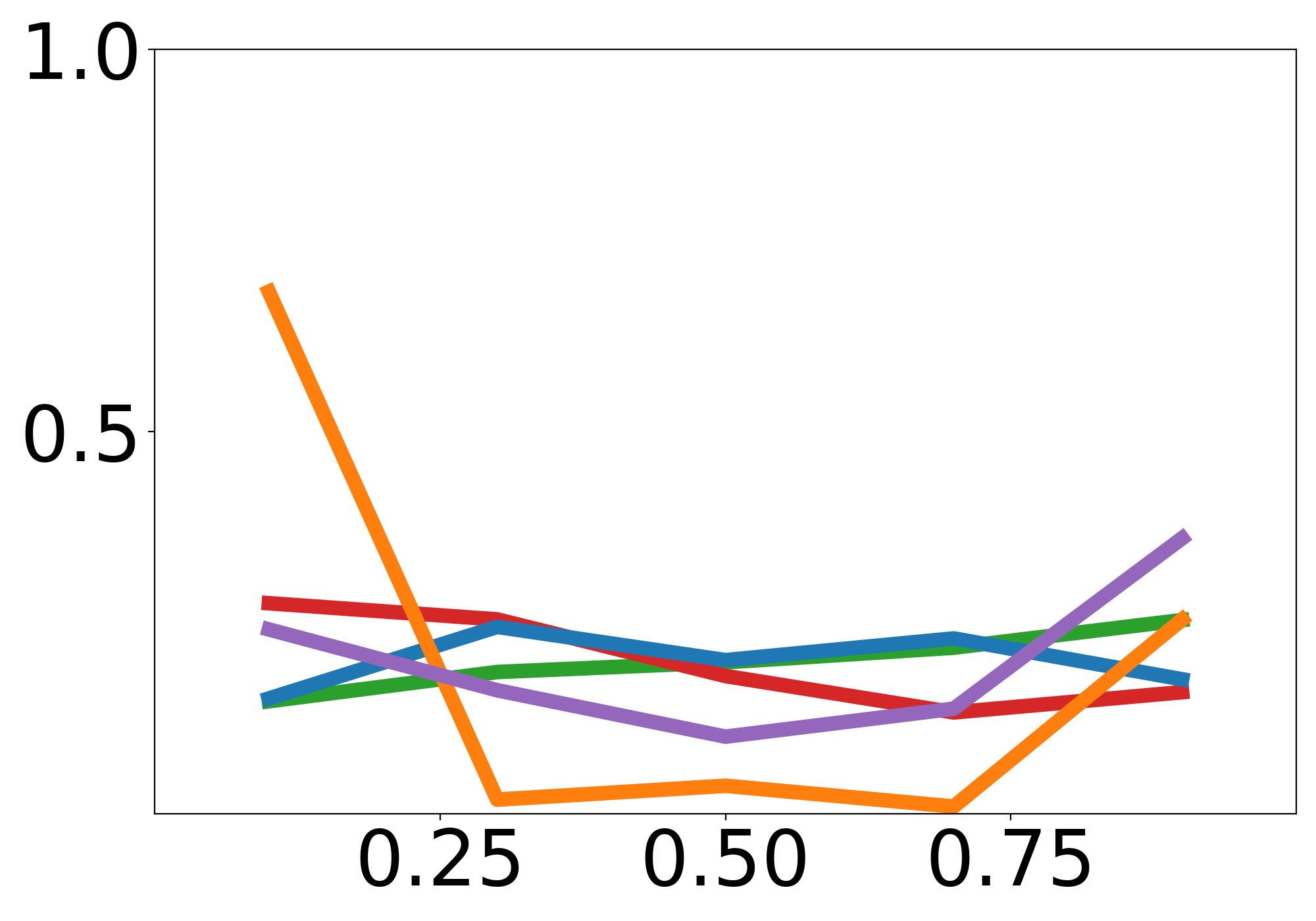}} &
\subcaptionbox{}{\includegraphics[width=\linewidth]{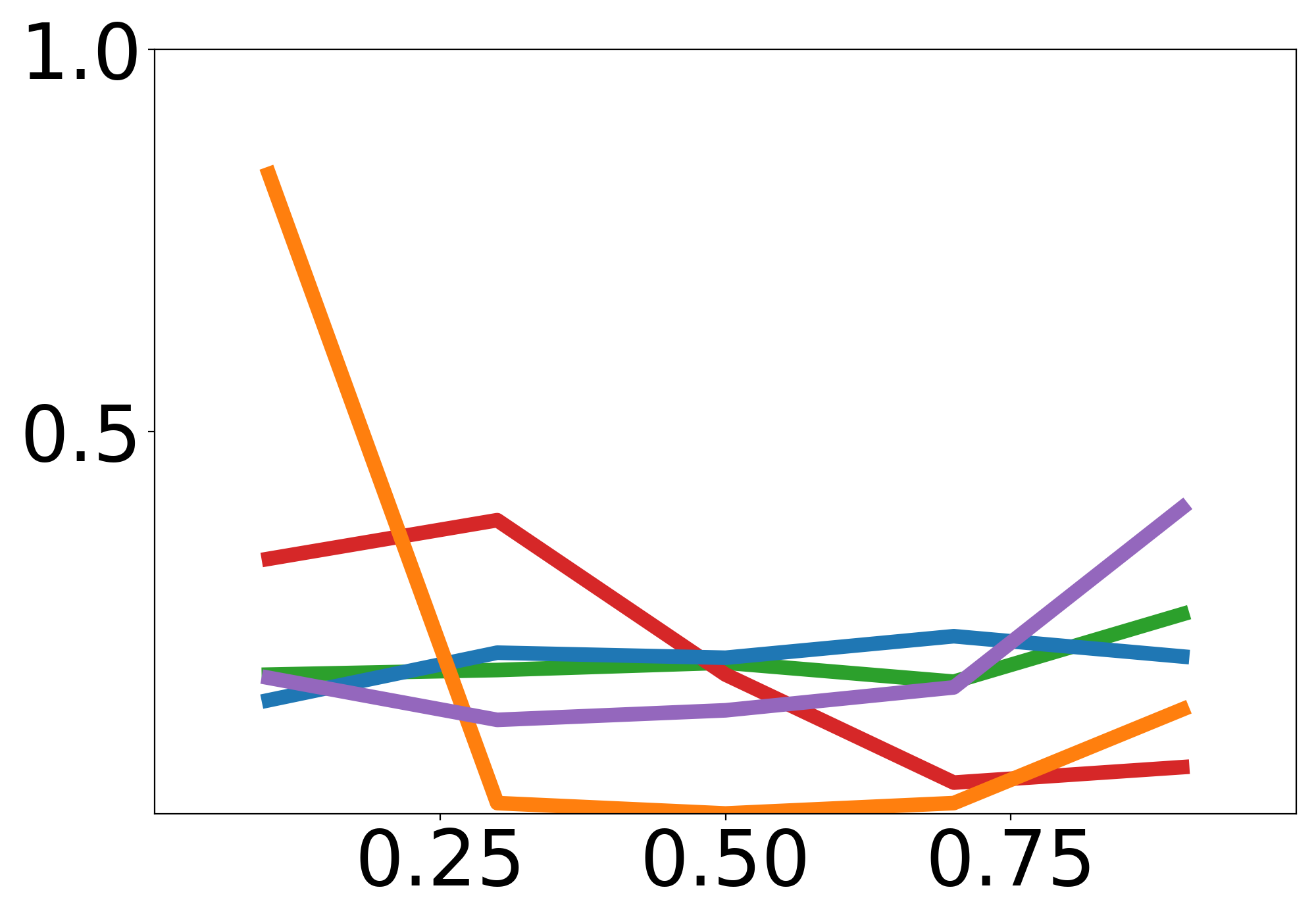}} &
\subcaptionbox{}{\includegraphics[width=\linewidth]{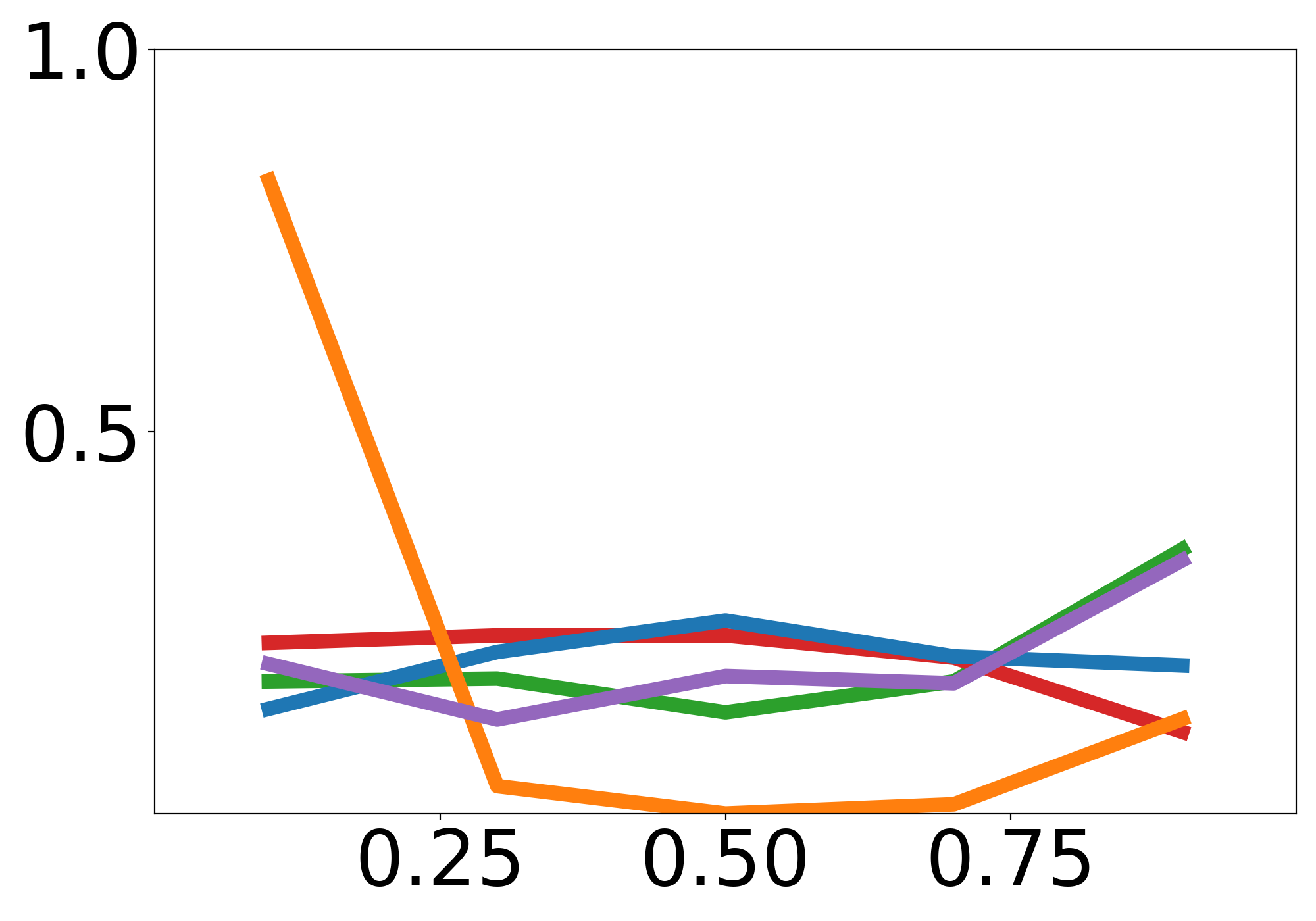}} &
\subcaptionbox{}{\includegraphics[width=\linewidth]{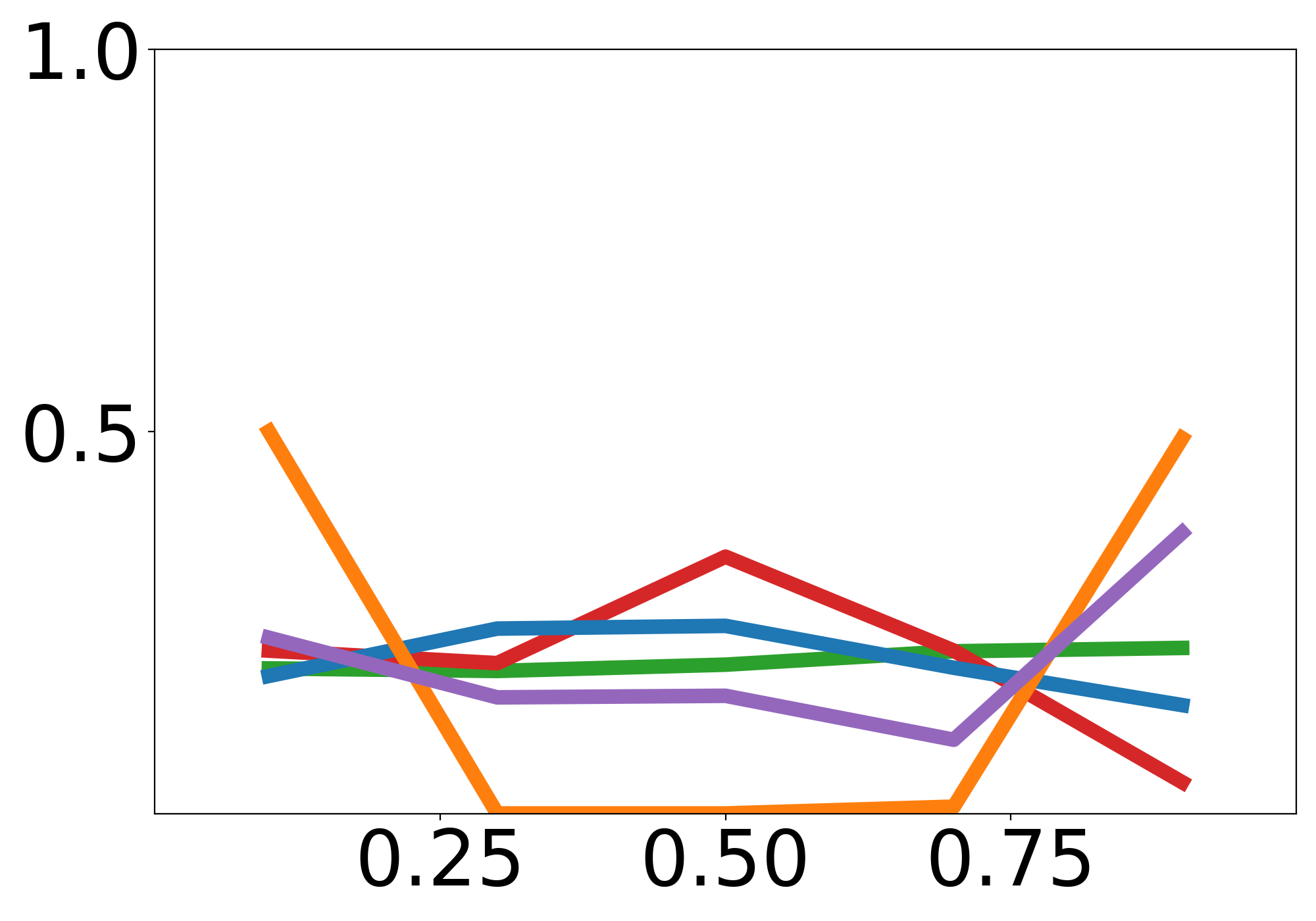}} \\

\RowStart
\subcaptionbox{}{\includegraphics[width=\linewidth]{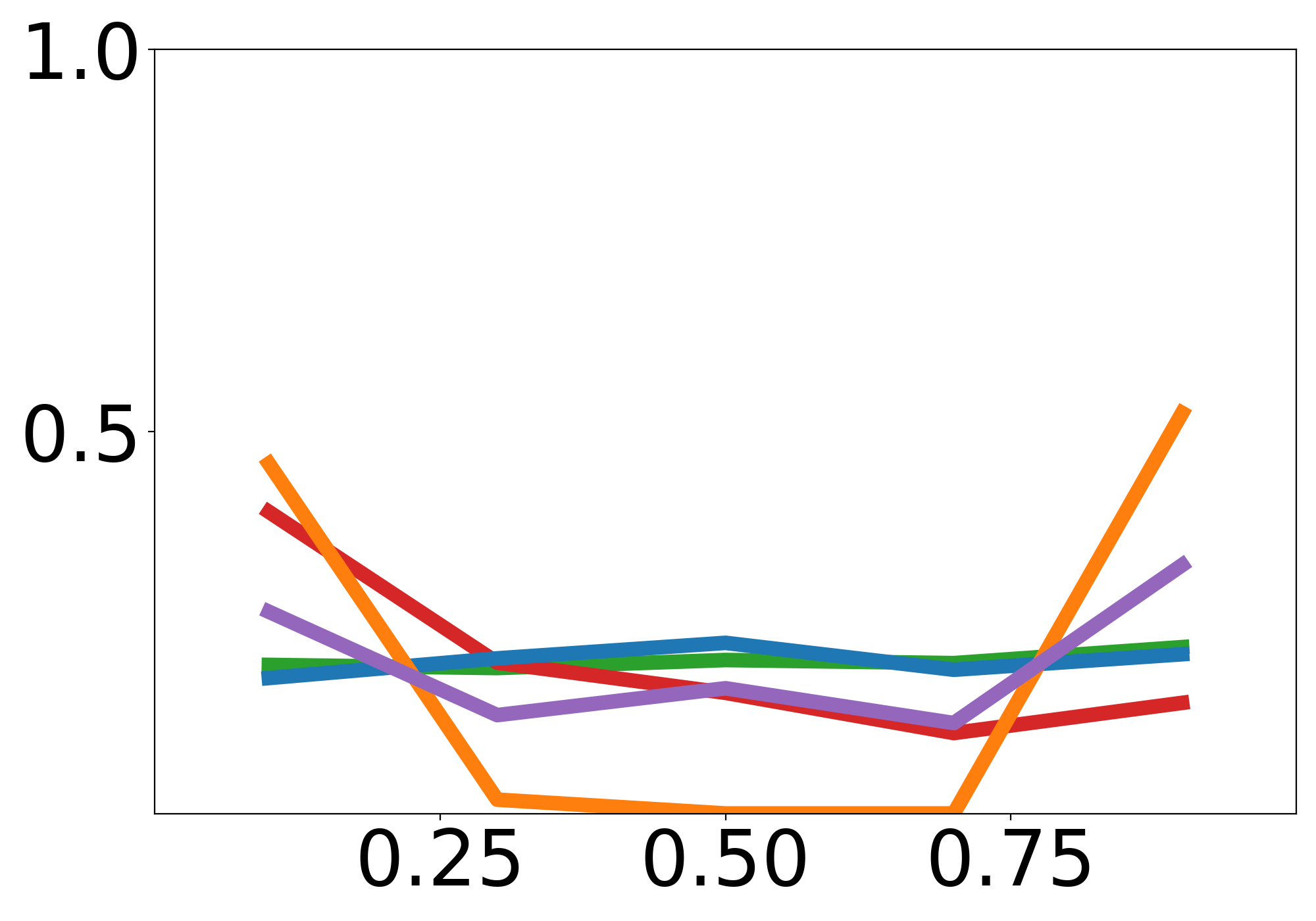}} &
\subcaptionbox{}{\includegraphics[width=\linewidth]{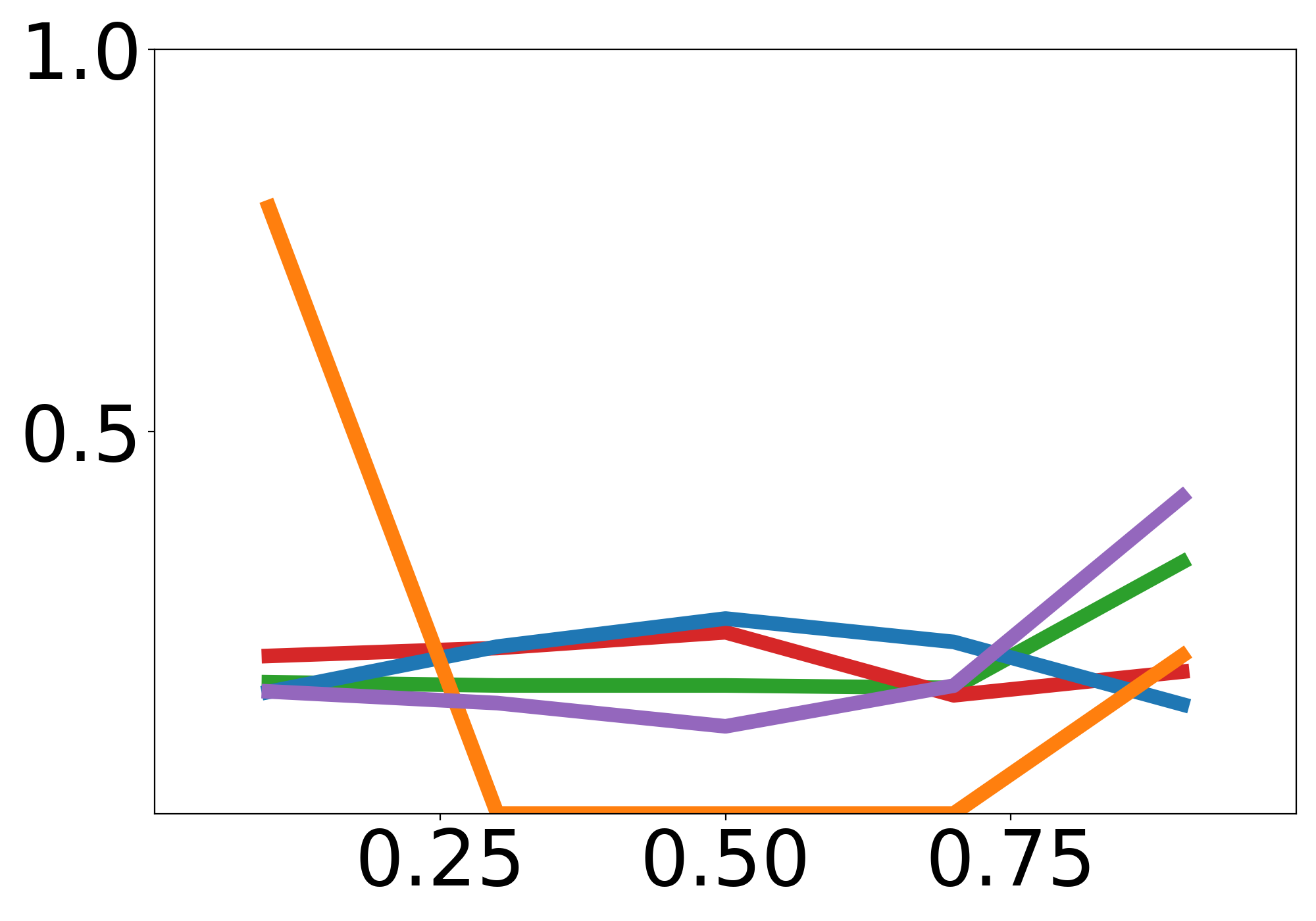}} &
\subcaptionbox{}{\includegraphics[width=\linewidth]{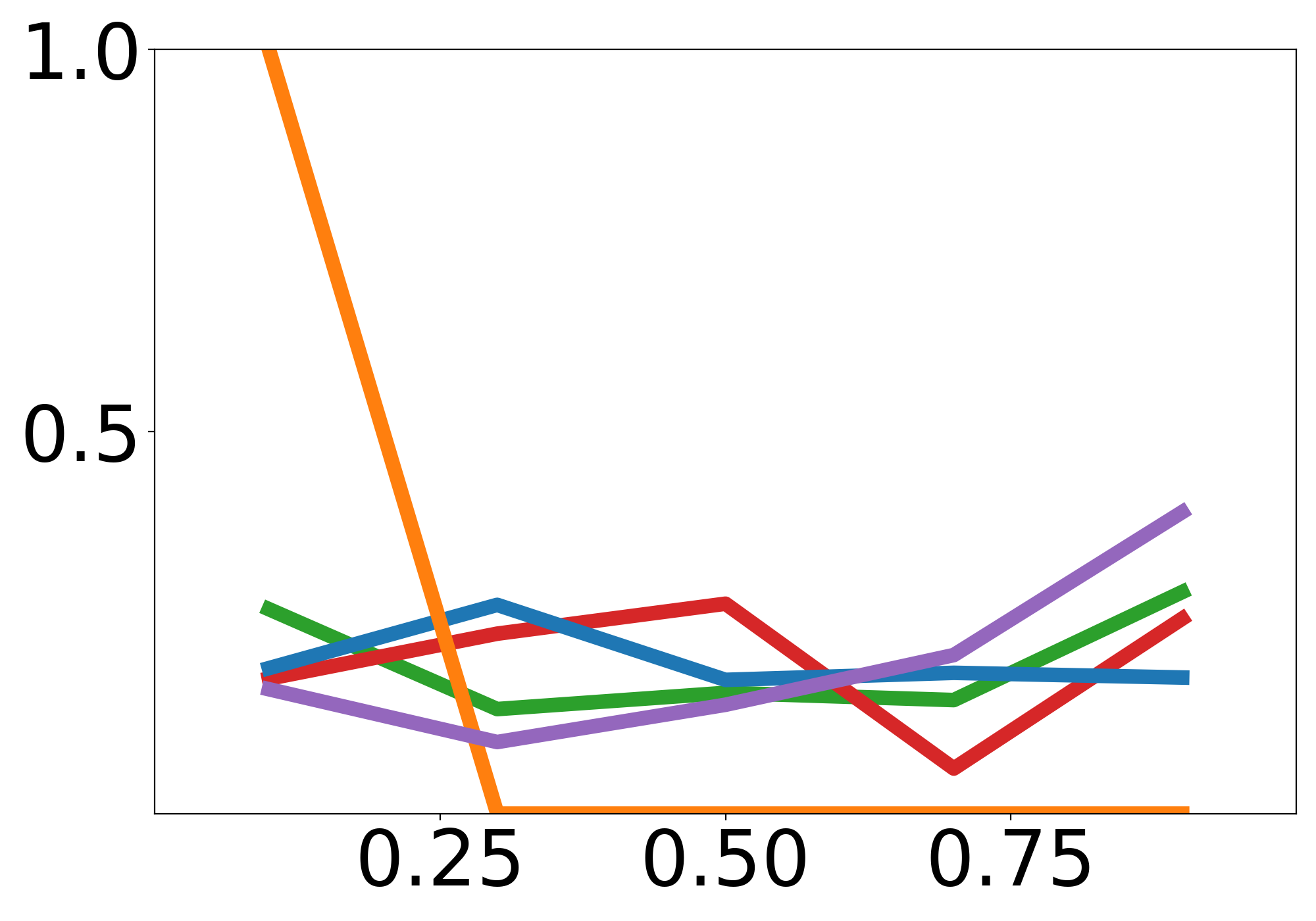}} &
\subcaptionbox{}{\includegraphics[width=\linewidth]{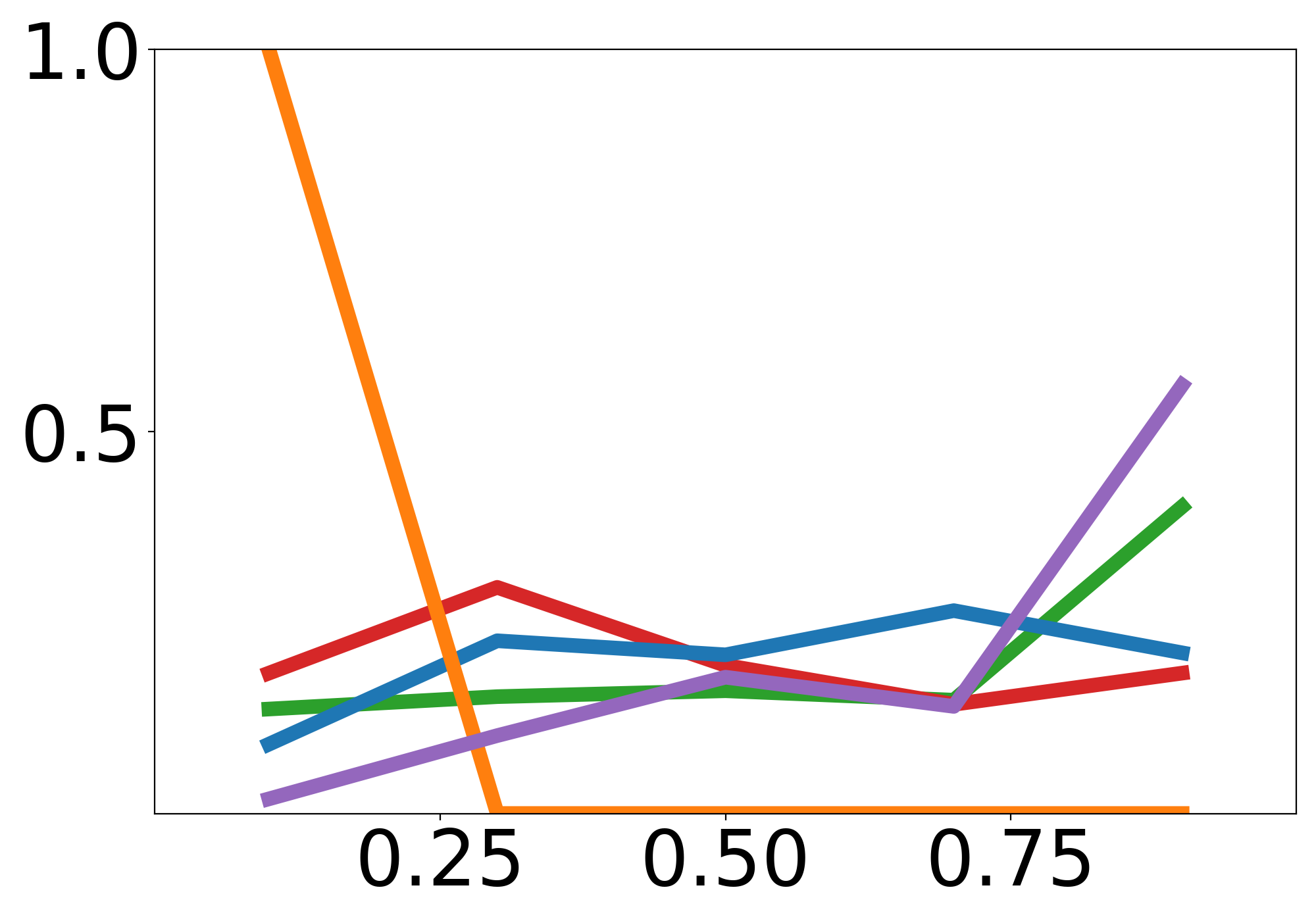}} &
\subcaptionbox{}{\includegraphics[width=\linewidth]{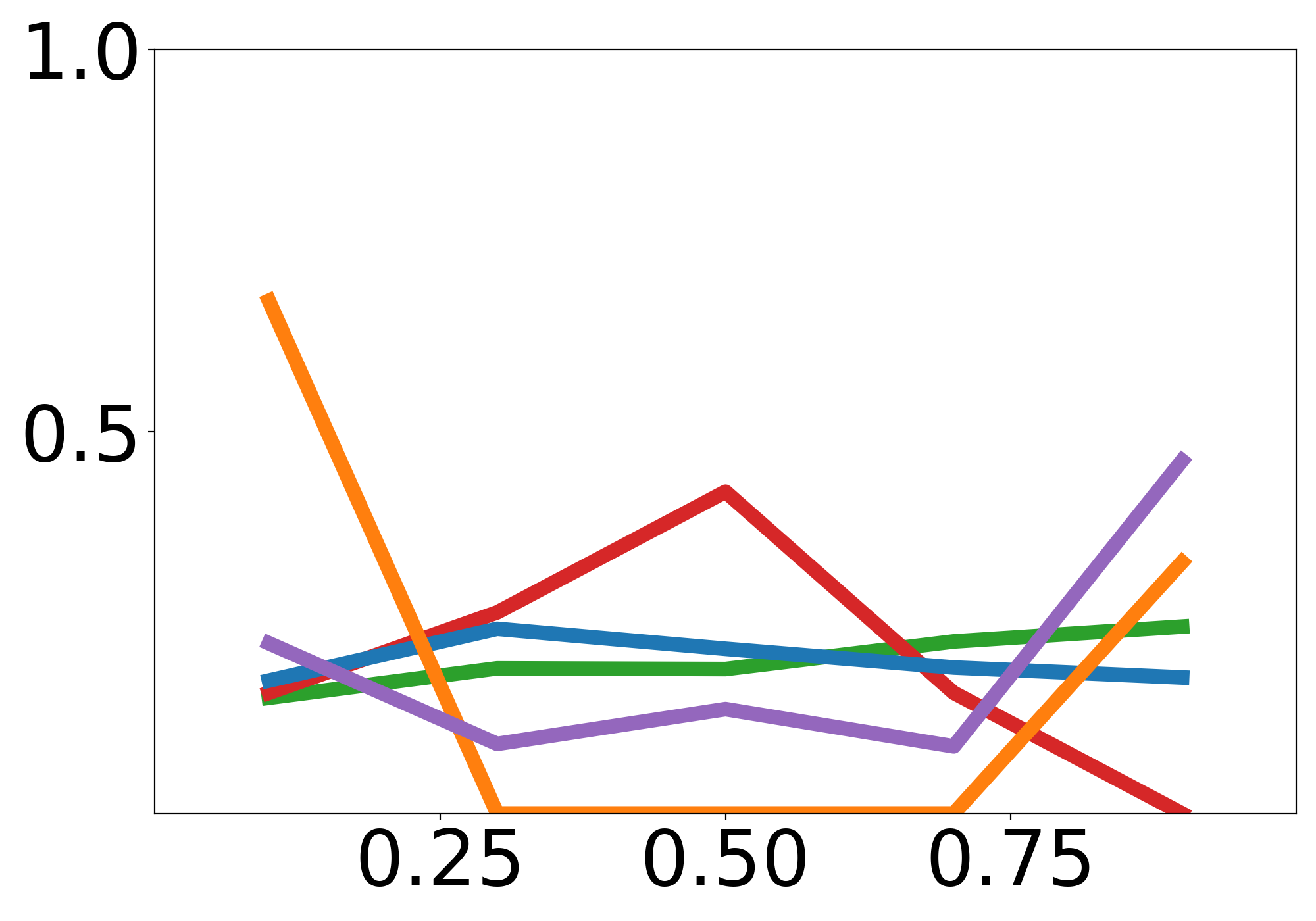}} \\

\RowStart
\subcaptionbox{}{\includegraphics[width=\linewidth]{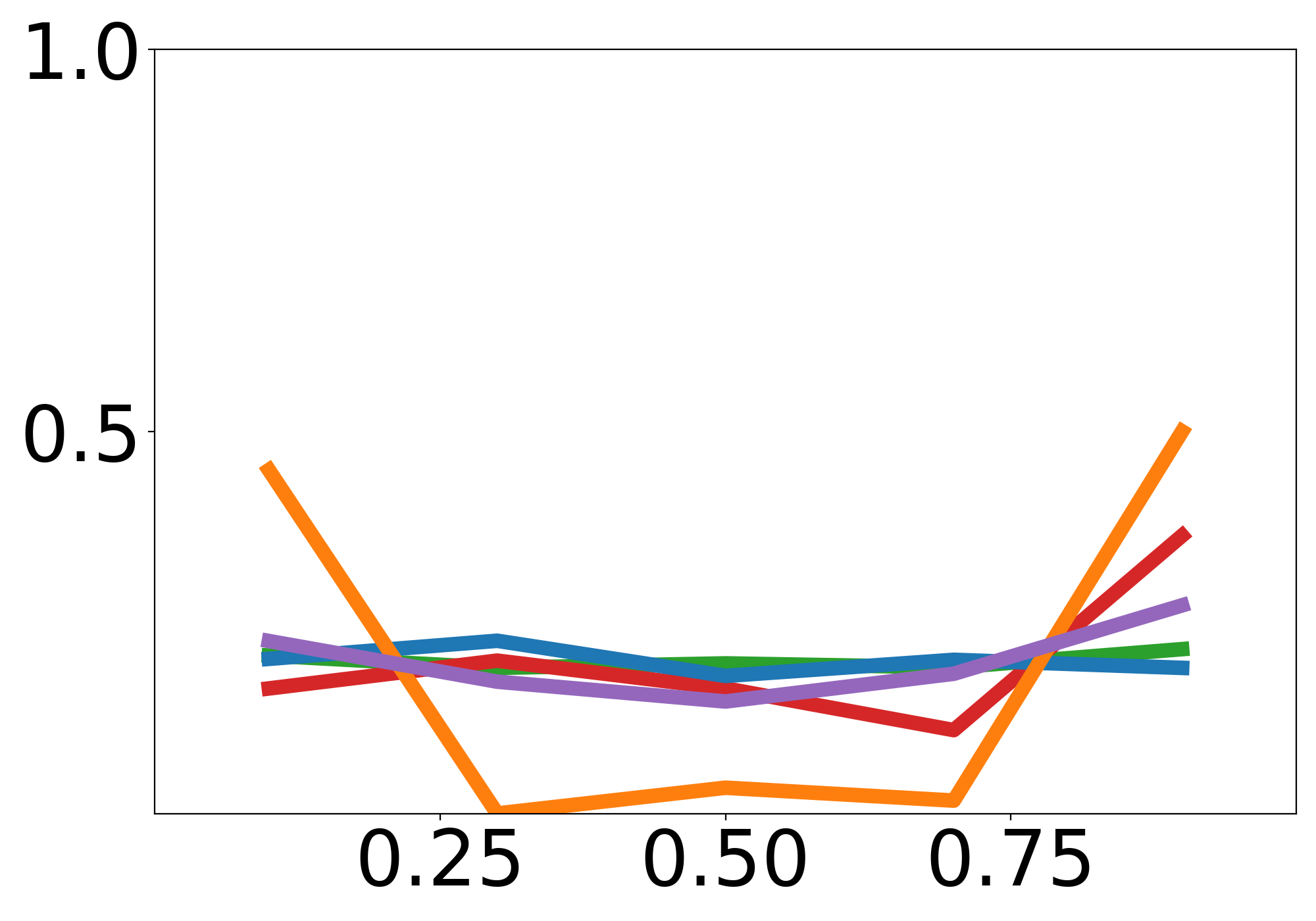}} &
\subcaptionbox{}{\includegraphics[width=\linewidth]{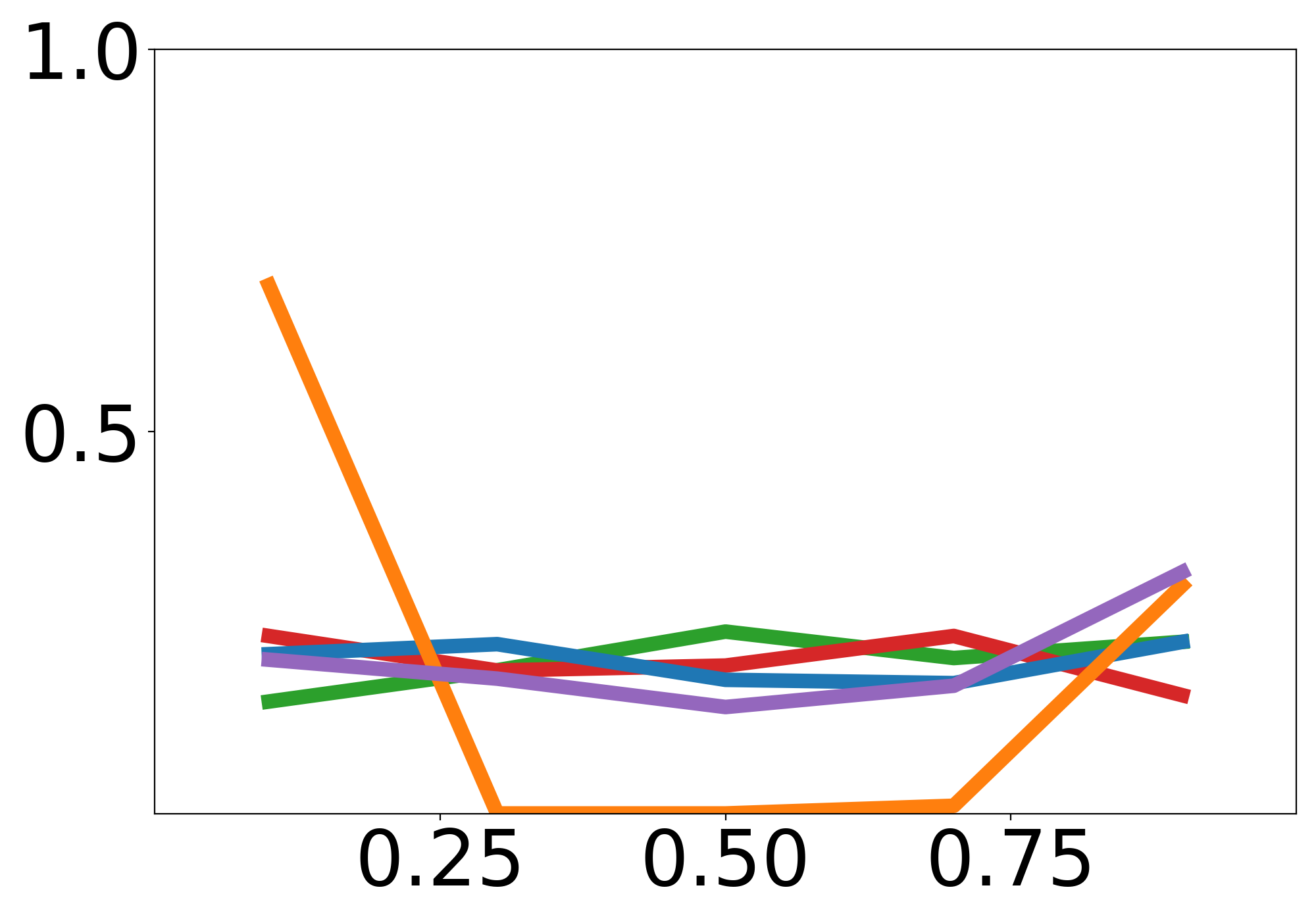}} &
\subcaptionbox{}{\includegraphics[width=\linewidth]{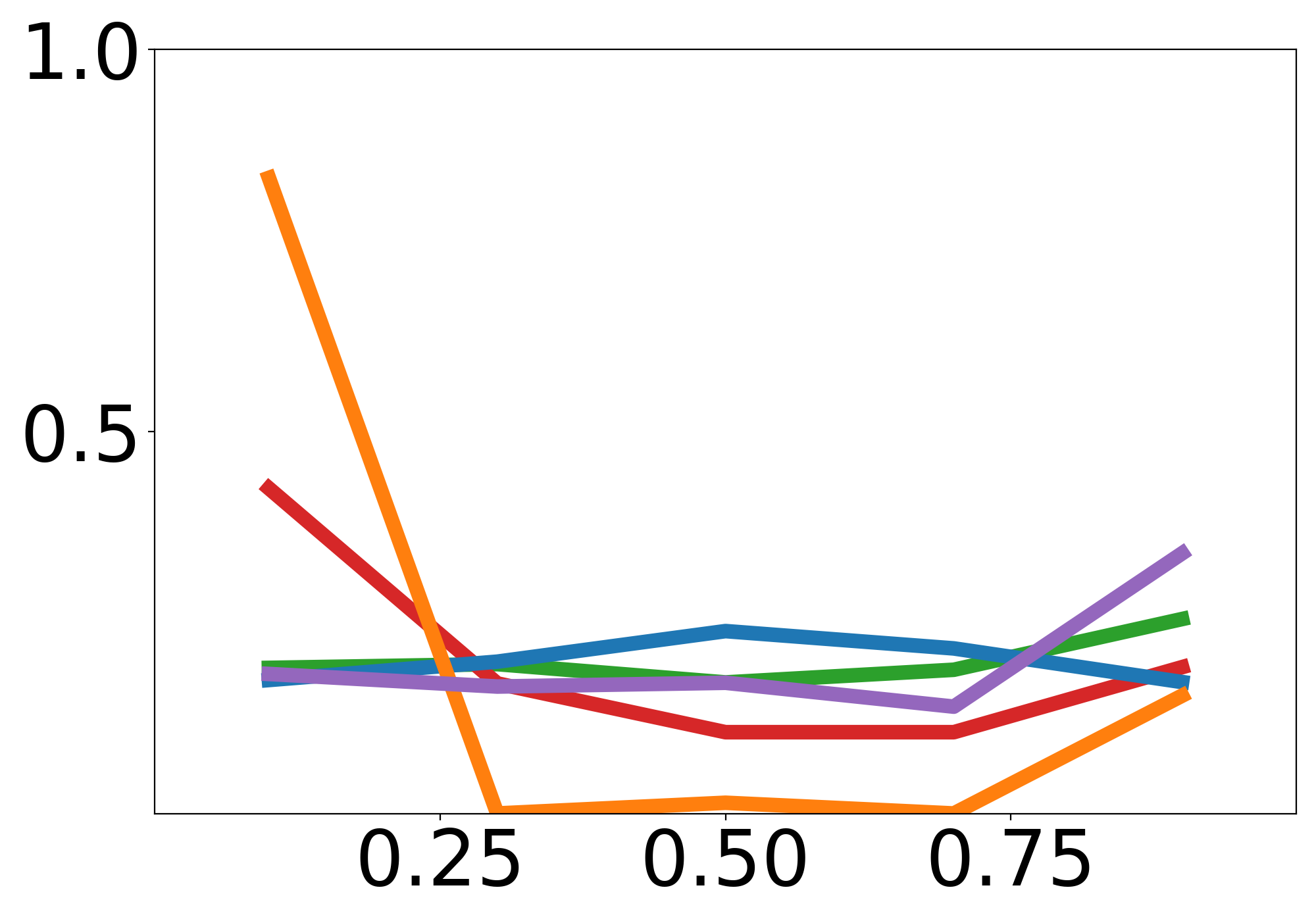}} &
\subcaptionbox{}{\includegraphics[width=\linewidth]{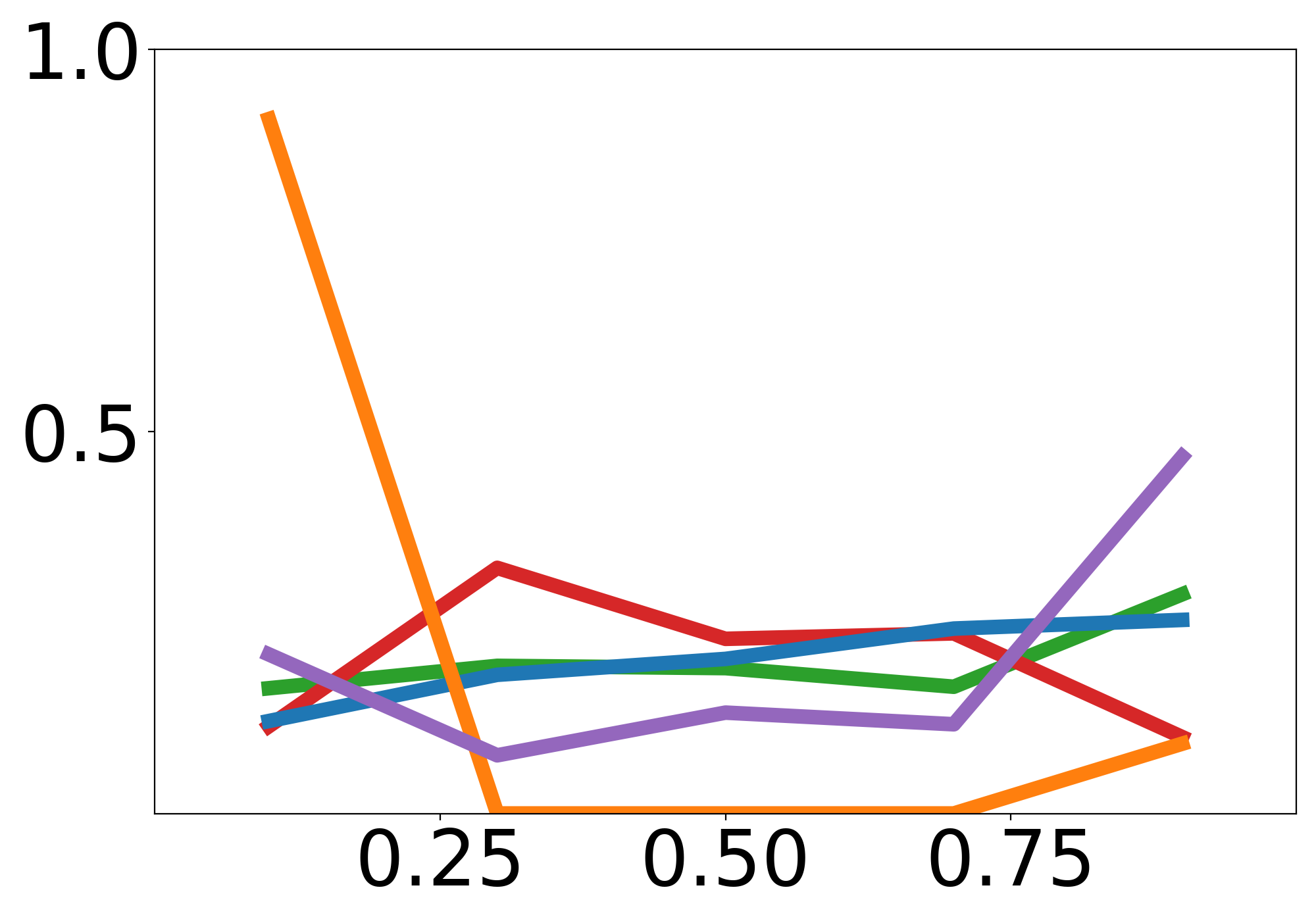}} &
\subcaptionbox{}{\includegraphics[width=\linewidth]{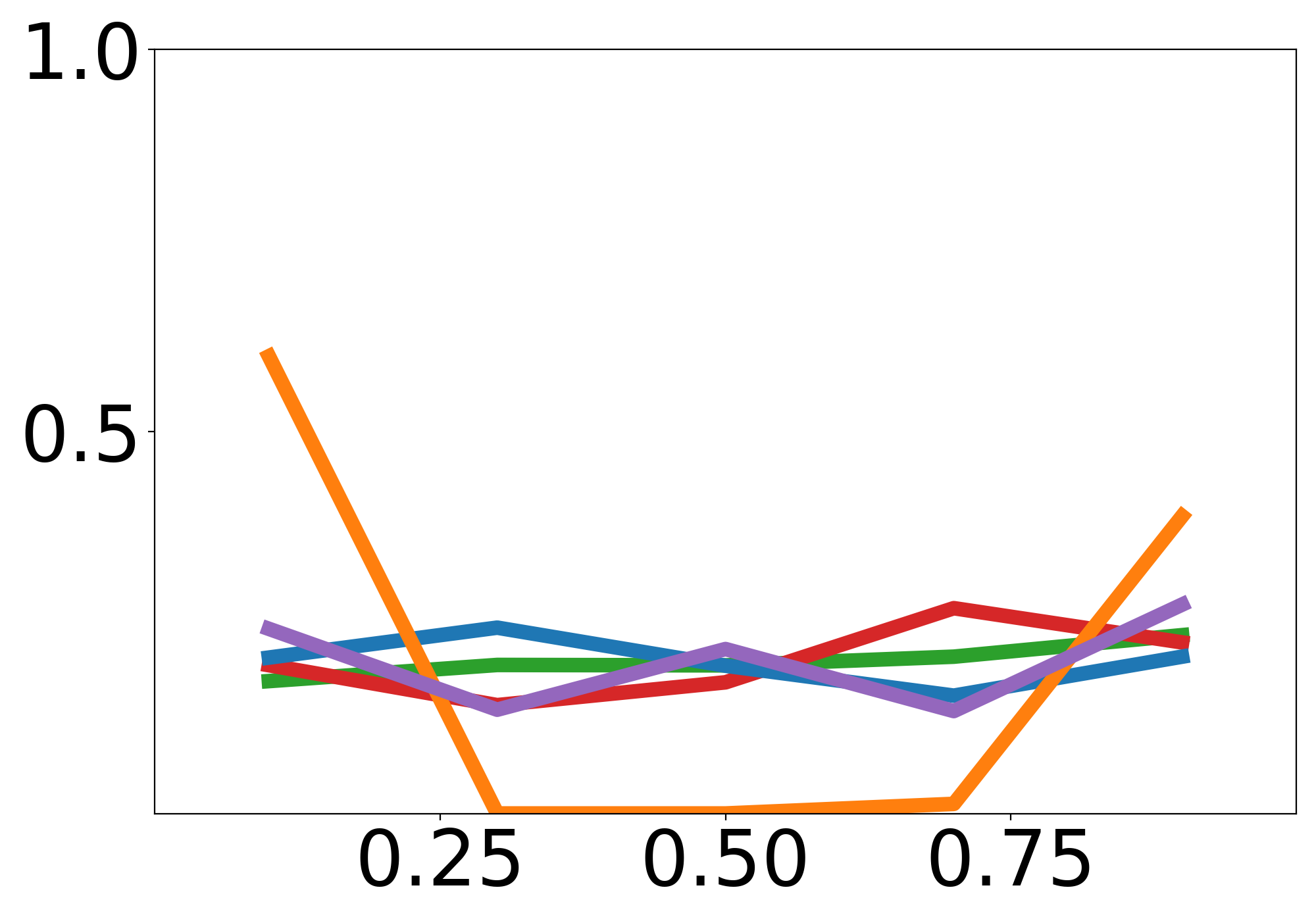}} \\
\bottomrule
\end{tabular}
\caption{Distributions of stance types across relative positions in author responses generated by five LLMs (a--e) under nine settings (1--9), together with authentic human responses (1.f). The x-axis indicates the relative position within a response, and the y-axis shows the proportion of each stance type at that position. Stance categories include cooperative (green), defensive (red), hedge (blue), social (orange), and other (purple).}
\label{fig:discourse_rel_pos}
\end{figure*}

\begin{figure*}[ht]
  \centering
  \includegraphics[width=0.99\textwidth]{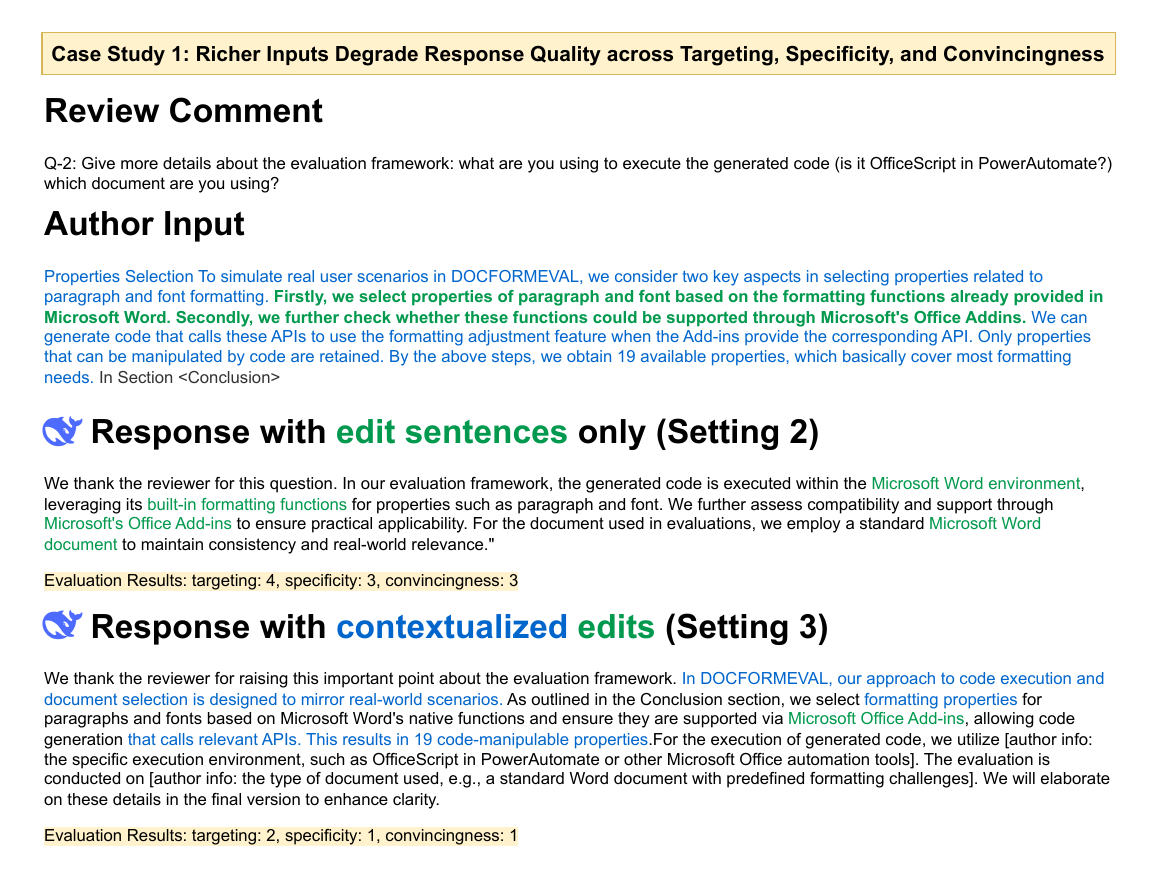}
  \caption{Case Study 1: Richer Inputs Degrade Response Quality Across Targeting, Specificity, and Convincingness.
In the Author Input block, sentences marked in green represent the sentence-level edit inputs used in Setting 2, while context marked in blue indicates the surrounding paragraph context included as additional author input in Setting 3. The final two blocks show responses generated by DeepSeek under Setting 2 (edit sentences only) and Setting 3 (richer contextualized edits), with information drawn from edit sentences marked in green and information drawn from paragraph context marked in blue.
In this case, the review poses a direct question requiring a clear and focused answer. The edit sentences alone already contain the key information needed to address it. Adding paragraph context introduces irrelevant details and creates ambiguity around the core answer, resulting in a response that is less targeted, less specific, and less convincing.}
  \label{fig:app_case_study_1}
\end{figure*}

\begin{figure*}[ht]
  \centering
  \includegraphics[width=0.99\textwidth]{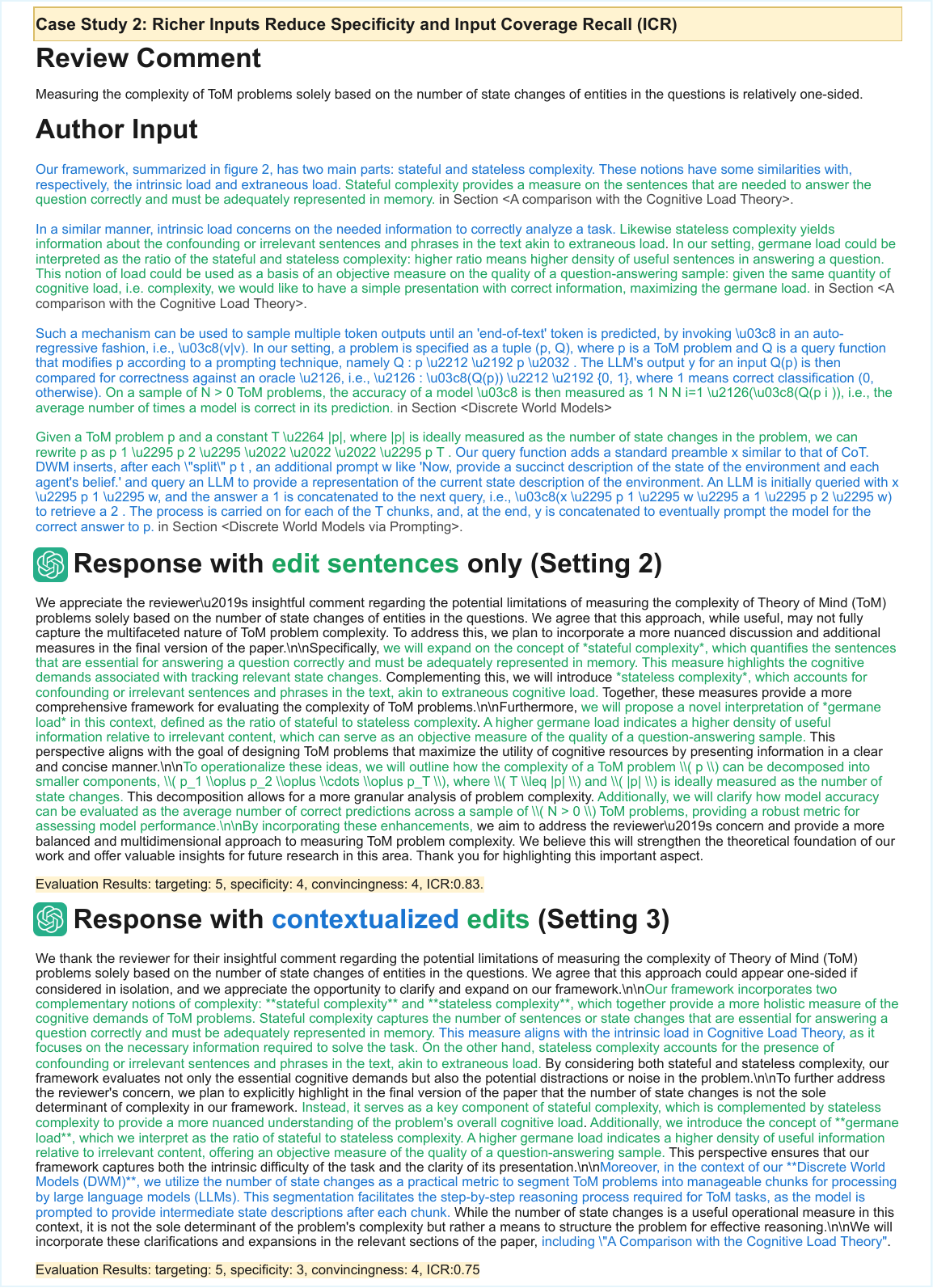}
  \caption{Case Study 2: Richer Inputs Reduce Specificity and ICR.
In Author Input, green indicates edit sentences (Setting 2) and blue indicates the additional paragraph context (Setting 3). The final two blocks show GPT-4o responses under each setting, with information sources color-coded accordingly. Here, the review raises a criticism requiring careful argumentation. The authors define stateful and stateless complexity, justify their necessity, and introduce notations for operationalization. In Setting 3, the richer context shifts focus and causes omission of the notations, reducing both specificity and ICR.}
  \label{fig:app_case_study_2}
\end{figure*}

\end{document}